\documentclass[12pt]{article}
\usepackage{graphicx} % Required for inserting images
\usepackage{amssymb}
\usepackage{amsmath}
\usepackage{color}
\usepackage{xcolor}
\usepackage{algorithm}
\usepackage{algpseudocode}
\usepackage{lineno}
\usepackage{hyperref}
\usepackage{xurl}

\usepackage{booktabs}
\usepackage{adjustbox}

\usepackage{amsthm}
\usepackage{authblk}
\theoremstyle{remark}
\newtheorem{remark}{Remark}
\newtheorem{definition}{Definition}

\newcommand{\code}[1]{\texttt{#1}}
\def\tsc#1{\csdef{#1}{\textsc{\lowercase{#1}}\xspace}}
\tsc{WGM}
\tsc{QE}

\usepackage[authoryear]{natbib}
\title{Robust LassoNet: Enhancing Feature Selection in Neural Networks via Robust Loss Functions}

\author[1]{Daniela De Canditiis}
\author[2]{Italia De Feis}
\author[1]{Paola Stolfi}

\affil[1]{Istituto per le Applicazioni del Calcolo ``Mauro Picone'', Consiglio Nazionale delle Ricerche, Roma, Italy}
\affil[2]{Istituto per le Applicazioni del Calcolo ``Mauro Picone'', Consiglio Nazionale delle Ricerche, Napoli, Italy}

\date{\today}

\begin{document}

\maketitle
\begin{abstract}
Feature selection in neural networks remains a challenging problem, particularly in the presence of noisy or contaminated data. LassoNet is a recent approach that addresses this issue by combining neural networks with hierarchical sparsity constraints, enabling simultaneous prediction and variable selection. However, its standard formulation relies on the mean squared error (MSE) loss, which is known to be highly sensitive to outliers.
In this paper, we present Robust LassoNet, an extension of LassoNet that incorporates robust loss functions—such as Huber, Cauchy, Tukey’s bisquare, and Nonnegative Garrote—to mitigate the effect of extreme observations. The proposed approach preserves the original optimization framework while improving stability under data contamination.
Through experiments on synthetic and real datasets, we show that robust LassoNet significantly improves both predictive performance and feature selection accuracy in the presence of outliers or heavy-tailed noise,
while maintaining comparable performance in standard gaussian settings.We also compare Robust LassoNet with robust benchmark methods, including robust Random Forest, robust XGBoost, and robust (groupwise) least angle regression. These results highlight the importance of robustness in neural network–based feature selection and suggest practical guidelines for choosing appropriate loss functions.
\end{abstract}

\vspace{0.5cm}
{\bf Keywords}: Robust regression, Feature selection, Neural networks, M-estimators, Hierarchical sparsity

% Main text
\section{Introduction}\label{sec:intro}
Machine learning methods, and in particular supervised learning approaches, play a central role in modern statistical inference and predictive modeling. In recent years, the increasing availability of high-dimensional and complex datasets has fostered the widespread adoption of highly flexible nonlinear models, among which deep neural networks have demonstrated remarkable predictive capabilities across a broad range of applications. Despite their success, however, interpretability and feature relevance remain important open challenges in deep learning models.

In supervised regression problems, a key objective is not only to achieve high predictive accuracy but also to identify which input variables are most informative for predicting the response. Feature selection improves model interpretability, reduces computational complexity, and enhances generalization performance, especially in high-dimensional settings where irrelevant or redundant variables may induce instability and overfitting.

Among sparse learning approaches, the Least Absolute Shrinkage and Selection Operator (LASSO) \citep{tibshirani1996lasso} represents one of the most influential methodologies for simultaneous estimation and variable selection. By introducing an $\ell_1$ penalty on the regression coefficients, the LASSO promotes sparse solutions and automatically removes irrelevant variables from the model. Owing to its interpretability and computational efficiency, $\ell_1$ regularization has become a cornerstone of modern statistical learning. Several extensions of the LASSO have subsequently been proposed to address some of its limitations. The Elastic Net \citep{zou2005elasticnet} combines $\ell_1$ and $\ell_2$ penalties to improve stability in the presence of correlated predictors. Non-convex penalties such as SCAD \citep{fan2001scad} and MCP \citep{zhang2010mcp} aim to reduce the estimation bias introduced by the LASSO on large coefficients while preserving sparsity. Other approaches, such as the group LASSO \citep{yuan2006grouplasso}, the group bridge \citep{Huang2009}  and non-convex grouped penalties \citep{Breheny2009}, extend sparse regularization to grouped variable structures.

Although these methodologies have proven highly successful in parametric and classical nonparametric models, their extension to deep neural networks is considerably more challenging.
In neural architectures, the relationship between model parameters and input variables is no longer one-to-one, since information propagates through multiple hidden layers and highly nonlinear transformations. Consequently, standard weight penalization strategies do not directly guarantee sparsity at the input-variable level. This limitation has motivated the development of neural architectures specifically designed to integrate feature selection within the learning process.

Among these approaches, LassoNet \citep{lemhadri2021lassonet} provides an elegant framework combining sparse linear modeling and deep neural networks. LassoNet introduces a residual architecture composed of a linear skip connection and a nonlinear feed-forward network, together with a hierarchical sparsity constraint linking the two components. This formulation allows a variable to contribute to the nonlinear component only if it is active in the linear part of the model, thereby enabling explicit feature selection while preserving the expressive power of neural networks.

Despite its flexibility and effectiveness, the original LassoNet formulation relies on the Mean Squared Error (MSE) loss function, which implicitly assumes Gaussian-distributed errors. This assumption may lead to substantial performance degradation in the presence of outliers, heavy-tailed noise, or corrupted observations, all of which frequently occur in real-world applications. Since the quadratic loss is highly sensitive to large residuals, even a small fraction of anomalous observations may significantly affect both prediction accuracy and variable selection stability. Robust statistics addresses this problem by developing estimation procedures that remain stable under small deviations from the assumed probabilistic model. A central idea in robust regression is to replace the quadratic loss with alternative loss functions that reduce the influence of large residuals.
 Foundational contributions were provided by Huber \citep{huber1964robust} and Hampel \citep{hampel1971general}, who introduced the principles of M-estimation and influence-function analysis; for a comprehensive view see \citep{rey1983}, \citep{hampel1986robust}, \citep{rousseeuw1987robust}, \citep{maronna2006} and \citep{maronna2019}.

To address the LassoNet limitation in contaminated scenarios, we propose \emph{Robust LassoNet} a framework obtained by replacing the classical MSE objective with robust loss functions. In particular, we investigate several established losses, including the Huber, Cauchy, Tukey’s bisquare, and Nonnegative Garrote, each characterized by different robustness-efficiency trade-offs. The proposed framework preserves the hierarchical sparsity structure and proximal optimization strategy of LassoNet while improving robustness against data contamination.

The main contributions of this work can be summarized as follows:
\begin{itemize}
    \item We introduce a robust extension of the LassoNet framework by integrating robust loss functions within its hierarchical sparse neural architecture.
    
    \item We demonstrate that these robust objectives can be naturally incorporated into the proximal gradient optimization framework underlying LassoNet training.
    
    \item We empirically evaluate the proposed methodology on both synthetic and real datasets, assessing its predictive performance, variable selection properties, and robustness under different contamination scenarios, and comparing it with benchmark robust methods.
\end{itemize}

The remainder of the paper is organized as follows. Section 2 reviews the main literature on sparse regularization, feature selection in neural networks, and robust learning methods. Section 3 introduces the Robust LassoNet framework and its optimization procedure. Section 4 presents the experimental results, and Section 5 contains conclusions.

\section{Background and Related Work}
\label{sec:background}
\subsection{Feature Selection in Neural Networks}

Feature selection in neural networks has attracted increasing attention in recent years owing to the growing demand for interpretable and parsimonious deep learning models. Early approaches mainly relied on pruning strategies and sensitivity analysis techniques \citep{Leray1999}. In pruning-based approaches, low-importance connections are removed after training; for comprehensive reviews, see  \citep{cheng_etal2024} and \citep{hoefler_etal2021}. In sensitivity analysis approaches, input relevance is quantified by measuring how the output sensitivity varies with respect to perturbations in the inputs, often through derivatives \citep{zurada1994sensitivity}. 

Although effective in reducing network complexity, these approaches are typically post hoc procedures and do not directly integrate variable selection into the learning process. More recent methodologies incorporate sparsity-inducing penalties within the neural network optimization framework itself. For instance, input-level regularization and sparse input layers have been proposed to promote feature sparsity during training \citep{Zhao2015, scardapane2017group,  feng_simon_2017_spinn, ZHANG2024, Fan2024_b}. These methods generally impose structured penalties on the weights connected to the input layer, thereby enabling feature selection to be optimized through backpropagation.

Among modern sparse neural network architectures, LassoNet \citep{lemhadri2021lassonet} represents one of the most principled approaches to feature selection in deep learning. By combining a linear skip connection with a nonlinear feed-forward network and imposing a hierarchical $\ell_1$ constraint, LassoNet achieves explicit variable-level sparsity while preserving the flexibility of nonlinear models. This hierarchical structure ensures that a feature may contribute to the nonlinear component only if it is active in the linear part of the architecture, thereby enhancing interpretability and improving the consistency of feature selection.

\subsection{Robustness in Neural Network}

The integration of robust loss functions into deep learning models has recently attracted considerable attention, particularly in the presence of noisy labels, corrupted observations, and heavy-tailed data distributions. Robust objective functions have been employed in both regression and classification settings to improve training stability and predictive performance under data contamination.

Among recent contributions, \citep{zhang2018gce} proposed robust learning strategies for deep neural networks in noisy-label classification settings, showing improved resilience to corrupted training data. Similarly, \citep{barron2019robust} introduced a general and adaptive family of robust loss functions that unifies several classical losses, including the mean squared error, Huber, and Cauchy losses, within a common framework applied to generative image synthesis and unsupervised monocular depth estimation.

In the context of robust nonparametric regression, several methods have been proposed, together with corresponding convergence rates for learning the regression function. For instance, \citep{Lederer2020, Shen2021, Fan2024} investigated non-asymptotic error bounds for estimators obtained by minimizing robust loss functions such as the least absolute deviation loss \citep{bassett1978}, the Huber loss \citep{huber1964robust}, the Cauchy loss \citep{Andrews1972}, and Tukey's biweight loss \citep{mosteller1977}. More recently, \citep{Wang2025} proposed a new estimator that is both efficient and robust, establishing its large-sample properties in terms of excess risk and near-optimal minimax convergence rates.

These developments suggest that robust loss functions can substantially improve the stability and generalization properties of neural networks. Nevertheless, most existing approaches primarily focus on predictive robustness and do not explicitly address feature selection or sparse neural network architectures.

\subsection{Research Gap and Motivation}

LassoNet provides an effective framework for nonlinear feature selection through hierarchical sparse regularization, whereas robust loss functions offer protection against outliers and heavy-tailed contamination. However, to the best of our knowledge, no previous work has systematically incorporated robust loss functions into the LassoNet framework while preserving its hierarchical sparsity structure and proximal optimization scheme.

This gap motivates the present work. By integrating robust loss functions within the LassoNet architecture, we aim to improve both predictive stability and feature selection reliability in contaminated data settings while maintaining the interpretability and computational tractability of the original formulation.

\section{Methods} \label{sec: methods}

In supervised regression, the goal is to learn a functional relationship between a set of covariates $\mathbf{x} \in \mathbb{R}^d$ and a response $y \in \mathbb{R}$, from a training dataset $\{ (\mathbf{x}_i, y_i) \}_{i=1}^n$. In the parametric setting, this relationship is represented by a fixed, pre-defined functional form $f_{\boldsymbol{\phi}}(\mathbf{x})$, parameterized by a finite-dimensional vector $\boldsymbol{\phi}$. The parameters $\boldsymbol{\phi}$ are estimated by minimizing an empirical loss function that measures the discrepancy between the model predictions and the observed responses:

\begin{equation}
L(\boldsymbol{\phi}) = \frac{1}{n} \sum_{i=1}^{n} \ell(f_{\boldsymbol{\phi}}(\mathbf{x}_i), y_i),
\end{equation}

\noindent  where $\ell(\cdot, \cdot)$ is a loss function and $\phi$ denotes the model parameters.  
In the standard least squares setting, the loss takes a quadratic form:
\begin{equation}
\ell(f_{\boldsymbol{\phi}}(\mathbf{x}_i), y_i) = (f_{\boldsymbol{\phi}}(\mathbf{x}_i) - y_i)^2.
\end{equation}

\noindent To prevent overfitting and improve interpretability, the empirical loss is augmented with regularization terms that penalize model complexity. A widely used approach is the \textbf{Lasso}, which introduces an $\ell_1$ penalty on the parameter vector, encouraging sparsity in the learned solution:

\begin{equation}
\hat{\boldsymbol{\phi}} = \arg\min_{\boldsymbol{\phi} \in \mathbb{R}^p} 
\left( \frac{1}{n} \sum_{i=1}^n (f_{\boldsymbol{\phi}}(\mathbf{x}_i) - y_i)^2 + \lambda \| \boldsymbol{\phi} \|_1 \right),
\label{eq:lasso}
\end{equation}

%In supervised regression, the goal is to learn a functional relationship between a set of covariates $\mathbf{x} \in \mathbb{R}^d$ and a response $y \in \mathbb{R}$, given a training dataset $\{ (\mathbf{x}_i, y_i) \}_{i=1}^n$. A parametric model $f_\phi(\mathbf{x})$ is fitted by minimizing an empirical loss function that measures the discrepancy between predictions and observations:

%\begin{equation}
%L(\phi) = \frac{1}{n} \sum_{i=1}^{n} \ell(f_\phi(\mathbf{x}_i), y_i),
%\end{equation}

%\noindent  where $\ell(\cdot, \cdot)$ is a loss function and $\phi$ denotes the model parameters.  
%In the standard least squares setting, the loss takes a quadratic form:
%\begin{equation}
%\ell(f_\phi(\mathbf{x}_i), y_i) = (f_\phi(\mathbf{x}_i) - y_i)^2.
%\end{equation}

%\noindent To prevent overfitting and improve interpretability, the empirical loss is augmented with regularization terms that penalize model complexity. A widely used approach is the \textbf{Lasso}, which introduces an $\ell_1$ penalty on the parameter vector, encouraging sparsity in the learned solution:

%\begin{equation}
%\hat{\phi} = \arg\min_{\phi \in \mathbb{R}^p} 
%\left( \frac{1}{n} \sum_{i=1}^n (f_\phi(\mathbf{x}_i) - y_i)^2 + \lambda \| \phi \|_1 \right),
%\label{eq:lasso}
%\end{equation}

\noindent where $\lambda > 0$ controls the amount of regularization.  
The $\ell_1$ norm induces sparsity by shrinking some coefficients exactly to zero, thereby performing variable selection. However, in nonlinear models such as neural networks, the parameters do not correspond one-to-one to the input variables, so $\ell_1$ regularization does not guarantee feature-level sparsity. This limitation motivated the development of LassoNet.

\subsection{LassoNet}
LassoNet extends the Lasso to neural networks by coupling a linear skip connection with a nonlinear feed-forward component, while enforcing a hierarchical $\ell_1$ constraint that controls the relationship between the two.  

\noindent Let $\boldsymbol{\theta} \in \mathbb{R}^d$ denote the weights of the linear component and $g_\mathbf{W}(\mathbf{x})$ a feed-forward neural network parameterized by $\mathbf{W}$, where $\mathbf{W}$ collectively denotes all the weight matrices of the input, hidden, and output layers. The model family is:

\begin{equation}
\mathcal{H} = \left\{ f_{\boldsymbol{\theta}, \mathbf{W}} : \mathbf{x} \mapsto \boldsymbol{\theta}^\top \mathbf{x} + g_\mathbf{W}(\mathbf{x}) \right\}.
\label{eq:classH}
\end{equation}

\noindent The learning problem is then defined as:

\begin{equation}
(\hat{\boldsymbol{\theta}}, \hat{\mathbf{W}}) = 
\arg\min_{\boldsymbol{\theta}, \mathbf{W}} 
\frac{1}{n} \sum_{i=1}^n (f_{\boldsymbol{\theta}, \mathbf{W}}(\mathbf{x}_i) - y_i)^2 + \lambda \| \boldsymbol{\theta} \|_1,
\label{eq:lassonet}
\end{equation}

\noindent subject to the hierarchical constraint

\begin{equation}
\left\| \mathbf{W}^{(1)}_{j, \cdot} \right\|_\infty \leq M |\theta_j|, 
\quad j = 1, \ldots, d
\label{eq:constraint}
\end{equation}

\noindent where $\mathbf{W}^{(1)} \in \mathbb{R}^{d \times K}$ is the weight matrix of the input layer of the neural network $g_{\mathbf{W}}(\mathbf{x})$.
This constraint implies that if $\theta_j = 0$, then the entire vector $\mathbf{W}^{(1)}_{j, \cdot} = 0$; that is, the input variable $x_j$ is excluded from the nonlinear component unless it is active in the linear term.
Consequently, the LASSO penalty on $\boldsymbol{\theta}$ induces variable-level sparsity across the entire network.  

The LassoNet optimization problem \eqref{eq:lassonet} is solved via a proximal gradient descent algorithm along a regularization path over a decreasing sequence of $\lambda$ values.  
    
For each fixed $\lambda$, the algorithm starts from a random initialization and iterates for $B$ epochs the following two steps: (i) a gradient descent update of the parameters, and (ii) application of a hierarchical proximal operator enforcing both the $\ell_1$ penalty and the constraint \eqref{eq:constraint}. See Algorithm~\ref{alg:lassonet} for the pseudocode. 

\begin{algorithm}[H]
\caption{LassoNet training procedure \citep{lemhadri2021lassonet}}
\label{alg:lassonet}
\begin{algorithmic}[1]
\Require Dataset $\mathbf{X} \in \mathbb{R}^{n \times d}$, $\mathbf{y} \in \mathbb{R}^n$; neural architecture $g_{\mathbf{W}}$; epochs per $\lambda$: $B$; learning rate $\alpha$; hierarchy parameter $M$; path parameter $\epsilon$; \texttt{lambda\_start}
\State Initialize network parameters $(\boldsymbol{\theta}, \mathbf{W})$ randomly
\State Pretrain model with $\lambda = 0$
\State Initialize $\lambda =$\texttt{lambda\_start}, active set size $k = d$

\While{$k > 0$}
    \State Update regularization parameter: $\lambda \leftarrow (1+\epsilon)\lambda$
    
    \For{$b = 1, \ldots, B$}
        \State Compute gradients of the smooth loss:
        \State \hspace{1em} $\nabla_{\boldsymbol{\theta}} \frac{1}{n}\sum_{i=1}^n (f_{\boldsymbol{\theta},\mathbf{W}}(\mathbf{x}_i)-y_i)^2$
        \State \hspace{1em} $\nabla_{\mathbf{W}} \frac{1}{n}\sum_{i=1}^n (f_{\boldsymbol{\theta},\mathbf{W}}(\mathbf{x}_i)-y_i)^2$
        
        \State Gradient descent step:
        \State \hspace{1em} $\boldsymbol{\theta} \leftarrow \boldsymbol{\theta} - \alpha \nabla_{\boldsymbol{\theta}}$
        \State \hspace{1em} $\mathbf{W} \leftarrow \mathbf{W} - \alpha \nabla_{\mathbf{W}}$
        
        \State Apply hierarchical proximal operator:
        \State \hspace{1em} $(\boldsymbol{\theta}, \mathbf{W}^{(1)}) \leftarrow \textsc{Hier-Prox}(\boldsymbol{\theta}, \mathbf{W}^{(1)}, \alpha \lambda, M)$ \label{alg:proxstep}
    \EndFor
    
    \State Update active set size:
    \State \hspace{1em} $k \leftarrow \#\{j : \theta_j \neq 0\}$
\EndWhile
\end{algorithmic}
\end{algorithm}

\noindent Line~\ref{alg:proxstep} of Algorithm \ref{alg:lassonet} applies the Hierarchical Proximal Operator \citep{lemhadri2021lassonet}. 
Since the proximal operator is separable across components, we can treat each variable independently. For completeness, we report the solution corresponding to a single fixed component $j$. For each variable $j$, the proximal operator solves:

\begin{equation}
\left( \boldsymbol{\theta}^{*}, \mathbf{W}^{(1)*} \right) =
\arg\min_{\boldsymbol{\theta}, \mathbf{W}^{(1)}_{j, \cdot}}
\frac{1}{2} \|\mathbf{v} - \boldsymbol{\theta}\|_2^2 +
\frac{1}{2} \left\|\mathbf{U} - \mathbf{W}^{(1)} \right \|_2^2 + \lambda \| \boldsymbol{\theta} \|_1, 
\end{equation}
subject to
\begin{equation}
\left \|\mathbf{W}^{(1)}_{j, \cdot} \right \|_\infty \le M |\theta_j|, j=1,\dots,d;
\end{equation}
whose solution is determined in Algorithm \ref{alg:hierprox}.

\begin{algorithm}[H]
\begin{small}
\caption{Hierarchical Proximal Operator, \citep{lemhadri2021lassonet}}
\label{alg:hierprox}
\begin{algorithmic}[1]
\Procedure{Hier-Prox}{$\boldsymbol{\theta}, \mathbf{W}^{(1)}; \lambda, M$}
    \For{$j \in \{1, \ldots, d\}$}
        \State Sort the entries of $\mathbf{W}^{(1)}_{j,\cdot}$ into $\left|{W}^{(1)}_{j,(1)}\right| \geq \ldots \geq \left|{W}^{(1)}_{j,(K)}\right|$
        \For{$m \in \{0, \ldots, K\}$}
            \State $w_m \gets \frac{1}{1 + m M^2} \cdot S_\lambda\left(|\theta_j| + M \cdot \sum_{i=1}^m\left |{W}^{(1)}_{j,(i)}\right|\right)$
        \EndFor
        \State Find $\tilde{m}$, the first $m \in \{0, \ldots, K\}$ such that $\left|{W}^{(1)}_{j,(m+1)}\right| \leq w_m \leq \left|{W}^{(1)}_{j,(m)}\right|$
        \State $\tilde{\theta}_j \gets \frac{1}{M} \cdot \operatorname{sign}(\theta_j) \cdot w_{\tilde{m}}$
        \State $\tilde{\mathbf{W}}^{(1)}_{j, \cdot} \gets \operatorname{sign}\left(\mathbf{W}^{(1)}_{j,\cdot}\right) \cdot \min\left(w_{\tilde{m}}, \left|\mathbf{W}^{(1)}_{j,\cdot}\right|\right)$, where the $\min$ is intended componentwise.
    \EndFor
    \State \Return $\left(\boldsymbol{\tilde{\theta}}, \tilde{\mathbf{W}}^{(1)}\right)$
\EndProcedure
\end{algorithmic}
\end{small}
\end{algorithm}

\subsection{Robust LassoNet}

The standard LassoNet employs the Mean Squared Error (MSE) loss, which is highly sensitive to large residuals and therefore to the presence of outliers. To improve robustness, we adopt an M-estimation framework of $\psi$-type and replace the quadratic loss with a robust loss function that down-weights large residuals.

We begin by recalling the definition of a $\psi$-function (\citep{robustbase, maronna2006}).

\begin{definition}
A $\psi$-function is a piecewise continuous function $\psi : \mathbb{R} \to \mathbb{R}$ such that:

\begin{enumerate}
\item $\psi$ is odd, i.e.,
\[
\psi(-x) = -\psi(x) \quad \forall x \in \mathbb{R}.
\]

\item $\psi(x) \ge 0$ for $x \ge 0$, and $\psi(x) > 0$ for $0 < x < x_r$, where
\[
x_r := \sup\{\tilde{x} : \psi(\tilde{x}) > 0\}, \quad (x_r > 0,\ \text{possibly } x_r = \infty).
\]

\end{enumerate}

\end{definition}

From this definition, the associated $\rho$-function can be defined as follows.

\begin{definition}
A $\rho$-function associated with a $\psi$-function is defined as
\begin{equation}
\rho(x) = \int_{0}^{x} \psi(u)\,du .
\label{eq:rho_integral}
\end{equation}
This representation implies that $\rho(0)=0$, $\rho(x) \geq 0$,  $\rho(x) > 0$ for $0 < |x| < x_r$ and that $\rho$ is an even function.
\end{definition}

A $\psi$-function is called \emph{redescending} if it has a finite rejection point $x_r$ such that
\[
\psi(x) = 0 \quad \text{for } |x| > x_r \quad \mathrm{for} \, x_r<\infty.
\]

Functions satisfying only 
\[
\psi(x) \to 0 \quad \text{as } |x| \to \infty
\]
but with $x_r = \infty$ are sometimes referred to as \emph{weakly redescending} (see \citep{maronna2006}, \citep{huber2009robust}). 

For completeness, we also distinguish \emph{monotone} $\psi$-functions, which are non-decreasing over the entire domain, from \emph{redescending} $\psi$-functions.
Monotone $\psi$-functions lead to convex $\rho$-functions such that the corresponding M-estimators are uniquely defined. 

In this paper, we considered the following robust loss.

\paragraph{Huber Loss:}
Historically, the ‘‘Huber function’’ has been the first $\psi$- monotone function, proposed by Huber (\citep{huber1964robust}). The family of Huber functions is defined as
\begin{equation}
\rho_C(x) =
\begin{cases}
\frac{1}{2}x^2, & |x| \le C, \\
C \left(|x| - \frac{C}{2}\right), & |x| > C,
\end{cases}
\qquad
\psi_C(x) =
\begin{cases}
x, & |x| \le C, \\
C\,\text{sign}(x), & |x| > C.
\end{cases}
\end{equation}
The constant $C$ is a tuning parameter that determines the threshold at which the Huber loss transitions from quadratic to linear behavior, thereby controlling the trade-off between efficiency and robustness. A commonly used choice is $C \approx 1.345$ \citep{rey1983}.

\paragraph{Cauchy Loss:}
The Cauchy loss has also been propagated as ‘‘Lorentzian merit function’’ in regression for outlier detection  (see \citep{Andrews1972}), it is a weak redescender loss and it is defined as
\begin{equation}
\rho_C(x) = \frac{C^2}{2} \log\!\left(1 + \frac{x^2}{C^2}\right),
\quad
\psi_C(x) = \frac{x}{1 + x^2 / C^2}.
\end{equation}
Note that $\rho$ is nonconvex. The constant $C$ is controls the scale of the loss function and therefore the level of robustness to outliers. A common choice is $C = 2.3849$ \citep{rey1983}.

\paragraph{Nonnegative Garrote (NNG) Loss:}
The Non-negative Garrote is a weak redescender loss and it  can be viewed through three equivalent lenses: as a robust \textit{M-estimator} (defined by its loss $\rho$), as a \textit{thresholding rule} (the proximal operator), or as an \textit{implicit penalty}, see \citep{Rodriguez2017} for details. 

\begin{equation}
\rho_C(x) =
\begin{cases}
\frac{1}{2}x^2, & |x| \le C,\\
\frac{1}{2}C^2\left(1 + 2\log\frac{|x|}{C}\right), & |x| > C,
\end{cases}
\qquad
\psi_C(x) =
\begin{cases}
x, & |x| \le C,\\
\frac{C^2}{x}, & |x| > C.
\end{cases}
\end{equation}
The constant $C$ determines the threshold at which the loss function departs from the quadratic behavior and reduces the influence of large residuals. A commonly used choice is $C = 2.0$.

\paragraph{Tukey’s Bisquare Loss:} Tukey’s bisquare (aka ‘‘biweight’’) family of functions (see \citep{mosteller1977}) is a redescender loss and it is defined as
\begin{equation}
\rho_C(x) =
\begin{cases}
\frac{C^2}{6} \left[ 1 - \left(1 - \left(\frac{x}{C}\right)^2\right)^3 \right], & |x| \le C,\\[1ex]
\frac{C^2}{6}, & |x| > C,
\end{cases}
\qquad
\psi_C(x) =
\begin{cases}
x \left[ - \left(1-\frac{x}{C}\right)^2\right]^2, & |x| \le C,\\
0, & |x| > C,
\end{cases}
\end{equation}
The tuning constant C is commonly set up to 4.685.

Using this construction, we define the \textit{Robust LassoNet} model by replacing the MSE loss in \eqref{eq:lassonet} with the robust loss function $\rho$:

\begin{equation}
(\hat{\boldsymbol{\theta}}, \hat{\mathbf{W}}) =
\arg\min_{\boldsymbol{\theta}, \mathbf{W}}
\frac{1}{n} \sum_{i=1}^n \rho\!\big(y_i - f_{\boldsymbol{\theta}, \mathbf{W}}(\mathbf{x}_i)\big)
+ \lambda \|\theta\|_1,
\label{eq:robust_lassonet}
\end{equation}

subject to the hierarchical constraint

\begin{equation}
\left\| \mathbf{W}^{(1)}_{j, \cdot} \right\|_\infty \leq M |\theta_j|, 
\quad j = 1, \ldots, d.
\label{eq:robust_constraint}
\end{equation}

\section{Implementation and Results}
\label{sec:res}

\subsection{Experimental Setup}

To evaluate the performance of Robust LassoNet, we conducted experiments on both synthetic and real-world datasets. The experiments were designed to assess the model's predictive accuracy, robustness to contamination, and variable selection consistency under varying levels of noise and types of contamination.

\noindent
We considered the LassoNet implementation available at \texttt{https://lasso-net.github.io/}, developed in Python using the \texttt{PyTorch} framework. We modified the original implementation to incorporate robust loss functions and implemented a custom cross-validation procedure along the regularization path to select the optimal regularization parameter $\lambda_{\mathrm{cv}}$. Specifically, validation scores were interpolated over a common logarithmic grid of $\lambda$ values, following the strategy adopted in the original LassoNet implementation. To ensure methodological consistency, cross-validation was performed for each candidate model using its own loss function as the validation score. 

All models were trained using the hierarchical proximal gradient algorithm described in Section \ref{sec: methods} (see Algorithms \ref{alg:lassonet} and \ref{alg:hierprox}), with the robust losses incorporated directly into the gradient computation.  
The smooth component of the objective function was optimized using Adam (\texttt{learning rate}$=1e-3$), during the initial dense phase ($\lambda=0$), and using SGD with momentum (\texttt{learning rate}$=1e-3$, \texttt{momentum}$=0.9$) during the regularization path. The activation function implemented is ReLU.
\begin{table}[H] 
	\centering
	\begin{tabular}{lll}
		\hline
		\textbf{Parameter} & \textbf{Default} & \textbf{Description} \\
		\hline
		\texttt{hidden\_dims} & (100,) & Dimensions of the hidden layers \\
		\texttt{lambda\_start} & 'auto' & Initial value of $\lambda$ along the path \\
		\texttt{lambda\_seq} & None & Explicit sequence of $\lambda$ values \\
		$\gamma$ & 0.0 & $L_2$ regularization applied to network weights \\
		$\gamma_{\text{skip}}$ & 0.0 & $L_2$ regularization applied to the skip connection \\
		\texttt{path\_multiplier} & 1.02 & Multiplicative factor along the path ($\lambda \leftarrow \lambda \cdot c$) \\
		$M$ & 10 & Parameter of the hierarchical constraint \\
		\texttt{groups} & None & Definition of groups for Group LassoNet \\
		\texttt{dropout} & 0 & Dropout rate \\
		\texttt{batch\_size} & None & Mini-batch size (None = full batch) \\
		\texttt{optim} & None & Optimizers for the initialization and path phases \\
		\texttt{n\_iters} & (1000, 100) & Maximum number of epochs (init, path) \\
		\texttt{patience} & (100, 10) & Early stopping patience (init, path) \\
		\texttt{tol} & 0.99 & Improvement threshold for early stopping \\
		\texttt{backtrack} & False & Restores the best model state if True \\
		\texttt{val\_size} & None & Proportion of data used for validation \\
		\texttt{device} & None & PyTorch device (CPU/GPU) \\
		\texttt{verbose} & 1 & Verbosity level \\
		\texttt{random\_state} & None & Seed for validation split \\
		\texttt{torch\_seed} & None & Seed for PyTorch initialization \\
		\hline
	\end{tabular}
	\caption{Parameters of \texttt{LassoNetRegressor}.}
	\label{tab:lasso_params}
\end{table}

Unless otherwise stated, we used the default parameter settings for all experiments (see Table~\ref{tab:lasso_params}), except for the following:
\begin{itemize}
    \item \textbf{\texttt{hidden\_dims}}$=(10,10)$, to maintain approximately the same number of degrees of freedom as the default configuration, while increasing the network depth to better capture nonlinearity and interactions \citep{Daubechies2022NonlinearDeep};
    \item \texttt{path\_multiplier}$=1.2$; to reduce the lenght of the grid of $\lambda$ values 
    \item \texttt{lambda\_start}$=10^{-3}$, in order to use the same grid starting point across all experimental settings.
\end{itemize}

The code implementing the methodology and reproducing the results of this study is publicly available at \url{https://github.com/paola-stolfi/Robust-LassoNet}.

To benchmark the performance of Robust LassoNet, we include several comparison methods. In the synthetic data experiments (Section \ref{subsec:expres}), we consider robust variants of Random Forest (\cite{breiman2001random}) and XGBoost (\cite{chen2016xgboost}) as representatives of ensemble-based nonlinear regression. Specifically, Random Forest is fitted using the mean absolute error (MAE) as the node-splitting criterion (criterion="absolute\_error" in sklearn.ensemble.RandomForestRegressor; \citep{pedregosa2011scikit}), which reduces sensitivity to outliers relative to the standard mean squared error criterion. XGBoost is fitted using the pseudo-Huber loss as the training objective (objective="reg:pseudohubererror", huber\_slope=1.345 in xgboost.XGBRegressor; \cite{chen2016xgboost}), where the slope parameter $\delta = 1.345$ is the same value adopted for the Huber loss in Robust LassoNet, corresponding to 95\% asymptotic efficiency under Gaussian errors. 
Since Tree ensembles assign a (typically nonzero) importance to almost every feature, a threshold is required to define a selected set. Per-feature importances
(impurity reduction for RandomForest; gain for XGBoost) are normalised to sum to
one and, where applicable, summed within each base variable. A variable, $j$, is
declared \emph{selected} when its (aggregated) importance, $\bar{I}_j$, is at least the mean of the positive importances,
\begin{equation}
  j \text{ selected} \iff \bar{I}_j \;\ge\; \operatorname{mean}\big(\{\bar{I}_k : \bar{I}_k > 0\}\big).
\end{equation}
The hyperparameters we used  are  reported in the Appendix B. 
For the real-data experiments, we additionally include two robust linear variable selection methods: Robust Groupwise Least Angle Regression (RGRPLARS \cite{Alfons2016}) for the Top Gear dataset, where predictors include groups of dummy variables, and Robust Least Angle Regression (RLARS)  for the Atherosclerosis dataset.
Both RGRPLARS and RLARS methods are implemented in the R package RobustHD  (\cite{Alfons2021}),  with the Bayesian Information Criterion (BIC) used for variable selection.

\subsection{Synthetic Data Experiments}
For all simulations we generated training data according to
$$
y_i=f(\mathbf{x}_i)+{\varepsilon_i}, \quad i=1,\dots,n
$$
with $\mathbf{x}_i \in \mathbb{R}^{d}$. We fixed $n=750$ and $d=200$.  We sampled each covariate $x_j \in \mathbb{R}$, $j=1,\dots,d$ independently from the standard uniform distribution. The true function $f$ will only depend on the first $| S|$ features, where $S=\left\{ j:  f \, \mathrm{depends \, on} \, x_j, j=1,\dots,d \right\}$ is the active set.
We considered the following $4$ regression models:

\begin{align*}
\textbf{Model 1:} \quad 
f(\mathbf{x}) &= 10 x_4 + 5 x_5 
+ 10 \sin\!\left(\pi x_1 x_2\right) 
+ 20 (x_3 - 0.5)^2
\\[1em]
\textbf{Model 2:} \quad 
f(\mathbf{x}) &= \tanh(x_1 + 2x_2 - 3x_3 + 2x_4) 
+ 2 \tanh(x_1 - 2x_5 + 2x_6) \\
&\quad + \tanh(-x_2 - x_3 - x_6) 
 + \tanh(x_5 - 0.5 x_3 + 0.5 x_6)
\\[1em]
\textbf{Model 3:} \quad 
f(\mathbf{x}) &= \sin\!\big(x_1 (x_1 + x_2)\big)
\, \cos\!\big(x_3 + x_4 x_5\big)
\, \sin\!\big(e^{x_5} + e^{x_6} - x_2\big)
\\[1em]
\textbf{Model 4:} \quad 
f(\mathbf{x}) &= \frac{5}{2} \Big(
x_1^2 x_2^3 
+ \log(1 + x_3)
+ \sqrt{1 + x_4 x_5}
+ e^{x_5/2} \\
&\qquad\quad
+ x_6^2 x_7^3
+ \log(1 + x_8)
+ \sqrt{1 + x_9 x_{10}}
+ e^{x_{10}/2}
\Big).
\end{align*} 

{\bf Model 1} was considered in \citep{friedman1991mars}, {\bf Models 2 and 3} were used in \citep{feng_simon_2017_spinn}, and {\bf Model 4} was introduced in \citep{yan2025semiparametric}. The four synthetic models were deliberately chosen to represent increasingly challenging
nonlinear regression structures. Model 1 combines additive linear terms, a nonlinear quadratic component, and a light interaction structure; Model 2 introduces multiple nonlinear transformations through additive hyperbolic tangent components; Model 3 combines highly oscillatory trigonometric terms with nested nonlinear interactions; finally, Model 4 represents the most complex
setting, involving multiple interaction effects, exponential transformations, logarithmic terms, and heterogeneous nonlinear contributions.

\begin{remark}
The four regression models considered above can be viewed as stress tests for LassoNet, as they are predominantly driven by nonlinear components and interaction effects, with only Model~1 containing a limited explicit linear structure. Consequently, within the LassoNet function class $\mathcal{H}$ (see Eq.~\eqref{eq:classH}), the nonlinear component is expected to account for most of the signal, while the linear coefficients $\boldsymbol{\theta}$ mainly act as gateways enabling the activation of nonlinear effects through the hierarchical constraint.

Recall that LassoNet enforces the hierarchical constraint
\[
\left\| W^{(1)}_{j, \cdot} \right\|_\infty \le M \, |\theta_j|,
\]
for each feature $j$, where $W^{(1)}_{j, \cdot}$ denotes the weights in the first hidden layer associated with covariate $x_j$. 
This condition implies that a feature can contribute to the nonlinear component only if its corresponding linear coefficient is nonzero. In strongly nonlinear settings, this mechanism becomes
particularly challenging because the model must identify relevant variables through weak linear proxies before exploiting nonlinear effects in the hidden layers. Therefore,
Models 2--4 provide particularly demanding benchmark for evaluating the ability of robust losses to stabilize both optimization and feature selection under hierarchical sparsity constraints.

In this context, the hyperparameter $M$ plays a crucial role. 
It controls how much nonlinear effect can be activated by a given linear coefficient:
\begin{itemize}
    \item If $M$ is small, the constraint is tight and the model is forced to behave almost linearly. 
    Since the true regression functions are strongly nonlinear, the network may not have sufficient flexibility to approximate them accurately.

    \item If $M$ is large, the constraint is relaxed: even a small nonzero linear coefficient can support a substantial nonlinear contribution. 
    This allows the model to recover complex nonlinear effects while still preserving the hierarchical sparsity structure.
\end{itemize}

\noindent
That's why, after some experimentations, we leave $M=10$ as default choice for the parameter $M$.
\end{remark}

\vspace{0.5cm}
We focused on robustness to vertical outliers, generating the error term according to the classical Huber contamination model.
More precisely, the distribution of the errors was defined as the mixture
\[
\varepsilon_i \sim (1-\pi)\, \mathcal{N}(0,\sigma^2)
\;+\;
\pi\, \mathcal{N}(\mu_c,\sigma_c^2),
\]
where $\pi \in [0,1)$ denotes the contamination proportion, and 
$(\mu_c,\sigma_c^2)$ characterize the mean and variance of the contaminating component. 

For each model, the variance parameter $\sigma^2$ was calibrated so that the signal-to-noise ratio,
\[
\mathrm{SNR} = \frac{\mathrm{Var}\!\big(f(\mathbf{x})\big)}{\sigma^2},
\]
is equal to $4$.

Let $q_{0.99}(f)$ denote the empirical $0.99$ quantile of $\{f(\mathbf{x}_i)\}_{i=1}^n$, 
and let $M_f = \max_{i=1,\dots,n} \left\{f(\mathbf{x}_i)\right\}$.
For each of the four regression models, we considered the following configurations:
\begin{enumerate}
	\item \textbf{Gaussian contamination:} $\pi = 0$.
	
	\item \textbf{Moderate mean contamination:} 
	$\pi = 0.1$, $\mu_c = q_{0.99}(f)$, $\sigma_c^2 = \sigma^2$.
	
	\item \textbf{Strong mean contamination:} 
	$\pi = 0.1$, $\mu_c = 1.5\,M_f$, $\sigma_c^2 = \sigma^2$.
	
	\item \textbf{Moderate variance contamination:} 
	$\pi = 0.1$, $\mu_c = 0$, $\sigma_c^2 = 100\sigma^2$.
	
	\item \textbf{Strong variance contamination:} 
	$\pi = 0.1$, $\mu_c = 0$, $\sigma_c^2 = 625\sigma^2$.
\end{enumerate} 
All covariates were standardized prior to model training. The response variable was also standardized, using a robust estimate of the scale.
Specifically, let $\tilde{y} = \mathrm{median}(y_1,\dots,y_n)$ denote the sample median of the training responses, and let
\[
\mathrm{MAD} = \mathrm{median}\bigl( |y_i - \tilde{y}| \bigr)
\]
be the median absolute deviation. 
We defined the robust scale estimate as
\[
\hat{\sigma}_{\mathrm{rob}} = 1.4826 \times \mathrm{MAD},
\]
which is consistent for the standard deviation under Gaussianity. 
The standardized response was then computed as
\[
y_{i,{\mathrm{scaled}}} = \frac{y_i - \tilde{y}}{\hat{\sigma}_{\mathrm{rob}}}.
\]
In order to evaluate predictive performance, we generated an additional test set with
$n_{\mathrm{test}} = 250$ observations $(y_i^{\mathrm{test}},\mathbf{x}_i^{\mathrm{test}})$, 
$i = 1,\dots,n_{\mathrm{test}}$, assuming
\[
\varepsilon_i \sim \mathcal{N}(0,\sigma^2),
\]
that is, without contamination.
Both the test covariates and responses were standardized using the mean and standard deviation computed from the training set.

\vspace{0.5cm}
\noindent Let us indicate with $S^c$ the complement of the active set $S$, i.e. 
$S^c=\left\{1,\dots,d\right\}-S$, and let be $\mathbf{\hat{y}^{\mathrm{test}}}$ the predicted response obtained for $\lambda_{\mathrm{cv}}$, i.e.
\[
\hat{y}_i^{\mathrm{test}}
=
{\boldsymbol{\hat{\theta}}}^\top \mathbf{x}_{i,\mathrm{scaled}}^{\mathrm{test}}
+
g_{\widehat{\mathbf{W}}}(\mathbf{x}_{i,\mathrm{scaled}}^{\mathrm{test}}),
\quad i = 1,\dots,n_{\mathrm{test}},
\]
where $\boldsymbol{\hat{\theta}}$ and $\widehat{\mathbf{W}}$ are the estimated parameters with $\lambda_{\mathrm{cv}}$.
We used the following quantitative metrics to evaluate the performance:

\begin{itemize}
	\item \textbf{Predictive Mean Squared Error (PE):}
	\[
	\text{PE} = \frac{1}{n_{test}} \sum_{i=1}^{n_{test}} \left(y_{i,\mathrm{scaled}}^{\mathrm{test}} - \hat{y}_i^{\mathrm{test}}\right)^2,
	\]    
	\item \textbf{False Positive Rate (FPR):}
	\[
	\text{FPR} = \frac{\left| \left\{j \in S^c: \hat{\theta}_j \neq 0\right\}\right|}{\left|S^c\right|}.
	\]
	\item \textbf{False Negative Rate (FNR):}
	\[
	\text{FNR} = \frac{\left| \left\{j \in S: \hat{\theta}_j = 0\right\}\right|}{\left|S\right|}.
	\]    
	\item \textbf{Number of Selected Rate (NSR):}
	\[
	\text{NSR} = \frac{\left|\left\{j: \hat{\theta}_j \neq 0 \right\}\right|}{d}.
	\]
\end{itemize}

\subsection{Results on Synthetic Data}
\label{subsec:expres}
In this section, we present and discuss the results obtained for the four synthetic regression models across the five contamination scenarios. All quantitative results: PE, FPR, FNR, and  NSR are reported as averages across 25 independent replications in Tables \ref{Table: model1_mean}--\ref{Table: model4_var} and as boxplots in Figures \ref{fig_boxplot_model1_mean}--\ref{fig_boxplot_model4_variance} in the Appendix. In the tables, the best result according to the PE metric is highlighted in bold. For improved readability, the results for the Gaussian contamination scenario are repeated in both the tables and the figures for the moderate/strong mean and moderate/strong variance contamination scenarios.

\medskip

We next consider the mean contamination scenario. Table \ref{Table: model1_mean} together with Figure \ref{fig_boxplot_model1_mean}, Table \ref{Table: model2_mean} together with Figure \ref{fig_boxplot_model2_mean}, Table \ref{Table: model3_mean} together with Figure \ref{fig_boxplot_model3_mean}, and Table \ref{Table: model4_mean} together with Figure \ref{fig_boxplot_model4_mean} present the performance results under the moderate and strong mean contamination scenarios for Models 1--4, respectively.

\noindent
We begin by commenting the uncontaminated Gaussian setting ($\pi=0$) for the MSE-based LassoNet and the robust LassoNet, which serves as a baseline for evaluating the efficiency cost associated with robustification.
In Models 1, 2, and 4, differences in PE are relatively small, indicating that the introduction of robust loss functions does not substantially reduce efficiency under ideal Gaussian conditions. Regarding variable selection, the robust losses generally produce sparser models, as reflected by lower NSR and FPR values compared with the standard MSE loss. In Models 2 and 4, this increased sparsity does not result in a loss of relevant information, since the false negative rate (FNR) remains essentially zero for most robust methods. In Model 1, however, the robust losses exhibit a slightly higher FNR, suggesting a mild tendency to exclude some relevant variables in exchange for improved false-positive control.
A more pronounced difference emerges in Model 3, where the standard MSE loss attains the lowest PE, although at the cost of selecting substantially larger models. Indeed, the MSE estimator exhibits considerably higher NSR and FPR values than the robust alternatives, suggesting a tendency toward over-selection in this highly nonlinear setting. By contrast, the robust losses produce sparser solutions, with only a moderate increase in FNR.

\noindent
A comparison with the tree-based benchmark methods reveals complementary behaviour. Random Forest exhibits the strongest variable selection performance in Models 1 and 2, achieving the lowest NSR and FPR while maintaining almost perfect recovery of the active set. In Model 3, it again produces markedly sparser models than the MSE-based LassoNet, with substantially lower NSR and FPR at the expense of only a moderate increase in FNR, while its predictive performance remains comparable to that of the MSE estimator. In the more challenging Model 4, however, the increased nonlinear complexity reduces the advantage of Random Forest, and the MSE-based LassoNet becomes more competitive in terms of predictive accuracy. By contrast, XGBoost consistently selects substantially larger models, yielding the highest NSR and FPR values across all four models, although its low FNR indicates that this behaviour stems primarily from over-selection rather than from failure to identify the relevant variables.
All these results confirm the classical robustness--efficiency trade-off: robust losses maintain competitive predictive performance under Gaussian noise while yielding more parsimonious models and improved control of false positives.

\medskip

\noindent
We now turn to the contaminated settings, where the advantages of robust approaches become substantially more evident. Indeed, the MSE-based LassoNet deteriorates rapidly across all models, with predictive error increasing particularly under strong mean contamination scenario, where the quadratic loss excessively amplifies the contribution of contaminated observations.
The impact of contamination is especially pronounced as the nonlinear complexity of the regression function increases. In Models 3 and 4, where the signal depends on higher-order interactions and nested nonlinear transformations, contaminated observations distort not only the estimation of the regression surface but also the activation of the hierarchical constraint. Consequently, the standard MSE-based LassoNet estimator frequently fails to recover the active set correctly.
By contrast, the robust LassoNet remain remarkably stable. The relative performance across methods, however, varies by model.

\begin{itemize}
\item Model 1 (Table \ref{Table: model1_mean}, Figure \ref{fig_boxplot_model1_mean}). Tukey’s bisquare achieves the lowest PE under both moderate and strong mean contamination (0.320 and 0.308, respectively). Regarding variable selection, under moderate contamination all robust LassoNet methods attain substantially lower NSR and FPR values than the standard MSE-based LassoNet, at the cost of a slight increase in FNR, while exhibiting broadly comparable behaviour among themselves. Under strong contamination, the robust LassoNet clearly outperform the MSE-based LassoNet estimator across all performance metrics. Among the robust losses, Huber provides particularly favourable variable selection properties, achieving the lowest NSR and FPR values, albeit with a higher PE than Tukey’s bisquare.
Compared with the tree-based benchmark methods, all robust LassoNet variants achieve substantially lower PEs while simultaneously selecting much sparser models than both Random Forest and XGBoost under both moderate and strong contamination. Although Random Forest exhibits better variable selection properties than XGBoost, with consistently lower NSR and FPR values, both methods maintain lower FNR values than the robust LassoNet variants, reflecting a tendency to retain relevant variables at the expense of selecting considerably more irrelevant predictors.

\item Model 2 (Table \ref{Table: model2_mean}, Figure \ref{fig_boxplot_model2_mean}). Huber and Tukey perform essentially equivalently in terms of PE (0.262 vs.\ 0.263 under moderate contamination and 0.265 vs.\ 0.265 under strong contamination). Variable selection performance also remains highly stable across the robust LassoNet methods. Under moderate contamination, Huber and Tukey achieve slightly lower NSR and FPR values than the MSE estimator while maintaining perfect recovery of the active set (FNR = 0). Under strong contamination, the robust LassoNet methods tend to select slightly larger models than MSE-based LassoNet, as reflected by moderately higher NSR and FPR values; however, they maintain substantially lower FNR values, indicating more reliable recovery of the relevant variables.
Compared with the tree-based benchmark methods, all robust LassoNet variants achieve substantially lower PEs under both moderate and strong contamination. Under moderate contamination, Random Forest produces the sparsest models, attaining the lowest NSR and FPR, although at the expense of a slightly higher FNR than the others robust methods. Under strong contamination, its variable selection performance becomes comparable to that of the best robust LassoNet variants. By contrast, XGBoost consistently selects substantially larger models, yielding the highest NSR and FPR values, while maintaining relatively low FNR.

\item Model 3 (Table \ref{Table: model3_mean}, Figure \ref{fig_boxplot_model3_mean}). Cauchy achieves the lowest PE under moderate contamination (0.428), while Tukey attains the best predictive performance under strong contamination (0.397), closely followed by Garrote (0.403). Under moderate contamination, all robust LassoNet variants outperform the MSE-baed LassoNet estimator across all performance metrics, while exhibiting broadly comparable variable selection behaviour among themselves. Under strong contamination, the robust LassoNet estimators continue to achieve substantially better PE and markedly lower FNR values than MSE, despite a moderate increase in NSR and FPR. Among the robust approaches, Tukey appears to provide the best overall balance between predictive accuracy and variable selection performance.
Compared with the tree-based benchmark methods, the robust LassoNet variants also achieve markedly lower PEs while providing improved control of NSR and FPR under both contamination levels. Under moderate contamination, both Random Forest and XGBoost exhibit slightly lower FNR values than the robust LassoNet methods, although at the expense of selecting larger models, markedly so for XGBoost and only marginally for Random Forest. Under strong contamination, this advantage is retained only by Random Forest, whereas XGBoost exhibits both poorer sparsity and higher FNR than the robust LassoNet variants.

\item Model 4 (Table \ref{Table: model4_mean}, Figure \ref{fig_boxplot_model4_mean}). This model exhibits the most dramatic failure of the MSE-based LassoNet estimator. Under moderate contamination, Cauchy and Tukey achieve the best PE values (0.238 and 0.240, respectively), while under strong contamination Tukey clearly attains the lowest PE (0.240), followed by Cauchy (0.277).
The deterioration of the MSE-based LassoNet estimator is particularly evident in terms of variable selection, with FNR increasing to 0.816 under moderate contamination and to 0.992 under strong contamination, indicating an almost complete failure to recover the active set.
Among the robust LassoNet variants, Tukey achieves perfect recovery of the active set (FNR = 0.000) under both contamination levels, while maintaining NSR and FPR values comparable to those of Cauchy under moderate contamination. Under strong contamination, Tukey exhibits slightly higher NSR and FPR values than Cauchy, but substantially improves both predictive accuracy and sparsity recovery. Overall, Tukey appears to provide the best compromise between predictive performance and reliable variable selection.
Compared with the tree-based benchmark methods, all robust LassoNet variants substantially outperform both Random Forest and XGBoost under both moderate and strong mean contamination. Both tree-based methods experience a pronounced deterioration in predictive accuracy and variable selection performance, exhibiting considerably higher PE, FPR, FNR, and NSR values than the robust approaches. This degradation is particularly severe for Random Forest in terms of predictive error, whereas XGBoost consistently produces the largest models and exhibits the weakest overall variable selection performance.
A notable anomaly emerges for the Nonnegative Garrote under strong mean contamination, where PE increases to 0.640 and FNR to 0.124, substantially worse than the other robust losses, although still far superior to MSE. This deterioration is likely attributable to the quadratic region of the Garrote loss ($|r| \leq C = 2.0$), which still assigns moderate weight to residuals that, while large, remain within the threshold.
\end{itemize}

\medskip

\begin{table}[htbp]
\centering
\small
\resizebox{\textwidth}{!}{
\begin{tabular}{l|cccc|cccc|cccc}
\toprule
& \multicolumn{4}{c|}{Gaussian contamination} 
& \multicolumn{4}{c|}{Moderate mean contamination} 
& \multicolumn{4}{c}{Strong mean contamination} \\
\cmidrule(lr){2-5}\cmidrule(lr){6-9}\cmidrule(lr){10-13}
Loss 
& PE & NSR & FPR & FNR
& PE & NSR & FPR & FNR
& PE & NSR & FPR & FNR \\
\midrule

mse 
& 0.398 & 0.122 & 0.103 & 0.152
& 0.532 & 0.116 & 0.098 & 0.184
& 0.850 & 0.046 & 0.030 & 0.320 \\

huber
& 0.399 & 0.087 & 0.068 & 0.176
& 0.348 & 0.062 & 0.043 & 0.192
& 0.347 & 0.034 & 0.015 & 0.200 \\

cauchy
& 0.402 & 0.102 & 0.083 & 0.160
& 0.342 & 0.072 & 0.053 & 0.192
& 0.331 & 0.061 & 0.042 & 0.192 \\

tukey
& 0.400 & 0.091 & 0.073 & 0.176
& \textbf{0.320} & 0.066 & 0.047 & 0.192
& \textbf{0.308} & 0.071 & 0.052 & 0.200 \\

garrote
& 0.400 & 0.090 & 0.071 & 0.176
& 0.338 & 0.058 & 0.039 & 0.192
& 0.331 & 0.042 & 0.023 & 0.200 \\

rf	
& 0.305	& 0.032	& 0.007	& 0.000
& 0.618	& 0.162	& 0.141	& 0.008
& 1.318	& 0.226	& 0.207	& 0.024\\

xgb	
& \textbf{0.286} & 0.281	& 0.263	& 0.000
& 0.583 & 0.395	& 0.381	& 0.056
& 0.948 & 0.383	& 0.371	& 0.144\\

%\midrule

%oracle
%& \textbf{0.193} & -- & -- & --
%& \textbf{0.146} & -- & -- & --
%& \textbf{0.146} & -- & -- & -- \\

\bottomrule
\end{tabular}
}
\caption{Performance under Gaussian and mean contamination scenarios for \textbf{Model 1}.}
\label{Table: model1_mean}
\end{table}

\begin{table}[htbp]
\centering
\small
\resizebox{\textwidth}{!}{
\begin{tabular}{l|cccc|cccc|cccc}
\toprule
& \multicolumn{4}{c|}{Gaussian contamination} 
& \multicolumn{4}{c|}{Moderate mean contamination} 
& \multicolumn{4}{c}{Strong mean contamination} \\
\cmidrule(lr){2-5}\cmidrule(lr){6-9}\cmidrule(lr){10-13}
Loss 
& PE & NSR & FPR & FNR
& PE & NSR & FPR & FNR
& PE & NSR & FPR & FNR \\
\midrule

mse 
& \textbf{0.253} & 0.180 & 0.155 & 0.000
& 0.282 & 0.131 & 0.105 & 0.007
& 0.320 & 0.055 & 0.028 & 0.067 \\

huber
& \textbf{0.253} & 0.104 & 0.076 & 0.000
& \textbf{0.262} & 0.115 & 0.087 & 0.000
& \textbf{0.265} & 0.092 & 0.065 & 0.007 \\

cauchy
& 0.264 & 0.129 & 0.102 & 0.000
& 0.266 & 0.139 & 0.112 & 0.000
& 0.268 & 0.111 & 0.084 & 0.007 \\

tukey
& 0.255 & 0.106 & 0.078 & 0.000
& 0.263 & 0.130 & 0.103 & 0.000
& \textbf{0.265} & 0.103 & 0.076 & 0.007 \\

garrote
& 0.257 & 0.122 & 0.095 & 0.000
& 0.274 & 0.140 & 0.114 & 0.007
& 0.286 & 0.100 & 0.072 & 0.007 \\

rf 
& 0.270 & 0.035 & 0.006 & 0.007 
& 0.317 & 0.054 & 0.026 & 0.027
& 0.394 & 0.096 & 0.069 & 0.027\\

xgb 
& 0.260 & 0.257 & 0.234 & 0.000 
& 0.304 & 0.343 & 0.323 & 0.007
& 0.378 & 0.385 & 0.367 & 0.033\\

%\midrule

%oracle
%& \textbf{0.182} & -- & -- & --
%& \textbf{0.165} & -- & -- & --
%& \textbf{0.155} & -- & -- & -- \\

\bottomrule
\end{tabular}
}
\caption{Performance under Gaussian and mean contamination scenarios for \textbf{Model 2}. }
\label{Table: model2_mean}
\end{table}

\begin{table}[htbp]
\centering
\small
\resizebox{\textwidth}{!}{
\begin{tabular}{l|cccc|cccc|cccc}
\toprule
& \multicolumn{4}{c|}{Gaussian contamination} 
& \multicolumn{4}{c|}{Moderate mean contamination} 
& \multicolumn{4}{c}{Strong mean contamination} \\
\cmidrule(lr){2-5}\cmidrule(lr){6-9}\cmidrule(lr){10-13}
Loss 
& PE & NSR & FPR & FNR
& PE & NSR & FPR & FNR
& PE & NSR & FPR & FNR \\
\midrule

mse 
& 0.498 & 0.266 & 0.244 & 0.027
& 0.539 & 0.145 & 0.125 & 0.227
& 0.894 & 0.077 & 0.060 & 0.367 \\

huber
& 0.523 & 0.137 & 0.114 & 0.100
& 0.433 & 0.113 & 0.092 & 0.180
& 0.447 & 0.122 & 0.100 & 0.187 \\

cauchy
& 0.532 & 0.106 & 0.082 & 0.107
& \textbf{0.428} & 0.108 & 0.086 & 0.180
& 0.418 & 0.118 & 0.096 & 0.160 \\

tukey
& 0.536 & 0.142 & 0.118 & 0.094
& 0.435 & 0.105 & 0.083 & 0.193
& \textbf{0.397} & 0.090 & 0.066 & 0.154\\

garrote
& 0.510 & 0.152 & 0.129 & 0.080
& 0.444 & 0.103 & 0.081 & 0.200
& 0.403 & 0.124 & 0.101 & 0.154 \\

rf 
& \textbf{0.495} & 0.040 & 0.013 & 0.080 
& 0.567 & 0.119 & 0.096 & 0.147
& 0.967 & 0.199 & 0.178 & 0.134\\

xgb 
& 0.595 & 0.310 & 0.291 & 0.087 
& 0.628 & 0.383 & 0.369 & 0.154 
& 0.947 & 0.394 & 0.382 & 0.213\\

%\midrule

%oracle
%& \textbf{0.311} & -- & -- & --
%& \textbf{0.215} & -- & -- & --
%& \textbf{0.214} & -- & -- & -- \\

\bottomrule
\end{tabular}
}
\caption{Performance under Gaussian and mean contamination scenarios for \textbf{Model 3}. }
\label{Table: model3_mean}
\end{table}

\begin{table}[htbp]
\centering
\small
\resizebox{\textwidth}{!}{
\begin{tabular}{l|cccc|cccc|cccc}
\toprule
& \multicolumn{4}{c|}{Gaussian contamination} 
& \multicolumn{4}{c|}{Moderate mean contamination} 
& \multicolumn{4}{c}{Strong mean contamination} \\
\cmidrule(lr){2-5}\cmidrule(lr){6-9}\cmidrule(lr){10-13}
Loss 
& PE & NSR & FPR & FNR
& PE & NSR & FPR & FNR
& PE & NSR & FPR & FNR \\
\midrule

mse 
& 0.282 & 0.188 & 0.146 & 0.000
& 1.841 & 0.018 & 0.009 & 0.816
& 3.300 & 0.005 & 0.004& 0.992 \\

huber
& 0.294 & 0.136 & 0.091 & 0.000
& 0.279 & 0.108 & 0.062 & 0.028
& 0.278 & 0.101 & 0.055 & 0.036 \\

cauchy
& 0.295 & 0.154 & 0.109 & 0.000
& \textbf{0.238} & 0.131 & 0.086 & 0.012
& 0.277 & 0.097 & 0.055 & 0.108 \\

tukey
& 0.296 & 0.153 & 0.109 & 0.004
& 0.240 & 0.136 & 0.091 & 0.000
& \textbf{0.240} & 0.136 & 0.091 & 0.000 \\

garrote
& 0.291 & 0.141 & 0.096 & 0.000
& 0.287 & 0.097 & 0.057 & 0.148
& 0.640 & 0.108 & 0.067 & 0.124 \\

rf 
& 0.356 & 0.052 & 0.008 & 0.112 
& 3.424 & 0.286 & 0.265 & 0.312
& 9.298 & 0.312 & 0.296 & 0.372\\

xgb 
& \textbf{0.281} & 0.272 & 0.234 & 0.000 
& 1.258 & 0.386 & 0.380 & 0.500
& 1.337 & 0.375 & 0.367 & 0.484\\

%\midrule

%oracle
%& \textbf{0.195} & -- & -- & --
%& \textbf{0.151} & -- & -- & --
%& \textbf{0.151} & -- & -- & -- \\

\bottomrule
\end{tabular}
}
\caption{Performance under Gaussian and mean contamination scenarios for \textbf{Model 4}. }
\label{Table: model4_mean}
\end{table}
%%%%%%%%%%%%%%%%%%%%%%%%%%%%%%%%%%%%%%%%%%%%%

\medskip

We next consider the variance contamination scenario.
Table \ref{Table: model1_var} together with Figure \ref{fig_boxplot_model1_variance} , Table \ref{Table: model2_var} together with Figure \ref{fig_boxplot_model2_variance}, Table \ref{Table: model3_var} together with Figure \ref{fig_boxplot_model3_variance}, and Table \ref{Table: model4_var} together with Figure \ref{fig_boxplot_model4_variance} present the performance results under the moderate and strong variance contamination scenarios for Models 1--4, respectively.

\noindent
Results under variance contamination exhibit a more nuanced
pattern. Under strong variance contamination, MSE-based LassoNet collapses across all models
(PE = 0.626, 0.639, 1.242, 0.666 for Models 1–4; FNR = 0.536, 0.573, 0.833, 0.756), confirming that the robust LassoNet variants provide substantially more stable prediction and variable
selection regardless of the specific loss function chosen. The relative performance of the robust LassoNet estimators, however, depends on the considered model.

\noindent
\begin{itemize}
\item Model 1 (Table \ref{Table: model1_var}, Figure \ref{fig_boxplot_model1_variance}). The robust Lassonet methods achieve highly comparable performance under variance contamination, with Tukey consistently attaining the lowest PE values under both moderate (0.336) and strong contamination (0.323). Regarding variable selection, under moderate contamination the robust LassoNet variants exhibit similar behaviour and uniformly outperform the MSE-based LassoNet estimator. Under strong contamination, the redescending estimators --- namely Cauchy, Tukey, and Garrote --- display highly comparable variable selection performance, all substantially improving over MSE and slightly outperforming Huber.
Compared with the tree-based benchmark methods, all robust LassoNet variants achieve lower PEs while providing markedly better control of NSR and FPR under both contamination levels; the PE gain is substantial against Random Forest throughout, and against XGBoost it is marginal under moderate variance but large under strong variance. Although both Random Forest and XGBoost attain lower FNR values, indicating improved recovery of the active set, this advantage comes at the expense of selecting considerably larger models. Between the two benchmarks, Random Forest consistently exhibits better variable selection performance than XGBoost, whereas XGBoost attains lower predictive errors — only marginally under moderate variance, but by a wide margin under strong variance, where Random Forest's PE collapses (1.755).

\item Model 2 (Table \ref{Table: model2_var}, Figure \ref{fig_boxplot_model2_variance}). Tukey’s bisquare achieves the best predictive performance under both moderate and strong variance contamination, attaining the lowest PE values (0.228 and 0.220, respectively). Under moderate contamination, Tukey also provides the best variable selection performance, achieving the lowest NSR and FPR values together with perfect recovery of the active set (FNR = 0.000). Under strong contamination, Huber attains slightly lower NSR and FPR values than Tukey, although at the cost of a small increase in FNR (0.007 versus 0.000). By contrast, Tukey maintains perfect recovery of the active set while preserving excellent predictive performance. Overall, all robust LassoNet methods substantially outperform the MSE-based LassoNet across all performance metrics.
Compared with the tree-based benchmark methods, all robust LassoNet variants achieve superior PEs under both moderate and strong variance contamination. Both benchmarks exhibit higher PE; XGBoost also shows much higher NSR and FPR, whereas Random Forest's NSR and FPR are comparable to the robust LassoNet methods under moderate variance and clearly higher only under strong variance. Overall, the robust LassoNet approaches match or outperform the tree-based benchmarks across all performance metrics, with XGBoost displaying the weakest variable selection performance.

\item Model 3 (Table \ref{Table: model3_var}, Figure \ref{fig_boxplot_model3_variance}). Garrote achieves the best predictive performance under both moderate (0.432) and strong variance contamination (0.417). Under moderate contamination, Garrote also exhibits variable selection properties comparable to those of Huber. More generally, all robust LassoNet methods substantially outperform the MSE estimator: although they exhibit a slight increase in NSR and FPR relative to MSE, they provide markedly better control of FNR, which for the MSE estimator is approximately twice as large as for the robust alternatives.
Under strong contamination, Garrote attains a lower FNR than Tukey, indicating improved recovery of the active set, although at the cost of moderately higher NSR and FPR values.
Compared with the tree-based benchmark methods, all robust LassoNet variants achieve substantially lower PEs while also providing better control of NSR and FPR under both moderate and strong variance contamination. Random Forest exhibits slightly better recovery of the active set under moderate contamination, as reflected by a lower FNR, and remains broadly comparable to the robust methods under strong contamination. By contrast, XGBoost consistently produces larger models, resulting in markedly higher NSR and FPR values. While its FNR is comparable to that of the robust methods under moderate contamination, it deteriorates under strong contamination, where the robust LassoNet variants provide more reliable recovery of the active variables.

\item Model 4 (Table \ref{Table: model4_var}, Figure \ref{fig_boxplot_model4_variance}). The Nonnegative Garrote achieves the lowest PE under both moderate and strong variance contamination (0.258 and 0.247, respectively). Under moderate contamination, the robust LasssoNet losses exhibit highly comparable behaviour across all metrics. In particular, Cauchy and Tukey achieve perfect recovery of the active set (FNR = 0.000), with Tukey exhibiting slightly higher NSR and FPR values.
Under strong contamination, Tukey and Garrote provide the best overall results, with Garrote exhibiting a slight advantage in terms of NSR and FPR, while both methods achieve perfect recovery of the active set.
Compared with the tree-based benchmark methods, all robust LassoNet variants substantially outperform both Random Forest and XGBoost in predictive accuracy under both moderate and strong variance contamination. XGBoost yields considerably higher NSR and FPR throughout, together with a much higher FNR, and thus the weakest overall variable selection performance. Random Forest deteriorates most sharply in predictive error — its PE reaching 1.623 under strong variance — while its NSR and FPR are comparable to the robust methods under moderate variance and clearly higher only under strong variance; in both cases its FNR remains well above that of the robust approaches.

\end{itemize}

\begin{table}[htbp]
	\centering
	\small
	\resizebox{\textwidth}{!}{
		\begin{tabular}{l|cccc|cccc|cccc}
			\toprule
			& \multicolumn{4}{c|}{Gaussian contamination} 
			& \multicolumn{4}{c|}{Moderate variance contamination} 
			& \multicolumn{4}{c}{Strong variance contamination} \\
			\cmidrule(lr){2-5}\cmidrule(lr){6-9}\cmidrule(lr){10-13}
			Loss 
			& PE & NSR & FPR & FNR
			& PE & NSR & FPR & FNR
			& PE & NSR & FPR & FNR \\
			\midrule
			
			mse 
			& 0.398 & 0.122 & 0.103 & 0.152
			& 0.407 & 0.135 & 0.118 & 0.192
			& 0.626 & 0.036 & 0.025 & 0.536 \\
			
			huber
			& 0.399 & 0.087 & 0.068 & 0.176
			& 0.339 & 0.081 & 0.063 & 0.192
			& 0.336 & 0.054 & 0.035 & 0.200 \\
			
			cauchy
			& 0.402 & 0.102 & 0.083 & 0.160
			& 0.339 & 0.081 & 0.062 & 0.192
			& 0.330 & 0.072 & 0.053 & 0.184 \\
			
			tukey
			& 0.400 & 0.091 & 0.073 & 0.176
			& \textbf{0.336} & 0.091 & 0.072 & 0.184
			& \textbf{0.323} & 0.078 & 0.059 & 0.192 \\
			
			garrote
			& 0.400 & 0.090 & 0.071 & 0.176
			& 0.338 & 0.079 & 0.060 & 0.192
			& 0.326 & 0.081 & 0.062 & 0.184 \\

			rf 
			& 0.305 & 0.032 & 0.007 & 0.000
			& 0.454 & 0.131 & 0.109 & 0.008 
			& 1.755 & 0.245 & 0.226 & 0.032\\

			xgb 
			& \textbf{0.286} & 0.281 & 0.263 & 0.000
			& 0.346 & 0.369 & 0.353 & 0.000
			& 0.443 & 0.374 & 0.359 & 0.024\\
			
			%\midrule
			
			%oracle
			%& \textbf{0.193} & -- & -- & --
			%& \textbf{0.159} & -- & -- & --
			%& \textbf{0.153} & -- & -- & -- \\
			
			\bottomrule
		\end{tabular}
	}
	\caption{Performance under Gaussian and variance contamination scenarios for \textbf{Model 1}.}
	\label{Table: model1_var}
\end{table}

\begin{table}[htbp]
	\centering
	\small
	\resizebox{\textwidth}{!}{
		\begin{tabular}{l|cccc|cccc|cccc}
			\toprule
			& \multicolumn{4}{c|}{Gaussian contamination} 
			& \multicolumn{4}{c|}{Moderate variance contamination} 
			& \multicolumn{4}{c}{Strong variance contamination} \\
			\cmidrule(lr){2-5}\cmidrule(lr){6-9}\cmidrule(lr){10-13}
			Loss 
			& PE & NSR & FPR & FNR
			& PE & NSR & FPR & FNR
			& PE & NSR & FPR & FNR \\
			\midrule
			
			mse 
			& \textbf{0.253} & 0.180 & 0.155 & 0.000
			& 0.321 & 0.147 & 0.123 & 0.067
			& 0.639 & 0.034 & 0.021 & 0.573 \\
			
			huber
			& \textbf{0.253} & 0.104 & 0.076 & 0.000
			& 0.235 & 0.117 & 0.090 & 0.000
			& 0.246 & 0.070 & 0.042 & 0.007 \\
			
			cauchy
			& 0.264 & 0.129 & 0.102 & 0.000
			& 0.237 & 0.130 & 0.103 & 0.000
			& 0.229 & 0.134 & 0.107 & 0.000\\
			
			tukey
			& 0.255 & 0.106 & 0.078 & 0.000
			& \textbf{0.228} & 0.114 & 0.086 & 0.000
			& \textbf{0.220} & 0.084 & 0.056 & 0.000 \\
			
			garrote
			& 0.257 & 0.122 & 0.095 & 0.000
			& 0.236 & 0.123 & 0.096 & 0.000
			& 0.227 & 0.133 & 0.107 & 0.000 \\

			rf
			& 0.270 & 0.035 & 0.006 & 0.007 
			& 0.413 & 0.129 & 0.102 & 0.013
			& 1.640 & 0.241 & 0.219 & 0.067\\

            xgb 
            & 0.260 & 0.257 & 0.234 & 0.000 
            & 0.340 & 0.369 & 0.351 & 0.047
            & 0.451 & 0.373 & 0.358 & 0.140\\
			
			%\midrule
			
			%oracle
			%& \textbf{0.182} & -- & -- & --
			%& \textbf{0.158} & -- & -- & --
			%& \textbf{0.153} & -- & -- & -- \\
			
			\bottomrule
		\end{tabular}
	}
	\caption{Performance under Gaussian and variance contamination scenarios for \textbf{Model 2}. }
	\label{Table: model2_var}
\end{table}

\begin{table}[htbp]
	\centering
	\small
	\resizebox{\textwidth}{!}{
		\begin{tabular}{l|cccc|cccc|cccc}
			\toprule
			& \multicolumn{4}{c|}{Gaussian contamination} 
			& \multicolumn{4}{c|}{Moderate variance contamination} 
			& \multicolumn{4}{c}{Strong variance contamination} \\
			\cmidrule(lr){2-5}\cmidrule(lr){6-9}\cmidrule(lr){10-13}
			Loss 
			& PE & NSR & FPR & FNR
			& PE & NSR & FPR & FNR
			& PE & NSR & FPR & FNR \\
			\midrule
			
			mse 
			& 0.498 & 0.266 & 0.244 &0.027
			& 0.687 & 0.126 & 0.107 & 0.280
			& 1.242 & 0.019 & 0.015 & 0.833 \\
			
			huber
			& 0.523 & 0.137 & 0.114 & 0.100
			& 0.449 & 0.157 & 0.135 & 0.140
			& 0.456 & 0.129 & 0.107 & 0.180 \\
			
			cauchy
			& 0.532 & 0.106 & 0.082 & 0.107
			& 0.444 & 0.133 & 0.111 & 0.160
			& 0.436 & 0.141 & 0.119 & 0.160 \\
			
			tukey
			& 0.536 & 0.142 & 0.118 & 0.094
			& 0.450 & 0.126 & 0.104& 0.154
			& 0.433 & 0.113 & 0.090 & 0.147 \\
			
			garrote
			& 0.510 & 0.152 & 0.129 & 0.080
			& \textbf{0.432} & 0.154 & 0.132 & 0.134
			& \textbf{0.417} & 0.122 & 0.099 & 0.134 \\
			
			rf 
			& \textbf{0.495} & 0.040 & 0.013 & 0.080 
			& 0.677 & 0.168 & 0.146 & 0.127 
			& 2.713 & 0.259 & 0.241 & 0.154\\
			
			xgb 
			& 0.595 & 0.310 & 0.291 & 0.087 
			& 0.633 & 0.376 & 0.362 & 0.160 
			& 0.719 & 0.371 & 0.358 & 0.200\\
			
			%\midrule
			
			%oracle
			%& \textbf{0.311} & -- & -- & --
			%& \textbf{0.245} & -- & -- & --
			%& \textbf{0.237} & -- & -- & -- \\
			
			\bottomrule
		\end{tabular}
	}
	\caption{Performance under Gaussian and variance contamination scenarios for \textbf{Model 3}. }
	\label{Table: model3_var}
\end{table}

\begin{table}[htbp]
	\centering
	\small
	\resizebox{\textwidth}{!}{
		\begin{tabular}{l|cccc|cccc|cccc}
			\toprule
			& \multicolumn{4}{c|}{Gaussian contamination} 
			& \multicolumn{4}{c|}{Moderate variance contamination} 
			& \multicolumn{4}{c}{Strong variance contamination} \\
			\cmidrule(lr){2-5}\cmidrule(lr){6-9}\cmidrule(lr){10-13}
			Loss 
			& PE & NSR & FPR & FNR
			& PE & NSR & FPR & FNR
			& PE & NSR & FPR & FNR \\
			\midrule
			
			mse 
			& 0.282 & 0.188 & 0.146 & 0.000
			& 0.350 & 0.173 & 0.136 & 0.120
			& 0.666 & 0.032 & 0.021 & 0.756 \\
			
			huber
			& 0.294 & 0.136 & 0.091 & 0.000
			& 0.259 & 0.158 & 0.114 & 0.004
			& 0.259 & 0.094 & 0.049 & 0.044 \\
			
			cauchy
			& 0.295 & 0.154 & 0.109 & 0.000
			& 0.259 & 0.160 & 0.115 & 0.000
			& 0.251 & 0.147 & 0.103 & 0.004 \\
			
			tukey
			& 0.296 & 0.153 & 0.109 & 0.004
			& 0.260 & 0.171 & 0.128 & 0.000
			& 0.249 & 0.145 & 0.099 & 0.000\\
			
			garrote
			& 0.291 & 0.141 & 0.096 & 0.000
			& \textbf{0.258} & 0.154 & 0.109 & 0.004
			& \textbf{0.247} & 0.137 & 0.092 & 0.000 \\
			
			rf 
			& 0.356 & 0.052 & 0.008 & 0.112 
			& 0.461 & 0.160 & 0.125 & 0.180
			& 1.623 & 0.254 & 0.229 & 0.272 \\
			
           xgb
           & \textbf{0.281} & 0.272 & 0.234 & 0.000 
           & 0.382 & 0.371 & 0.345 & 0.136
           & 0.481 & 0.374 & 0.355 & 0.256\\
			
			%\midrule
			
			%oracle
			%& \textbf{0.195} & -- & -- & --
			%& \textbf{0.166} & -- & -- & --
			%& \textbf{0.160} & -- & -- & -- \\
			
			\bottomrule
		\end{tabular}
	}
	\caption{Performance under Gaussian and variance contamination scenarios for \textbf{Model 4}. }
	\label{Table: model4_var}
\end{table}

Across all contamination scenarios, no single robust LassoNet methods uniformly dominates the others. Nevertheless, the experiments consistently show that robust losses considerably improve the stability of MSE-based LassoNet under contamination, leading to substantial gains in predictive accuracy and more reliable variable selection, particularly in highly nonlinear settings and under strong contamination.
On the other hand, the tree-based benchmarks, while competitive under Gaussian noise, are less resilient to contamination: Random Forest suffers the largest losses in predictive accuracy, whereas XGBoost shows a milder loss in predictive accuracy but selects substantially larger models than all LassoNet variants in every scenario.

\subsection{Real Data Applications}
In this section, we analyze two real-world datasets from the literature, namely Top Gear and Atherosclerosis, which are described in detail in the following subsections.

To assess both predictive performance and variable selection capabilities, each dataset was randomly split into a training set (80\%) and a test set (20\%). The models were trained on the training set, and predictive accuracy was evaluated on the test set using a trimmed prediction error (trimmed PE). Specifically, the largest 5\% of residuals in absolute value were excluded from the calculation of the PE. At the same time, we recorded the variables selected by each method when fitted on the training data.
This random splitting procedure was repeated 50 times. The distribution of the trimmed PE and the Number of Selected Rate (NSR) across the repetitions were then compared among the proposed methods using boxplots. For completeness we also report  the average over the 50 runs of both metrics (trimmed PE and NSR) in a table.
Regarding variable selection, for each method we computed the relative selection frequency of each variable across the 50 splits. We report only those variables that were selected in more than 70\% of the repetitions.

\noindent
The results obtained for the two datasets are presented below.

\subsection{Top Gear}
The Top Gear dataset is described in \citep{Alfons2016} and contains information on cars scraped from the website of the popular BBC television show Top Gear (\url{http://www.topgear.com/uk/}). It is included in the R package robustHD \citep{Alfons2021}, and consists of $n = 242$ complete observations on $d = 29$ numerical and categorical variables. Specifically, it includes 4 categorical variables with two levels and 12 categorical variables with three levels, as detailed in Table~\ref{Table:topgear_description}.

\noindent
In this study, we use MPG (fuel consumption) as the response variable, while the remaining variables are treated as predictors. The resulting design matrix comprises 12 numerical variables, 4 individual dummy variables, and 12 groups of two dummy variables each, for a total of 40 covariates. Following \citep{Alfons2016}, we apply a logarithmic transformation to the Price (list price) variable to account for skewness.

\noindent
Furthermore, this dataset is particularly suitable for analysis using non-linear techniques, as its non-linear structure is extensively investigated in \citep{Bottmer2022}.

As a linear robust benchmark specifically designed for grouped predictors, we include the RGRPLARS method of \citep{Alfons2016}, which applies robust groupwise least angle regression and is implemented in the R package robustHD (\cite{Alfons2021}).

\begin{table}[htbp]
\centering
\caption{Description of the Top Gear car data.}
\label{Table:topgear_description}
\small % Riduce leggermente la dimensione per far stare tutto comodamente
\begin{tabular}{lll}
\toprule
\textbf{Variable} & \textbf{Description} & \textbf{Possible outcomes} \\
\midrule
MPG & Fuel consumption (in miles per gallon) & \\
Fuel & Type of fuel & Diesel, Petrol \\
Price & List price (in UK pounds) & \\
Cylinders & Number of cylinders in the engine & \\
Displacement & Displacement of the engine (in cc) & \\
DriveWheel & Type of drive wheel & 4WD, Front, Rear \\
BHP & Power of the engine (in bhp) & \\
Torque & Torque of the engine (in lb/ft) & \\
Acceleration & Time from 0 to 62 mph (in seconds) & \\
TopSpeed & Top speed (in mph) & \\
Weight & Curb weight (in kg) & \\
Length & Length (in mm) & \\
Width & Width (in mm) & \\
Height & Height (in mm) & \\
AdaptiveHeadlights & Whether the car has adaptive headlights & No, Optional, Standard \\
AdjustableSteering & Whether the car has adjustable steering & No, Standard \\
AlarmSystem & Whether the car has an alarm system & No/optional, Standard \\
Automatic & Whether the car has an automatic transmission & No, Optional, Standard \\
Bluetooth & Whether the car has bluetooth & No, Optional, Standard \\
ClimateControl & Whether the car has climate control & No, Optional, Standard \\
CruiseControl & Whether the car has cruise control & No, Optional, Standard \\
ElectricSeats & Whether the car has electric seats & No, Optional, Standard \\
Leather & Whether the car has a leather interior & No, Optional, Standard \\
ParkingSensors & Whether the car has parking sensors & No, Optional, Standard \\
PowerSteering & Whether the car has power steering & No, Standard \\
SatNav & Whether the car has a satellite navigation system & No, Optional, Standard \\
ESP & Whether the car has ESP & No, Optional, Standard \\
Verdict & Review score & \\
Origin & Origin of the car maker & Asia, Europe, USA \\
\bottomrule
\end{tabular}
\end{table}

\begin{figure*}[h]
\centering
\includegraphics[width=0.48\textwidth]{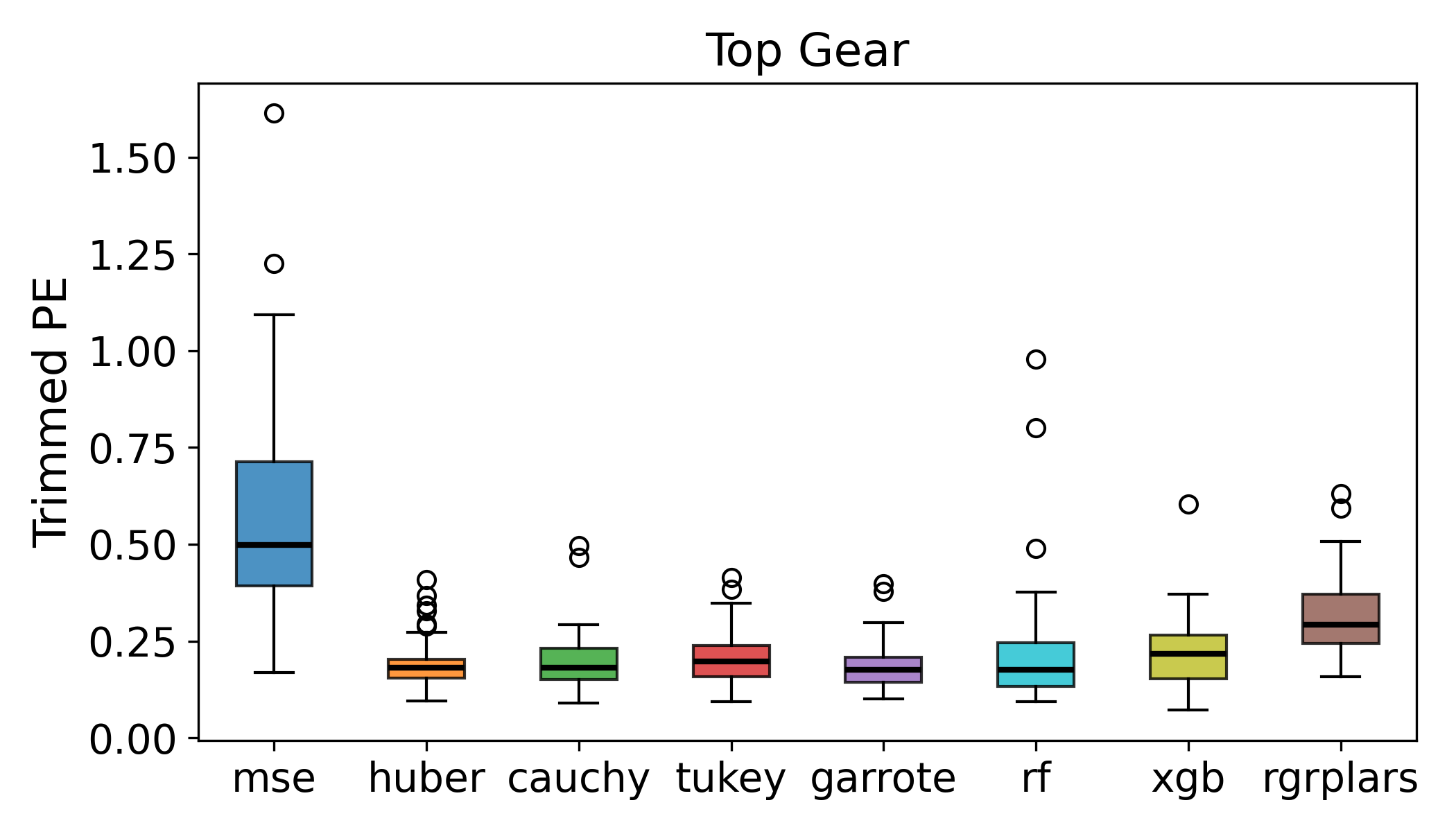}
\hfill
\includegraphics[width=0.48\textwidth]{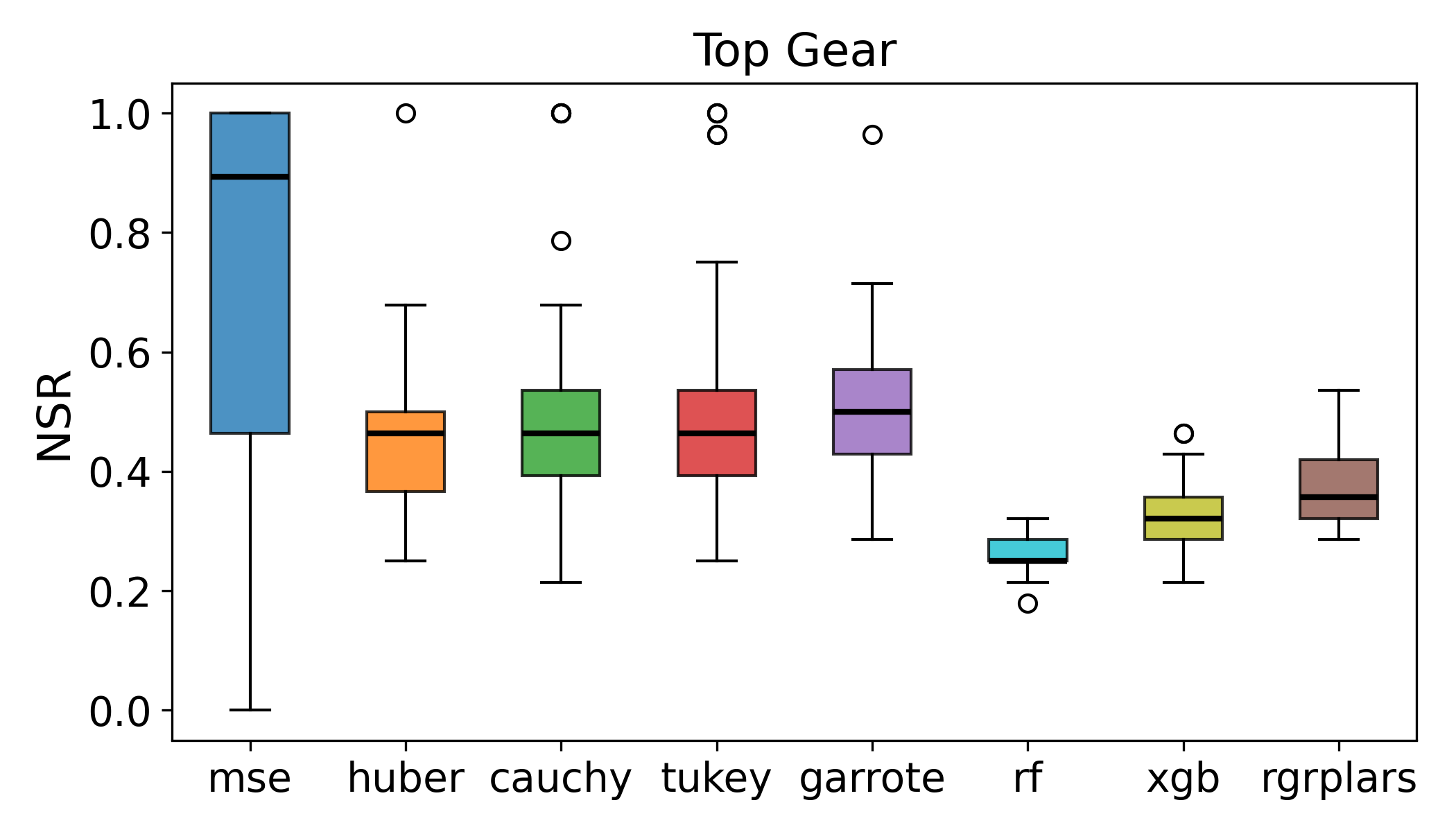}
\caption{Boxplots of performance metrics for the {\tt Top Gear} dataset. The left panel reports trimmed PE, while the right panel reports NSR at the variable level, across the methods.}
\label{fig_boxplot_topgear}
\end{figure*}

The results highlight a clear advantage of robust LassoNet methods over the classical MSE-based LassoNet. In terms of predictive performance (Table \ref{Table: TopGear}, Figure \ref{fig_boxplot_topgear}), all robust LassoNet
losses (Huber, Cauchy, Tukey, and Garrote) achieve lower and more stable trimmed prediction error, whereas the MSE exhibits high variability and extreme values.
Differences also emerge in model complexity (Figure \ref{fig_boxplot_topgear}: right panel) because they  tend to select more parsimonious models with respect to MSE, with only minor differences among them.

Compared with the tree-based benchmark methods, the robust LassoNet variants achieve slightly better predictive performance, although Random Forest and XGBoost produce considerably sparser models, as reflected by their lower NSR values. RGRPLARS also yields a relatively sparse solution, but with a substantially larger trimmed prediction error than both the robust LassoNet variants and the tree-based methods. Overall, the robust LassoNet methods offer a favourable balance between predictive accuracy and model complexity on this dataset.

\begin{table}[h]
\begin{tabular}{lrr}
\toprule
{\bf Loss} & {\bf trimmed PE} & {\bf NSR} \\
\midrule
mse & 0.570 & 0.734 \\
huber & 0.193 & 0.455 \\
cauchy & 0.197 & 0.500 \\
tukey & 0.204 & 0.508 \\
garrote & 0.186 & 0.511 \\
rf & 0.214 & 0.259 \\
xgb & 0.213 & 0.321 \\
rgrplars & 0.312 & 0.369 \\
\bottomrule
\end{tabular}
\caption{Performance for {\tt Top Gear} dataset. All metrics have been averaged over 50 independent runs.}
\label{Table: TopGear}
\end{table}

% boxplots Topgear

\begin{figure*}[h]
\centering
\includegraphics[width=\textwidth]{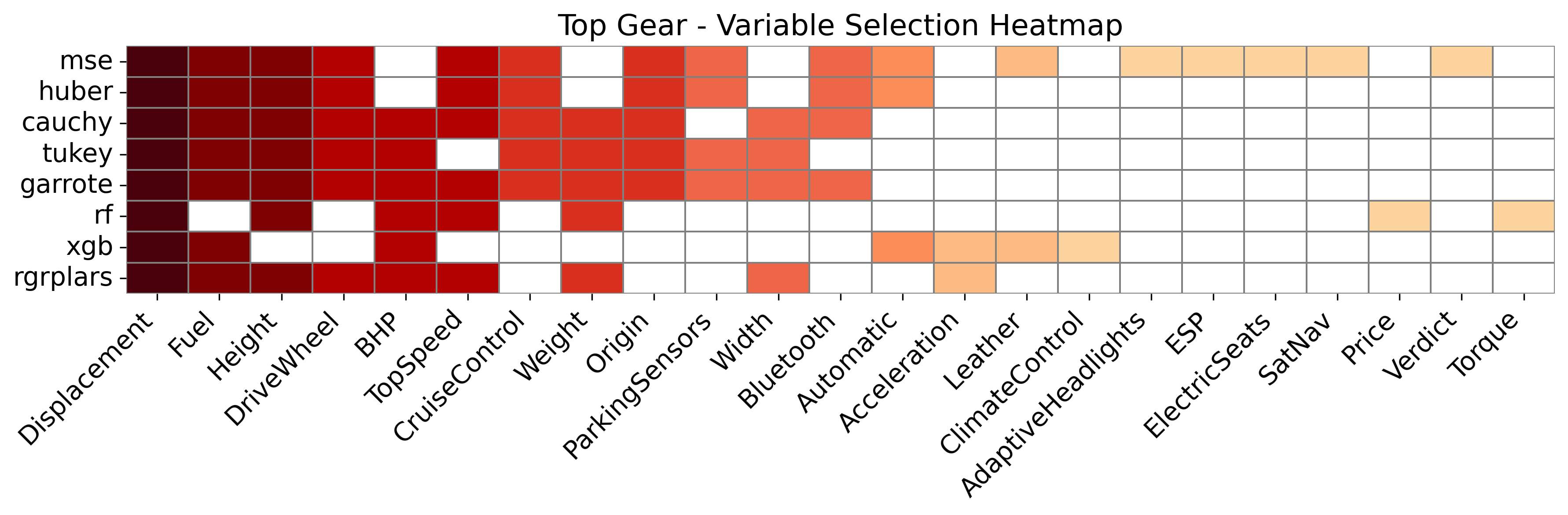}
\caption{Heatmap of variable selection for the {\tt Top Gear} dataset. Rows correspond to the considered methods, and columns to variables selected in at least 70\% of the 50 runs by at least one method. White cells indicate absence of selection, while colored cells denote selected variables. Color intensity reflects the level of agreement across methods, ranging from light yellow (selection by a single method) to red (selection by all eight methods).}
\label{fig_heatmap_topgear}
\end{figure*}

The heatmap of variable selection (Figure \ref{fig_heatmap_topgear}) shows a clear consistency in variable selection across the four robust LassoNet methods. In particular, the variables Displacement, CruiseControl, Height, Fuel, DriveWheel, and Origin are selected by all four methods in more than 70\% of the runs, and represent a stable core of variables. The variables ParkingSensors, Bluetooth, BHP, Weight, TopSpeed, and Width are selected by three out of the four methods considered. Finally, the variable Automatic appears more method dependent being selected only by the Huber loss.

\noindent
These findings are consistent with the results reported in Table 4 of \citep{Alfons2016} and in Table 6 of \citep{Bottmer2022}, where the same dataset was analyzed using different methodologies.

The benchmark methods exhibit a more heterogeneous selection pattern. Random Forest and XGBoost each recover only part of the stable core identified by the robust LassoNet methods — Random Forest selects Displacement and Height, while XGBoost selects Displacement, and Fuel. Each also introduces predictors that the robust LassoNet methods do not retain (Price and Torque for Random Forest; Leather, Acceleration and ClimateControl for XGBoost), so their sparser overall models (NSR 0.259 and 0.321, respectively) reflect a partly different variable set rather than a subset of the robust selection.  RGRPLARS, by contrast, recovers a large share of the robust LassoNet selected predictors — including all of the core except Origin and CruiseControl — while additionally including Acceleration, and is therefore among the denser benchmark models (NSR 0.369). Overall, the four robust LassoNet variants achieve the greatest mutual agreement in variable selection, identifying a stable core set of predictors while preserving a moderate level of sparsity.

% Variable Selection Heatmap Topgear

\subsection{Atherosclerosis}
The atherosclerosis dataset used in this study was obtained from the repository provided by \citep{Fan2017} (available at https://fan.princeton.edu/publications/software, under the “Matlab code for Adaptive Huber estimation”, in the *RA-lasso/realdata* directory). The data originate from the microarray study of \citep{Huang2011}, who investigated the role of the innate immune system in the development of atherosclerosis by analyzing gene expression profiles from 119 individuals.

\noindent
Gene expression levels were measured using the Illumina HumanRef-8 v2.0 BeadChip and are publicly available through the Gene Expression Omnibus (GEO). The original study identified a prominent role of innate immune activation, highlighting in particular the involvement of the Toll-like receptor (TLR) signaling pathway in atherosclerosis.

\noindent
Building on this dataset, \citep{Fan2017} considered a regression framework in which the expression level of the gene TLR8 was treated as the response variable, and 464 genes from 12 biologically related pathways (including TLR, IFNG, MAPK, NOD, and apoptosis-related pathways) were used as covariates. This formulation provides a high-dimensional setting for evaluating robust regression methods.

\noindent
 Unlike \citep{Fan2017}, we focus on a reduced and data-driven subset of covariates. Specifically, we restrict the analysis to the 60 genes exhibiting the strongest association with TLR8, as measured by a robust correlation estimator. The selection is based on the *corrHuber* function implemented in the R package *robustHD* \citep{Alfons2021}, which mitigates the influence of outliers and heavy-tailed distributions. This preprocessing step aims to retain the most relevant predictors while improving stability and interpretability in the subsequent analysis. In order to reduce the number of free parameters in the LassoNet models, the network architecture is configured with a single hidden layer of five units, i.e., $\texttt{hidden\_dims}=(5,)$. 

\begin{figure*}[h]
\centering
\includegraphics[width=0.48\textwidth]{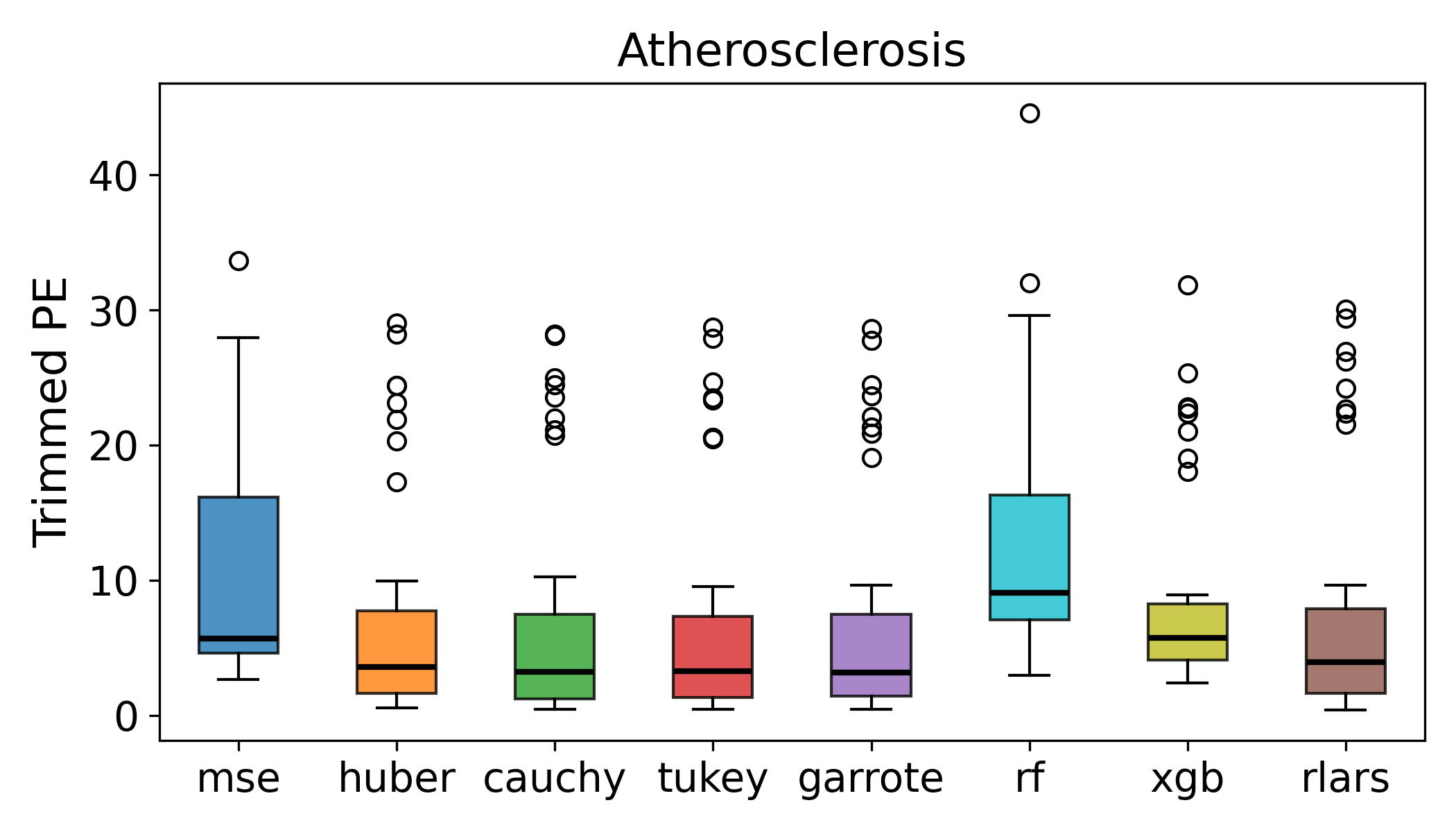}
\hfill
\includegraphics[width=0.48\textwidth]{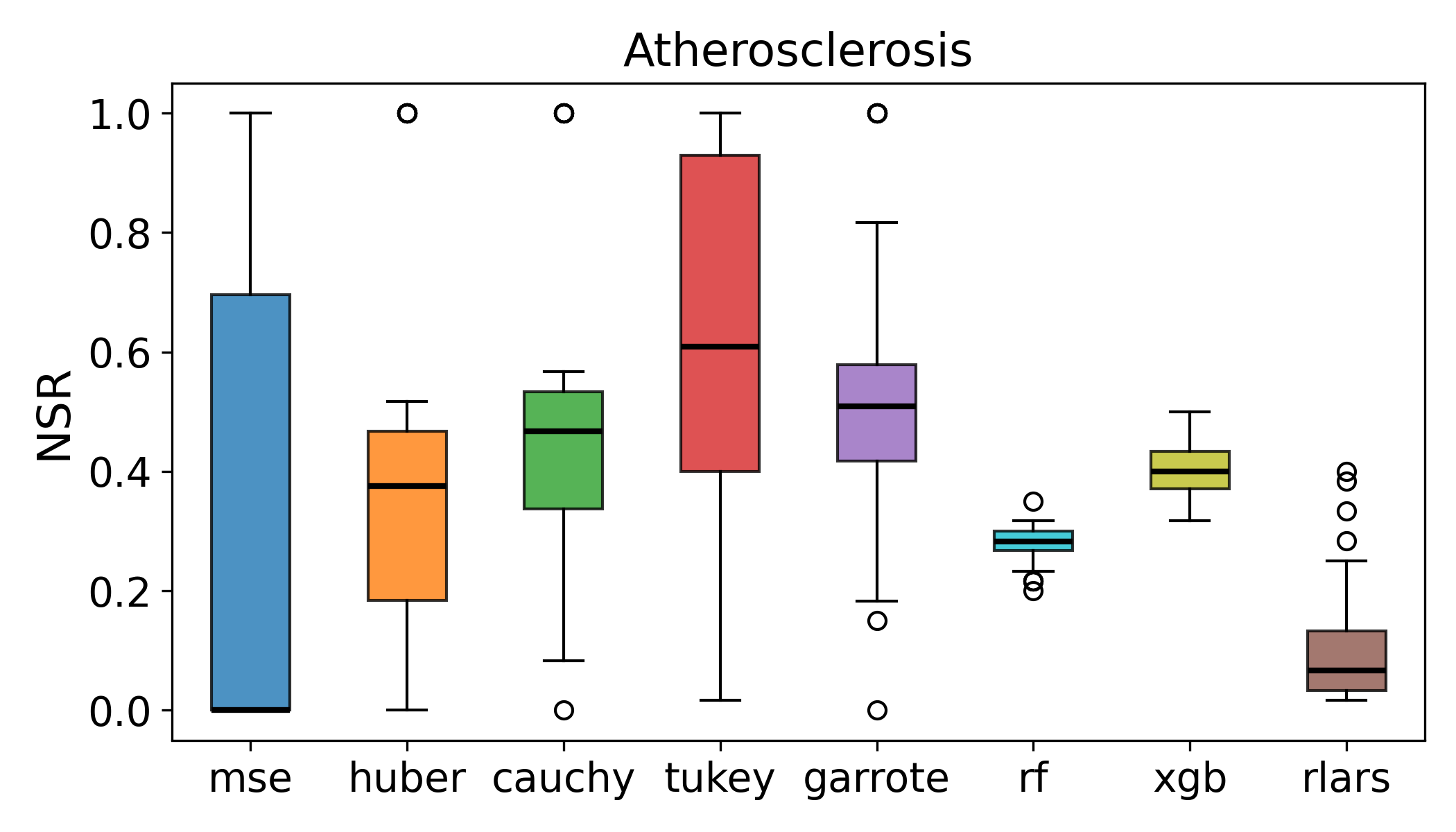}
\caption{Boxplots of performance metrics for the {\tt Atherosclerosis} dataset. The left panel reports trimmed PE, while the right panel reports NSR at the variable level, across the methods.}
\label{fig_boxplot_atherosclerosis}
\end{figure*}

\noindent
The results highlight a clear advantage of robust LassoNet methods over the classical MSE-based LassoNet. In terms of predictive performance (Table \ref{Table: atherosclerosis}, Figure \ref{fig_boxplot_atherosclerosis}), all robust losses (Huber, Cauchy, Tukey, and Garrote) achieve lower and more stable trimmed prediction error, whereas the MSE exhibits high variability and extreme values, indicating the presence of heavy-tailed noise or outliers in the data.

\noindent
Differences also emerge in model complexity (Figure \ref{fig_boxplot_atherosclerosis}): Huber tends to select the most parsimonious models, Tukey identifies larger and more variable sets of predictors, while Cauchy and Garrote provide an intermediate trade-off between sparsity and prediction.

\noindent
Compared with the benchmark methods, the robust LassoNet variants consistently achieve the lowest trimmed prediction errors. Random Forest exhibits the poorest predictive performance, while XGBoost provides an improvement over both Random Forest and the MSE estimator but remains clearly inferior to the robust losses. RLARS yields by far the sparsest models, although this pronounced sparsity is accompanied by an increase in trimmed prediction error.
Overall, the robust LassoNet methods deliver the strongest predictive accuracy while retaining an explicit, interpretable feature-selection mechanism; although their models are less sparse than those of RLARS or Random Forest, the selected variables — as discussed below — recover a biologically coherent core, making them the most informative choice on this dataset.

\begin{figure*}[h]
\centering
\includegraphics[width=\textwidth]{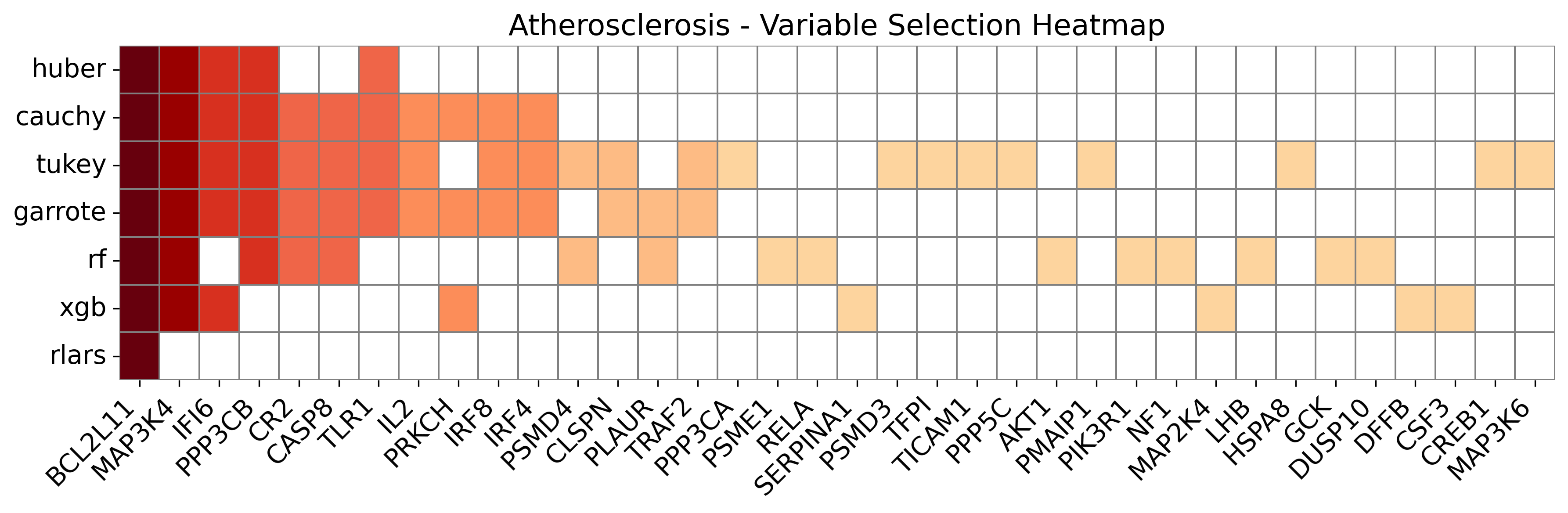}
\caption{Heatmap of variable selection for the {\tt Atherosclerosis} dataset. Rows correspond to the considered methods, and columns to variables selected in at least 70\% of the 50 runs by at least one method. White cells indicate absence of selection, while colored cells denote selected variables. Color intensity reflects the level of agreement across methods, ranging from light yellow (selection by a single method) to red (selection by all seven methods). Note that LassoNet with the MSE loss is not included, as no variable was selected in at least 70\% of the 50 runs.}
\label{fig_heatmap_atherosclerosis}
\end{figure*}

\noindent
The heatmap of variable selection (Figure \ref{fig_heatmap_atherosclerosis}) reveals a highly structured pattern, with a small group of core genes - BCL2L11, IFI6, PPP3CB, MAP3K4, and TLR1 - consistently selected across all robust LassoNet methods, indicating a strong and stable signal. A second group of genes, including IL2, CR2, CASP8, IRF4, and IRF8, is selected by the most robust procedures (Cauchy, Garrote and Tukey), suggesting moderately stable but biologically relevant effects, while the remaining selected genes appear more method-dependent.

The benchmark methods exhibit markedly different selection patterns. Random Forest identifies several of the strongest signals but also includes a number of isolated predictors that are not consistently selected by the robust LassoNet methods, whereas XGBoost retains only a small subset of the core genes while omitting many of the moderately stable predictors. RLARS produces the sparsest solution, selecting essentially only BCL2L11. 

\noindent
From a biological perspective, the selected genes map coherently to key processes involved in atherosclerosis, namely innate immune activation through the Toll-like receptor pathway, interferon signaling, and apoptosis. These findings are consistent with Huang et al. (2011), who identified the central role of TLR signaling.

\noindent
%Importantly, by restricting the analysis to a reduced subset of 60 genes selected via robust correlation, the proposed approach preserves the essential biological structure while improving interpretability and stability. Overall, these results show that robust and parsimonious models can effectively capture the core mechanisms underlying complex genomic data.

\begin{table}[h]
\begin{tabular}{lrr}
\toprule
{\bf Loss} & {\bf trimmed PE} & {\bf NSR} \\
\midrule
mse & 10.356 & 0.278 \\
huber & 6.759 & 0.402 \\
cauchy & 6.598 & 0.474 \\
tukey & 6.535 & 0.631 \\
garrote & 6.532 & 0.528 \\
rf & 12.398 & 0.278 \\
xgb & 8.193 & 0.401 \\
rlars & 7.117 & 0.103 \\
\bottomrule
\end{tabular}
\caption{Performance for {\tt Atherosclerosis} dataset. All metrics have been averaged over 50 independent runs.}
\label{Table: atherosclerosis}
\end{table}

% boxplots Atherosclerosis

% Variable Selection Heatmap Atherosclerosis

\section{Conclusions}

In this work we introduced robust LassoNet, an extension of the LassoNet framework (Lemhadri et al., 2021) in which the MSE loss is replaced by robust M-estimation losses: Huber, Cauchy, Nonnegative Garrote  and  Tukey losses, to deal with outliers and extreme observations. 

\noindent
We evaluated the methodology on four synthetic nonlinear regression models under Gaussian noise and four contamination scenarios, and on two real datasets. The comparisons included robust versions of Random Forest and XGBoost and, for the real data, robust (groupwise) least angle regression.

Under uncontaminated Gaussian noise, the simulations show that robustness costs little in efficiency. The picture changes markedly under contamination, especially vertical outliers generated by mean contamination. The breakdown of the MSE loss becomes more severe as the regression function moves from Model 1 to Model 4, with the exception of Model 2. In Model 1, which has an explicit linear and additive component, MSE degrades but still recovers a good part of the active set. In Model 2, the effect is milder. In Models 3 and 4, where the signal is driven by nested nonlinearities and interactions, the MSE estimator degrades much more sharply. In Model 4 it fails almost completely.

We conjecture that this behaviour stems from the interaction between the
contamination and the hierarchical constraint. In LassoNet, a variable can contribute to the nonlinear component only if its linear coefficient
$\theta_j$ is nonzero, so $\theta_j$ acts as a gate that gives $x_j$ access to the hidden layers. Whether this gate stays open along the regularization
path depends on how strongly $x_j$ is \emph{linearly} associated with the
residuals.

When the regression function contains an explicit linear structure, as in
Model~1, or is close to linear over the covariate range, as in Model~2, this
association is strong and the gates open easily. In Models~3 and~4, by
contrast, the signal is driven by interactions and nested nonlinear
transformations. The relevant variables therefore have only weak linear
proxies, even when their overall effect on the response is large.

Under the squared loss, the gradient with respect to $\theta_j$ is dominated
by the large residuals of the contaminated observations, which can mask this
weak linear signal. The $\ell_1$ proximal step then sets $\theta_j$ to zero,
and the hierarchical constraint~\eqref{eq:constraint} removes $x_j$ from
the whole network before its nonlinear effect can be exploited. Robust losses
bound or eliminate the influence of large residuals, so the weak linear
signal is preserved and the relevant variables can enter the nonlinear
component.

Variance contamination leads to similar conclusions: MSE LassoNet collapses under strong contamination in every model, while all robust LassoNet variants keep stable prediction and selection. The tree-based benchmarks also deteriorate considerably as nonlinear complexity grows, most notably Random Forest in Model 4.

The real-data applications confirm the indications of the simulation study. On both datasets, the robust LassoNet variants show a degree of stability in variable selection that the MSE-based LassoNet and the benchmark methods do not reach. The four robust losses largely agree with one another and identify a consistent core of predictors across random splits, and this core is in line with previous analyses of the same data and, for the Atherosclerosis dataset, with the known biology. This stability does not come at the expense of prediction, since the robust variants remain competitive with or better than
the alternatives in terms of predictive error.

Overall, Robust LassoNet provides a reliable tool when the goal is not only accurate prediction but also reproducible and interpretable variable selection in the presence of contaminated data.

\appendix
\section{Appendix}
In this section, we report the boxplots of the performance measures obtained from the synthetic experiments presented in Section \ref{subsec:expres}.

% boxplots model 1 mixture mean
\begin{figure*}[t]
	\centering
	
	% --- Row 1 ---
	\includegraphics[width=0.32\textwidth,height=0.25\textwidth]{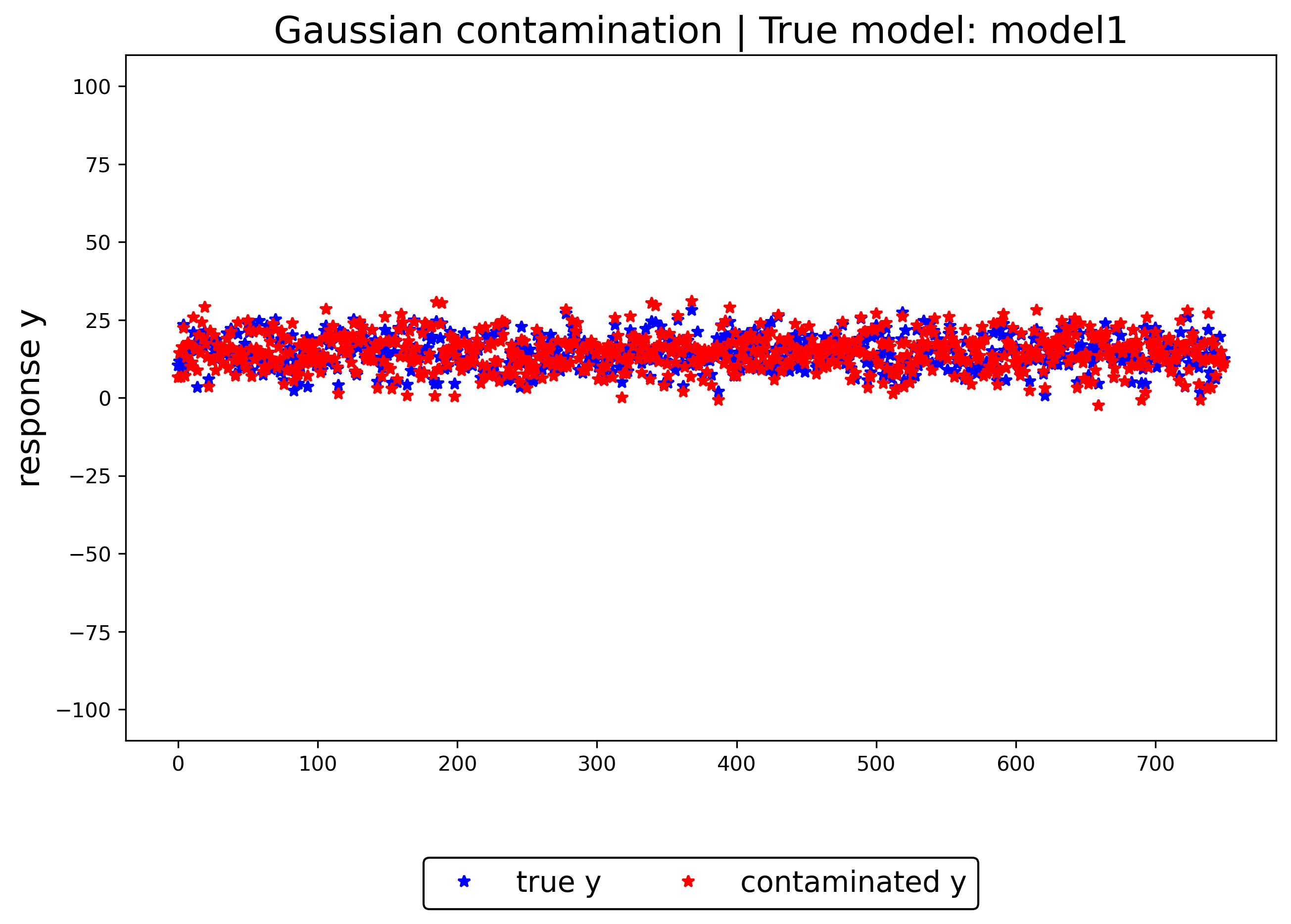}
	\includegraphics[width=0.32\textwidth,height=0.25\textwidth]{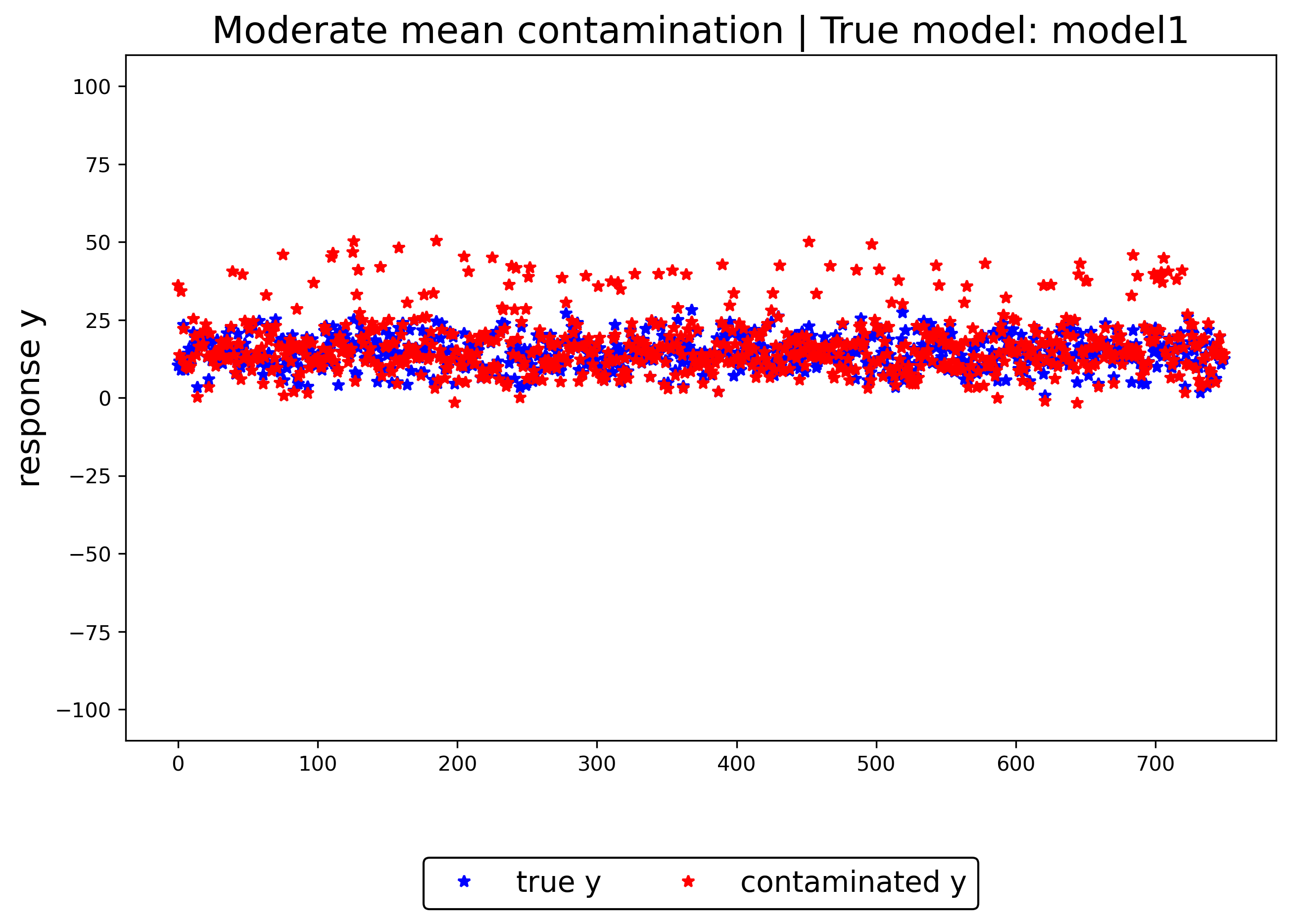}
	\includegraphics[width=0.32\textwidth,height=0.25\textwidth]{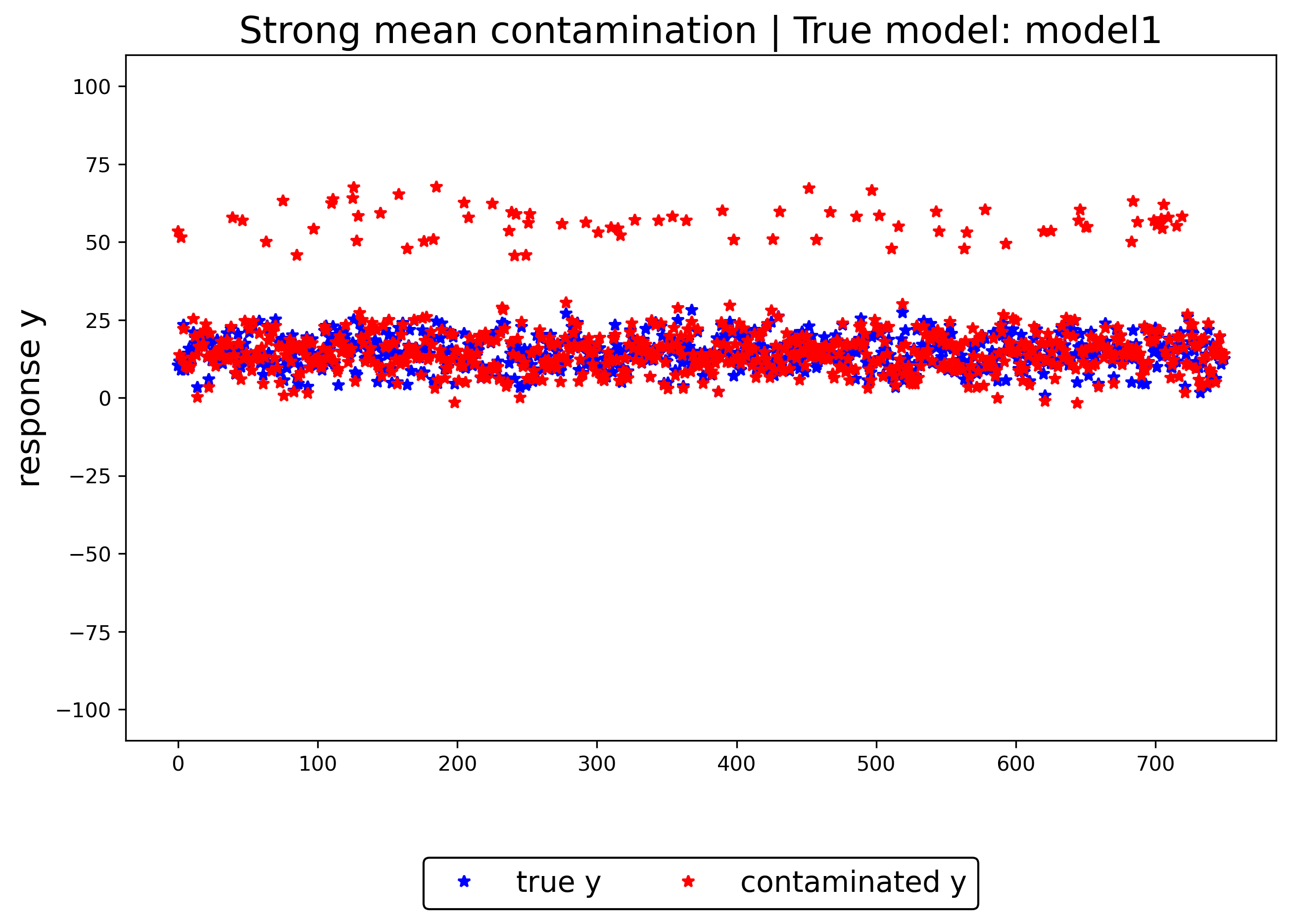}

	% --- Row 2 ---
	\includegraphics[width=0.32\textwidth,height=0.25\textwidth]{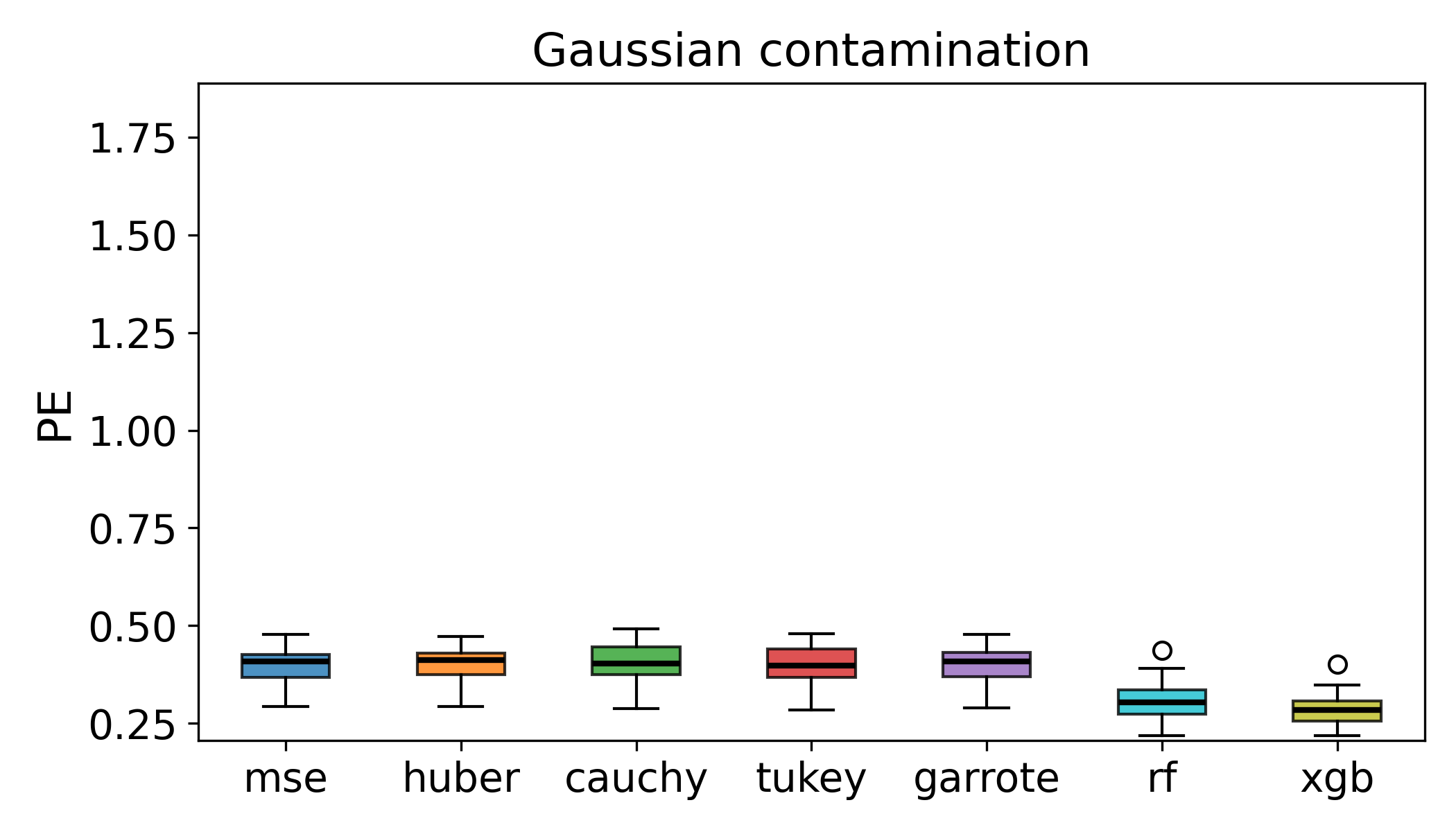}
	\includegraphics[width=0.32\textwidth,height=0.25\textwidth]{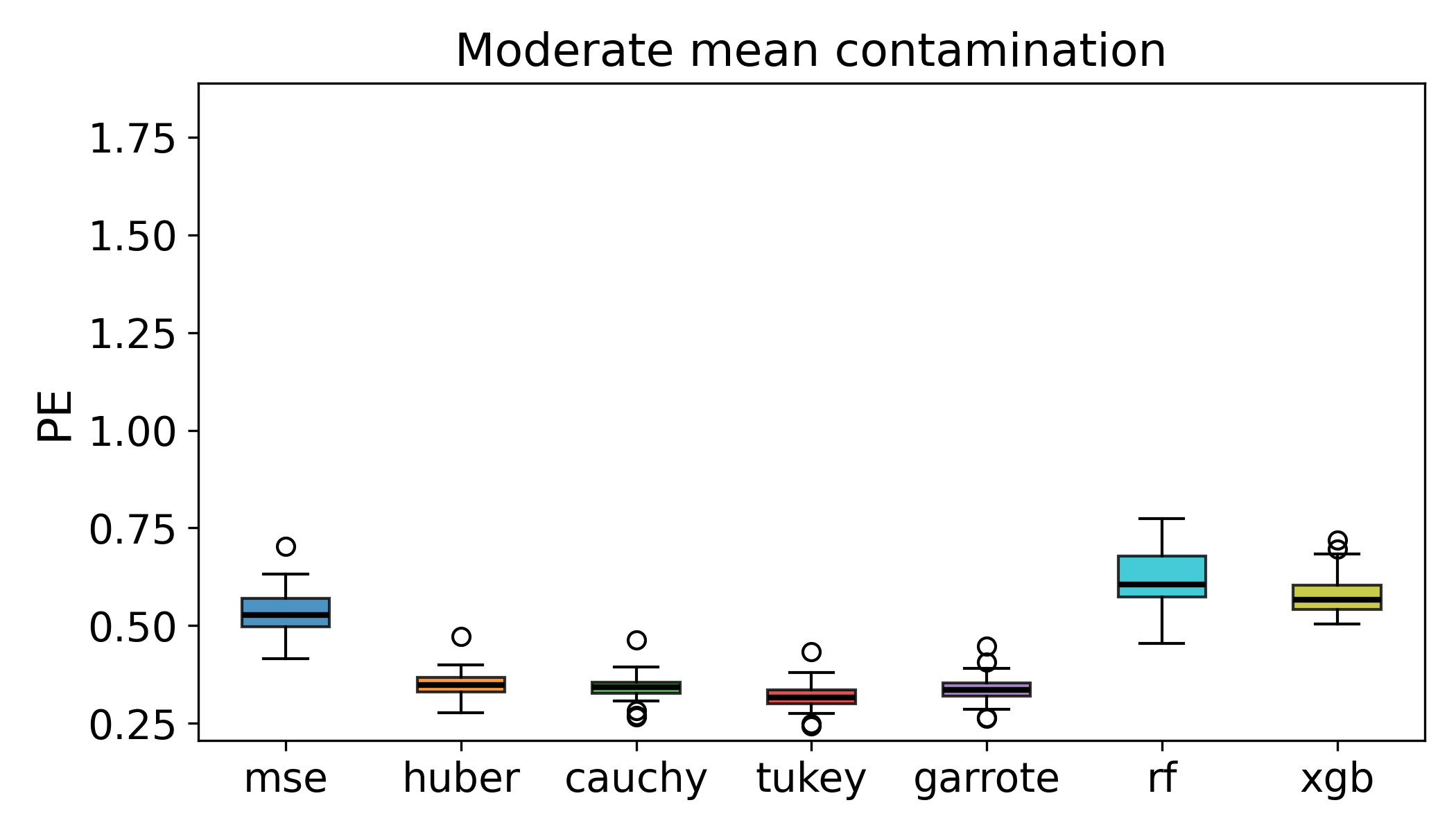}
	\includegraphics[width=0.32\textwidth,height=0.25\textwidth]{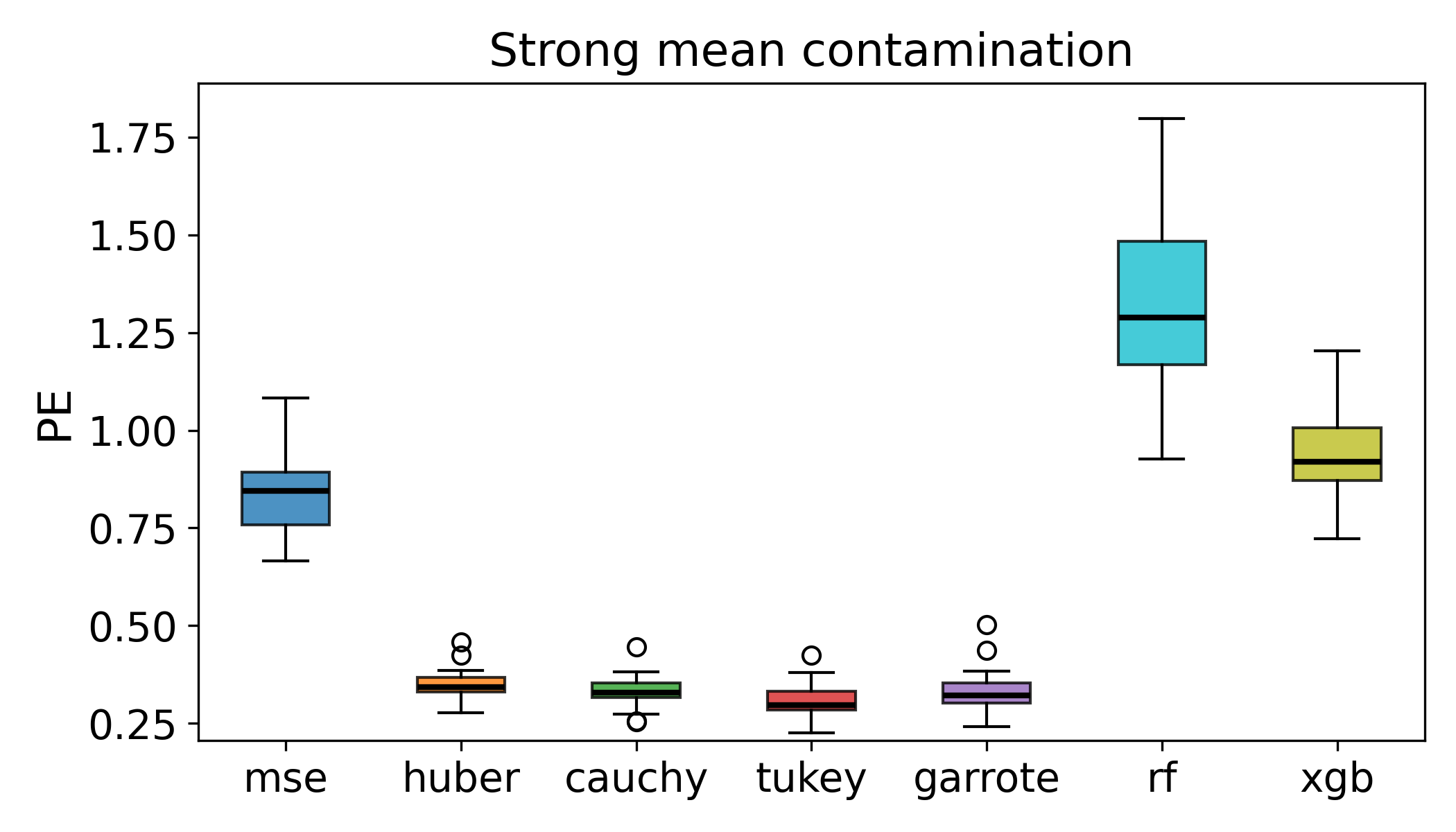}
	
	\vspace{0.2cm}
	
	% --- Row 3 ---
	\includegraphics[width=0.32\textwidth,height=0.25\textwidth]{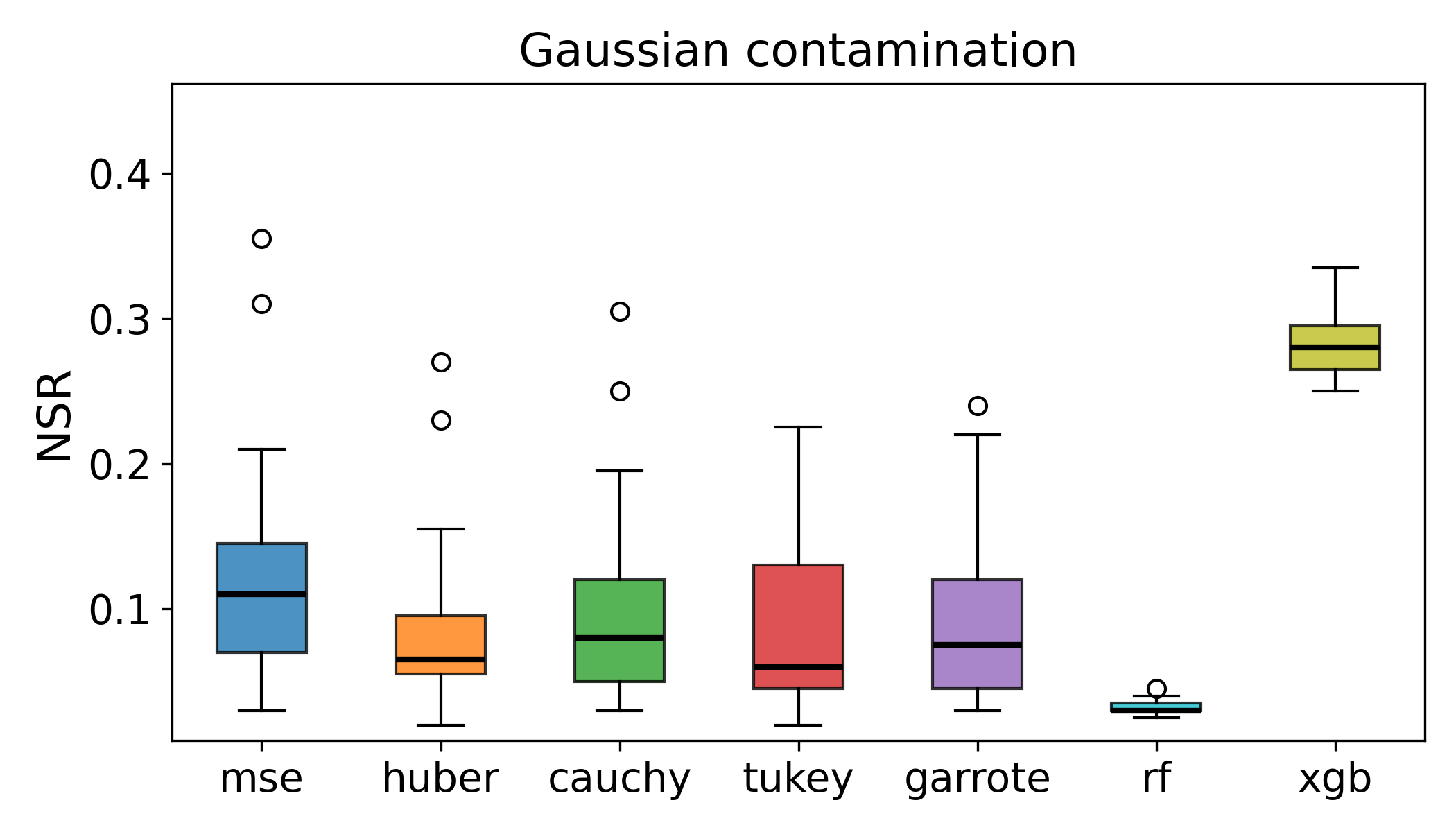}
	\includegraphics[width=0.32\textwidth,height=0.25\textwidth]{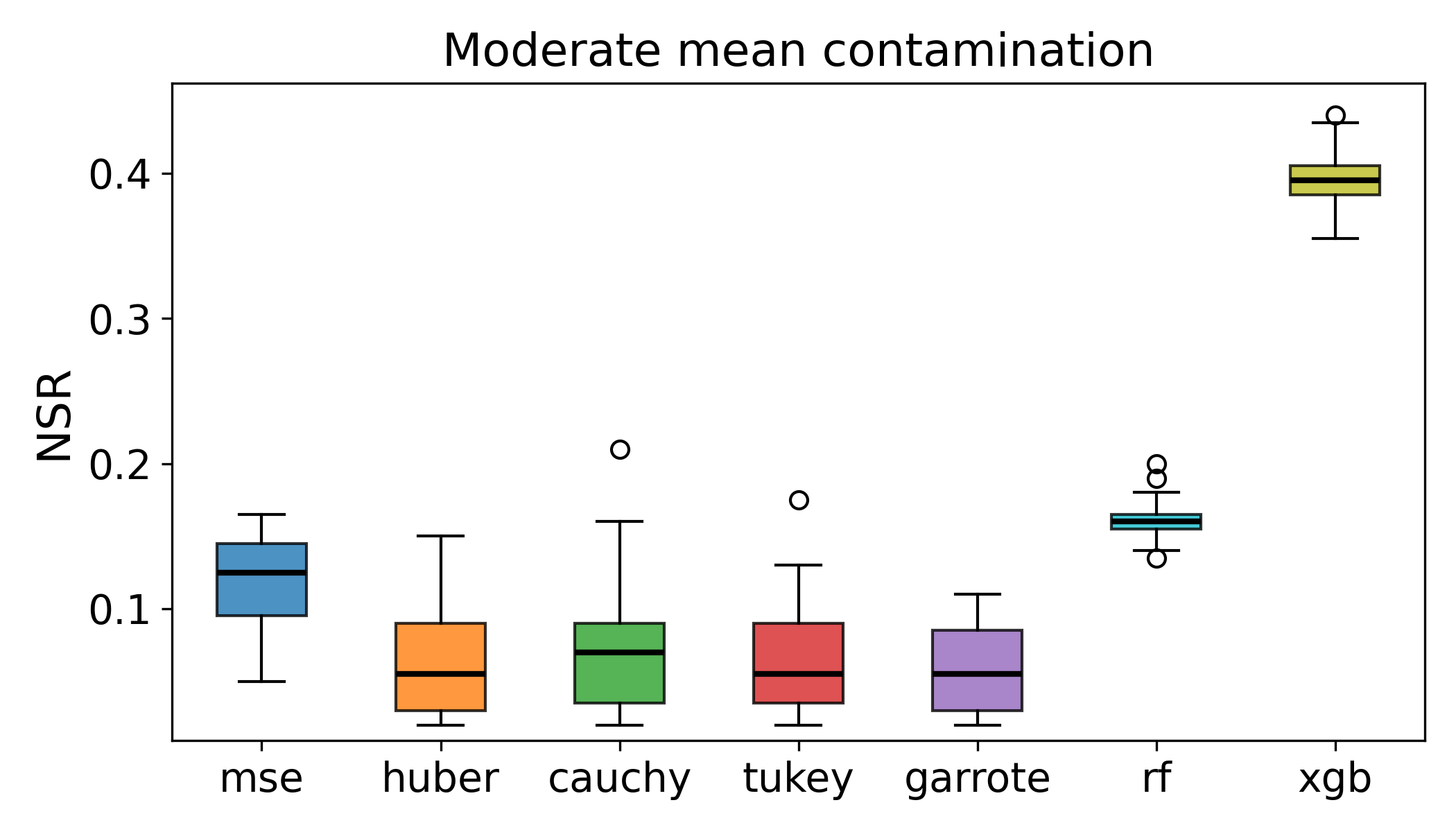}
	\includegraphics[width=0.32\textwidth,height=0.25\textwidth]{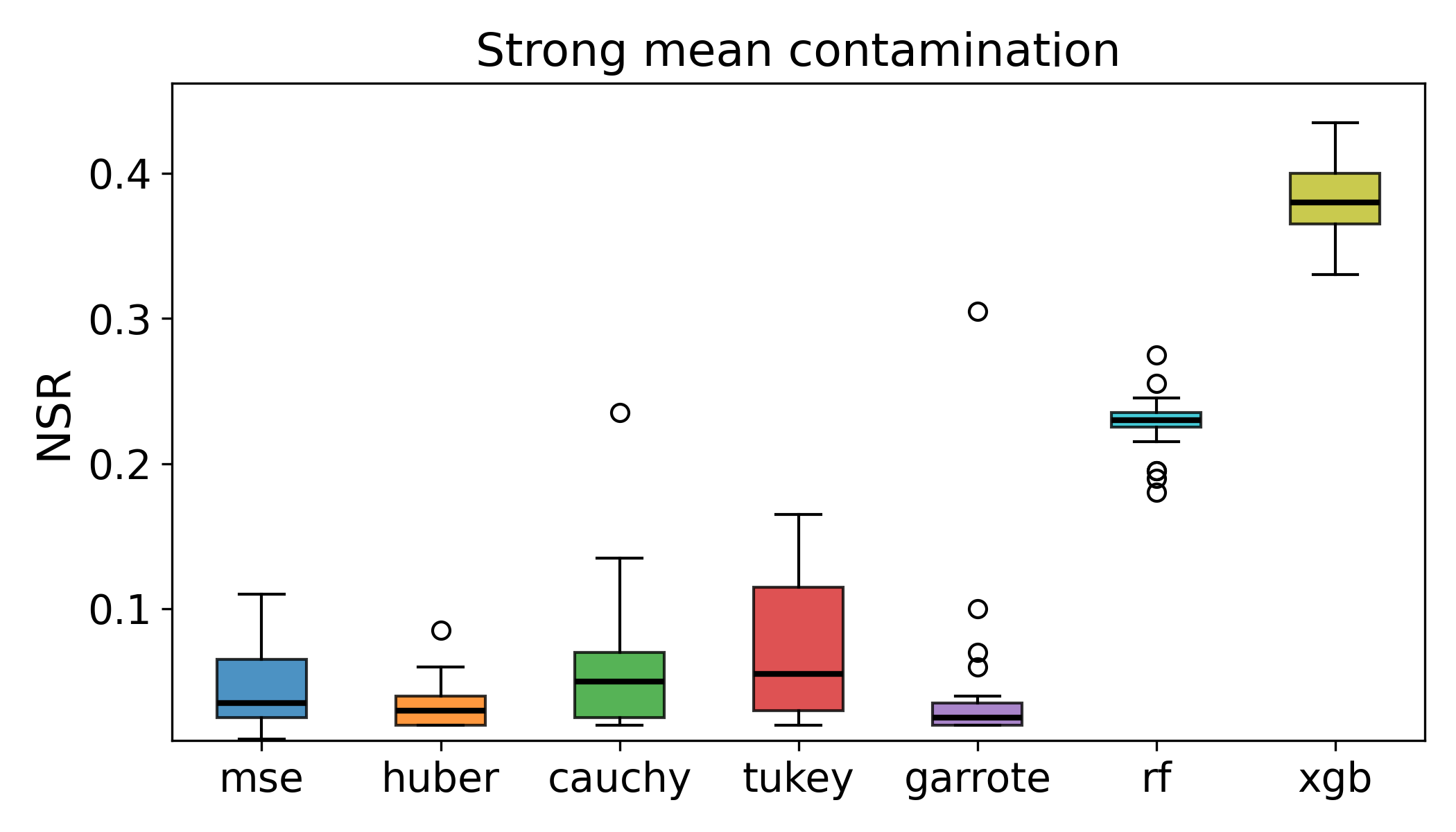}
	
	\vspace{0.2cm}
	
	% --- Row 4 ---
	\includegraphics[width=0.32\textwidth,height=0.25\textwidth]{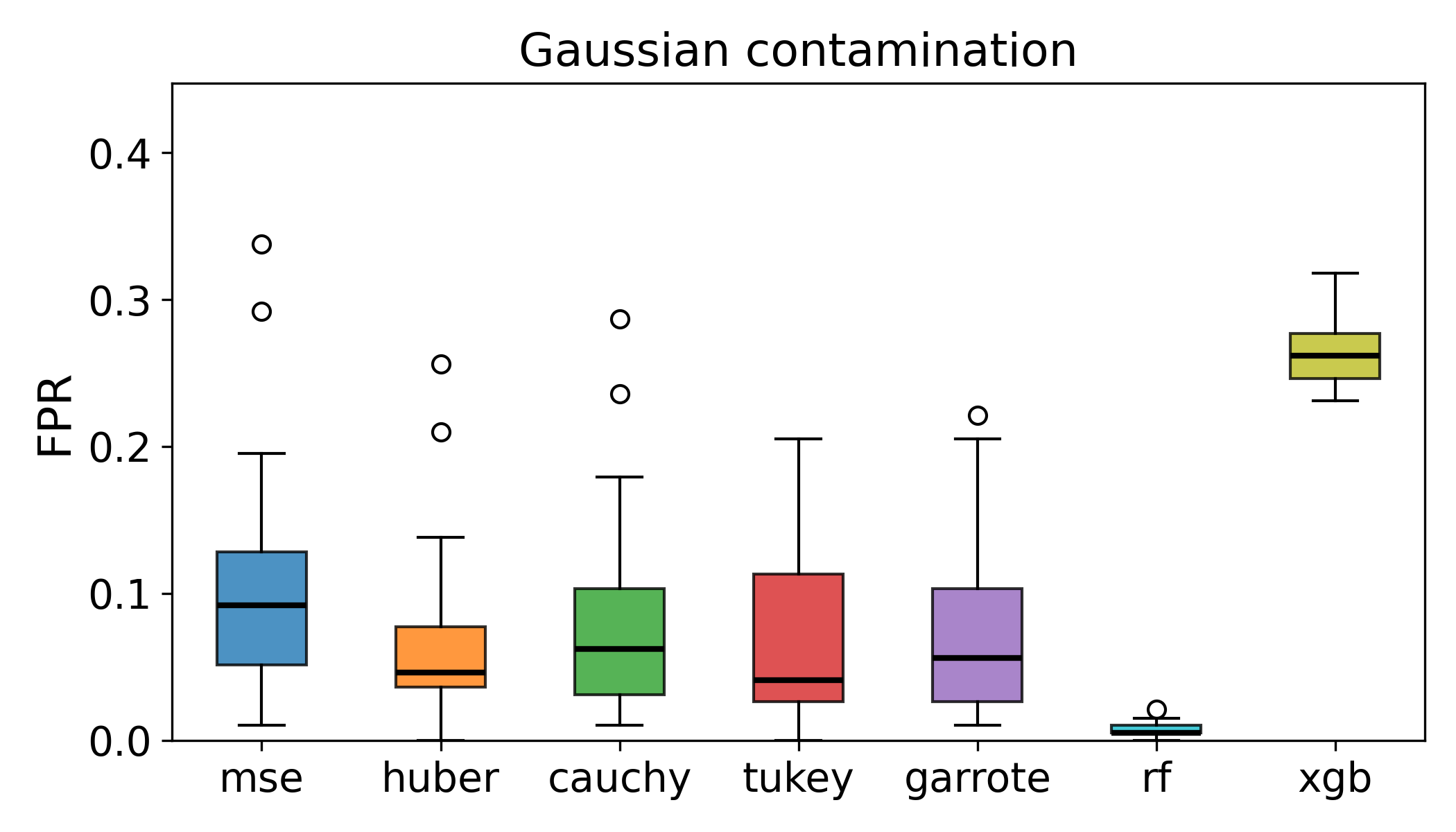}
	\includegraphics[width=0.32\textwidth,height=0.25\textwidth]{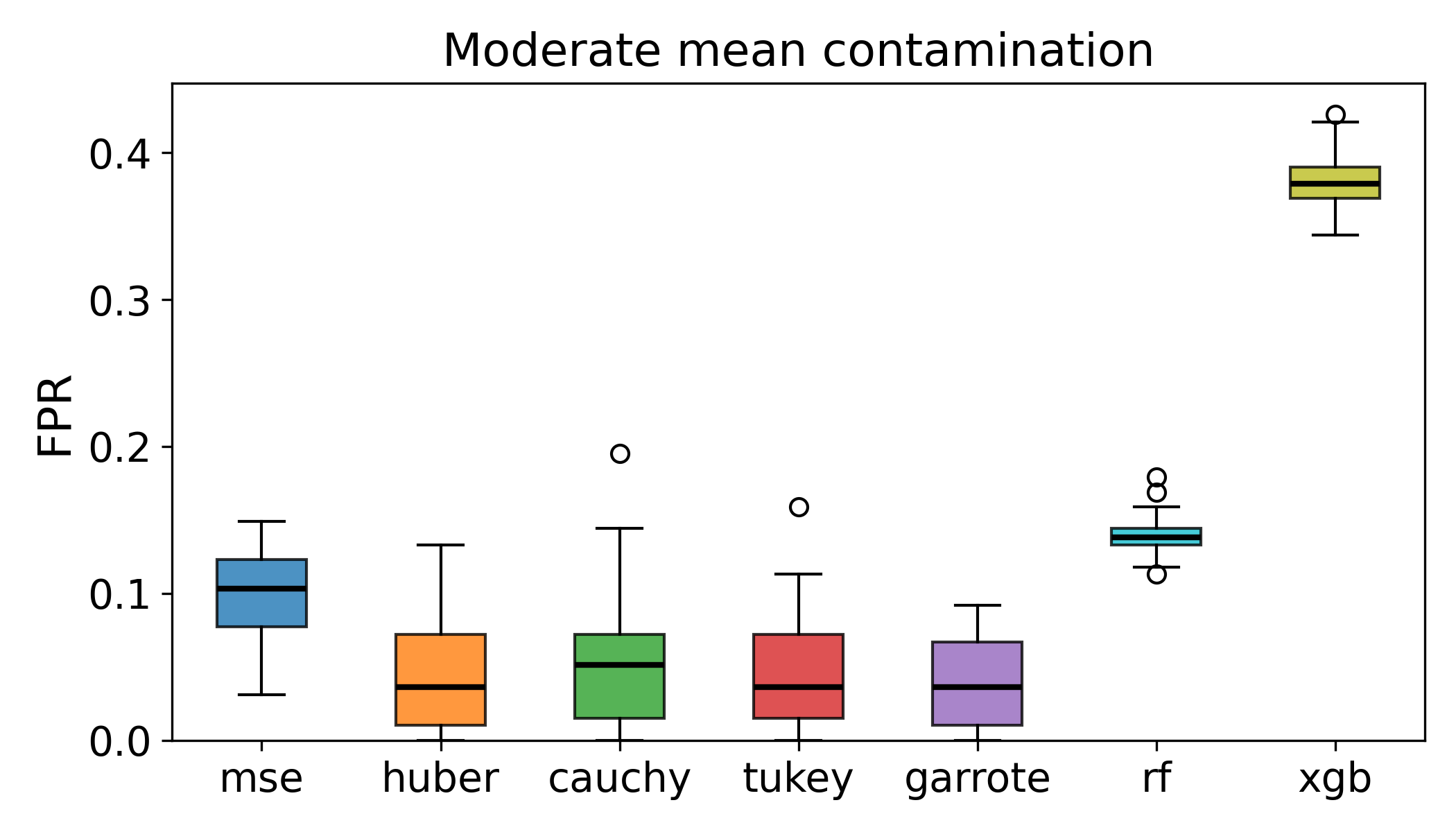}
	\includegraphics[width=0.32\textwidth,height=0.25\textwidth]{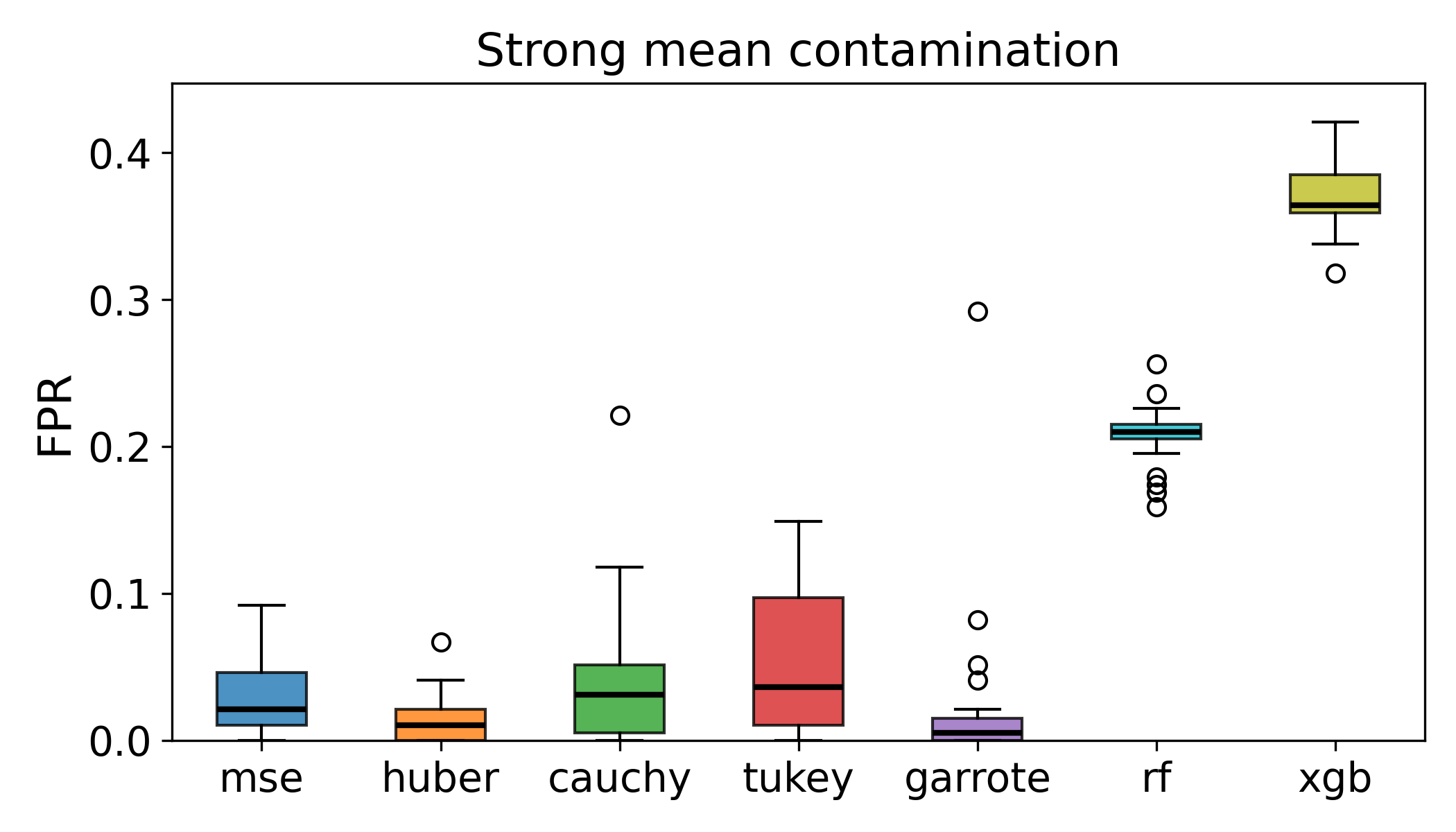}
	
	\vspace{0.2cm}
	
	% --- Row 5 ---
	\includegraphics[width=0.32\textwidth,height=0.25\textwidth]{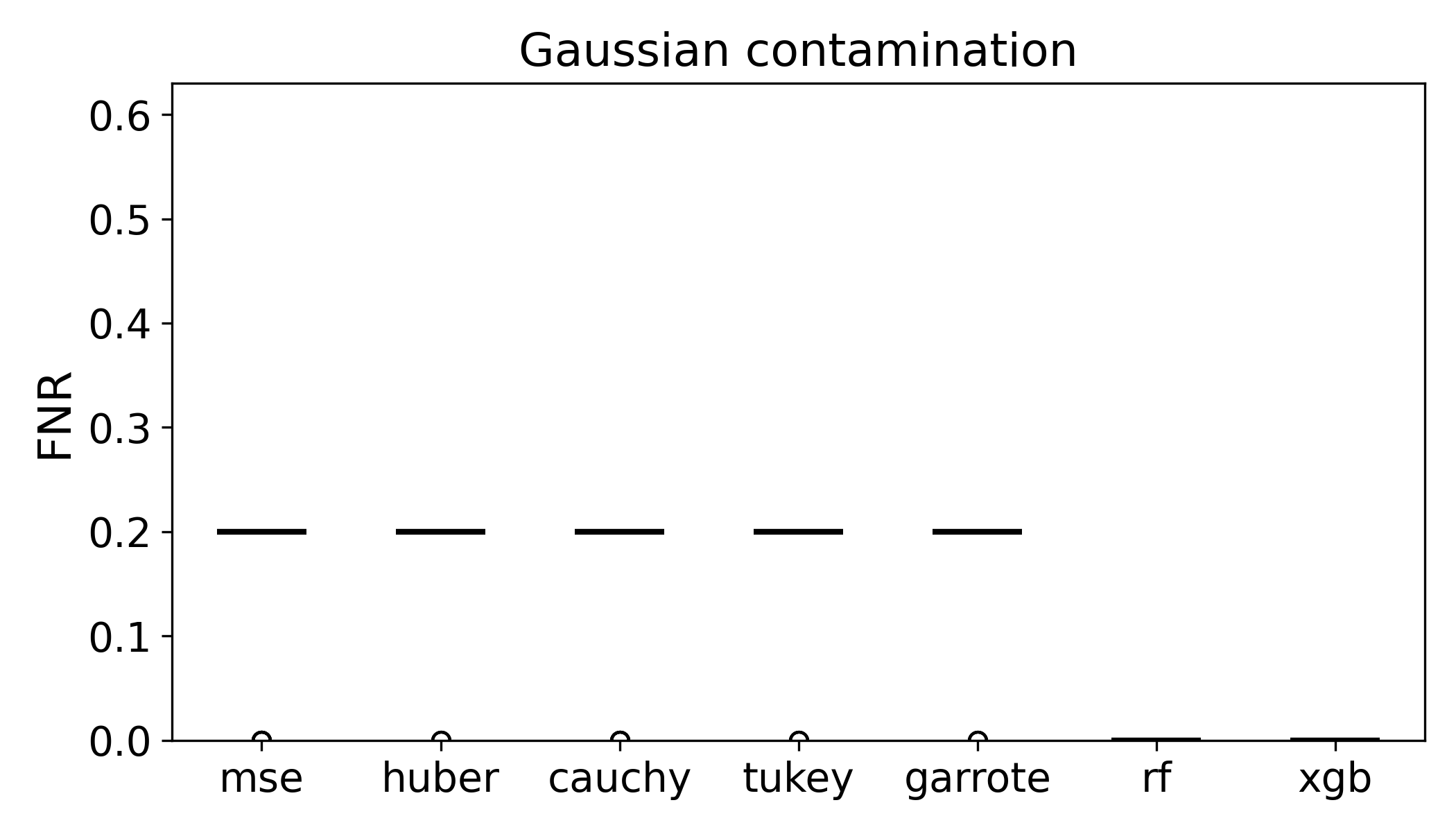}
	\includegraphics[width=0.32\textwidth,height=0.25\textwidth]{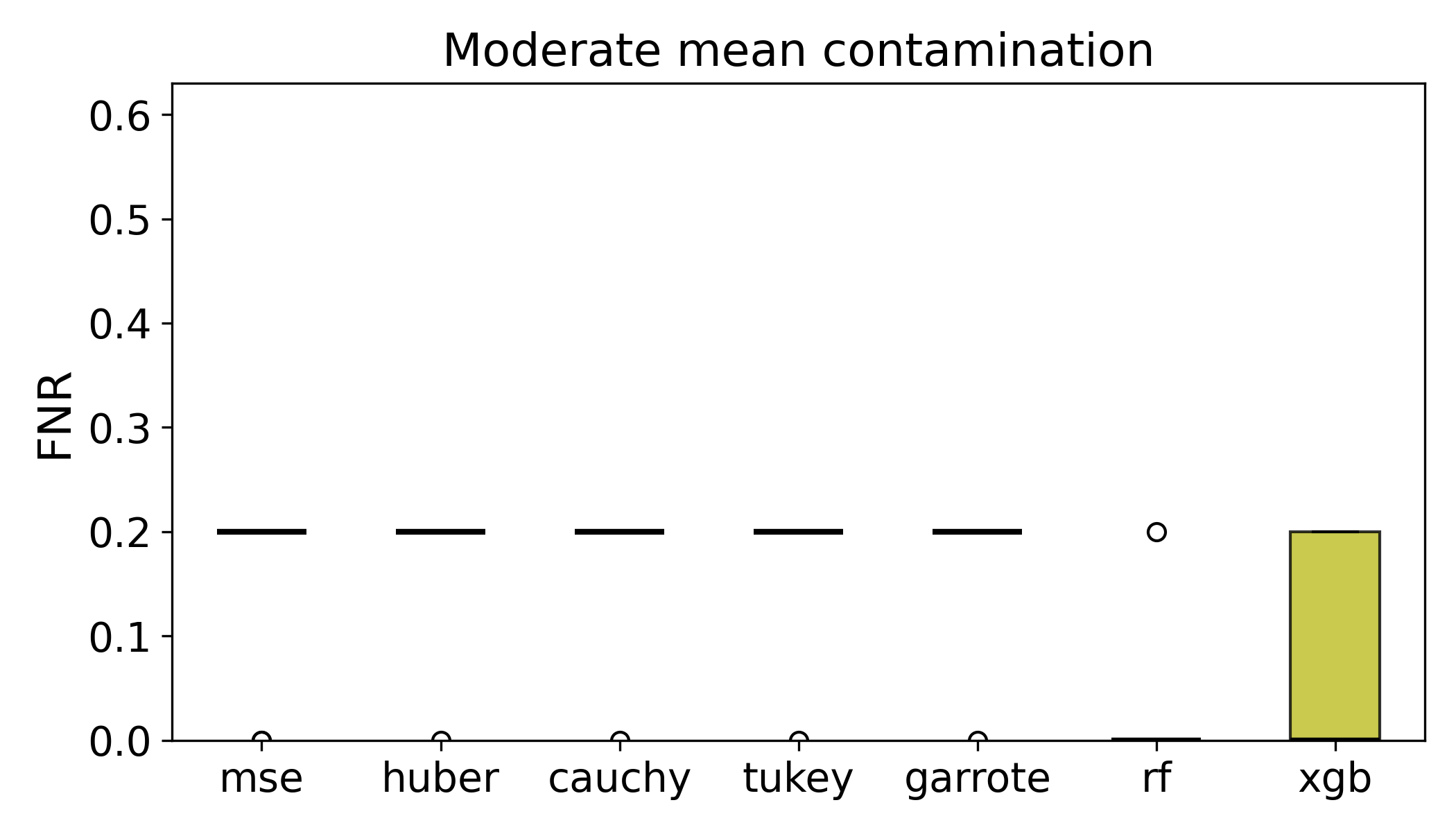}
	\includegraphics[width=0.32\textwidth,height=0.25\textwidth]{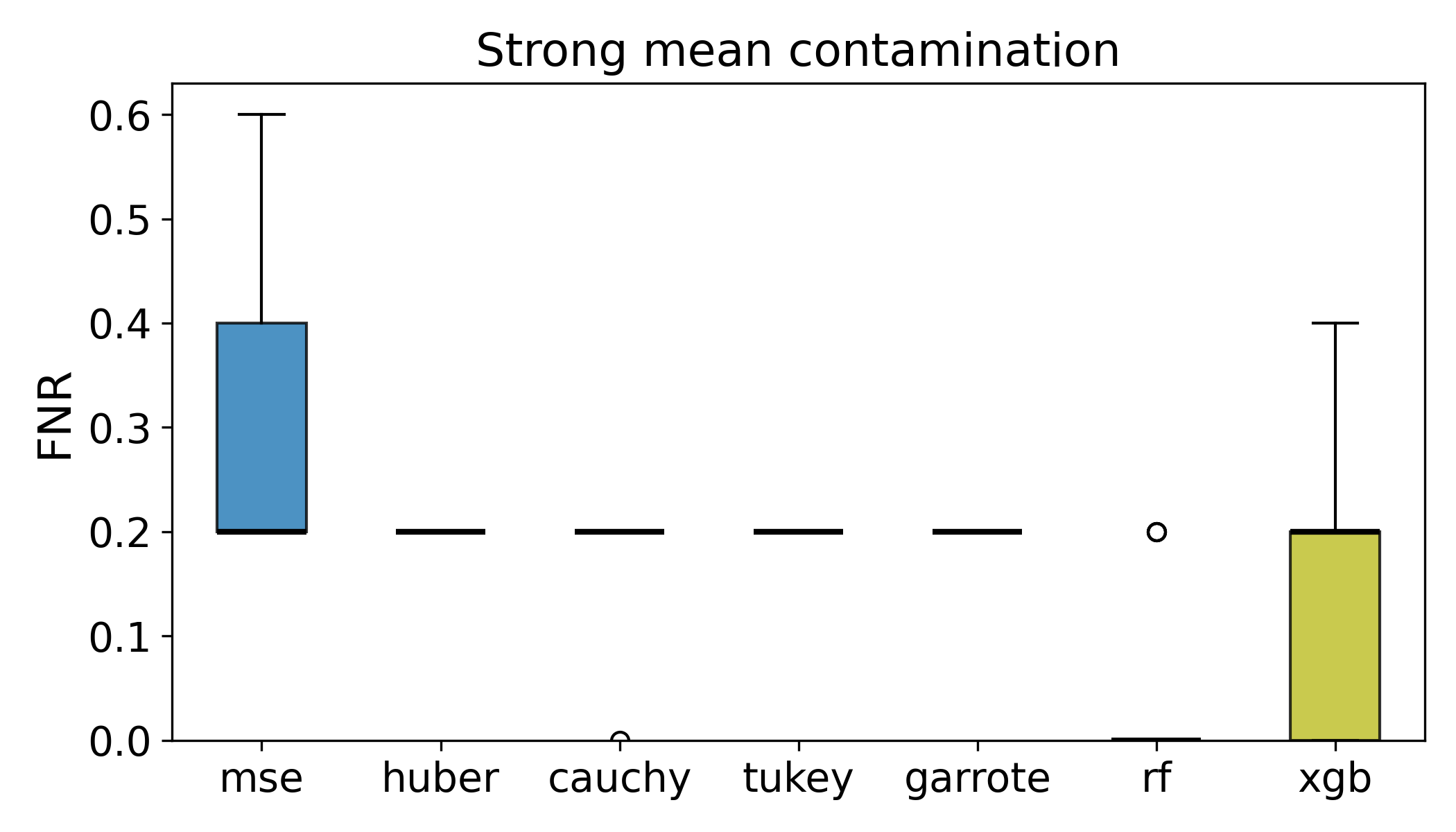}
	
	\caption{Boxplots of the performance metrics results for {\tt model 1} with $10\%$ contamination. First column refers to {\tt gaussian} scenario; second column to {\tt moderate mean} scenario; third column to {\tt strong mean} scenario. The first row shows a realization of the true and contaminated training response $\mathbf{y}$ for all the scenarios.}
	\label{fig_boxplot_model1_mean}
	
\end{figure*}

% boxplots model 2 mixture mean
\begin{figure*}[t]
	\centering
	
	% --- Row 1 ---
	\includegraphics[width=0.32\textwidth,height=0.25\textwidth]{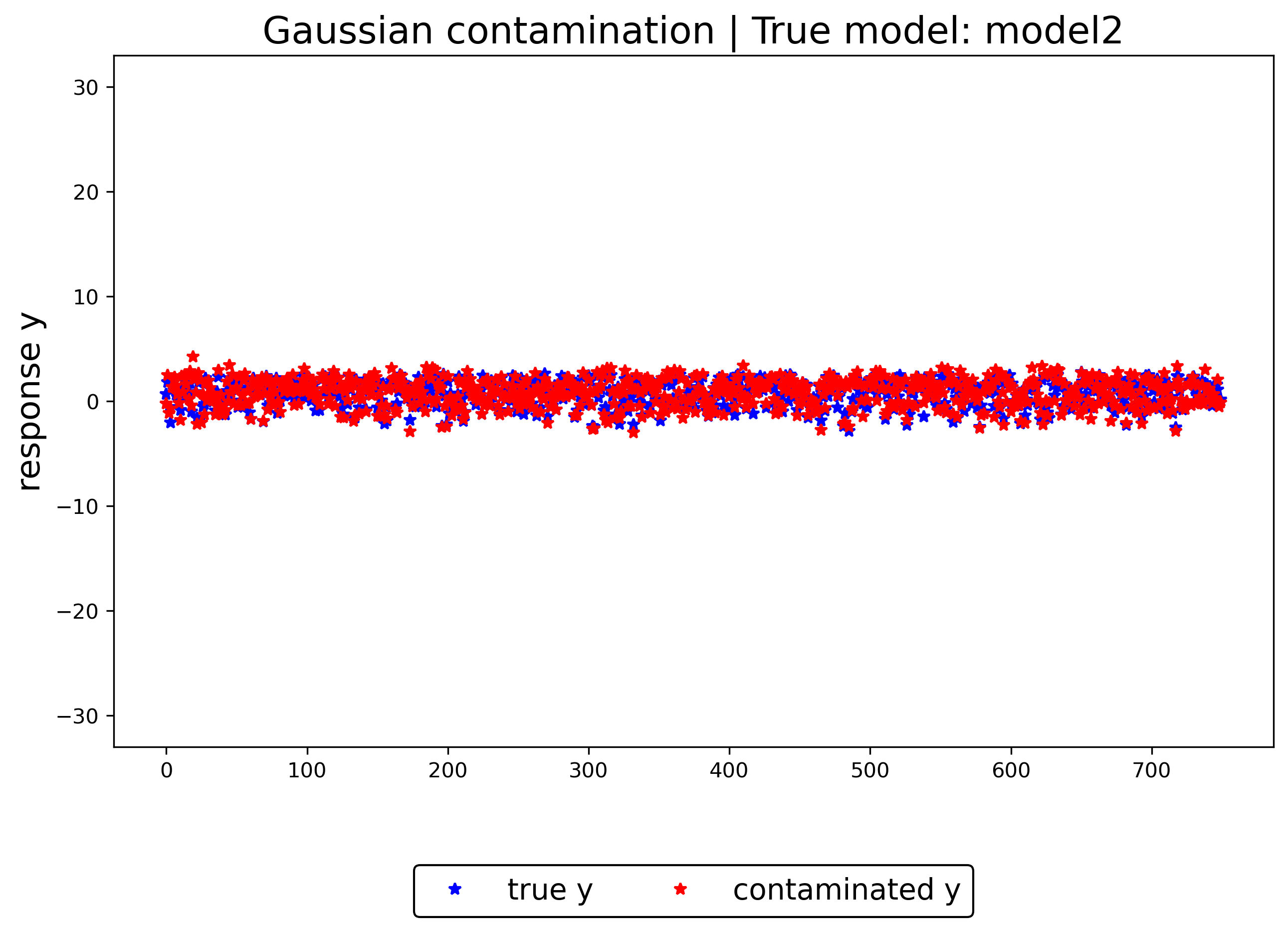}
	\includegraphics[width=0.32\textwidth,height=0.25\textwidth]{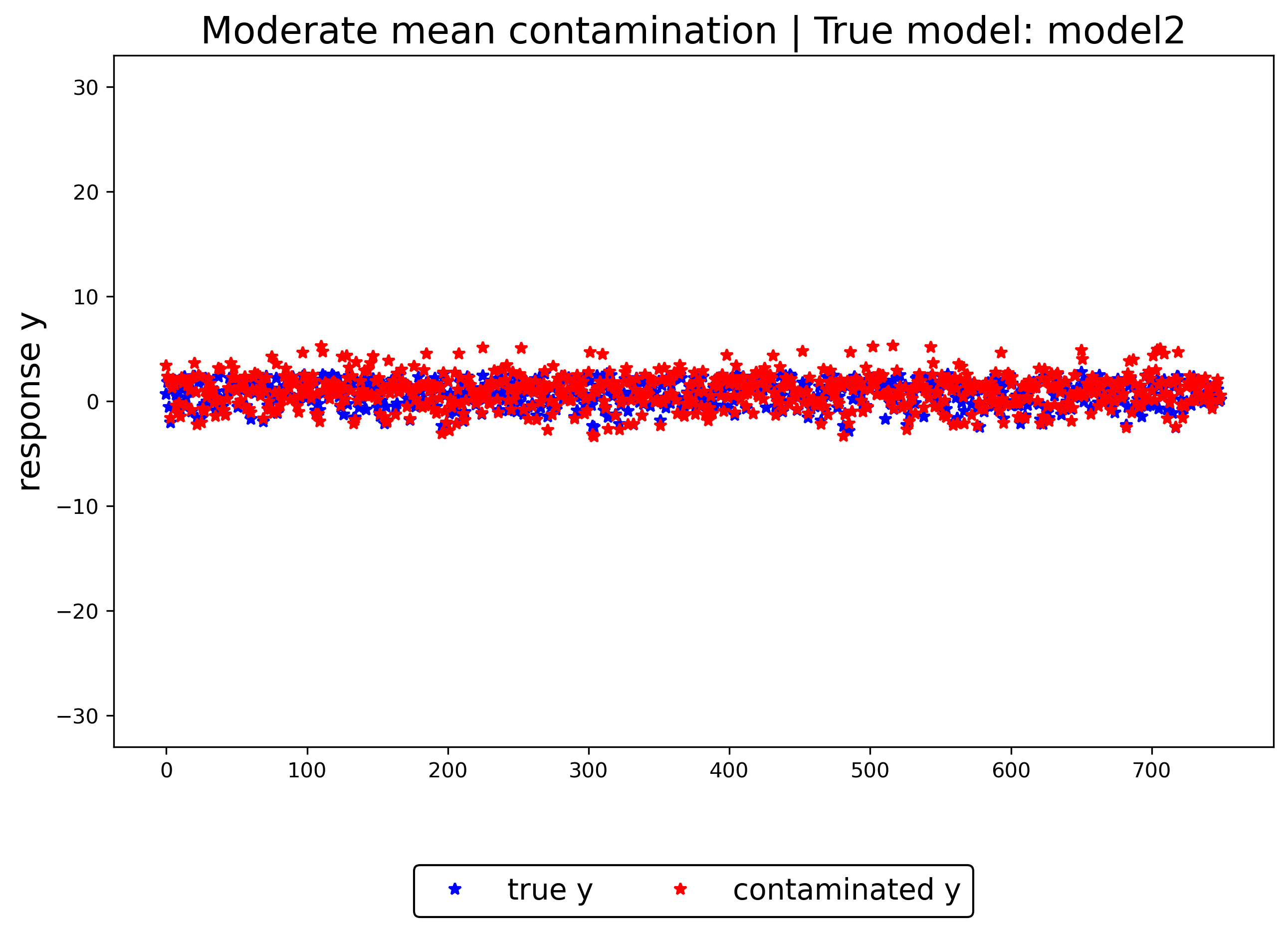}
	\includegraphics[width=0.32\textwidth,height=0.25\textwidth]{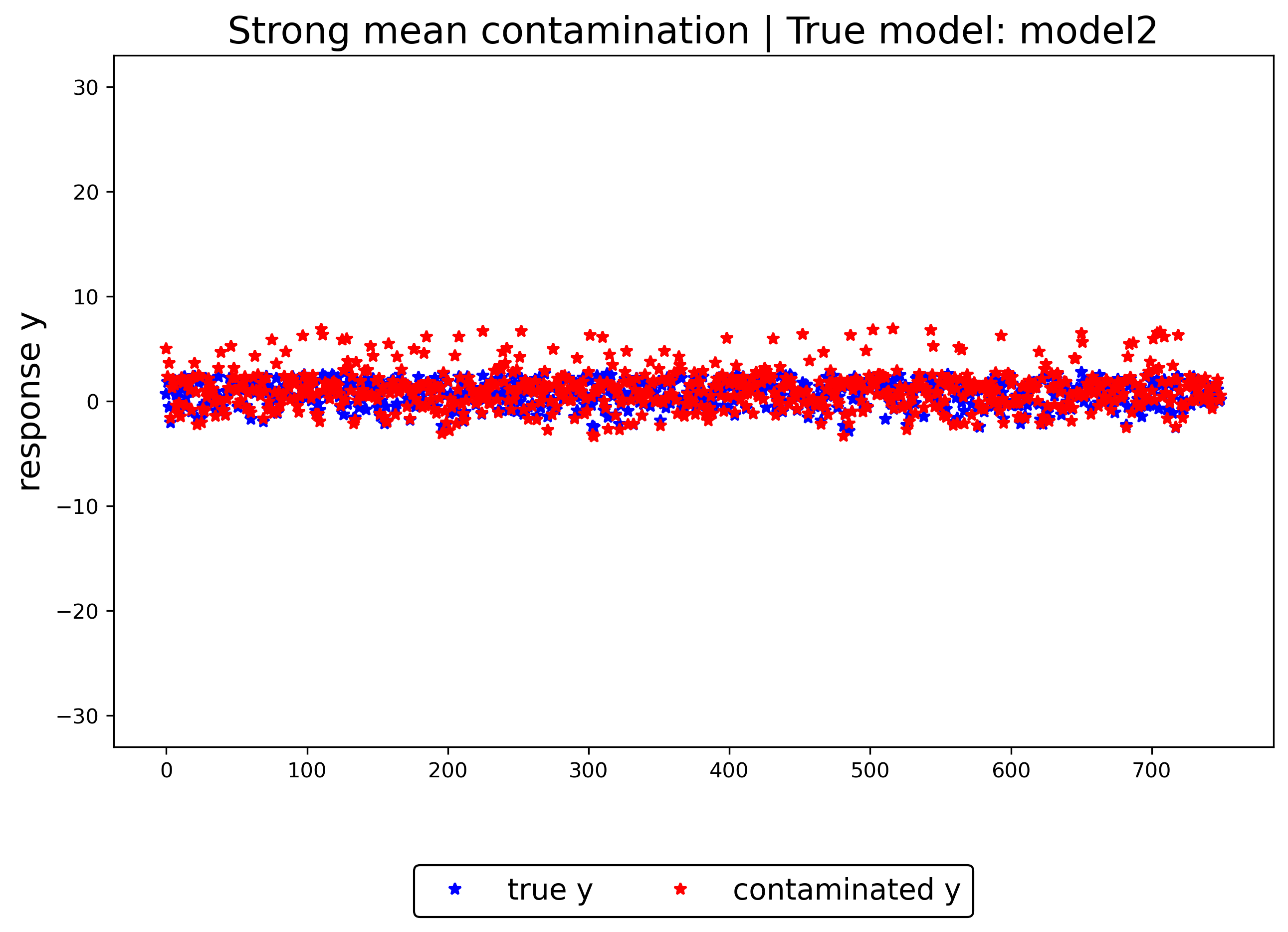}

	% --- Row 2 ---
	\includegraphics[width=0.32\textwidth,height=0.25\textwidth]{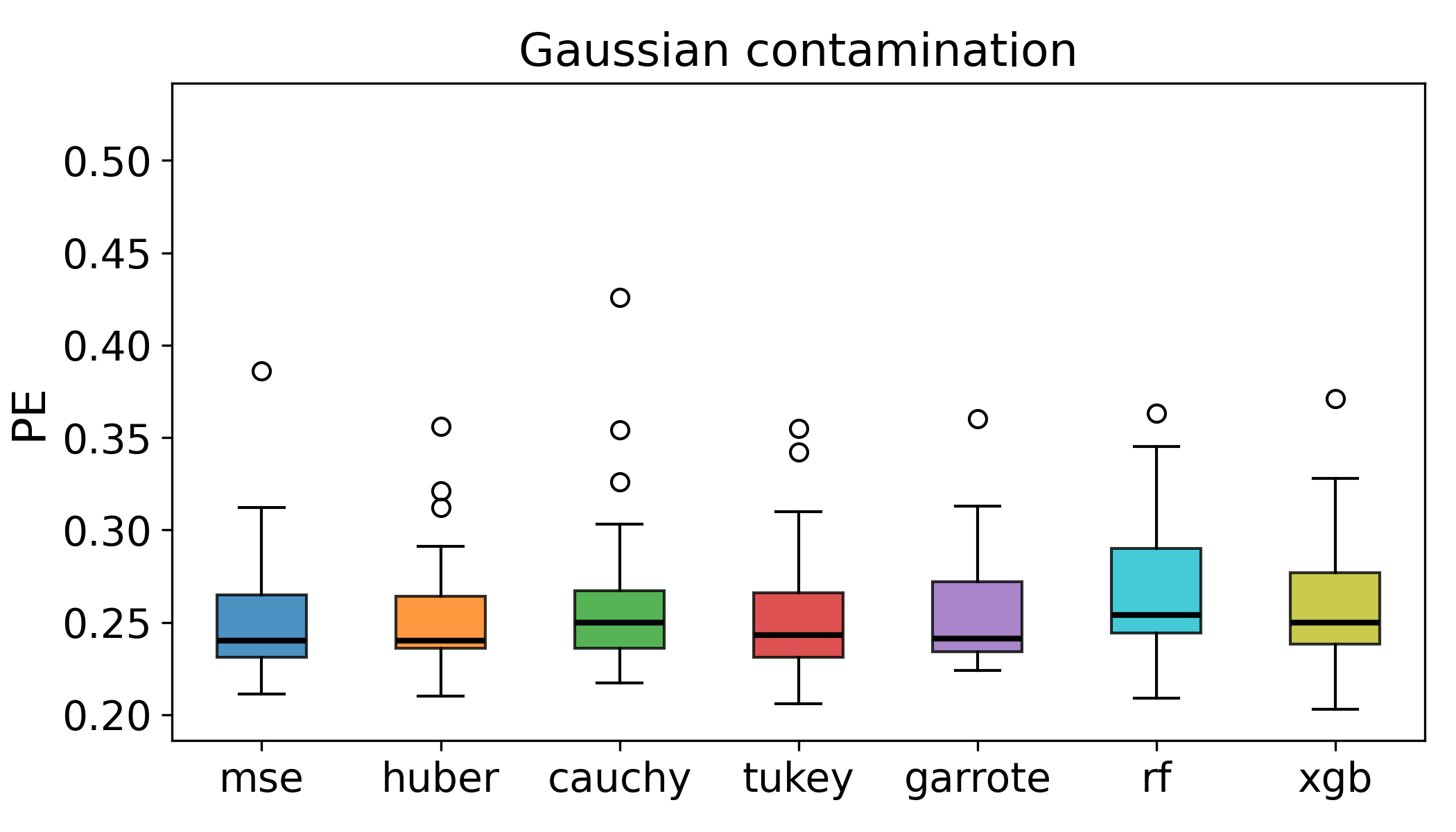}
	\includegraphics[width=0.32\textwidth,height=0.25\textwidth]{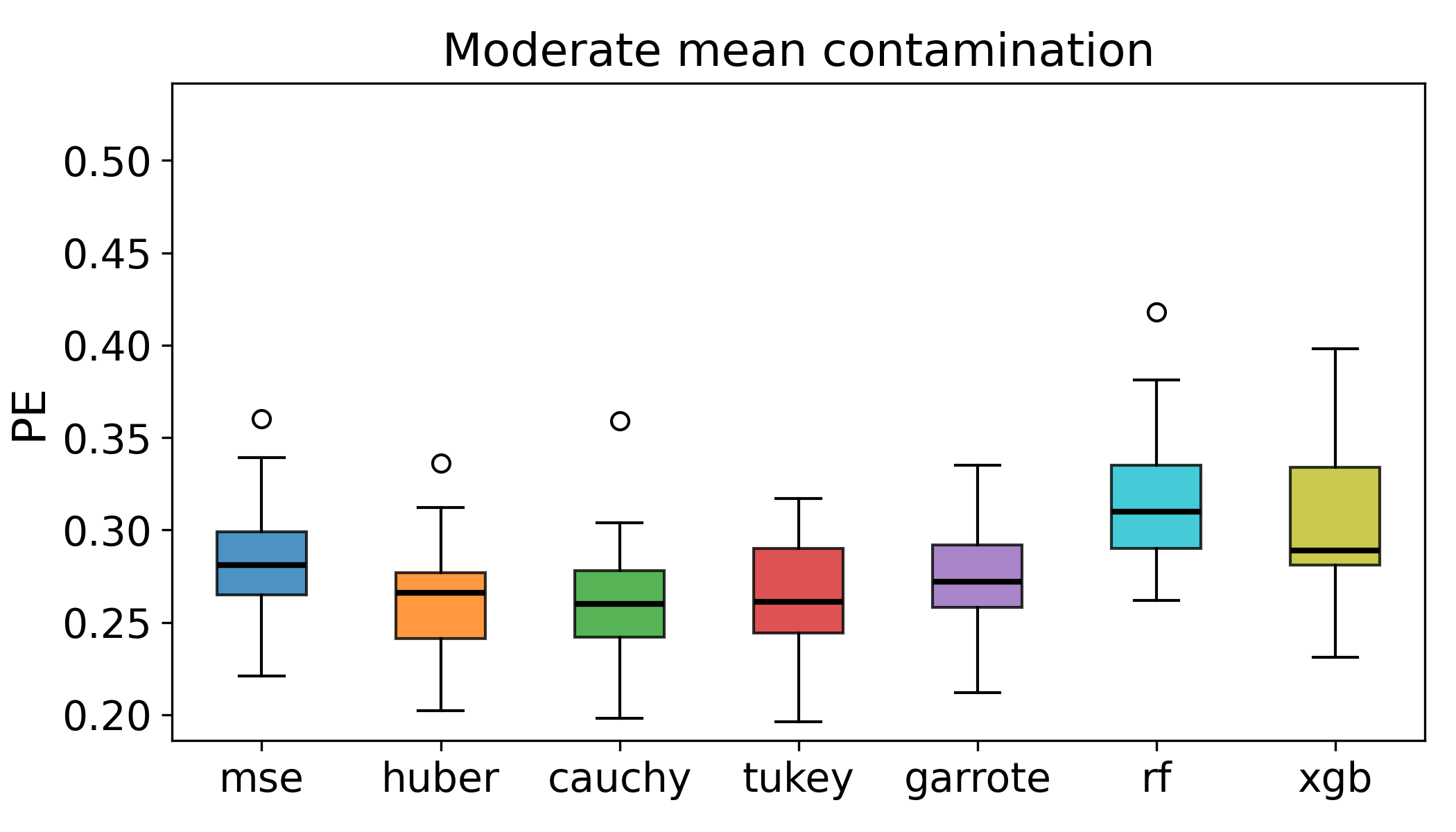}
	\includegraphics[width=0.32\textwidth,height=0.25\textwidth]{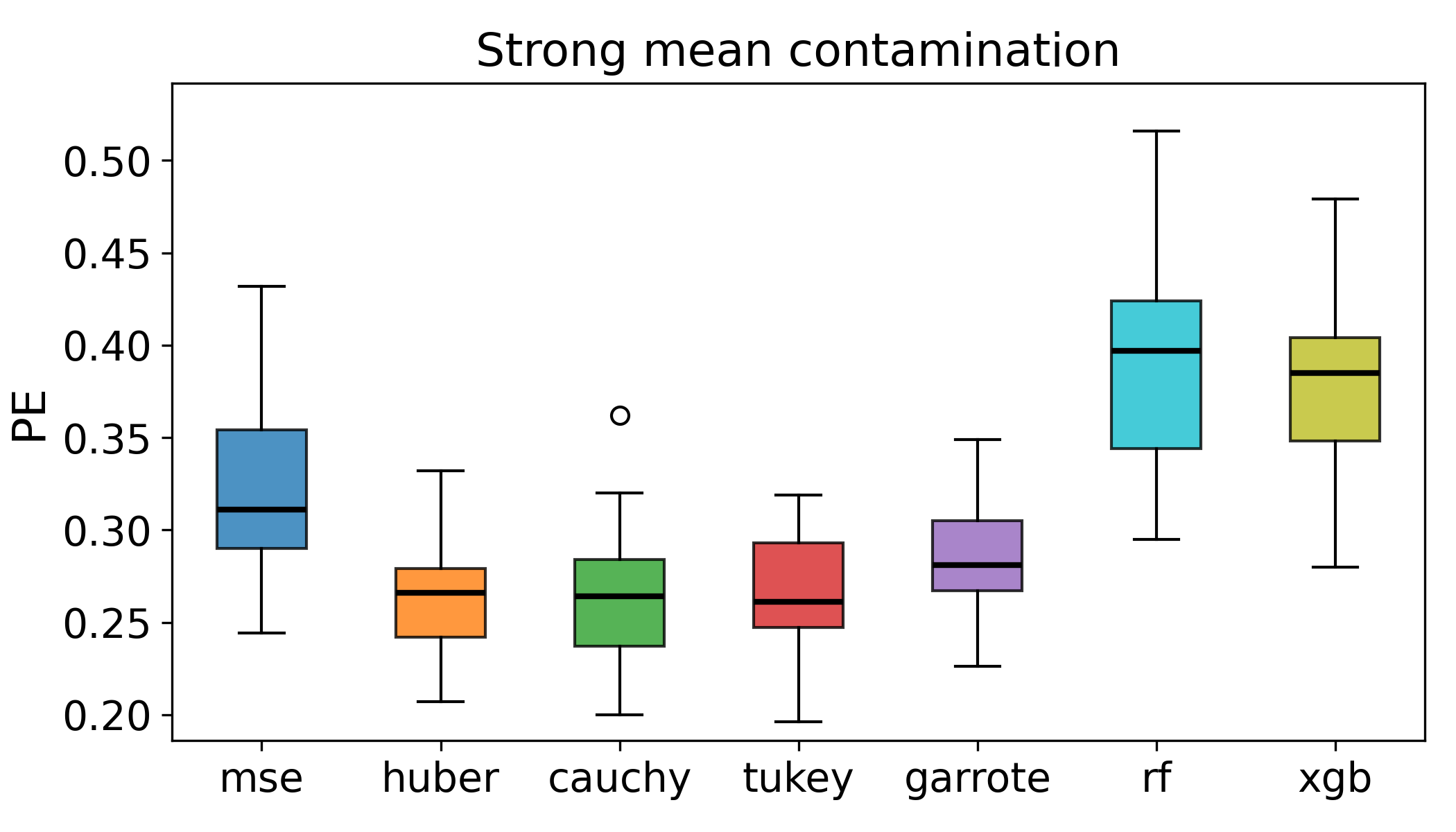}
	
	\vspace{0.2cm}
	
	% --- Row 3 ---
	\includegraphics[width=0.32\textwidth,height=0.25\textwidth]{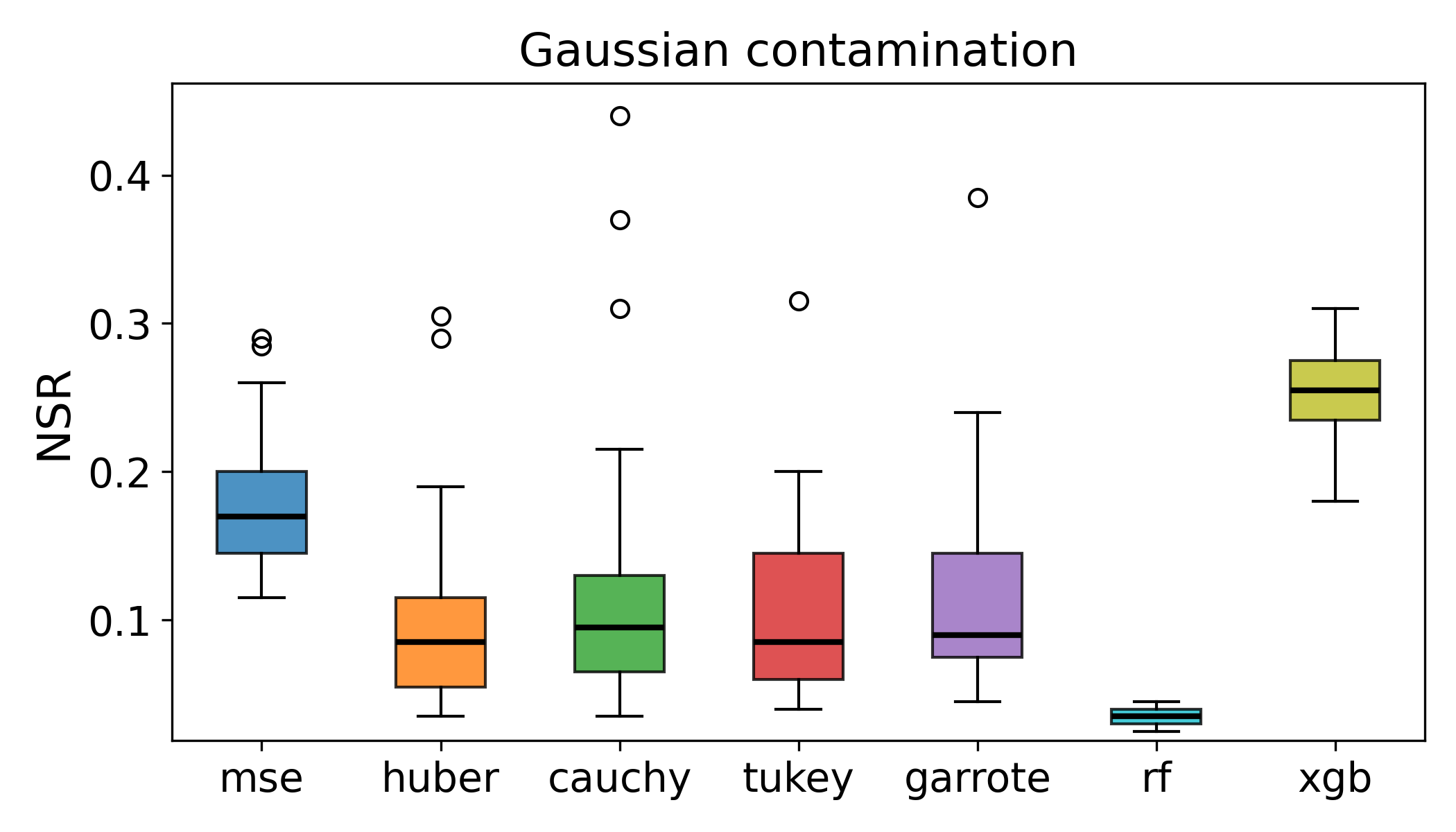}
	\includegraphics[width=0.32\textwidth,height=0.25\textwidth]{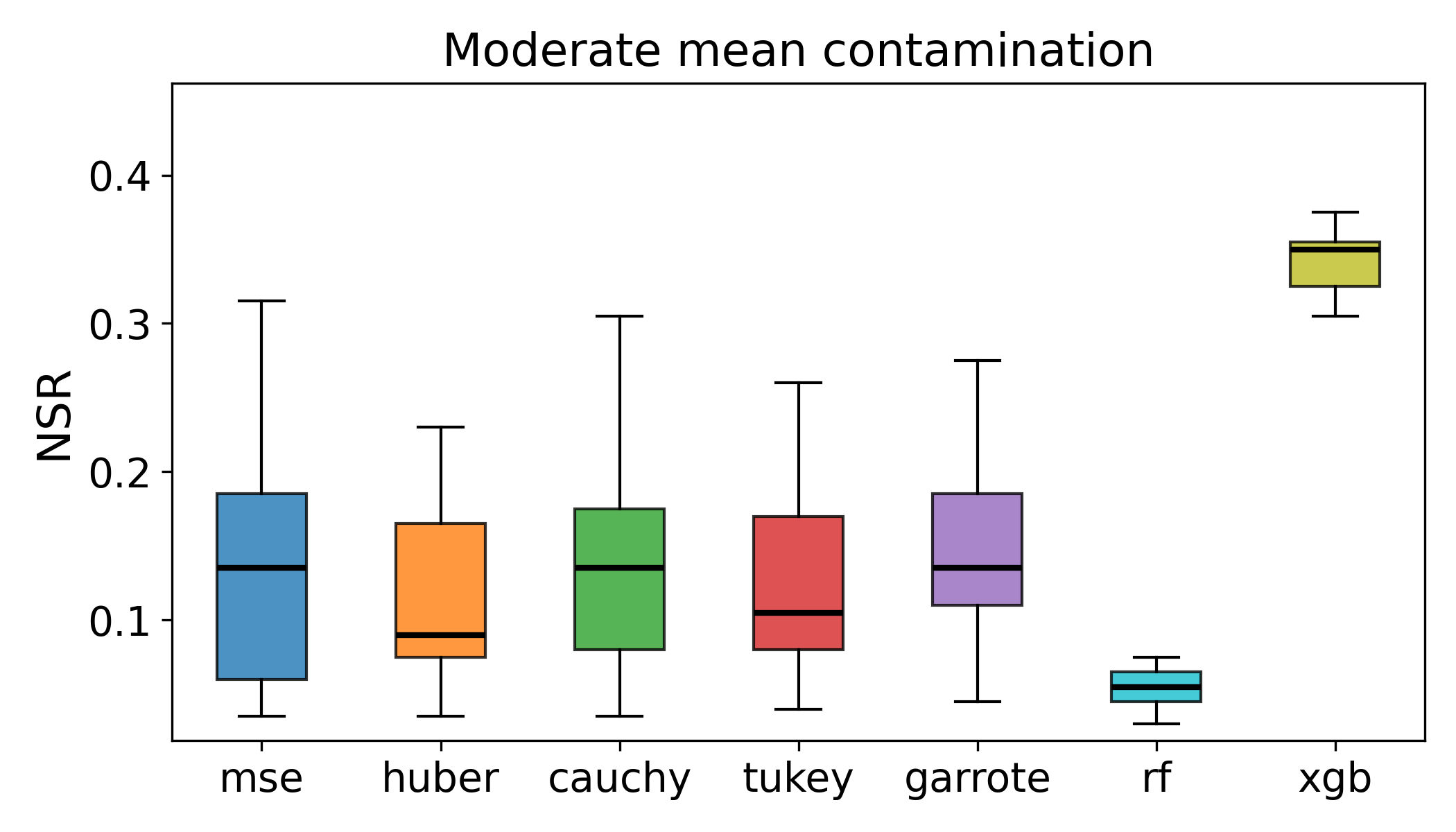}
	\includegraphics[width=0.32\textwidth,height=0.25\textwidth]{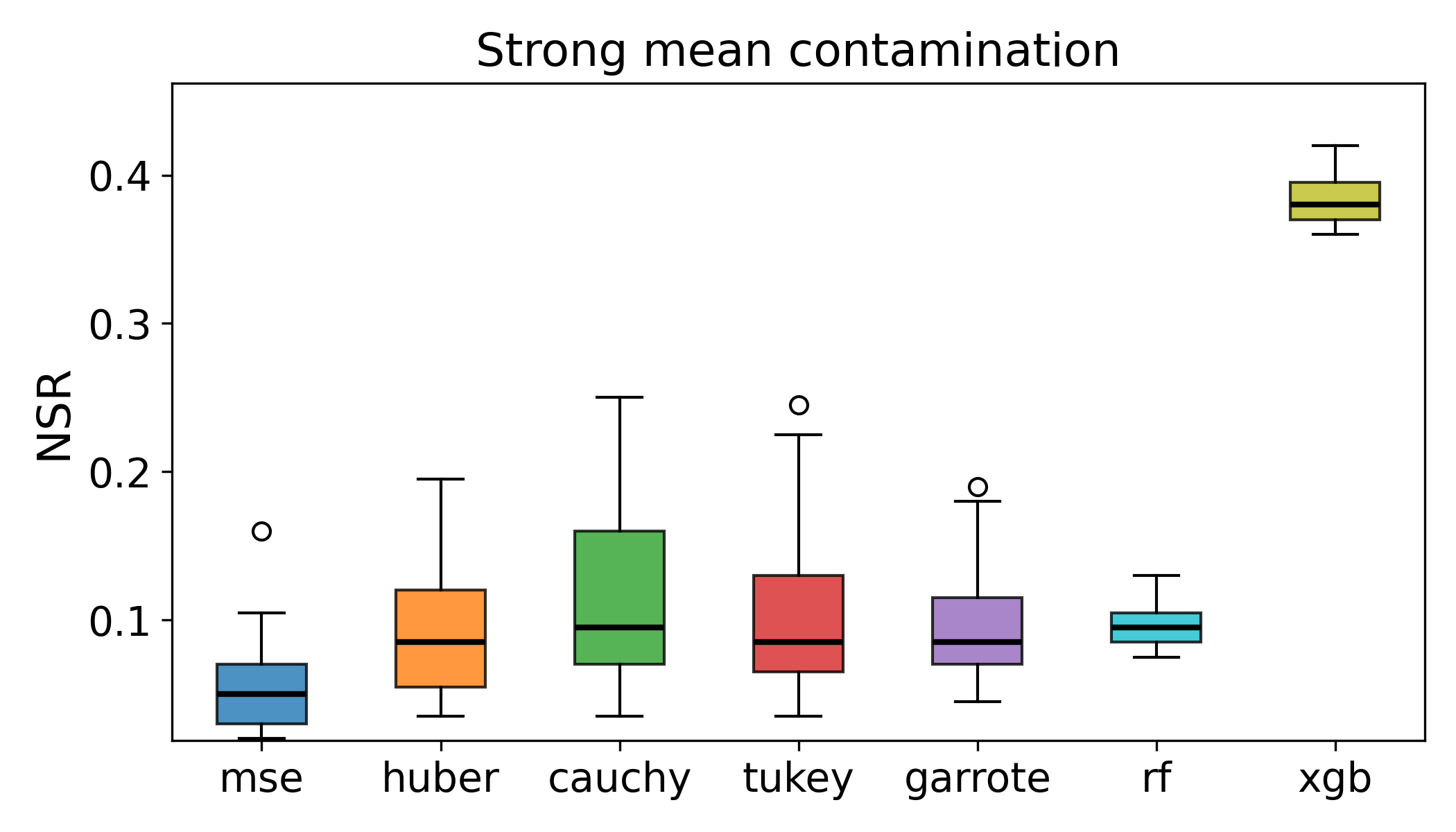}
	
	\vspace{0.2cm}
	
	% --- Row 4 ---
	\includegraphics[width=0.32\textwidth,height=0.25\textwidth]{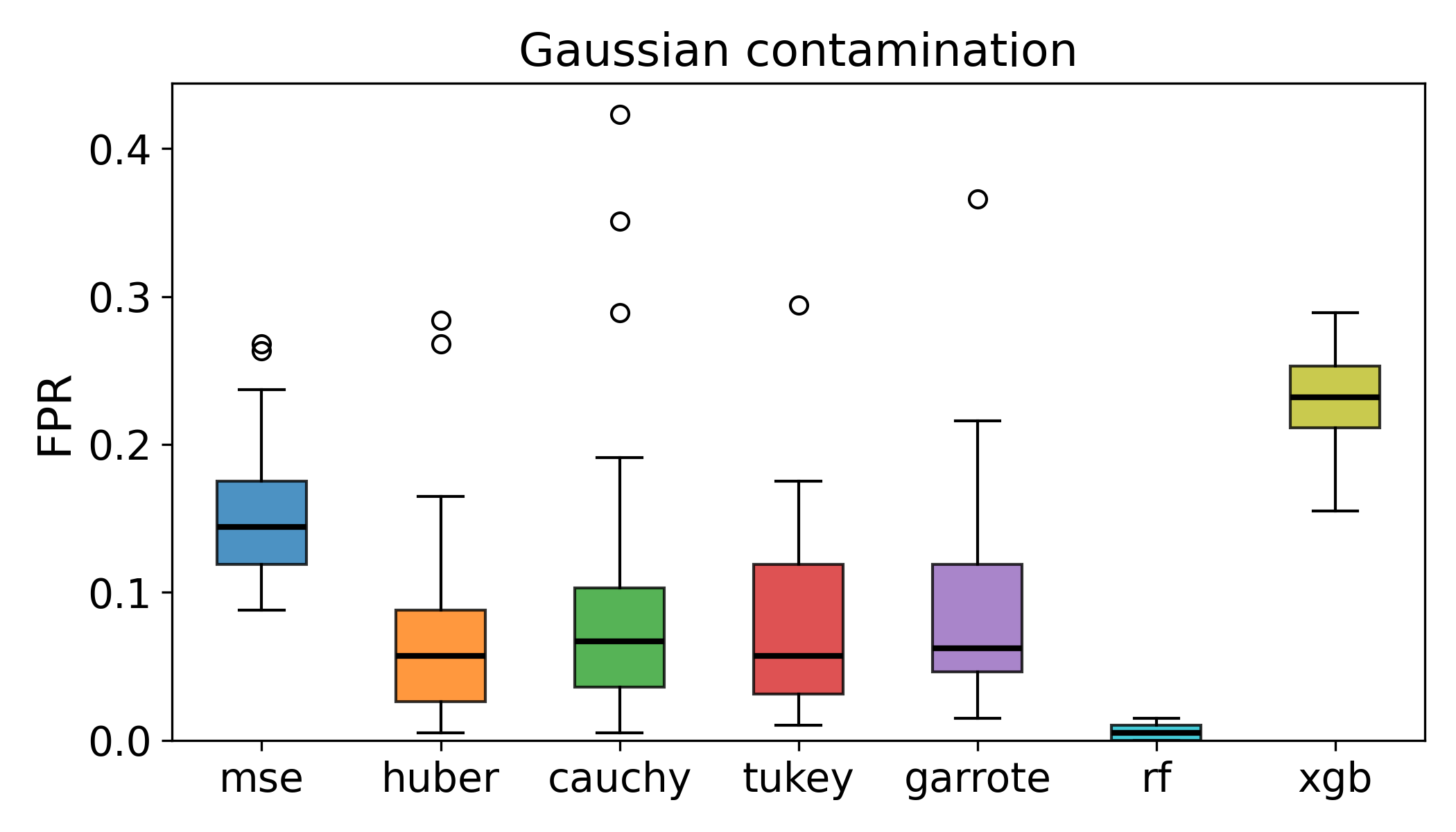}
	\includegraphics[width=0.32\textwidth,height=0.25\textwidth]{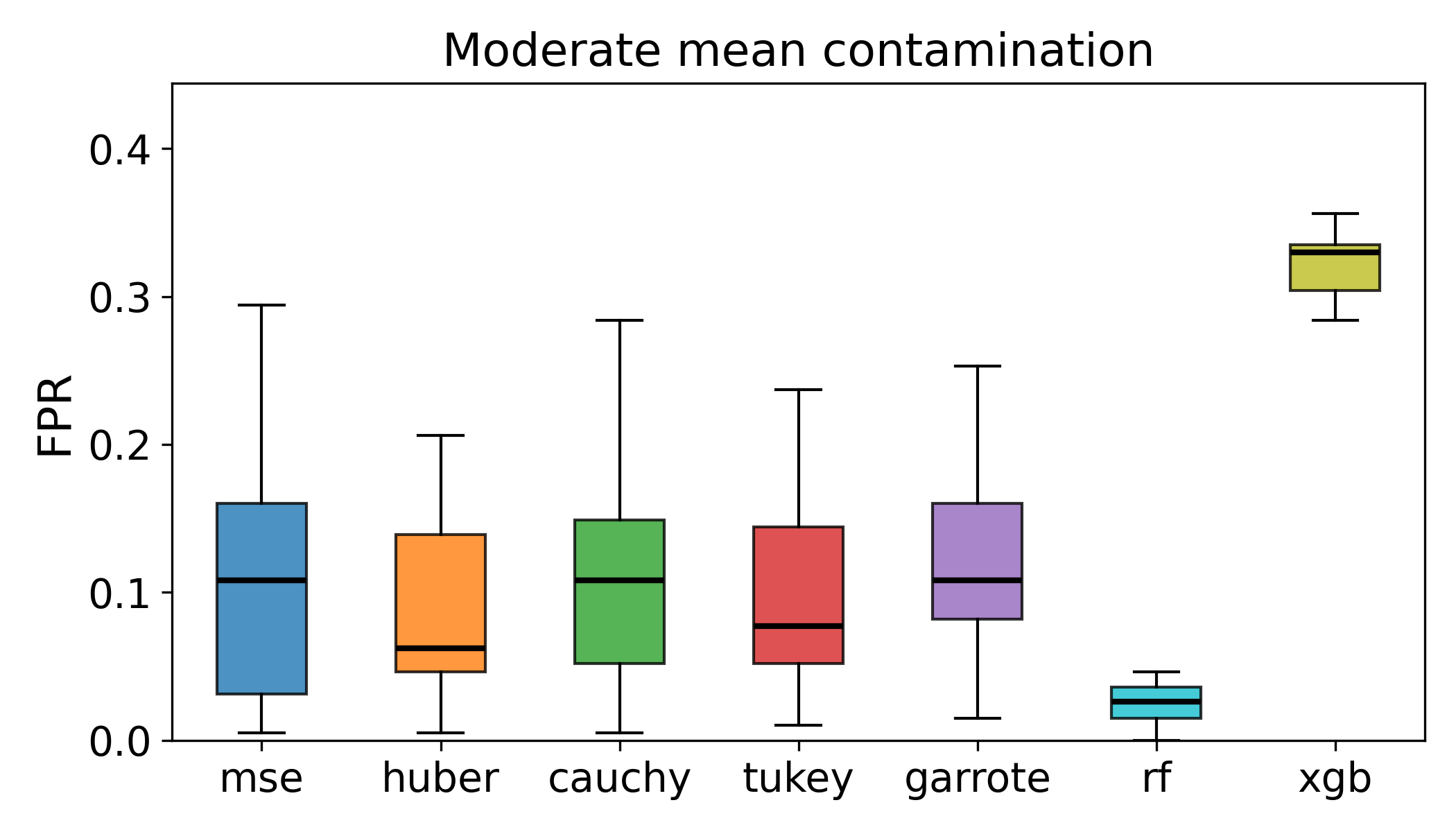}
	\includegraphics[width=0.32\textwidth,height=0.25\textwidth]{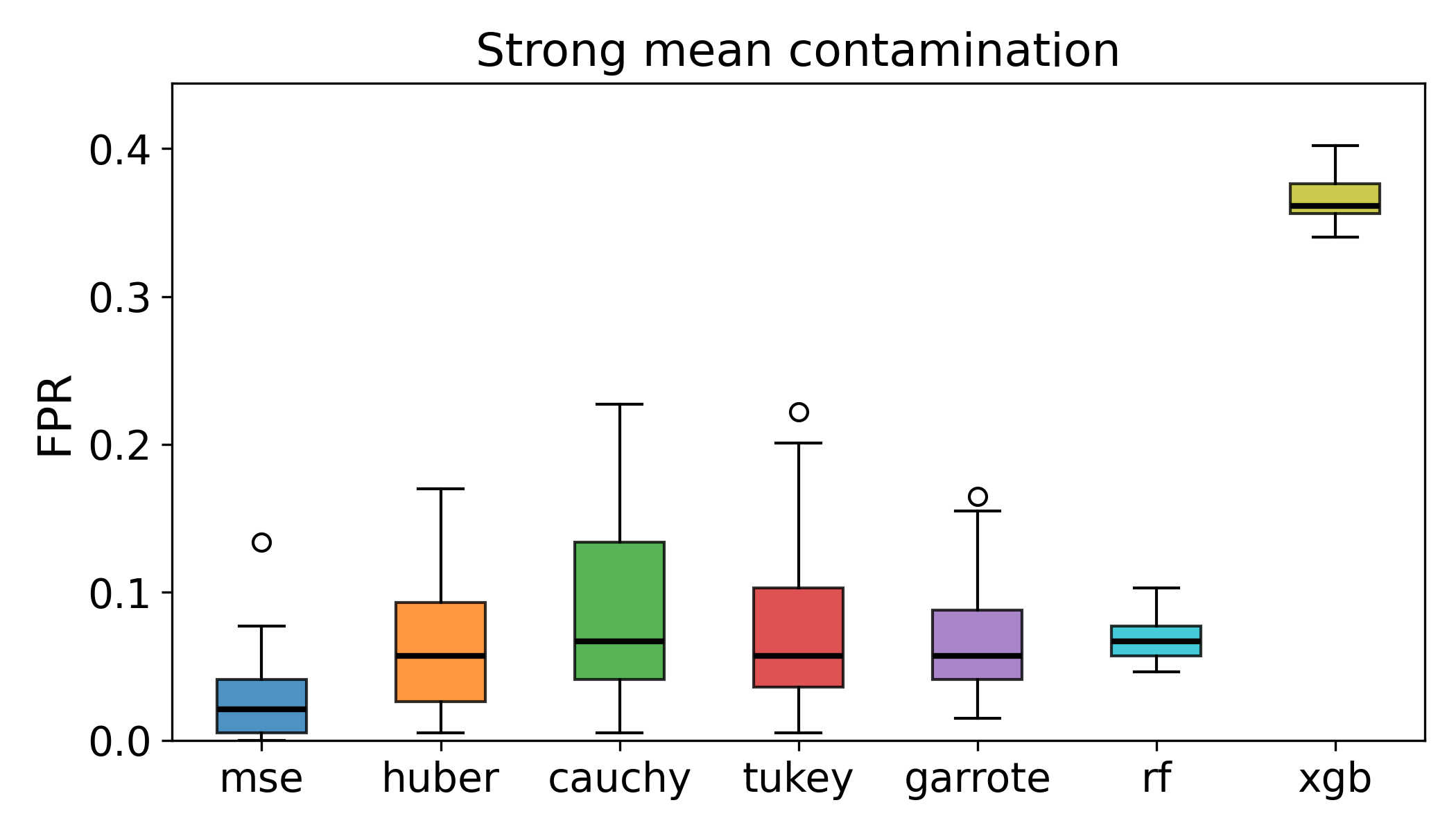}
	
	\vspace{0.2cm}
	
	% --- Row 5 ---
	\includegraphics[width=0.32\textwidth,height=0.25\textwidth]{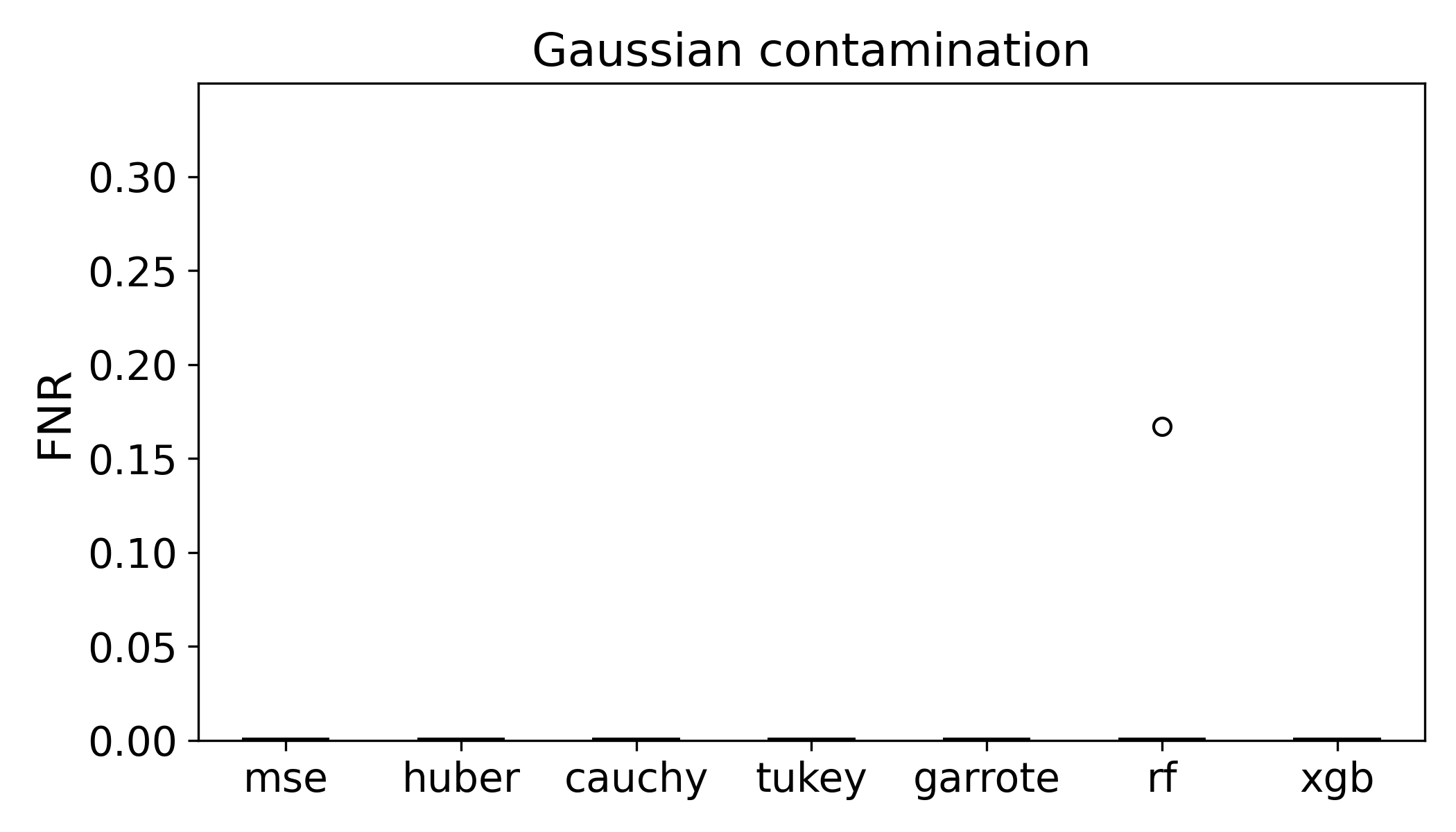}
	\includegraphics[width=0.32\textwidth,height=0.25\textwidth]{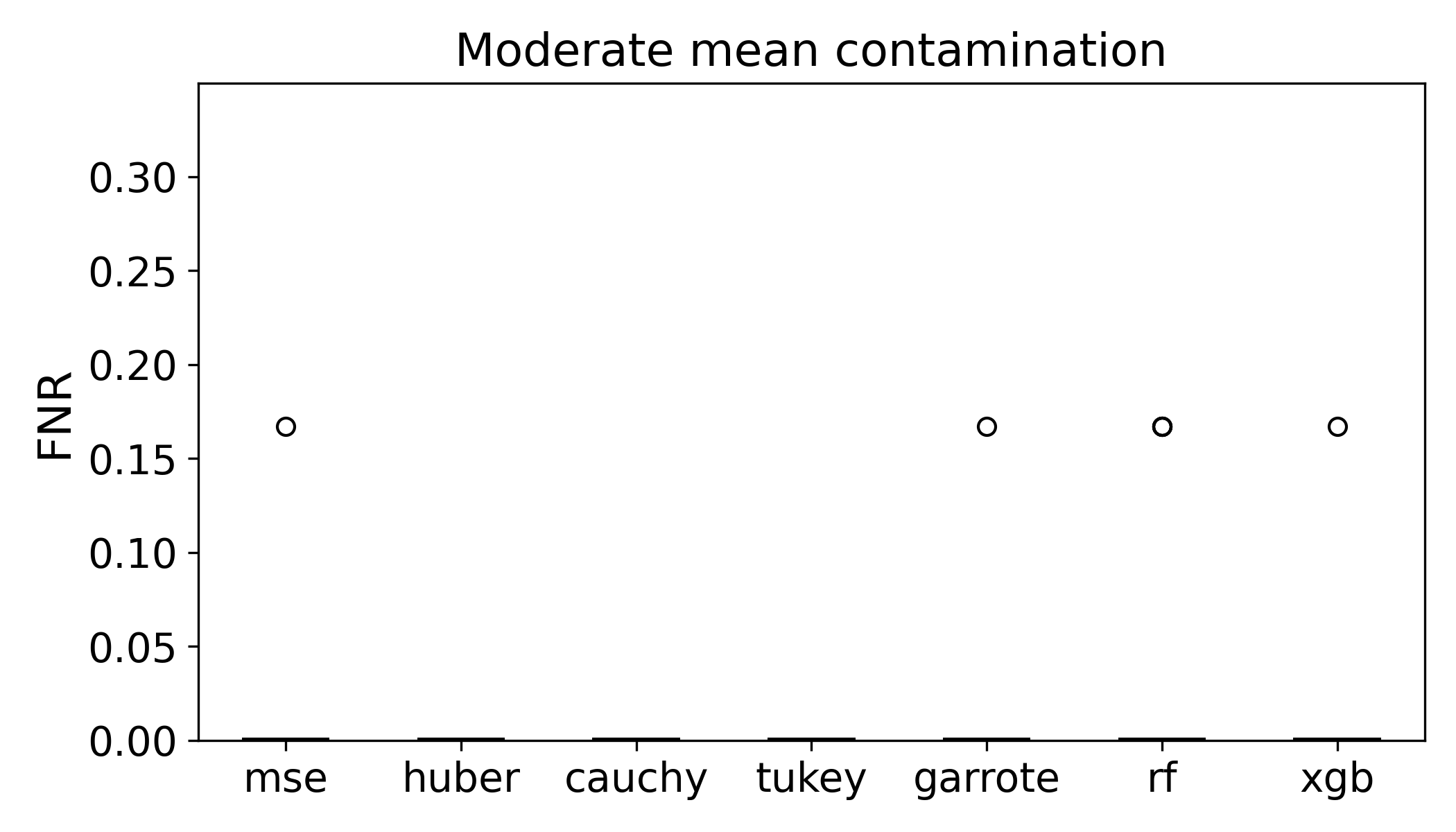}
	\includegraphics[width=0.32\textwidth,height=0.25\textwidth]{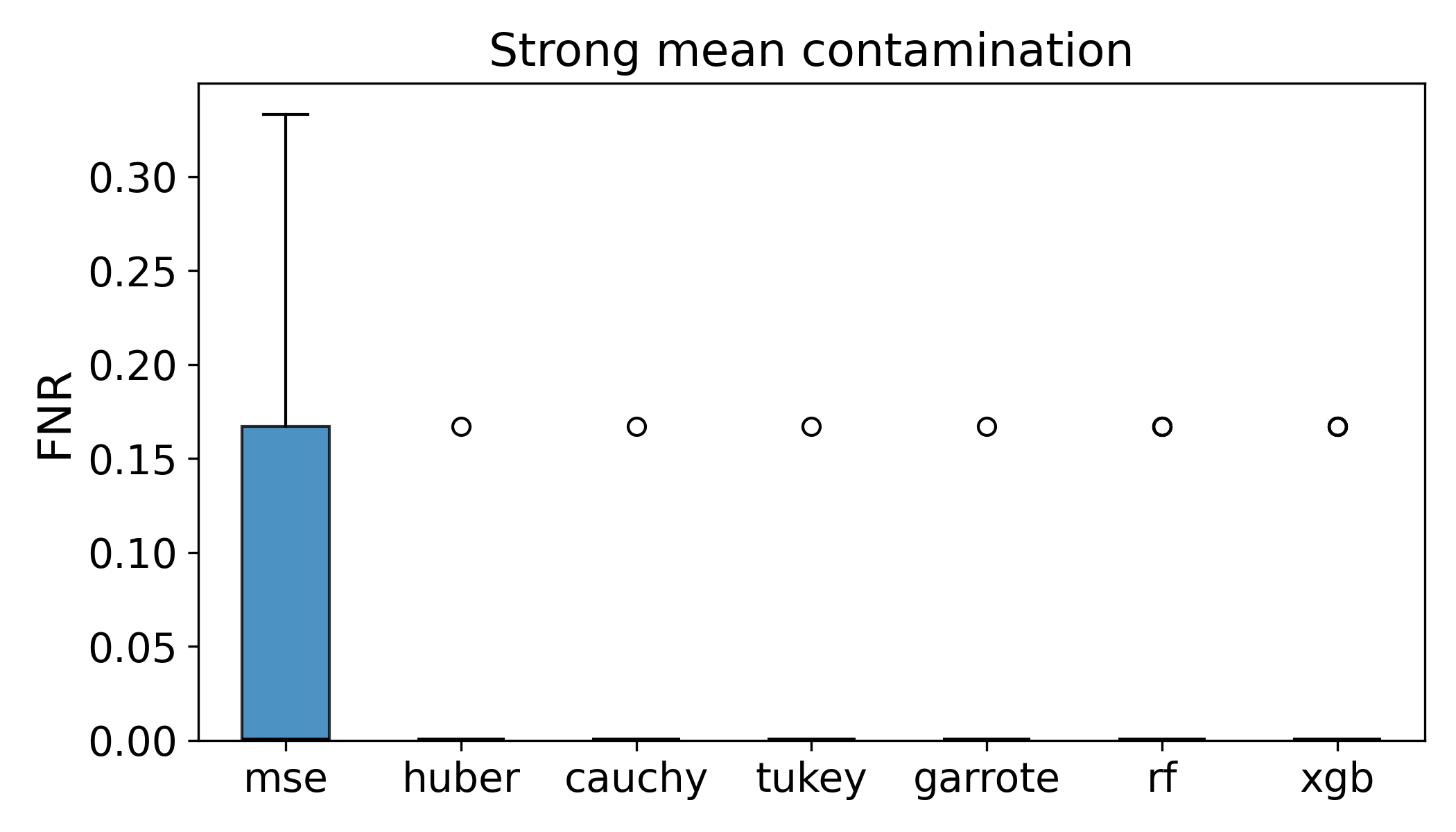}
	
	\caption{Boxplots of the performance metrics results for {\tt model 2} with $10\%$ contamination. First column refers to {\tt gaussian} scenario; second column to {\tt moderate mean} scenario; third column to {\tt strong mean} scenario. The first row shows a realization of the true and contaminated training response $\mathbf{y}$ for all the scenarios.}
	\label{fig_boxplot_model2_mean}
	
\end{figure*}

% boxplots model 3 mixture mean
\begin{figure*}[t]
	\centering
	
	% --- Row 1 ---
	\includegraphics[width=0.32\textwidth,height=0.25\textwidth]{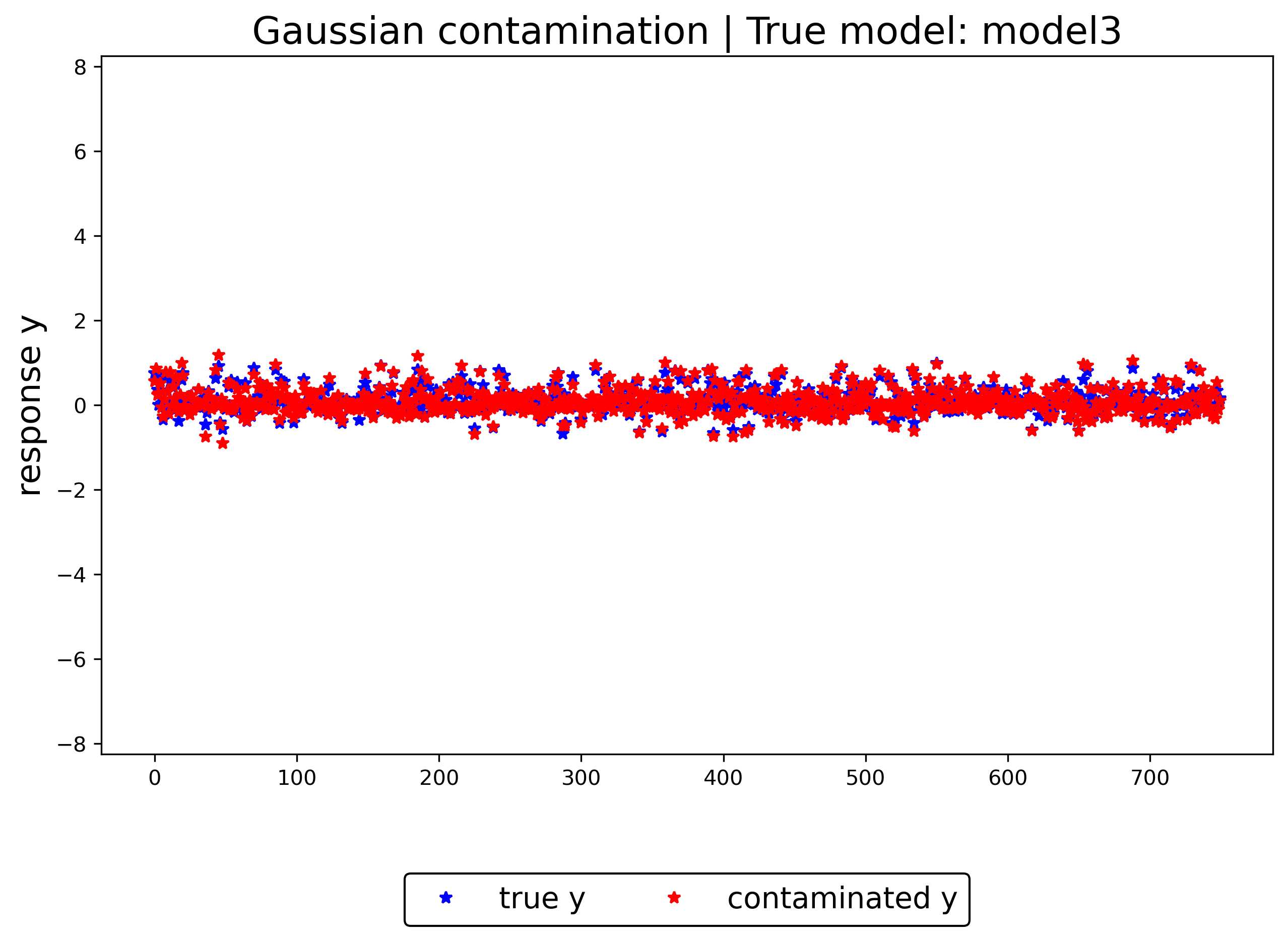}
	\includegraphics[width=0.32\textwidth,height=0.25\textwidth]{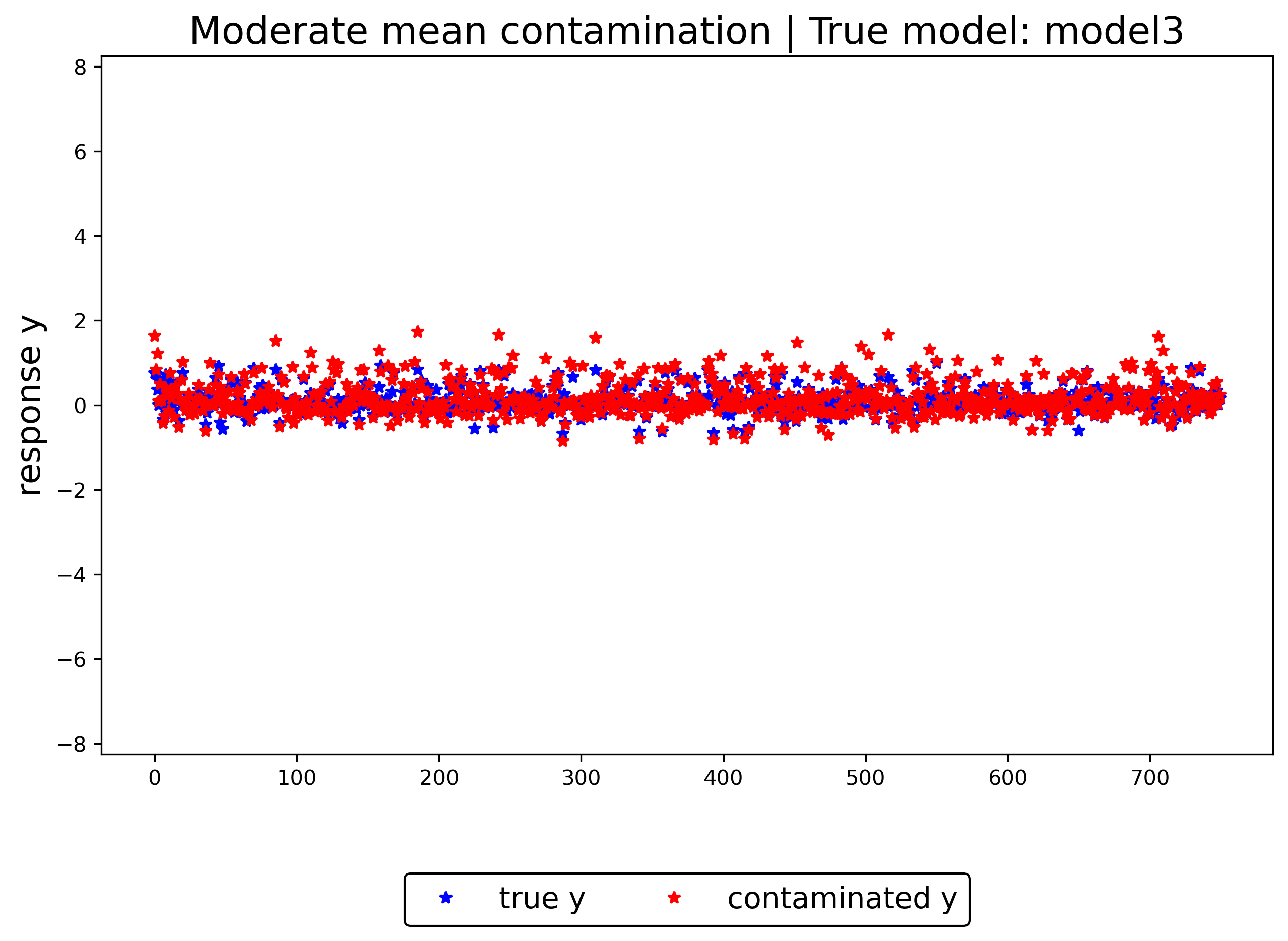}
	\includegraphics[width=0.32\textwidth,height=0.25\textwidth]{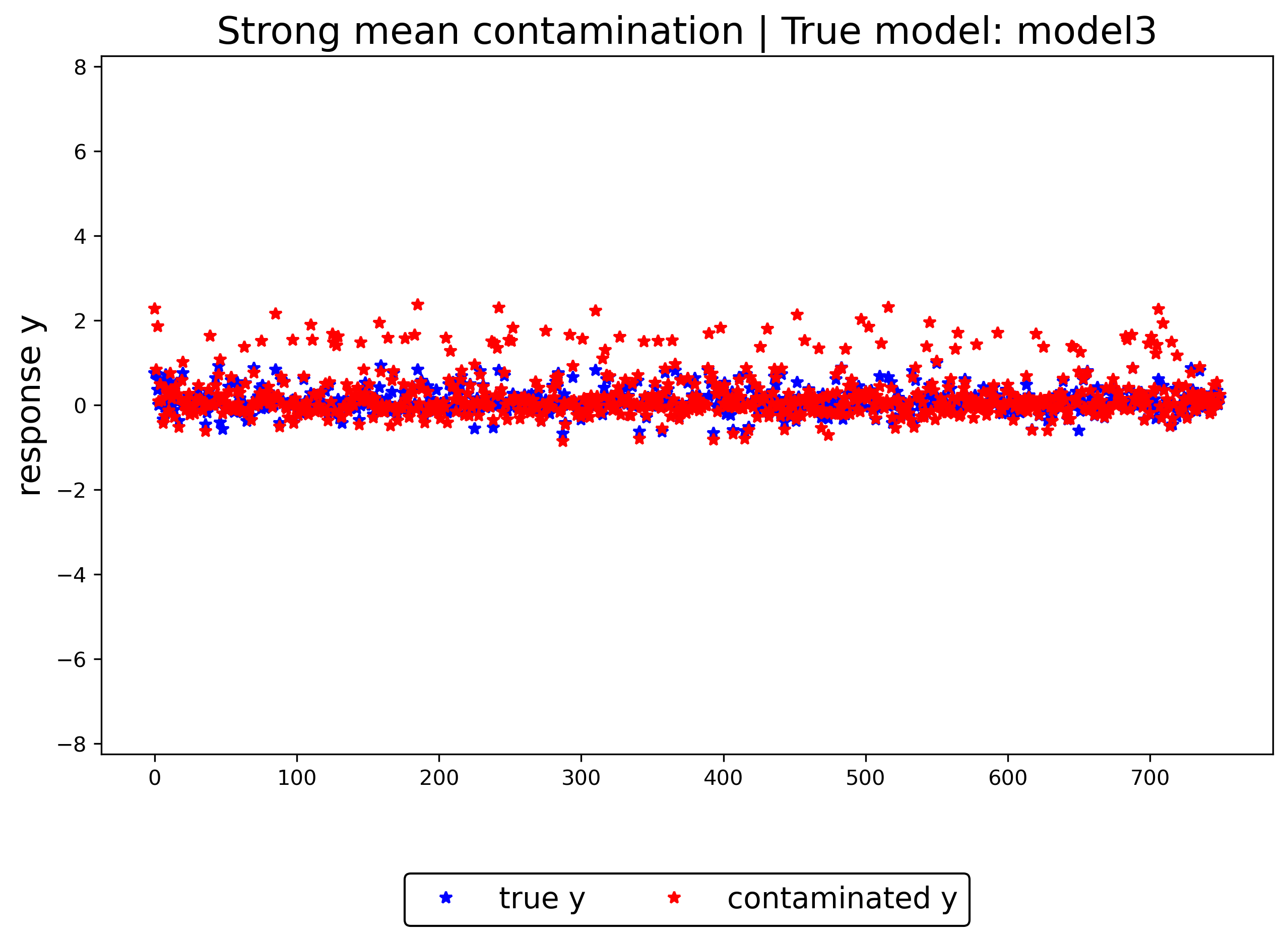}

	% --- Row 2 ---
	\includegraphics[width=0.32\textwidth,height=0.25\textwidth]{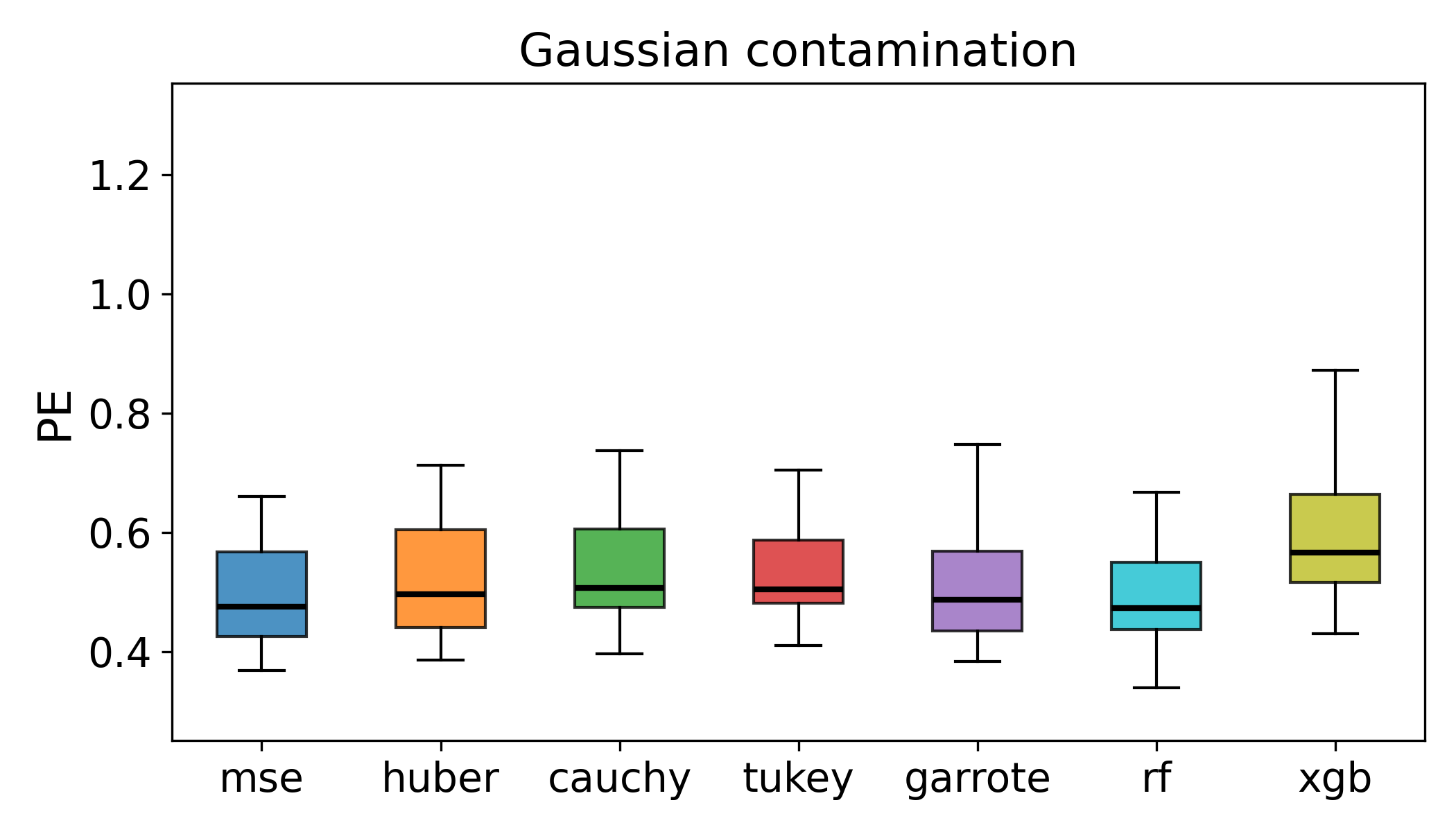}
	\includegraphics[width=0.32\textwidth,height=0.25\textwidth]{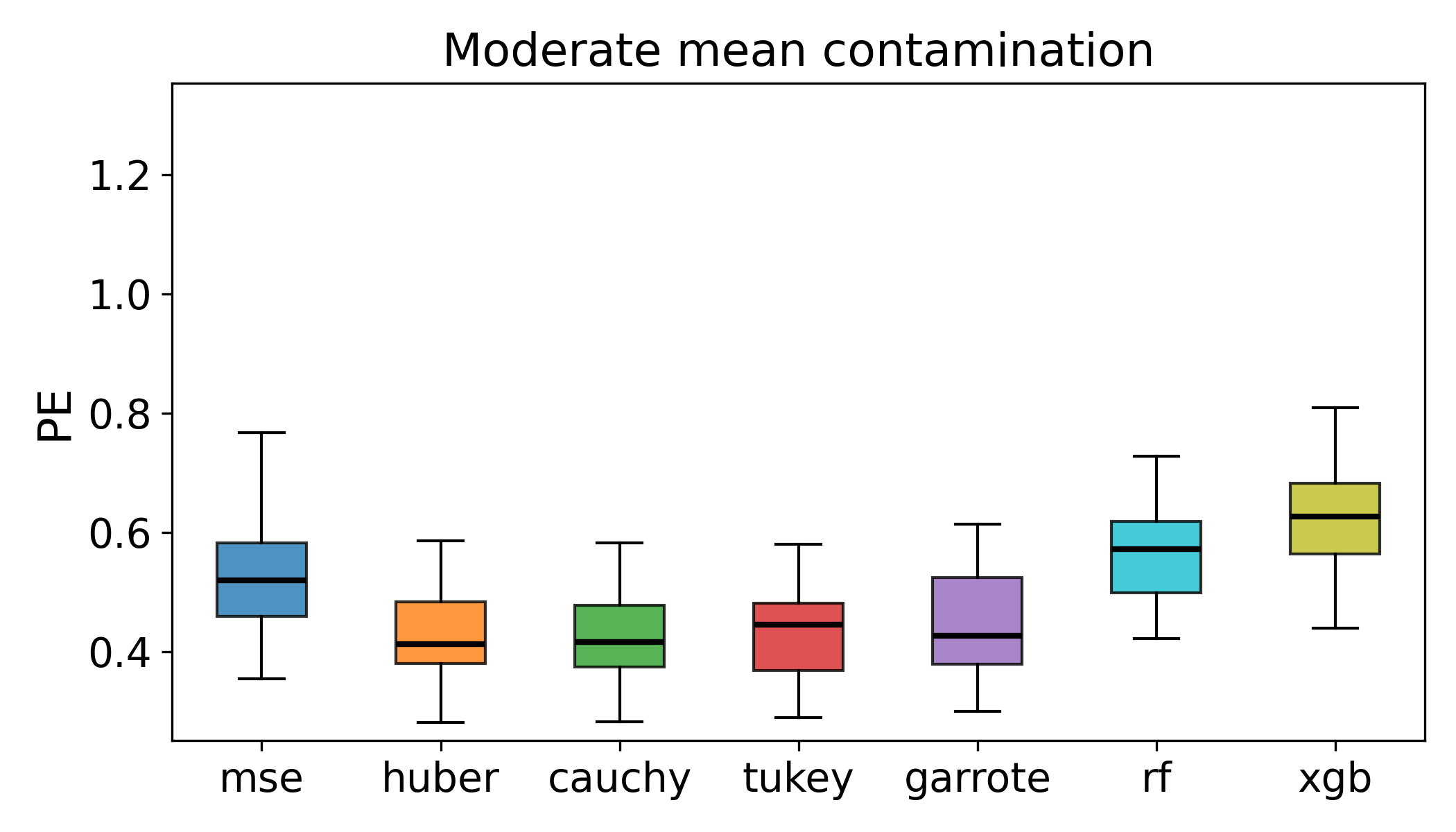}
	\includegraphics[width=0.32\textwidth,height=0.25\textwidth]{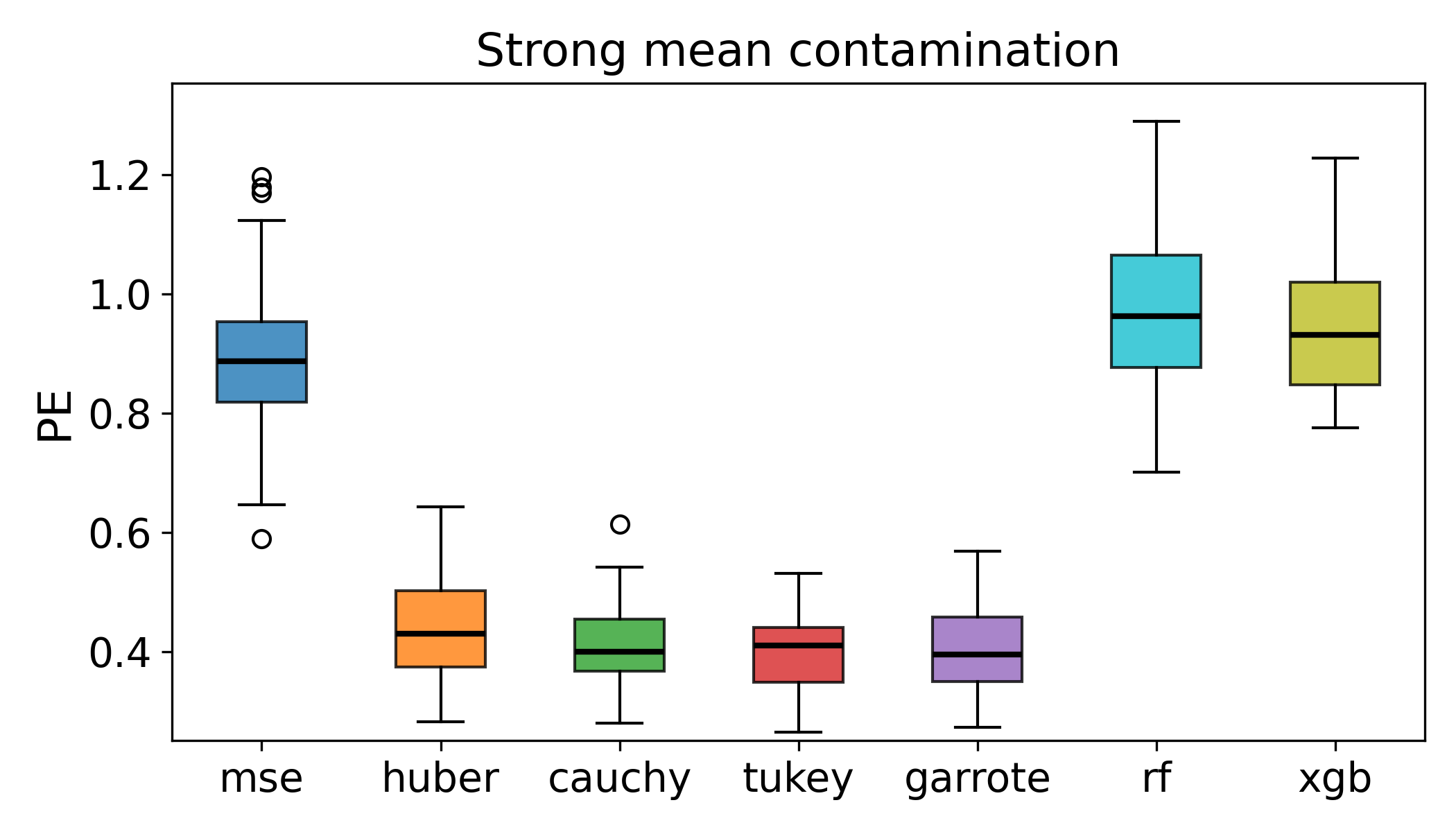}
	
	\vspace{0.2cm}
	
	% --- Row 3 ---
	\includegraphics[width=0.32\textwidth,height=0.25\textwidth]{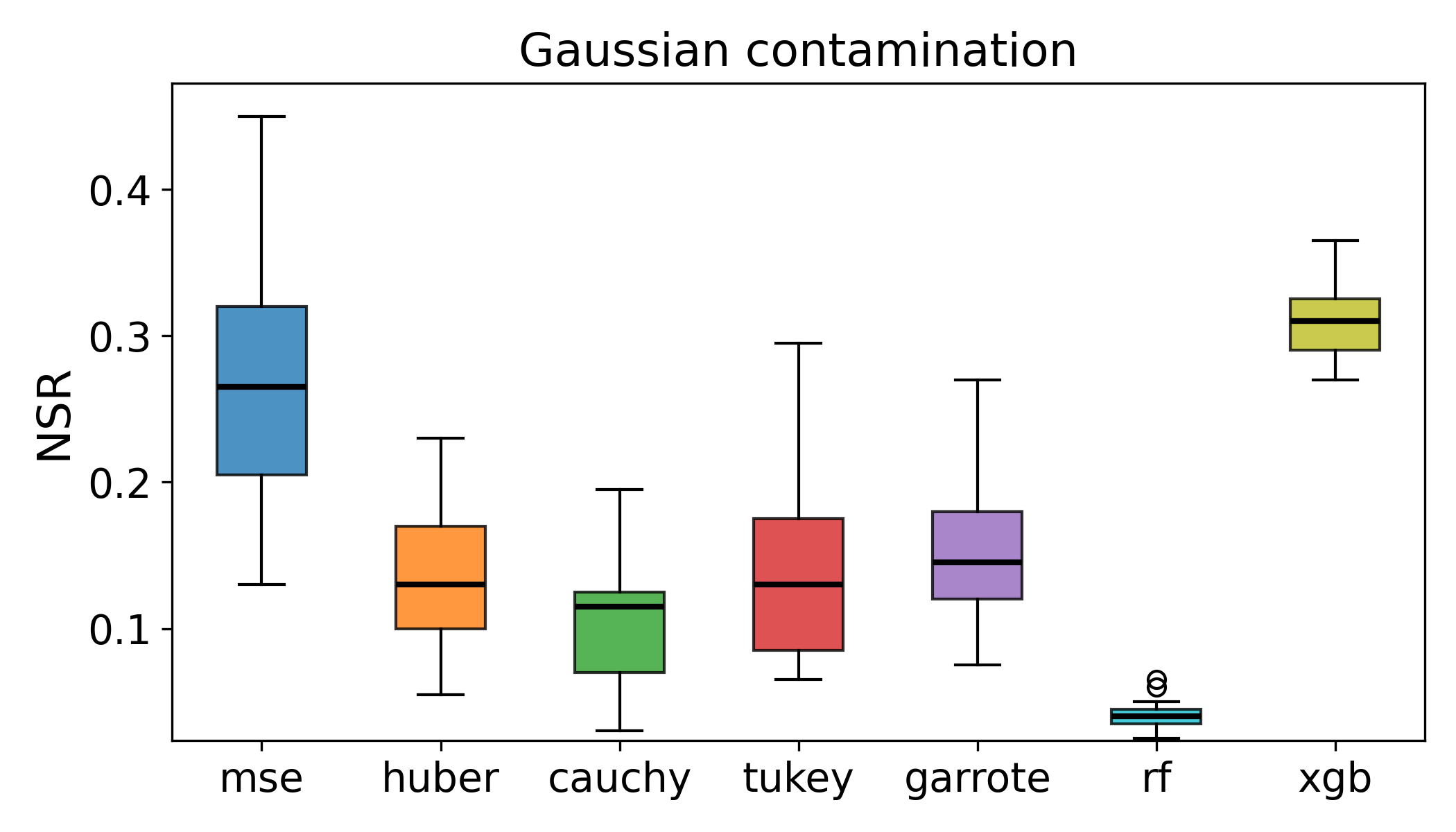}
	\includegraphics[width=0.32\textwidth,height=0.25\textwidth]{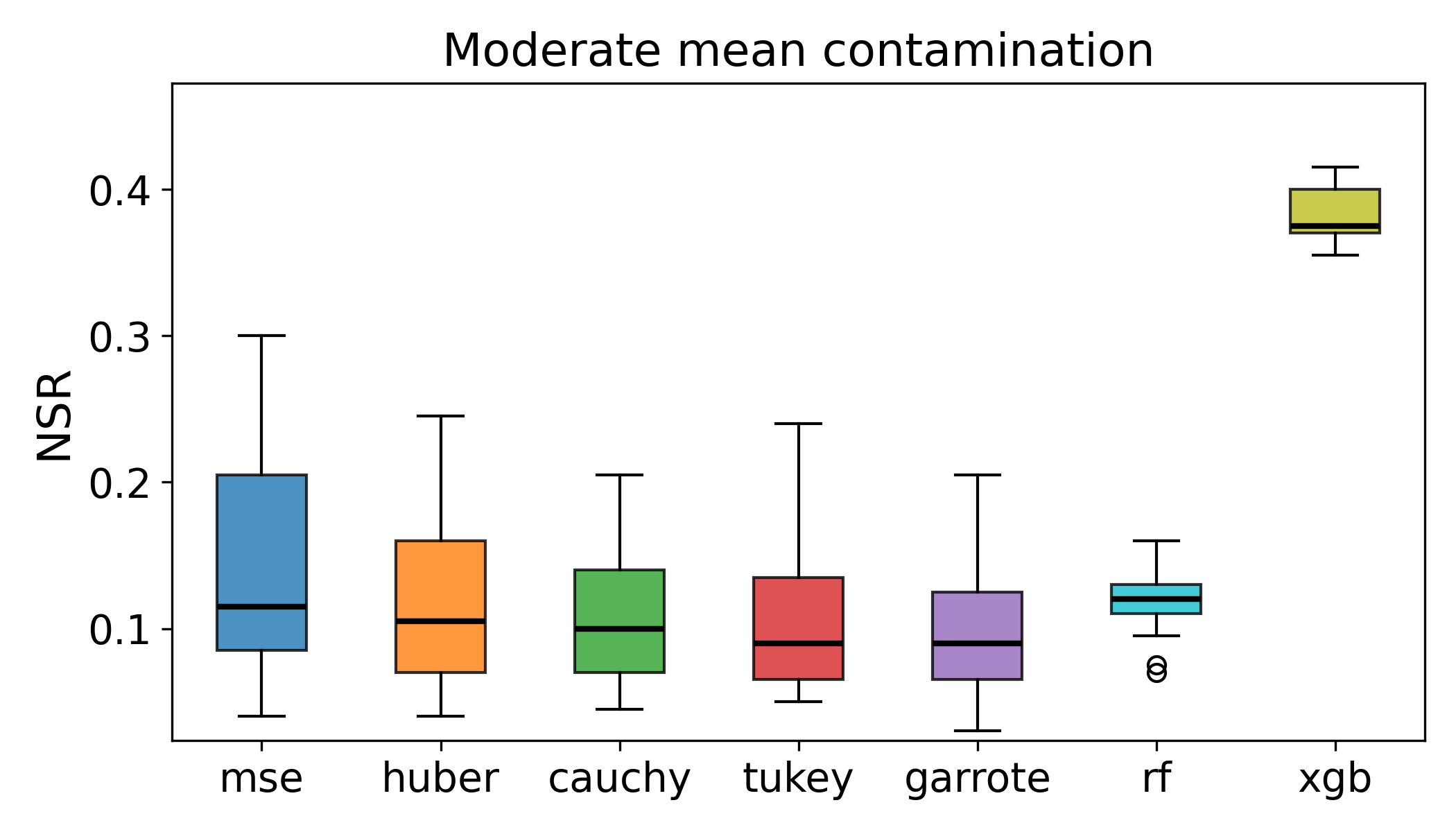}
	\includegraphics[width=0.32\textwidth,height=0.25\textwidth]{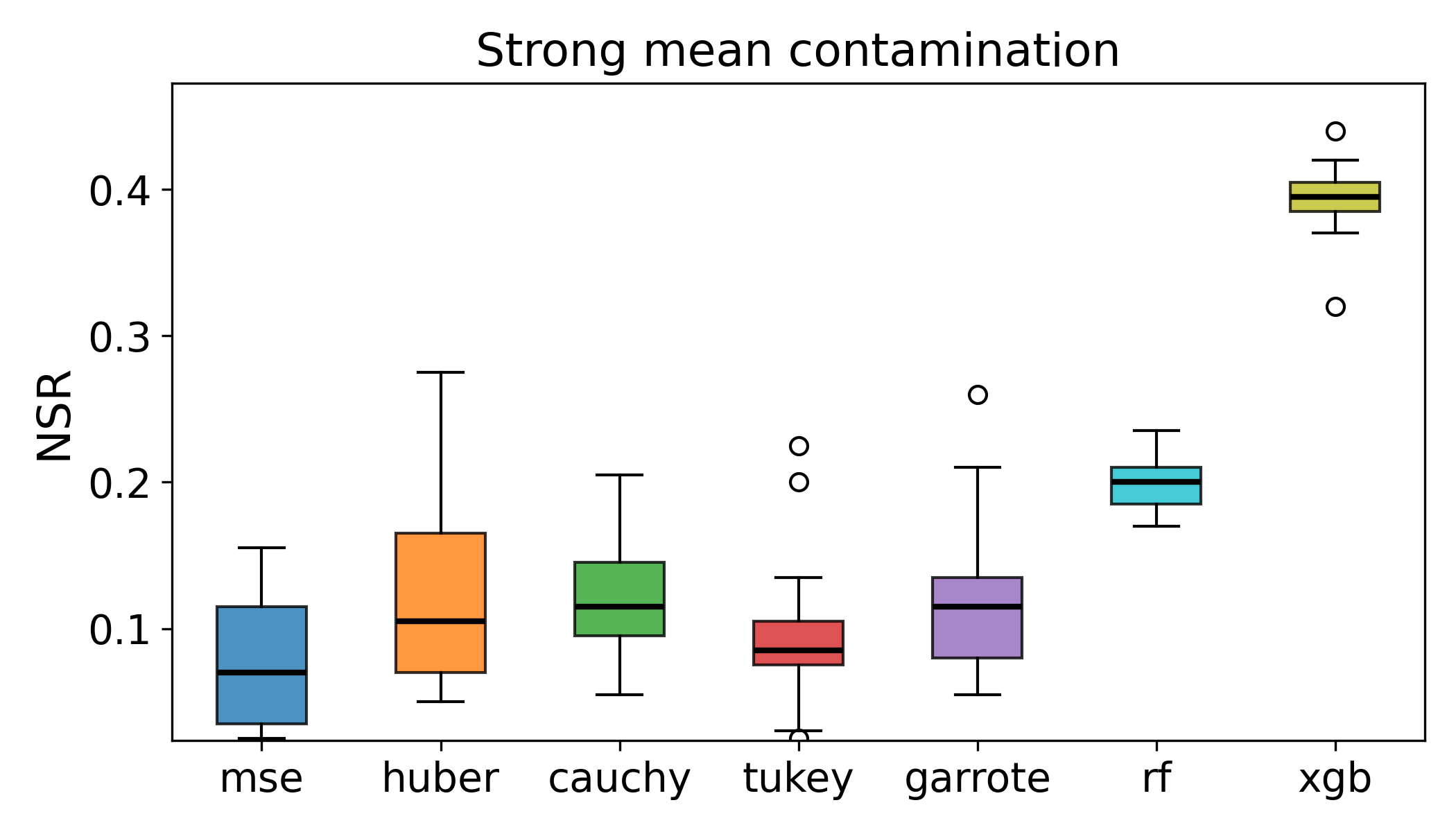}
	
	\vspace{0.2cm}
	
	% --- Row 4 ---
	\includegraphics[width=0.32\textwidth,height=0.25\textwidth]{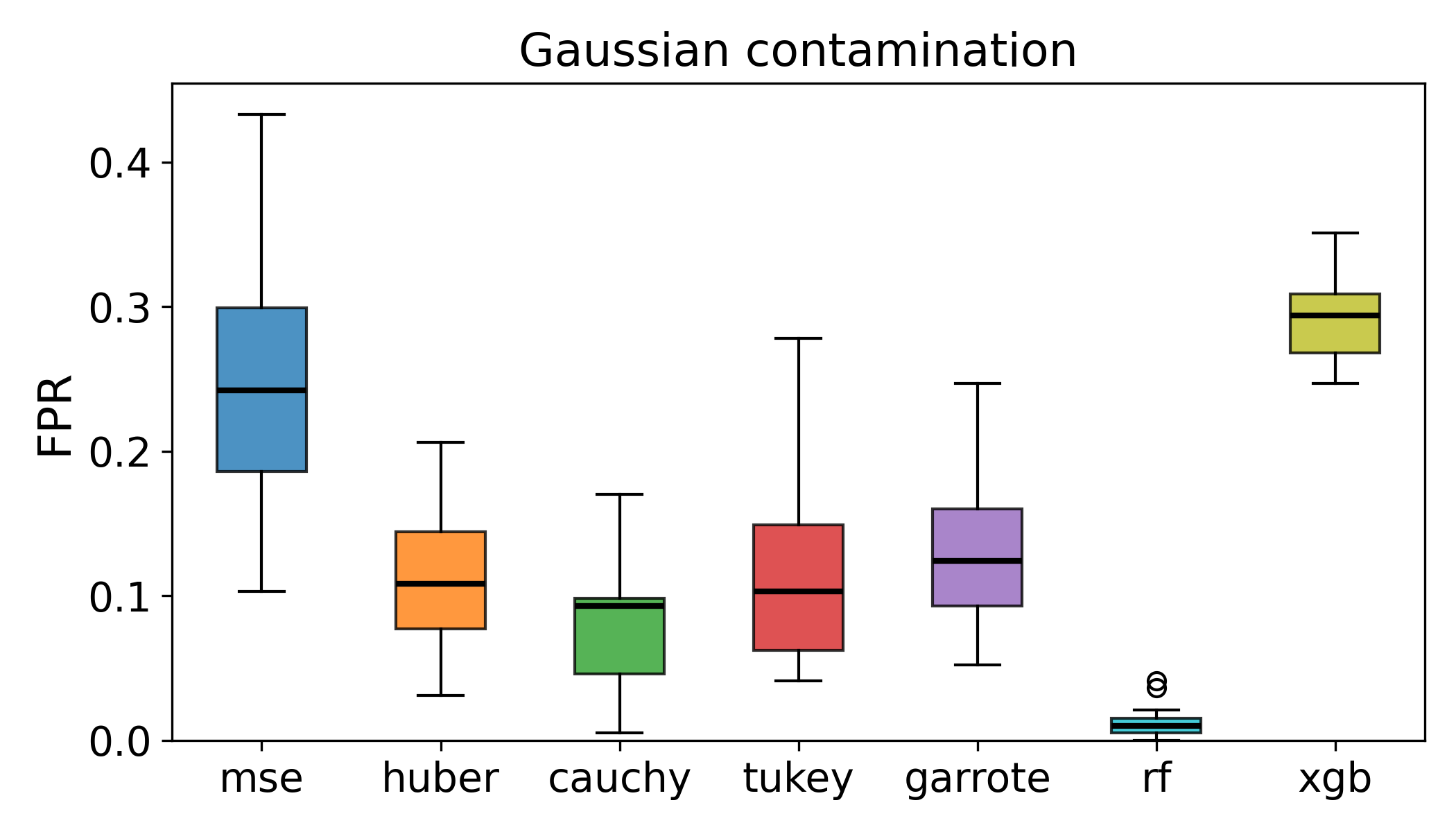}
	\includegraphics[width=0.32\textwidth,height=0.25\textwidth]{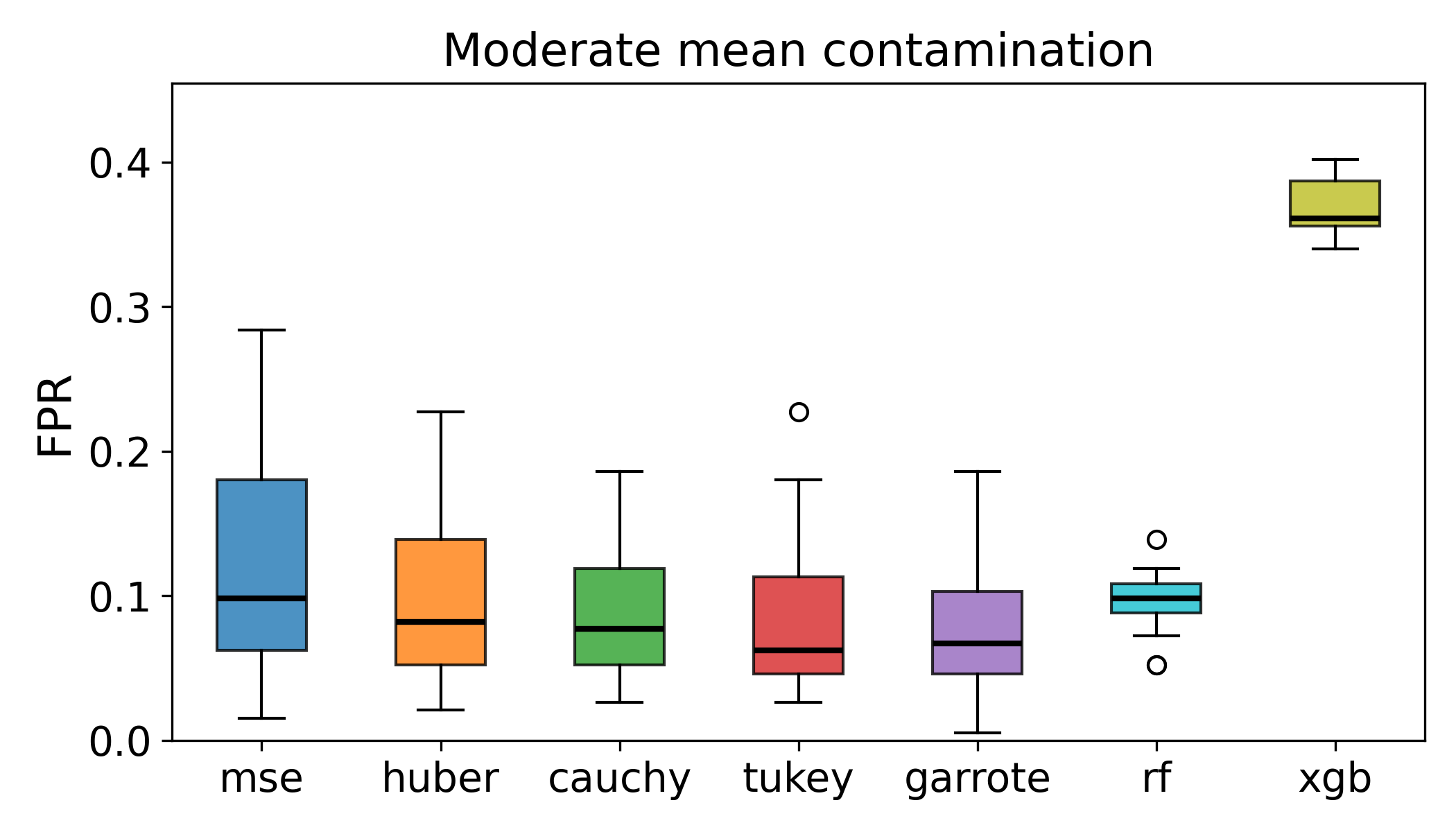}
	\includegraphics[width=0.32\textwidth,height=0.25\textwidth]{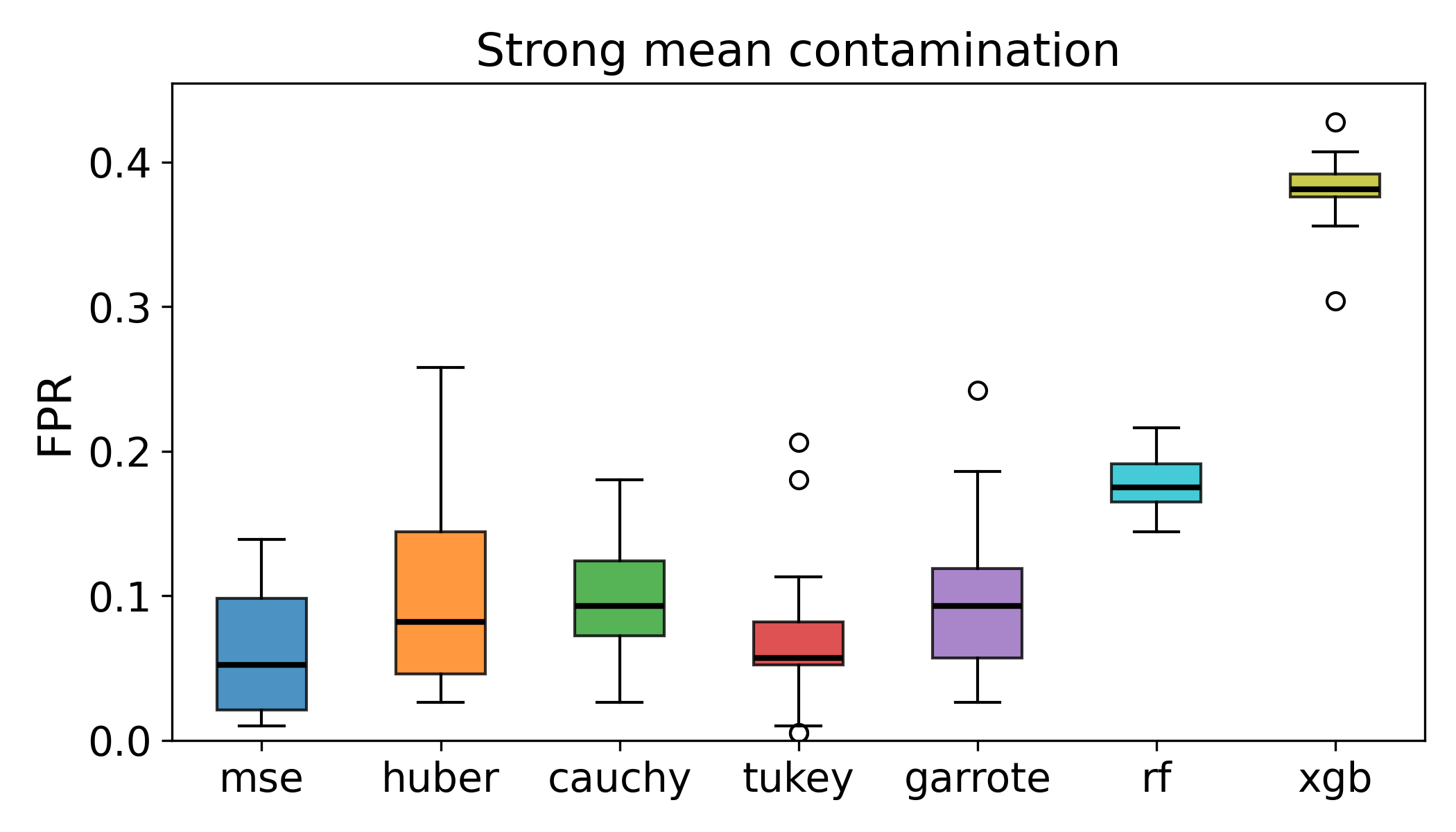}
	
	\vspace{0.2cm}
	
	% --- Row 5 ---
	\includegraphics[width=0.32\textwidth,height=0.25\textwidth]{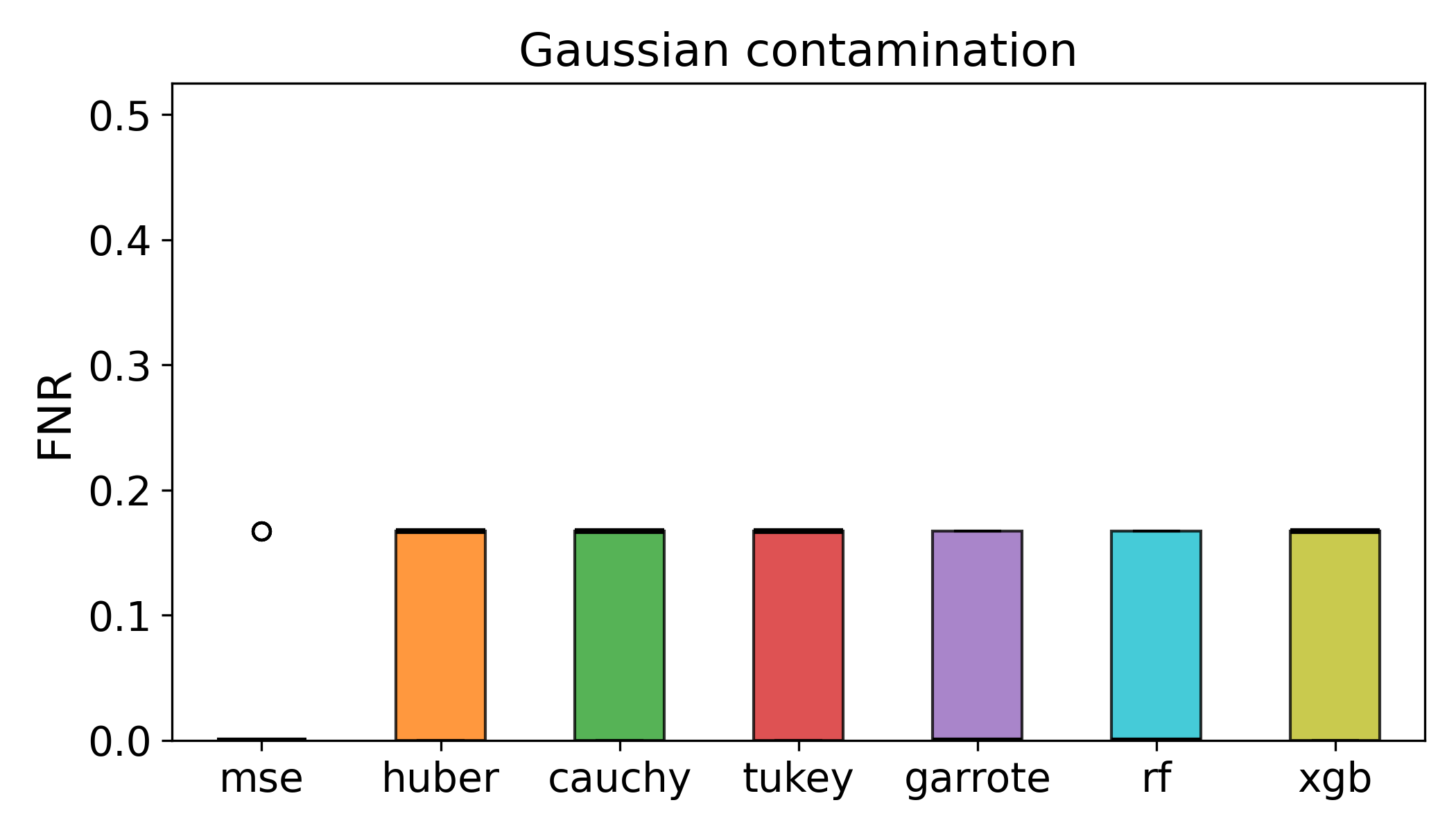}
	\includegraphics[width=0.32\textwidth,height=0.25\textwidth]{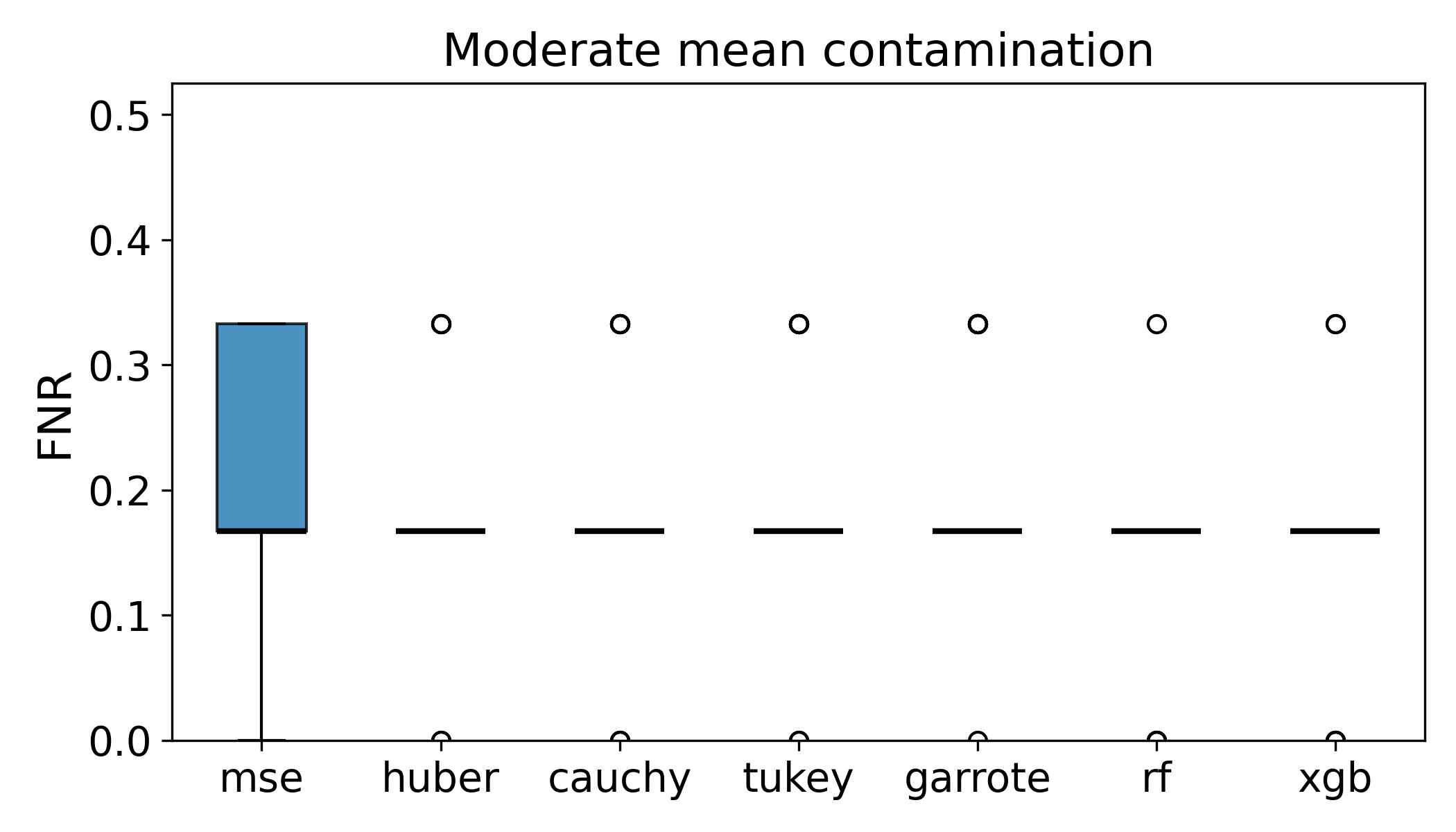}
	\includegraphics[width=0.32\textwidth,height=0.25\textwidth]{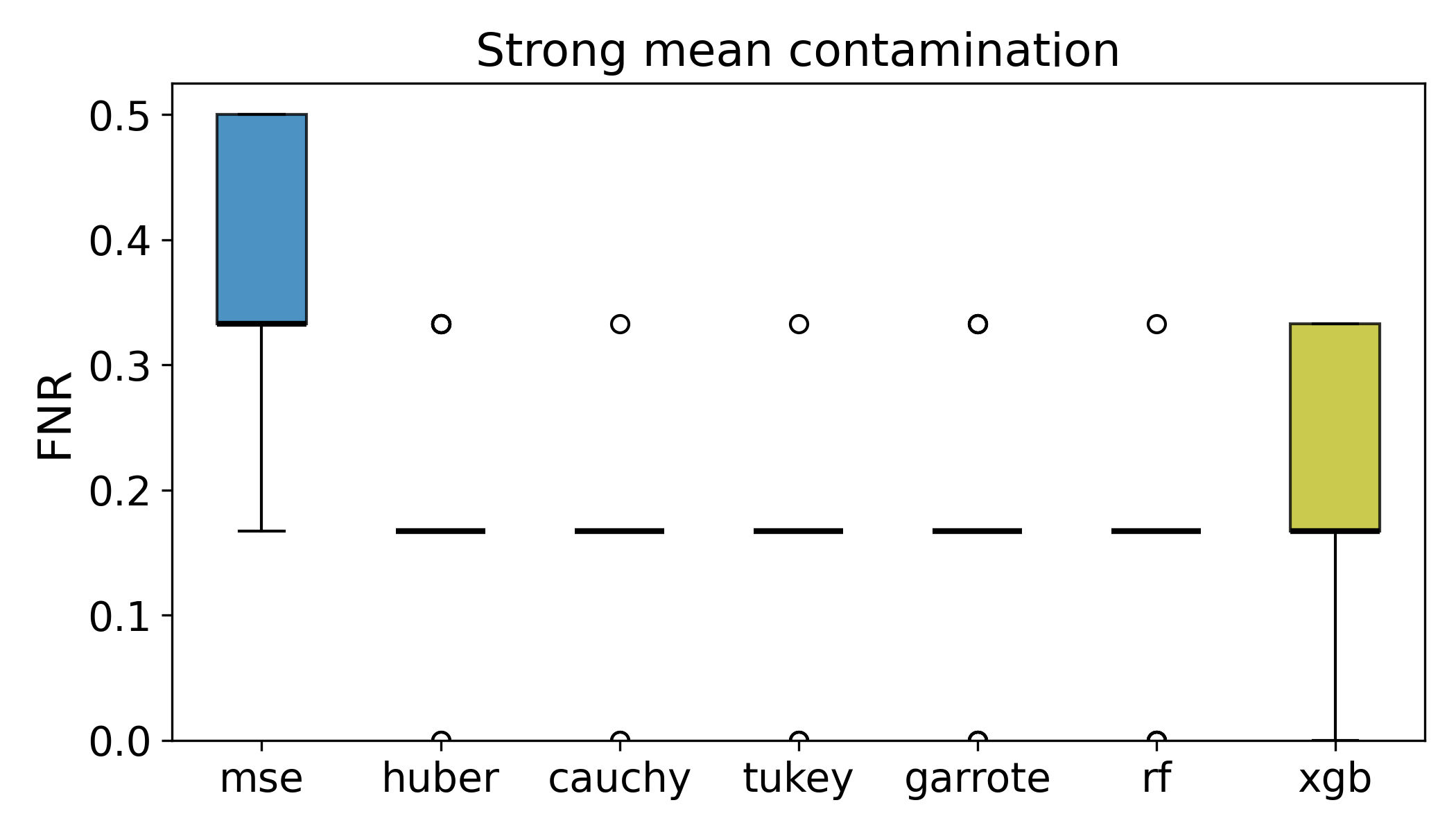}
	
	\caption{Boxplots of the performance metrics results for {\tt model 3} with $10\%$ contamination. First column refers to {\tt gaussian} scenario; second column to {\tt moderate mean} scenario; third column to {\tt strong mean} scenario. The first row shows a realization of the true and contaminated training response $\mathbf{y}$ for all the scenarios.}
	\label{fig_boxplot_model3_mean}
	
\end{figure*}

% boxplots model 4 mixture mean
\begin{figure*}[t]
	\centering
	
	% --- Row 1 ---
	\includegraphics[width=0.32\textwidth,height=0.25\textwidth]{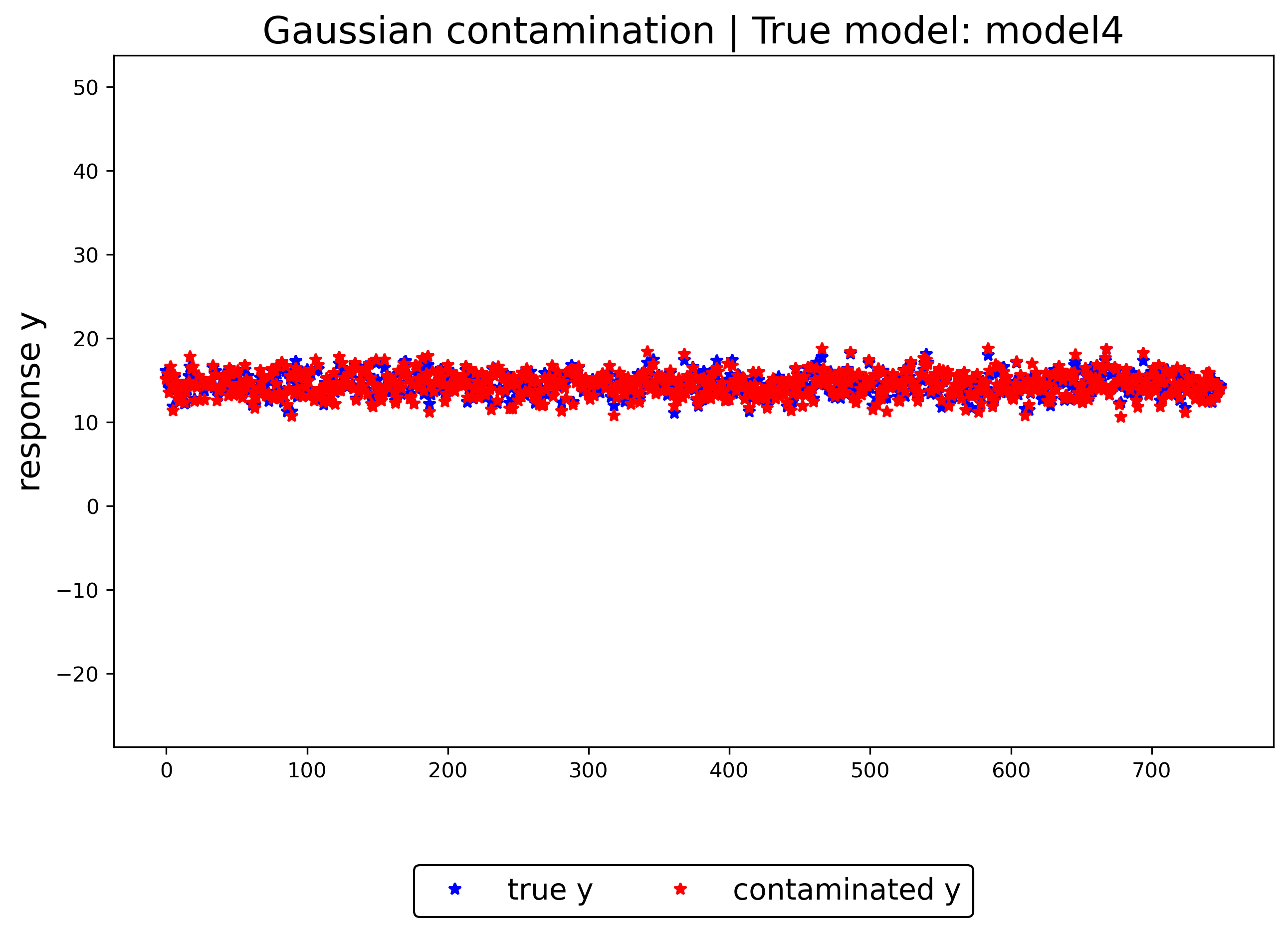}
	\includegraphics[width=0.32\textwidth,height=0.25\textwidth]{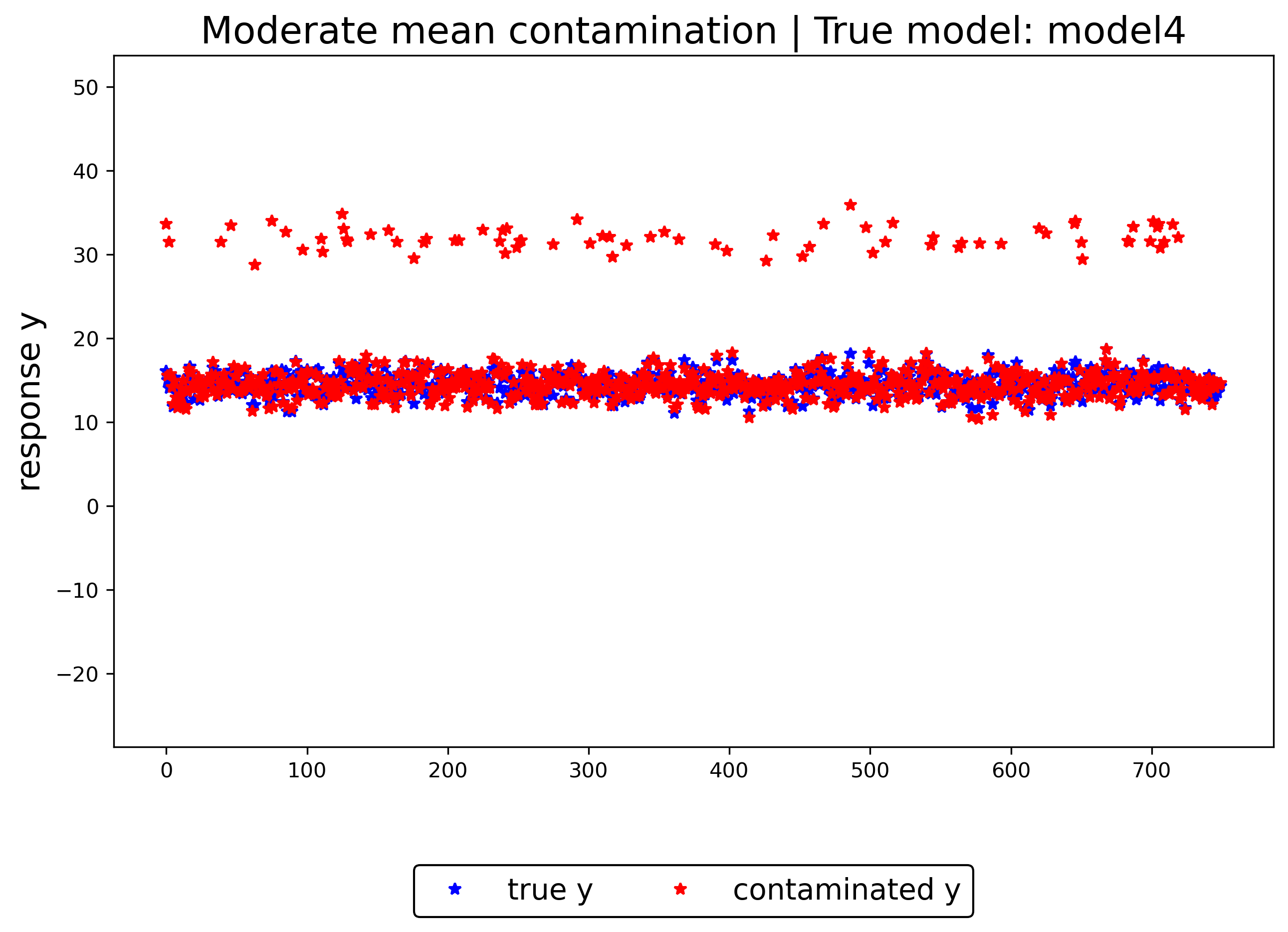}
	\includegraphics[width=0.32\textwidth,height=0.25\textwidth]{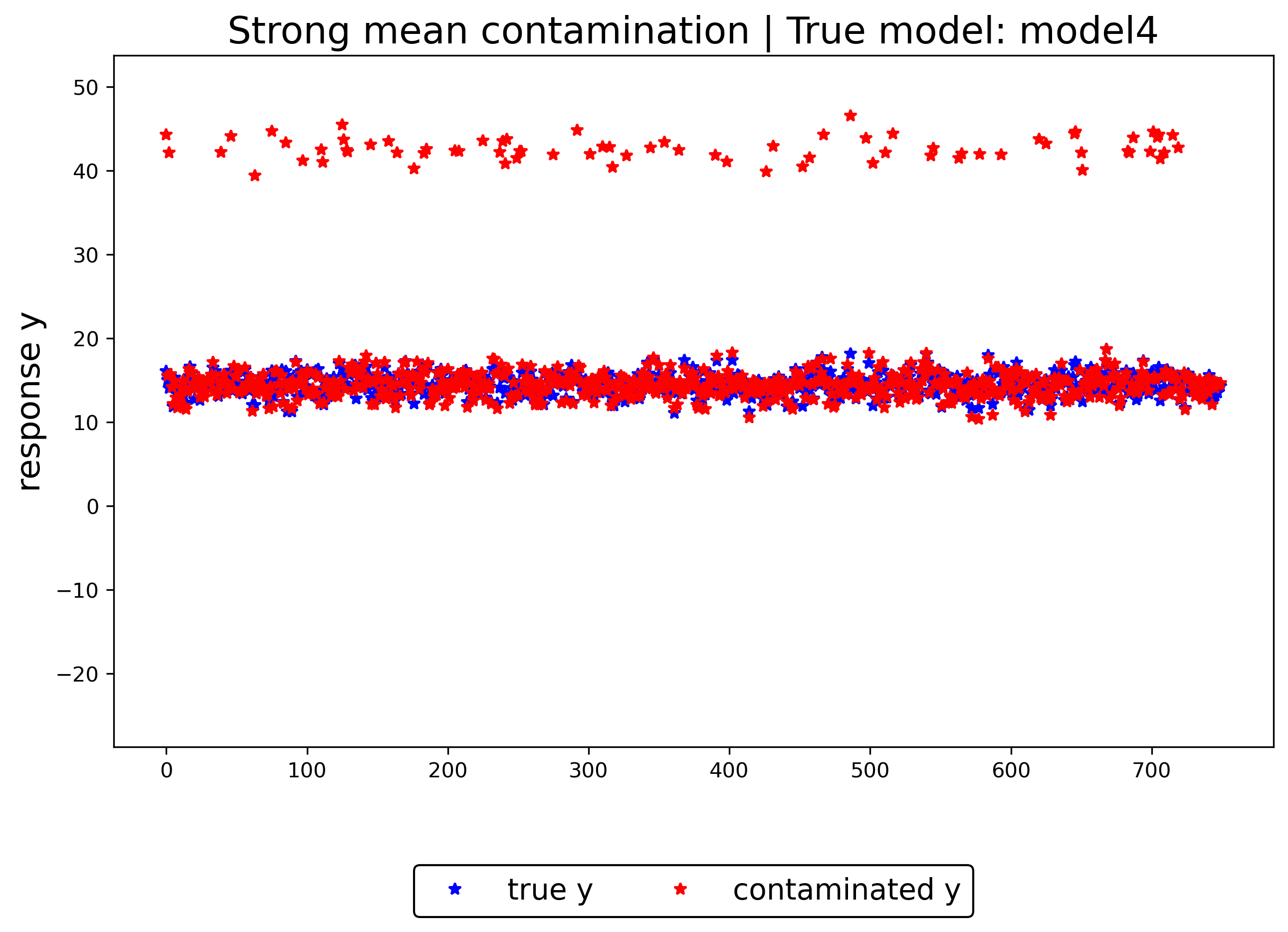}

	% --- Row 2 ---
	\includegraphics[width=0.32\textwidth,height=0.25\textwidth]{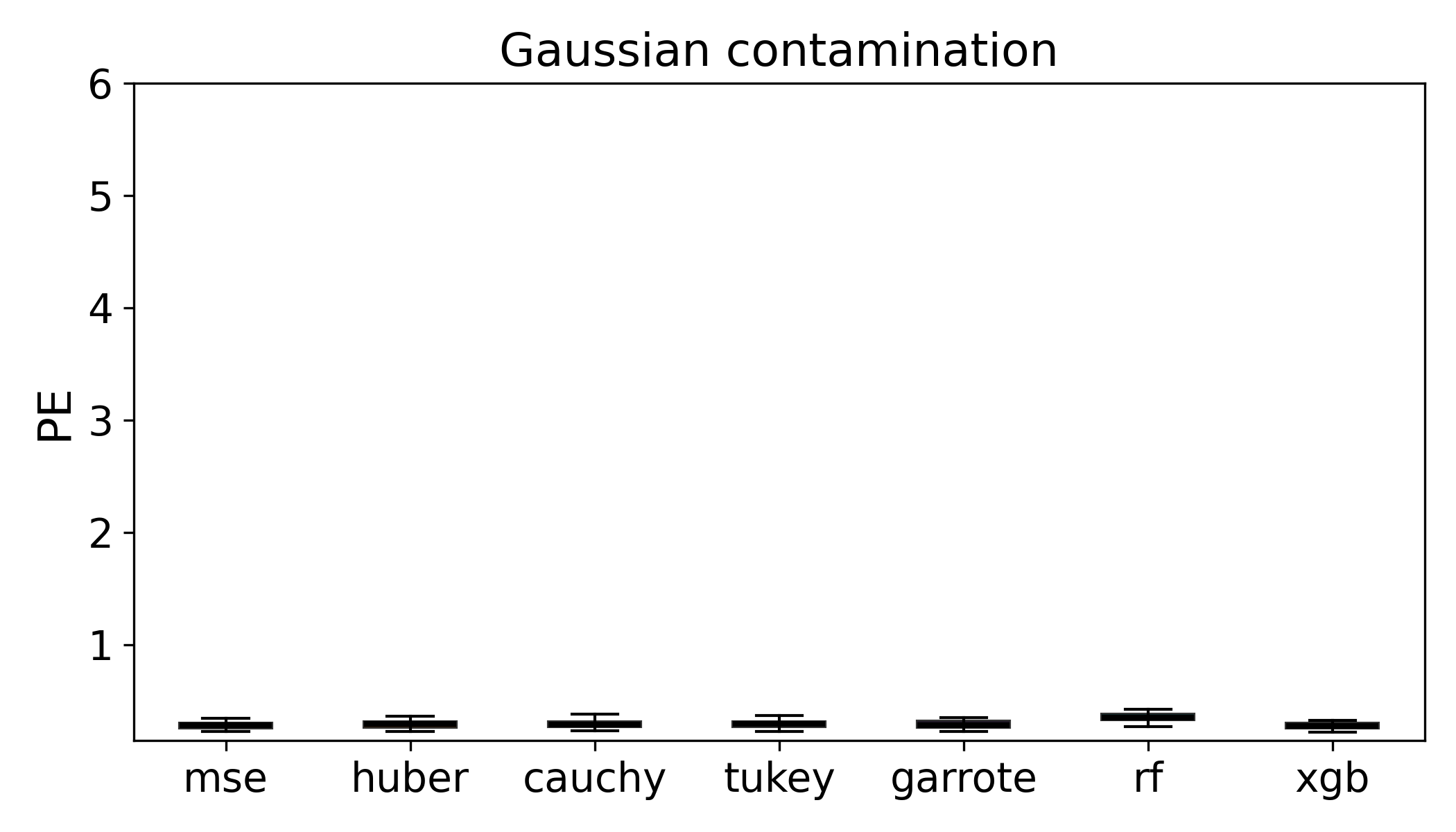}
	\includegraphics[width=0.32\textwidth,height=0.25\textwidth]{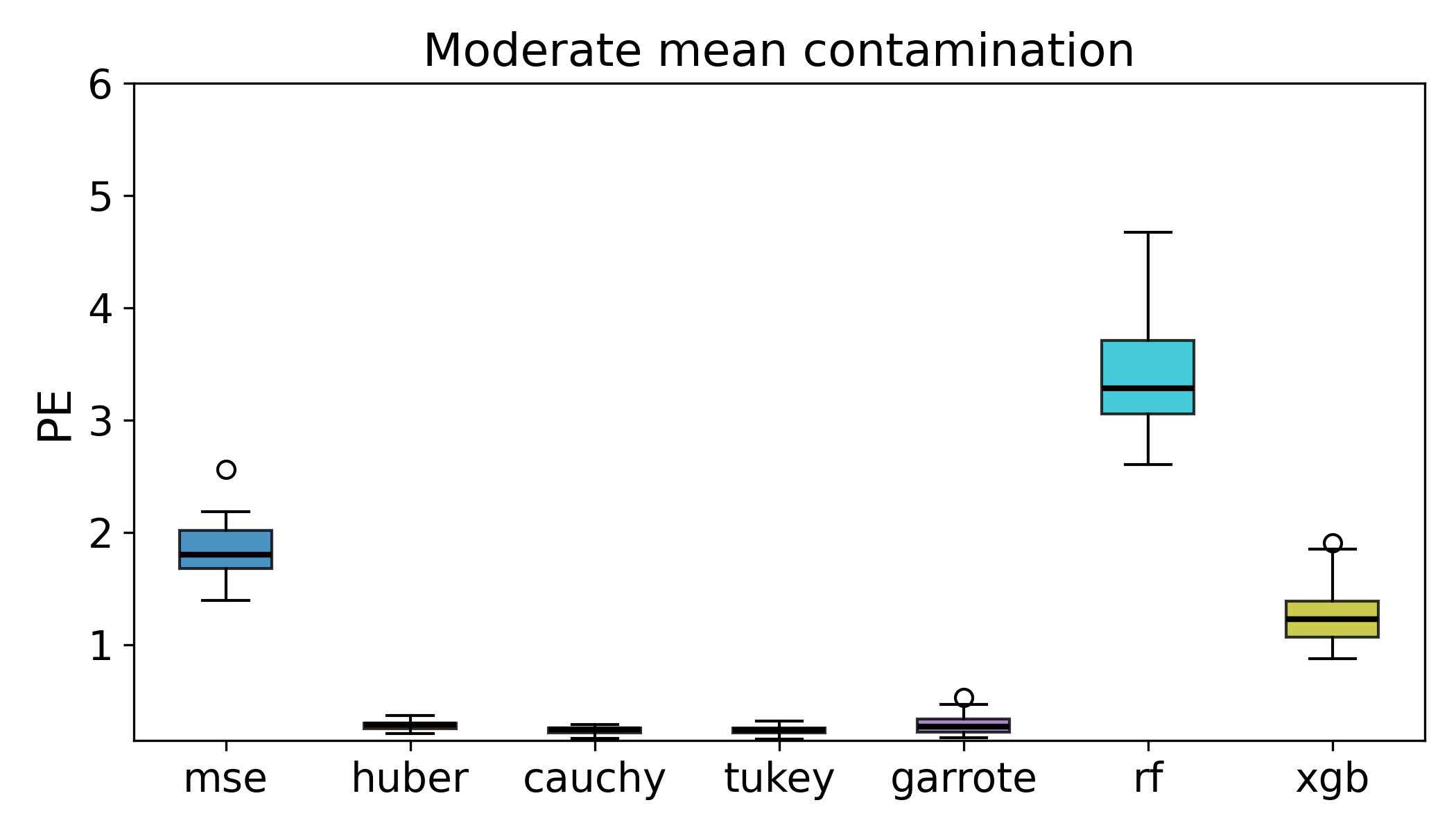}
	\includegraphics[width=0.32\textwidth,height=0.25\textwidth]{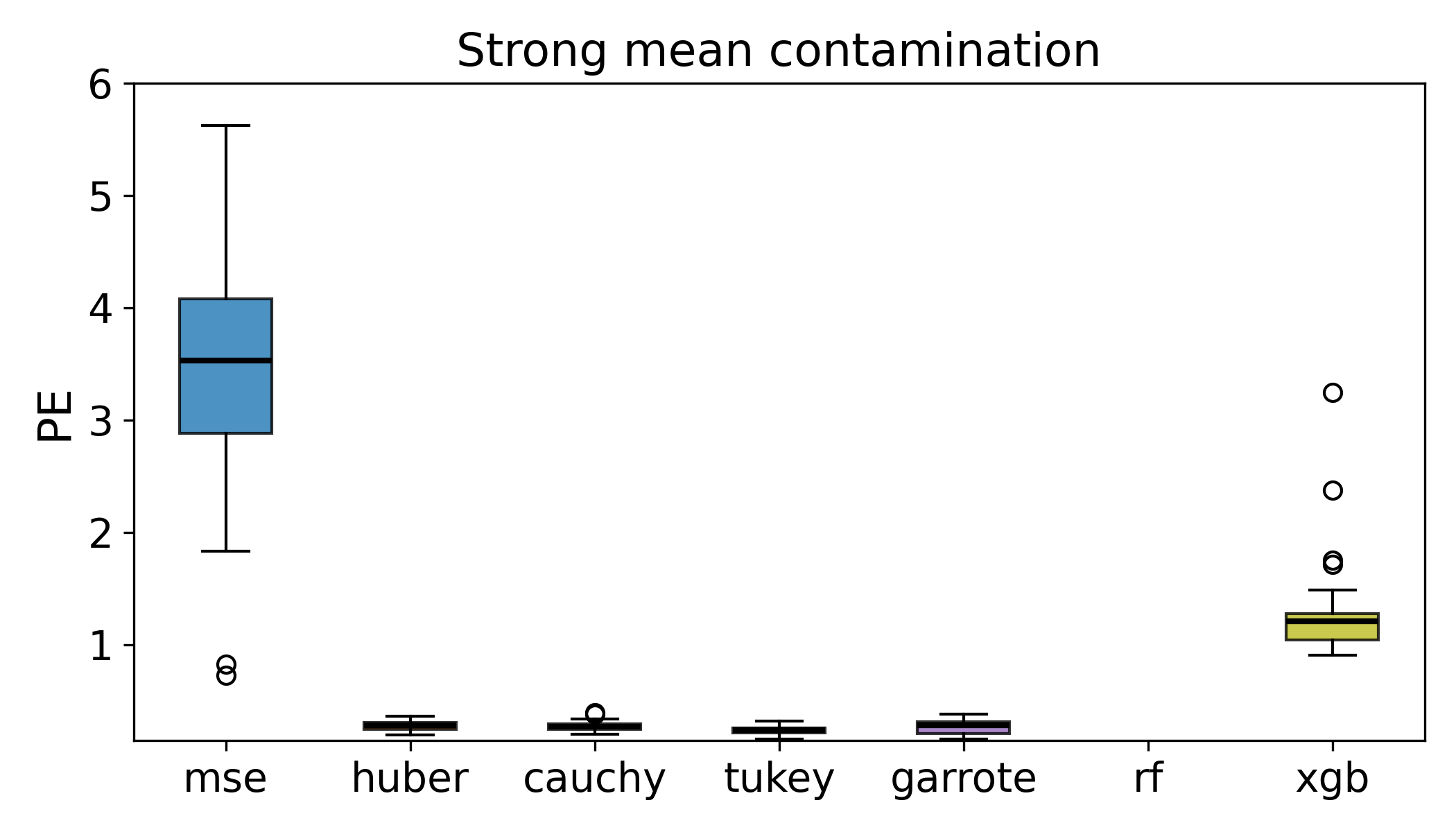}
	
	\vspace{0.2cm}
	
	% --- Row 3 ---
	\includegraphics[width=0.32\textwidth,height=0.25\textwidth]{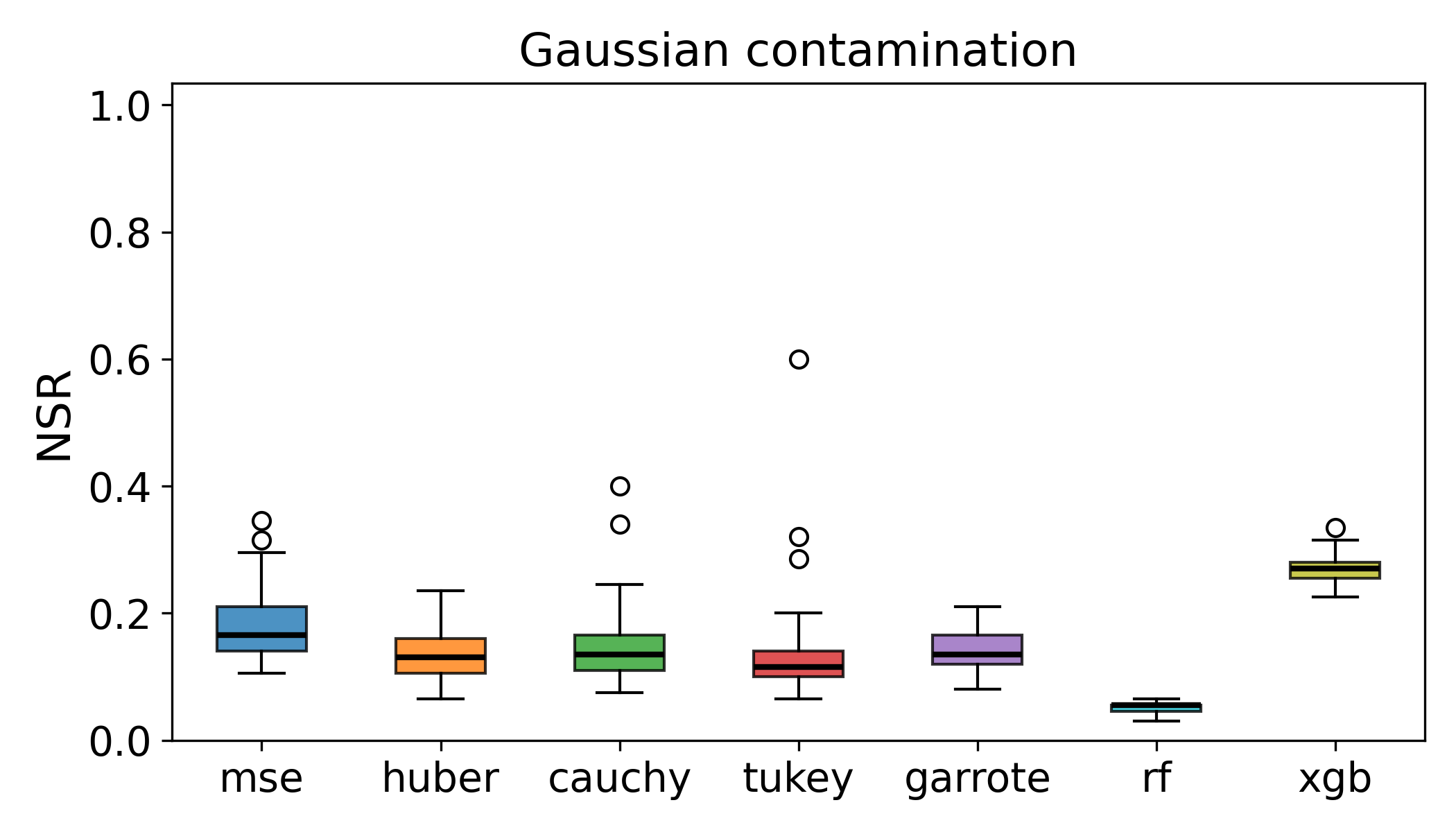}
	\includegraphics[width=0.32\textwidth,height=0.25\textwidth]{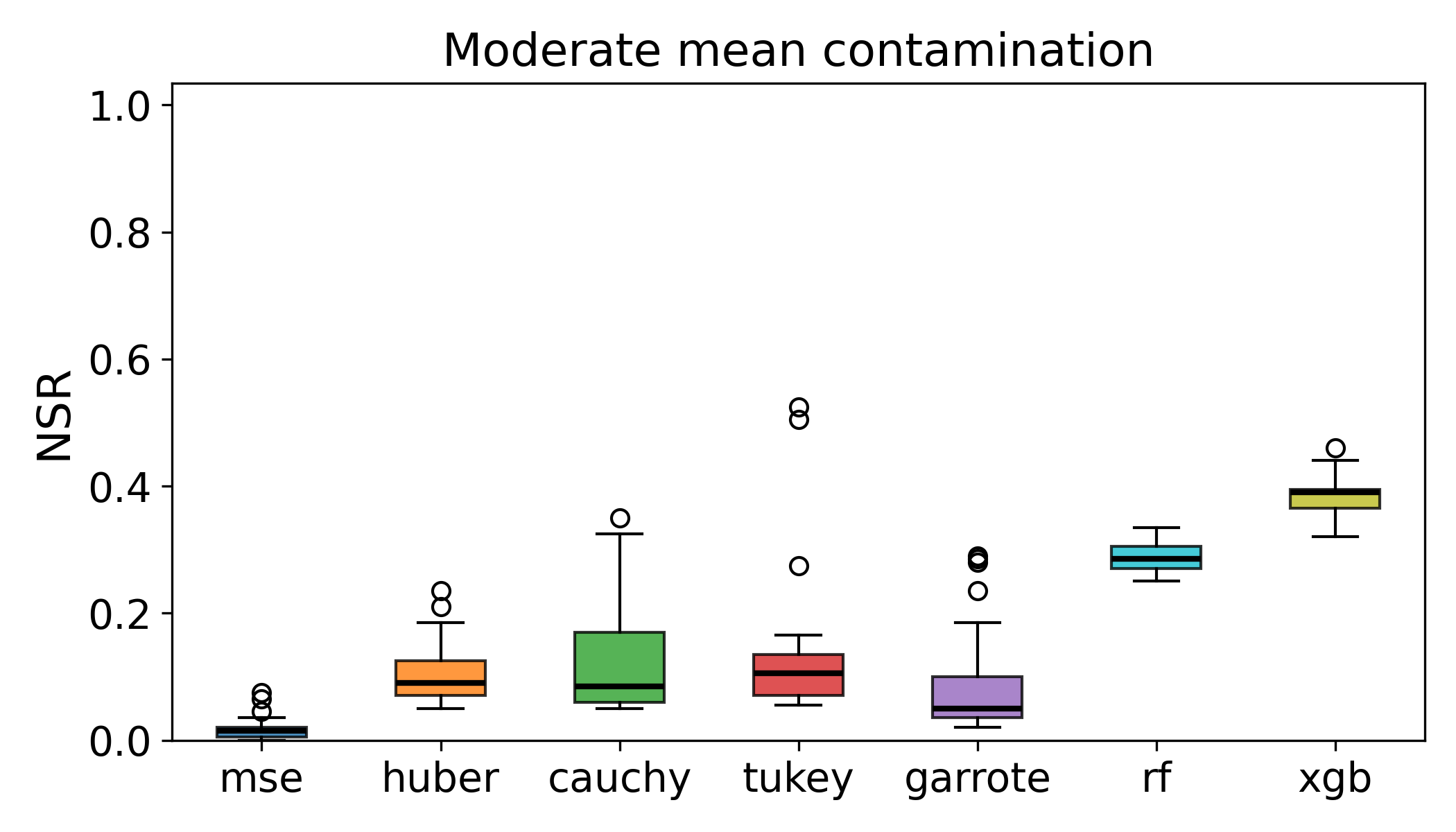}
	\includegraphics[width=0.32\textwidth,height=0.25\textwidth]{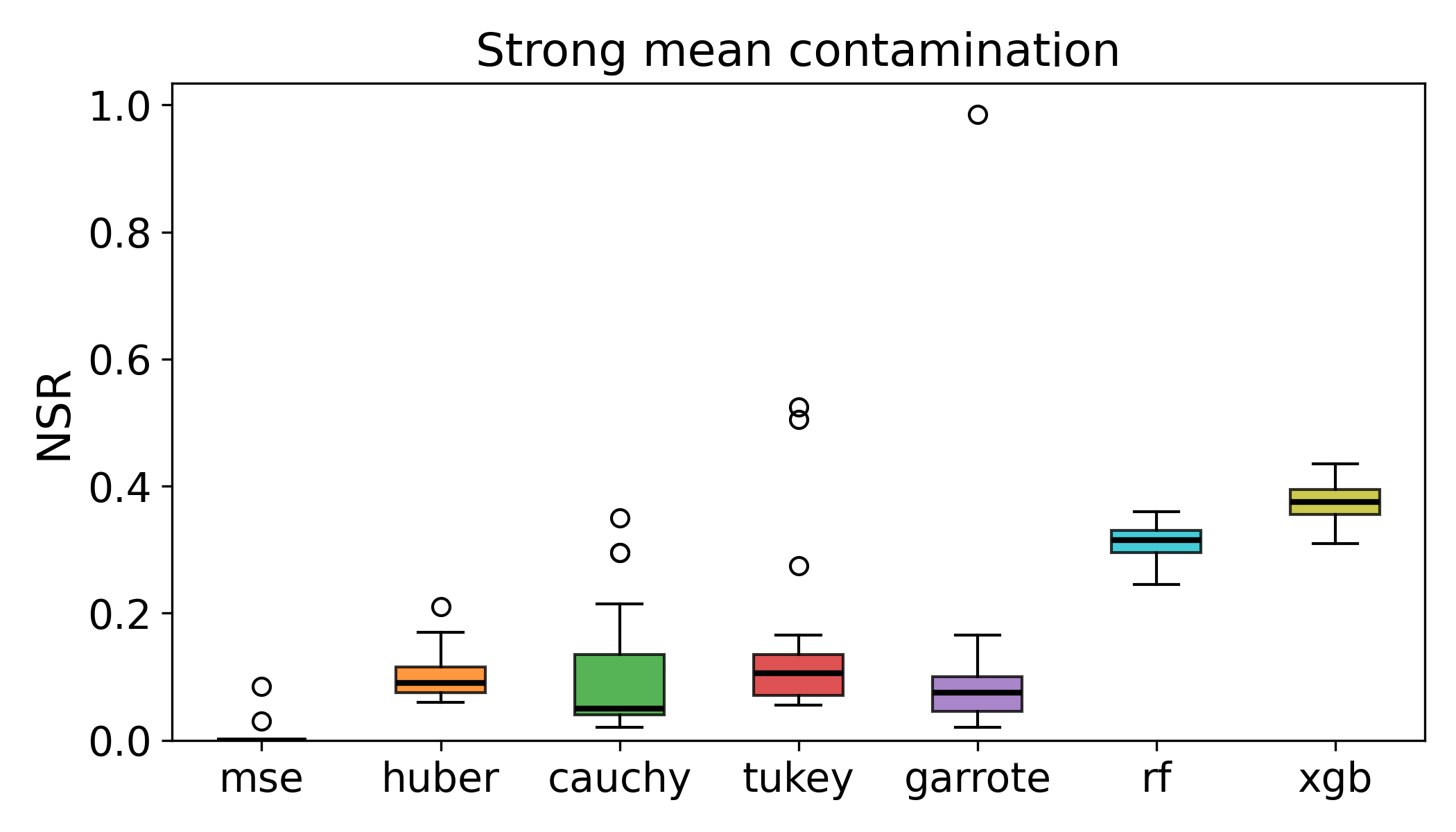}
	
	\vspace{0.2cm}
	
	% --- Row 4 ---
	\includegraphics[width=0.32\textwidth,height=0.25\textwidth]{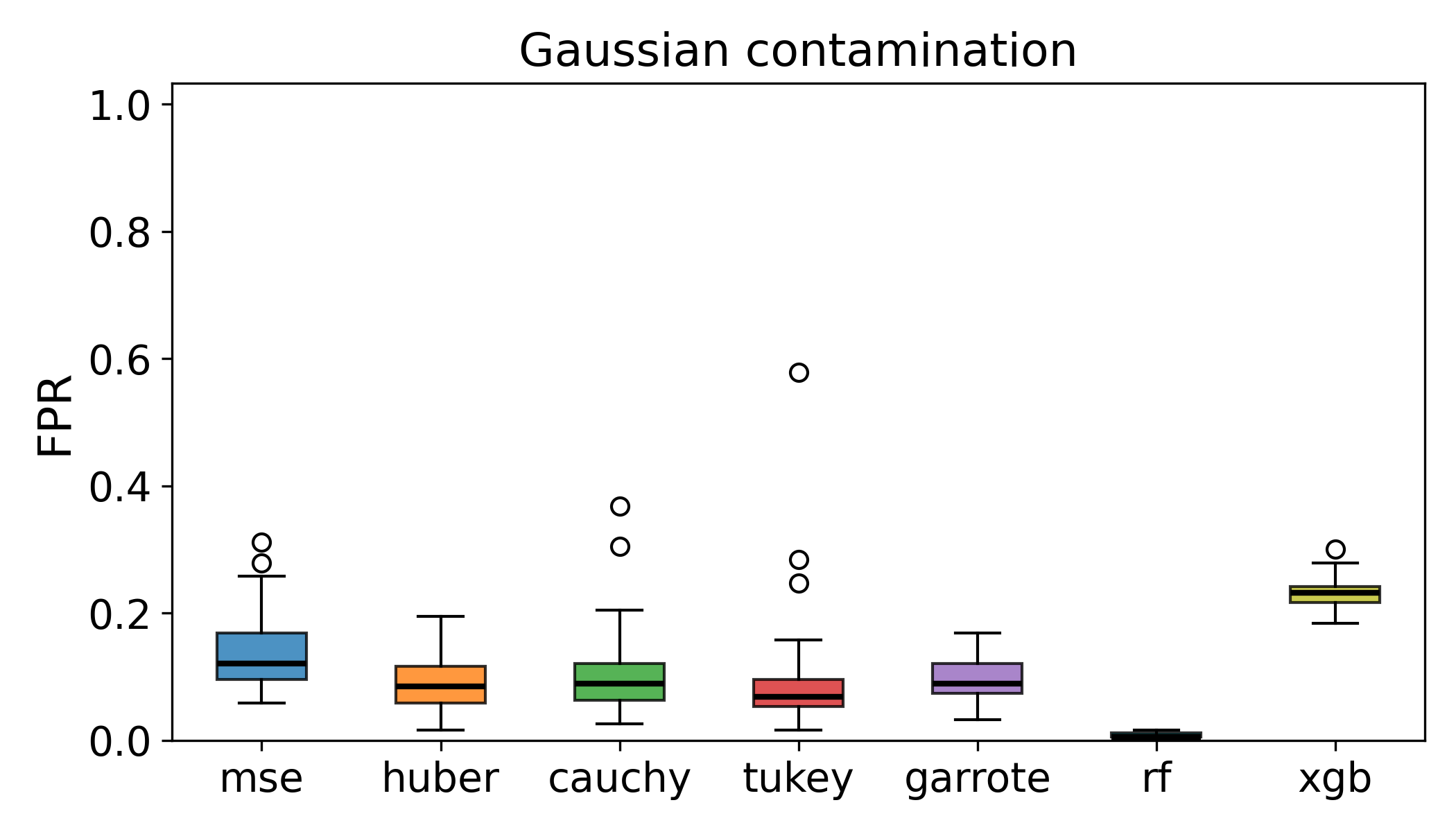}
	\includegraphics[width=0.32\textwidth,height=0.25\textwidth]{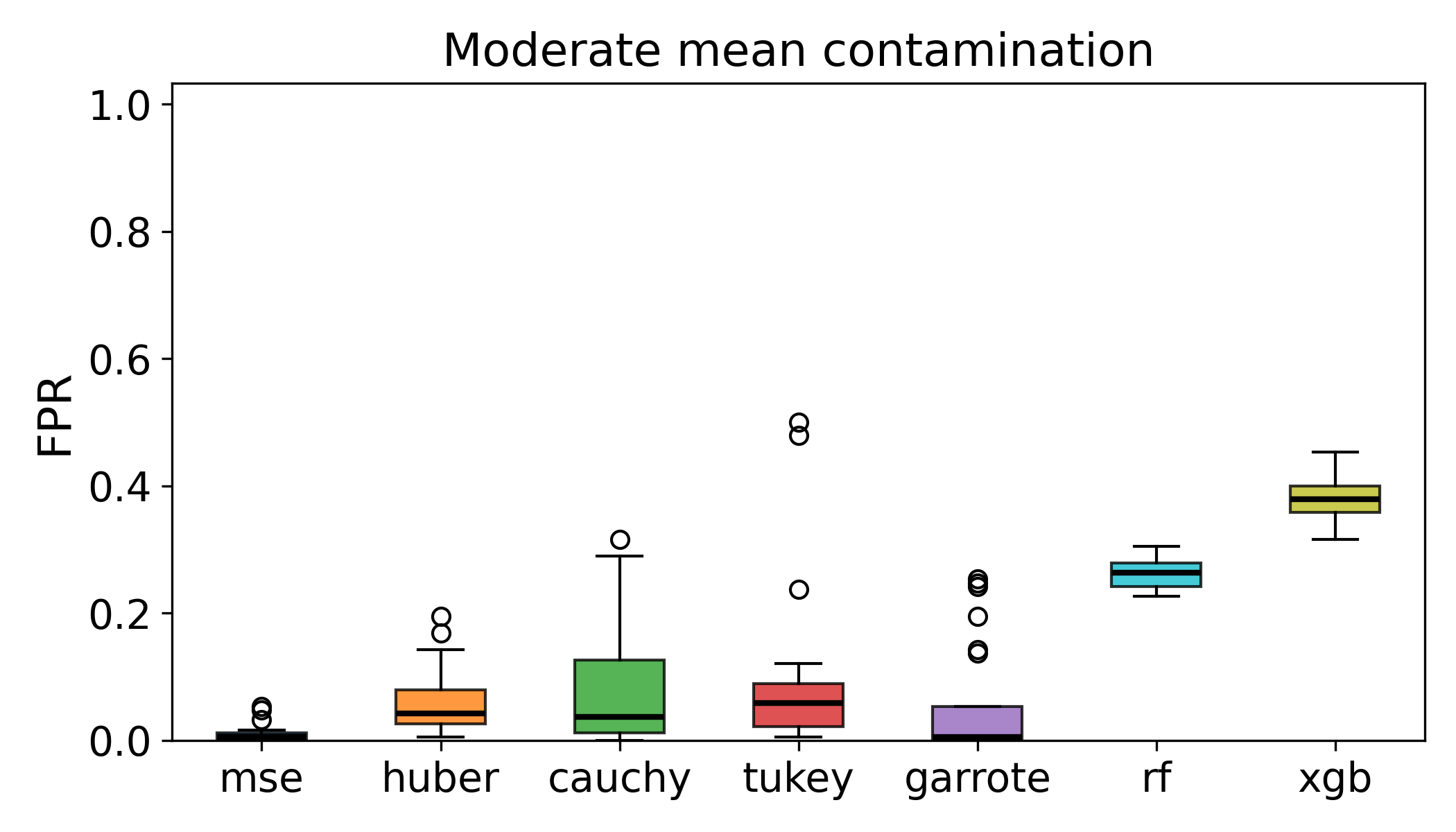}
	\includegraphics[width=0.32\textwidth,height=0.25\textwidth]{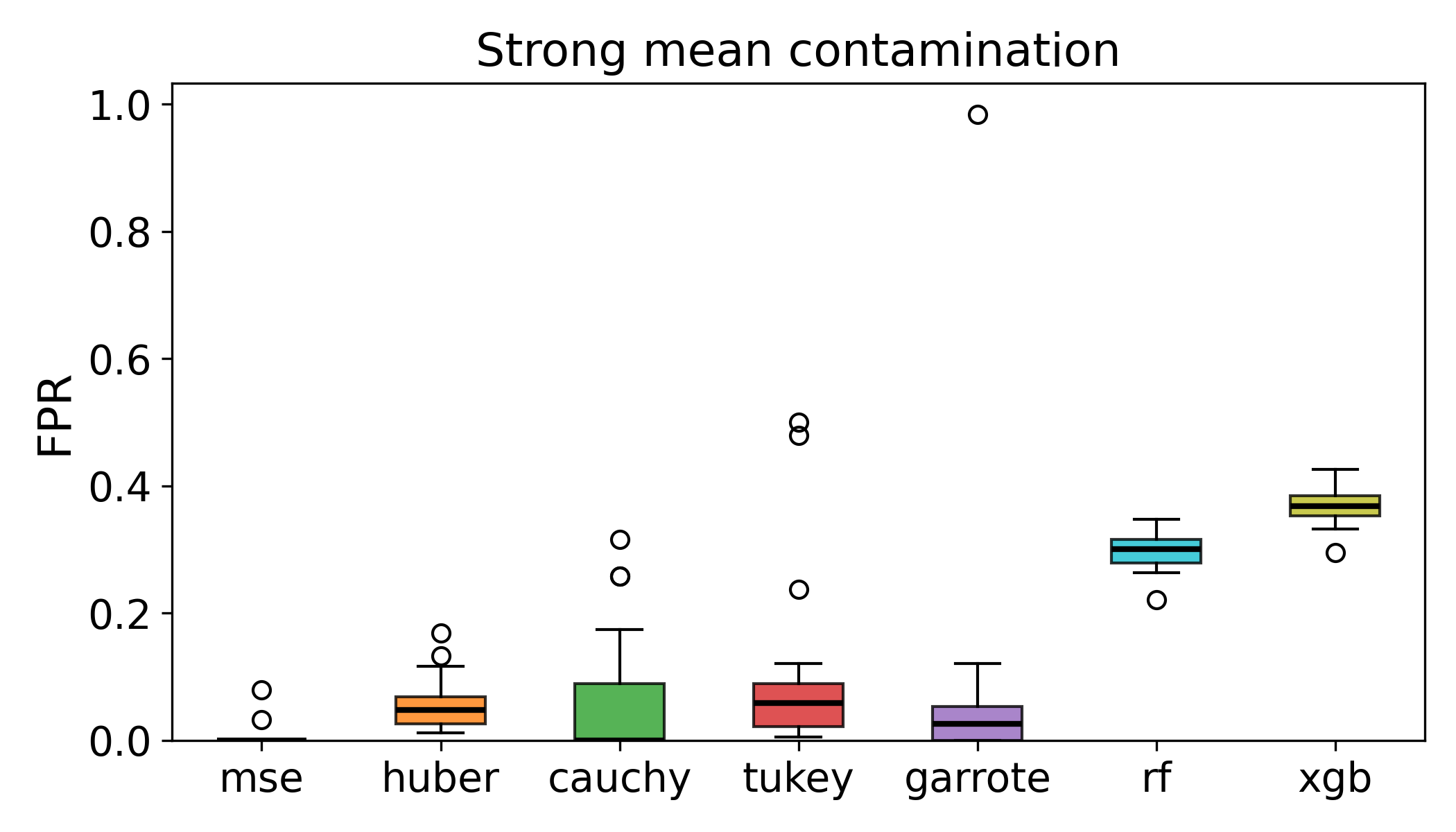}
	
	\vspace{0.2cm}
	
	% --- Row 5 ---
	\includegraphics[width=0.32\textwidth,height=0.25\textwidth]{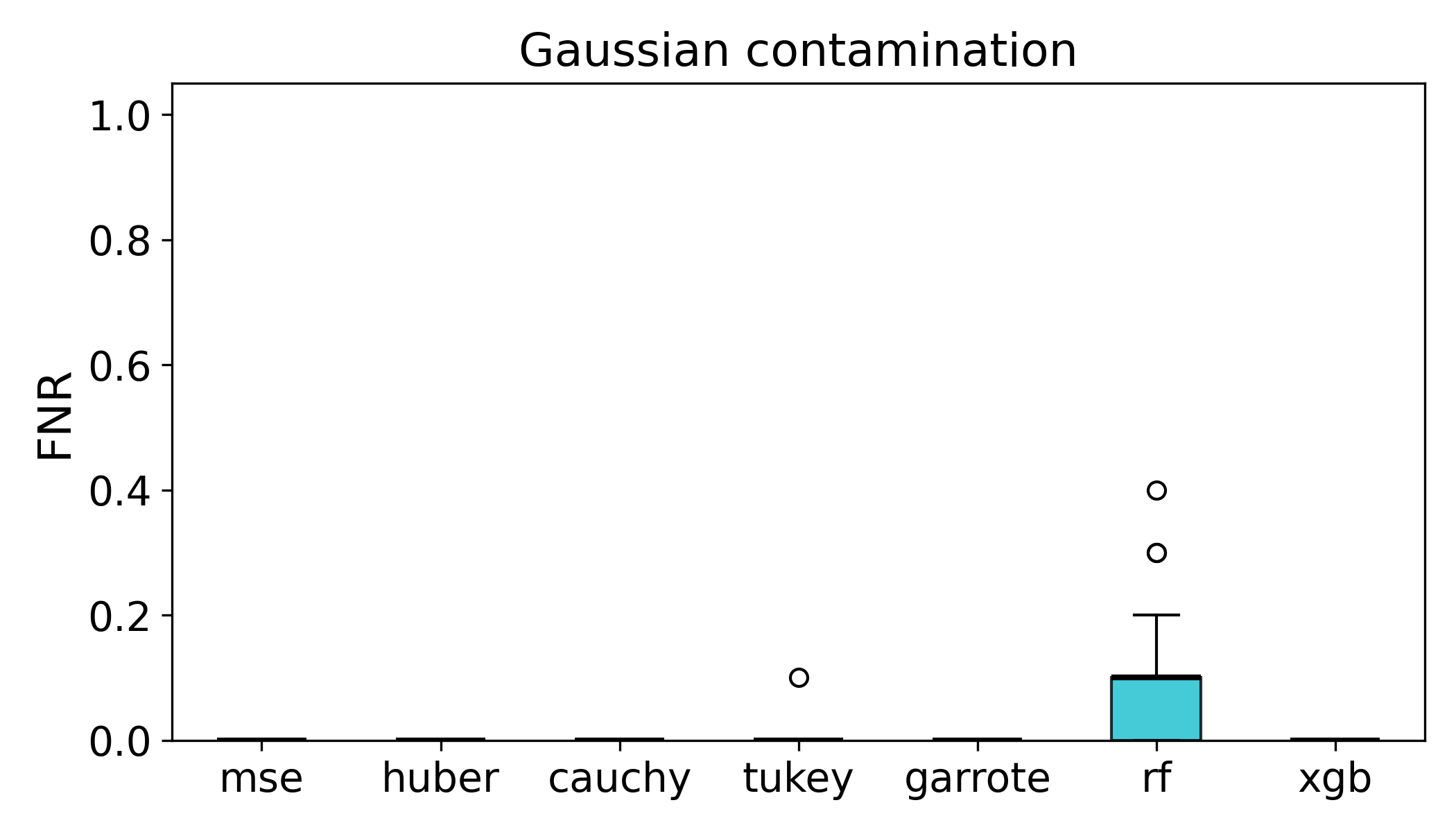}
	\includegraphics[width=0.32\textwidth,height=0.25\textwidth]{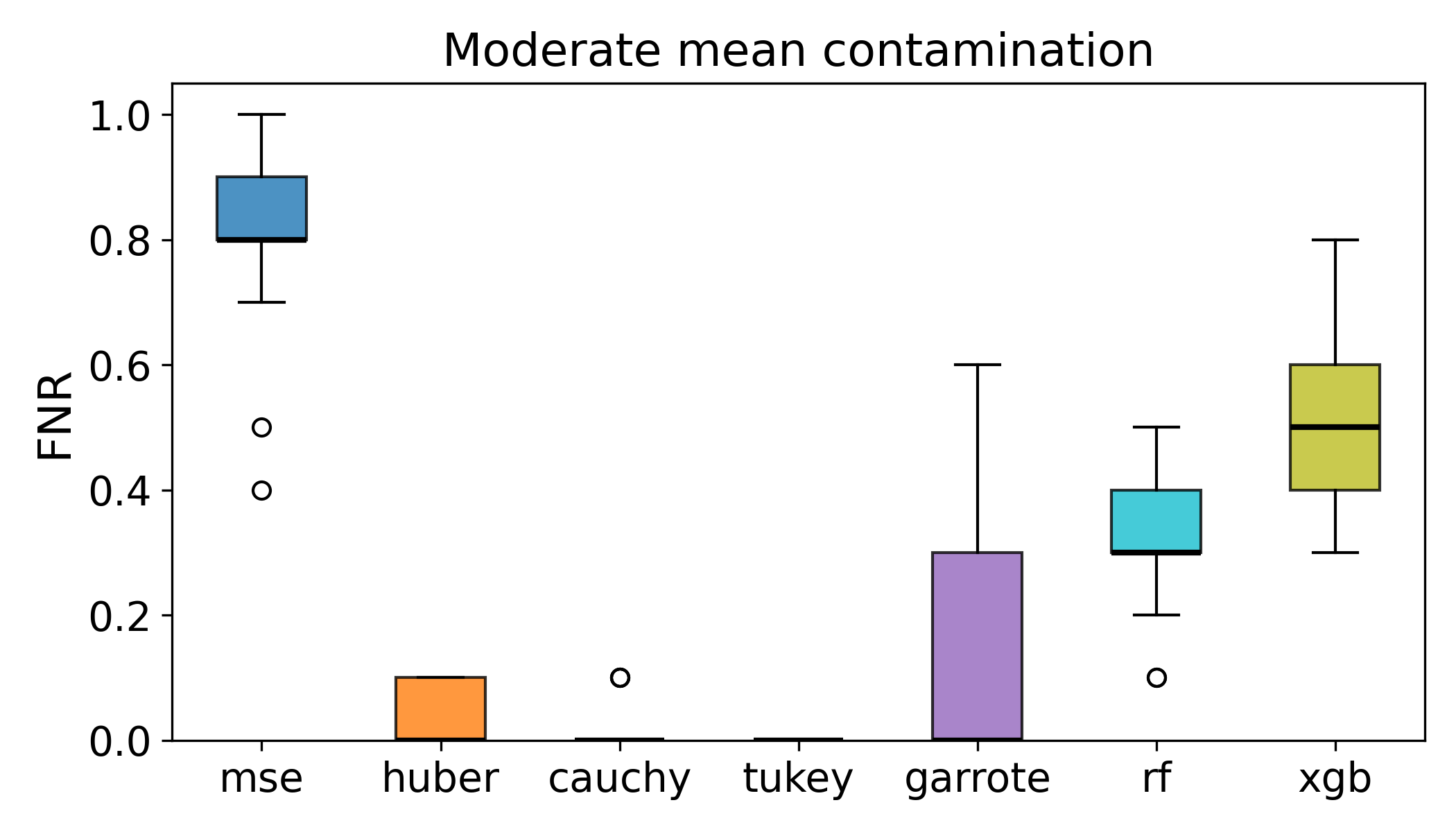}
	\includegraphics[width=0.32\textwidth,height=0.25\textwidth]{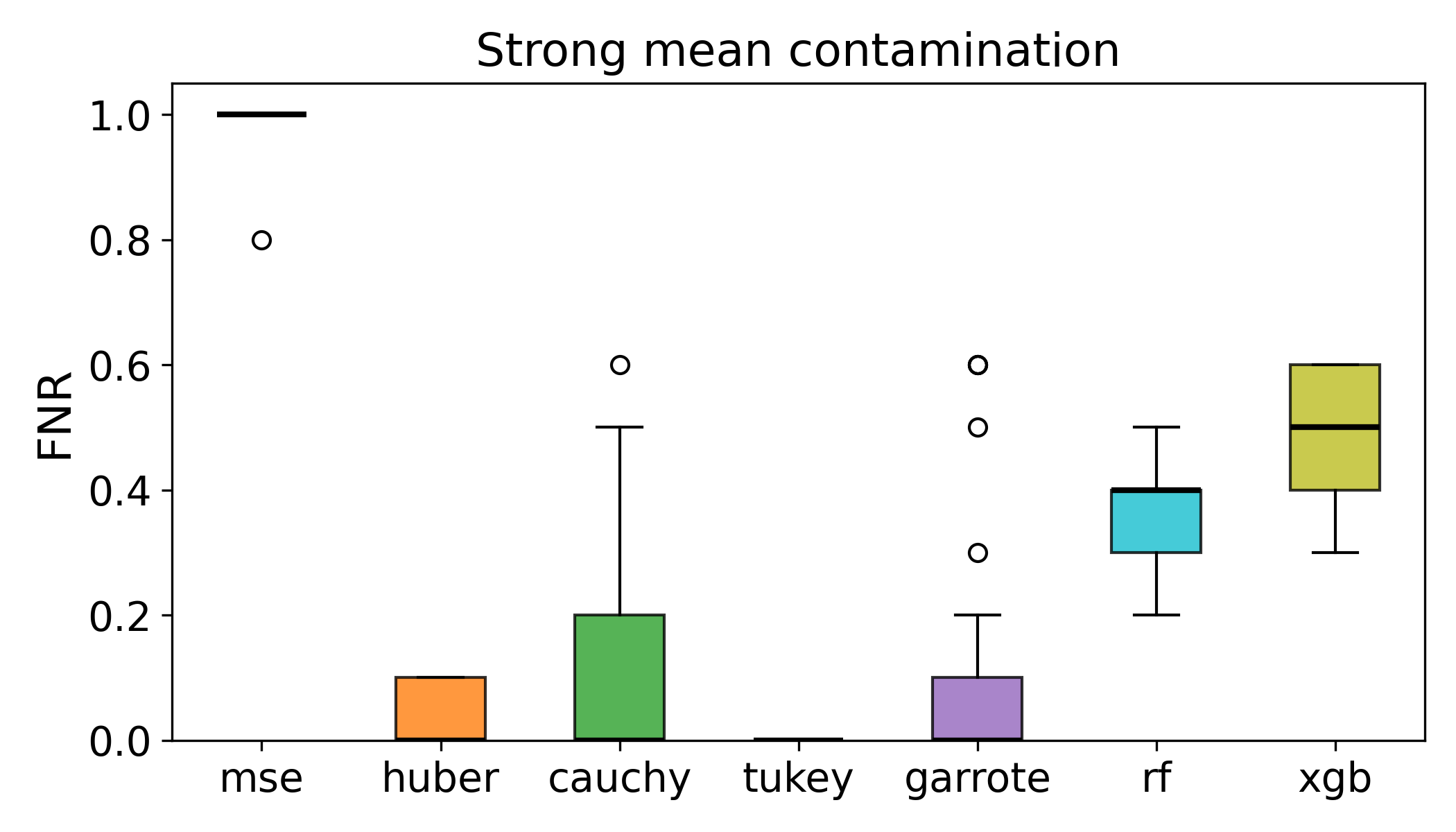}
	
	\caption{Boxplots of the performance metrics results for {\tt model 4} with $10\%$ contamination. First column refers to {\tt gaussian} scenario; second column to {\tt moderate mean} scenario; third column to {\tt strong mean} scenario. The first row shows a realization of the true and contaminated training response $\mathbf{y}$ for all the scenarios.}
	\label{fig_boxplot_model4_mean}
	
\end{figure*}

%%%%%%%%%%%%%%%%%%%%%%%%%%%%%%%%%%%%%%%%%%%%%%%%%%%%%%%%%%%%%%%
% boxplots model 1 mixture variance
\begin{figure*}[t]
	\centering
	
	% --- Row 1 ---
	\includegraphics[width=0.32\textwidth,height=0.25\textwidth]{model1/response_n1000_p200_N_RUNS25_M10_gaussian10.png}
	\includegraphics[width=0.32\textwidth,height=0.25\textwidth]{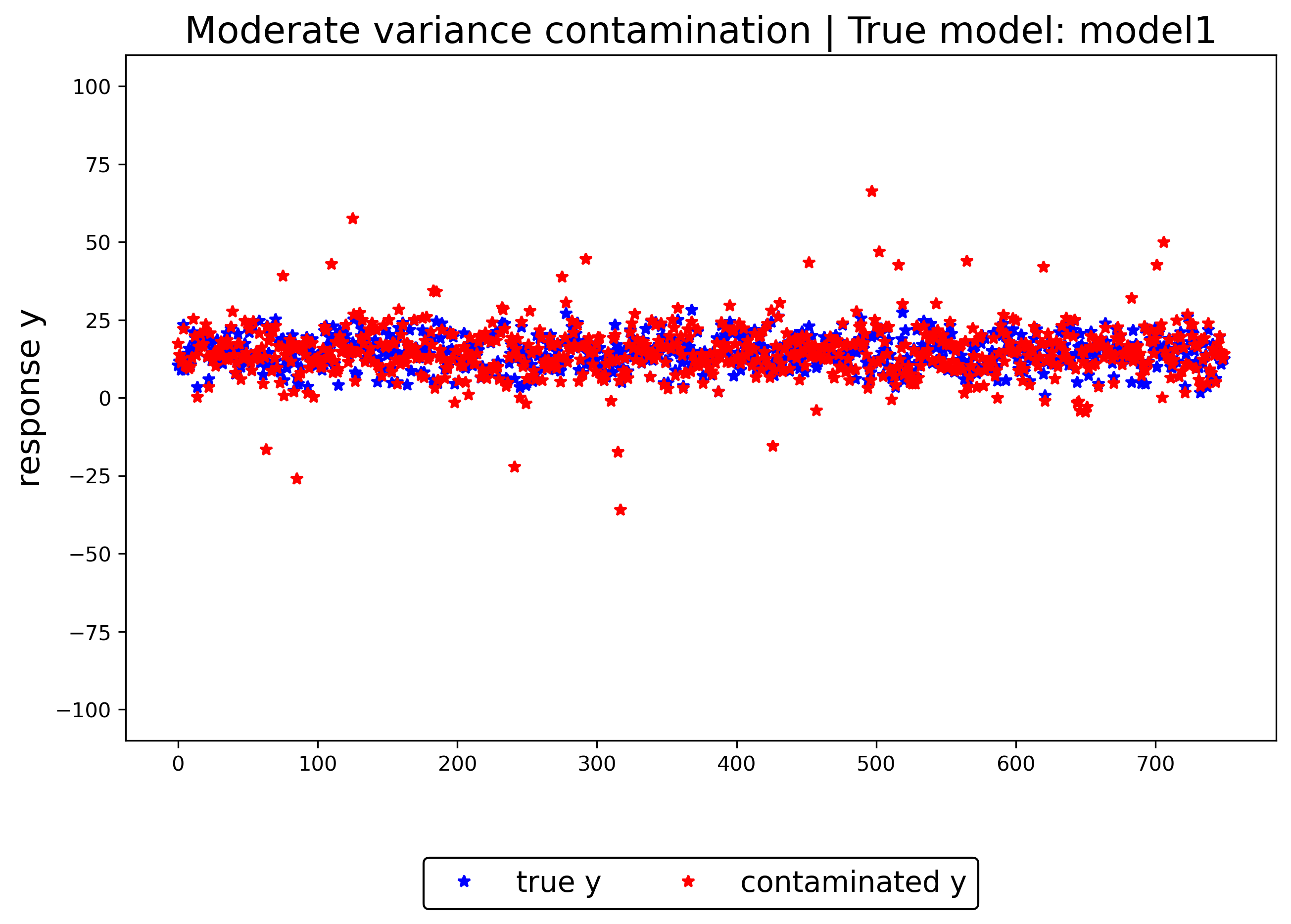}
	\includegraphics[width=0.32\textwidth,height=0.25\textwidth]{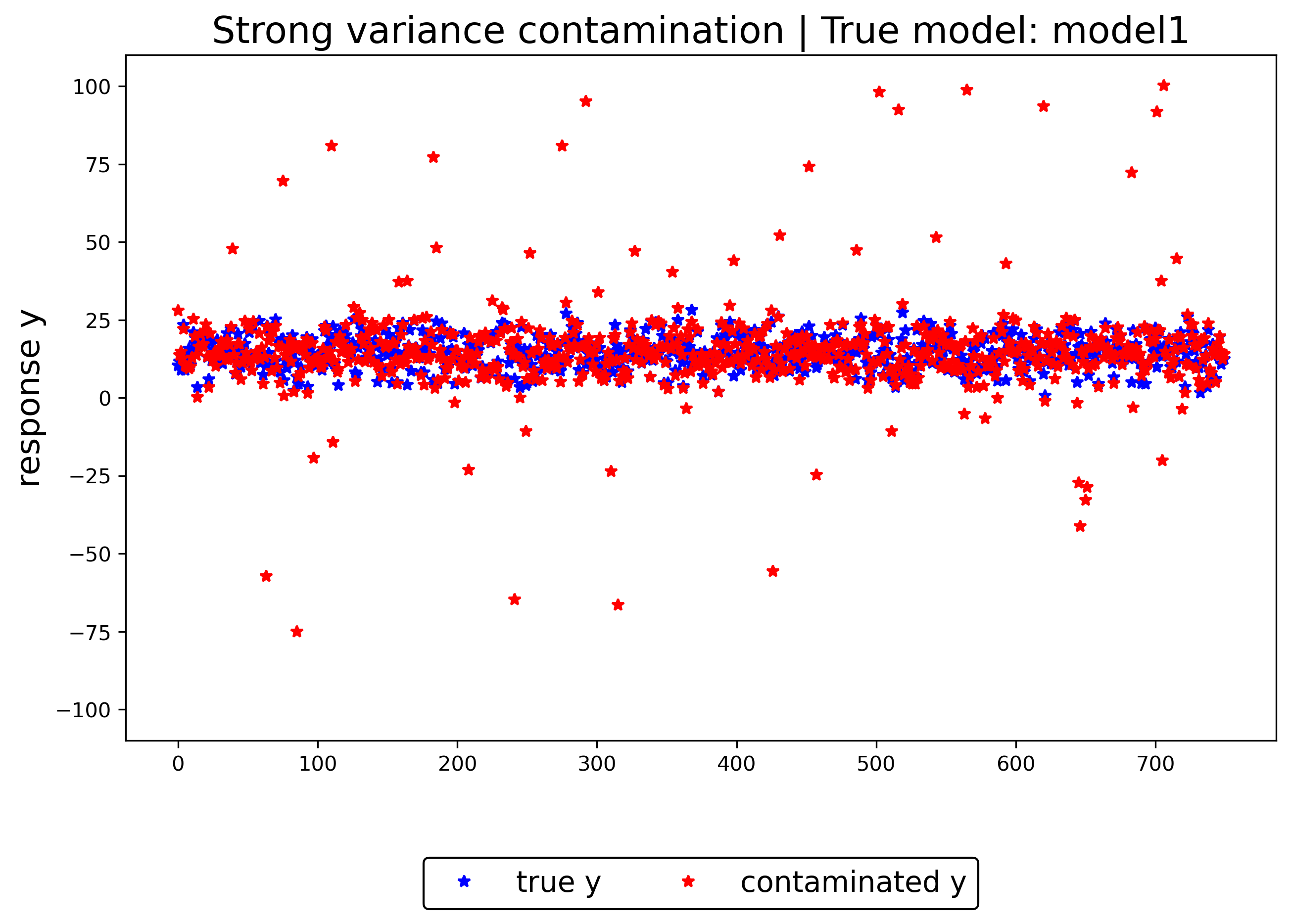}

	% --- Row 2 ---
	\includegraphics[width=0.32\textwidth,height=0.25\textwidth]{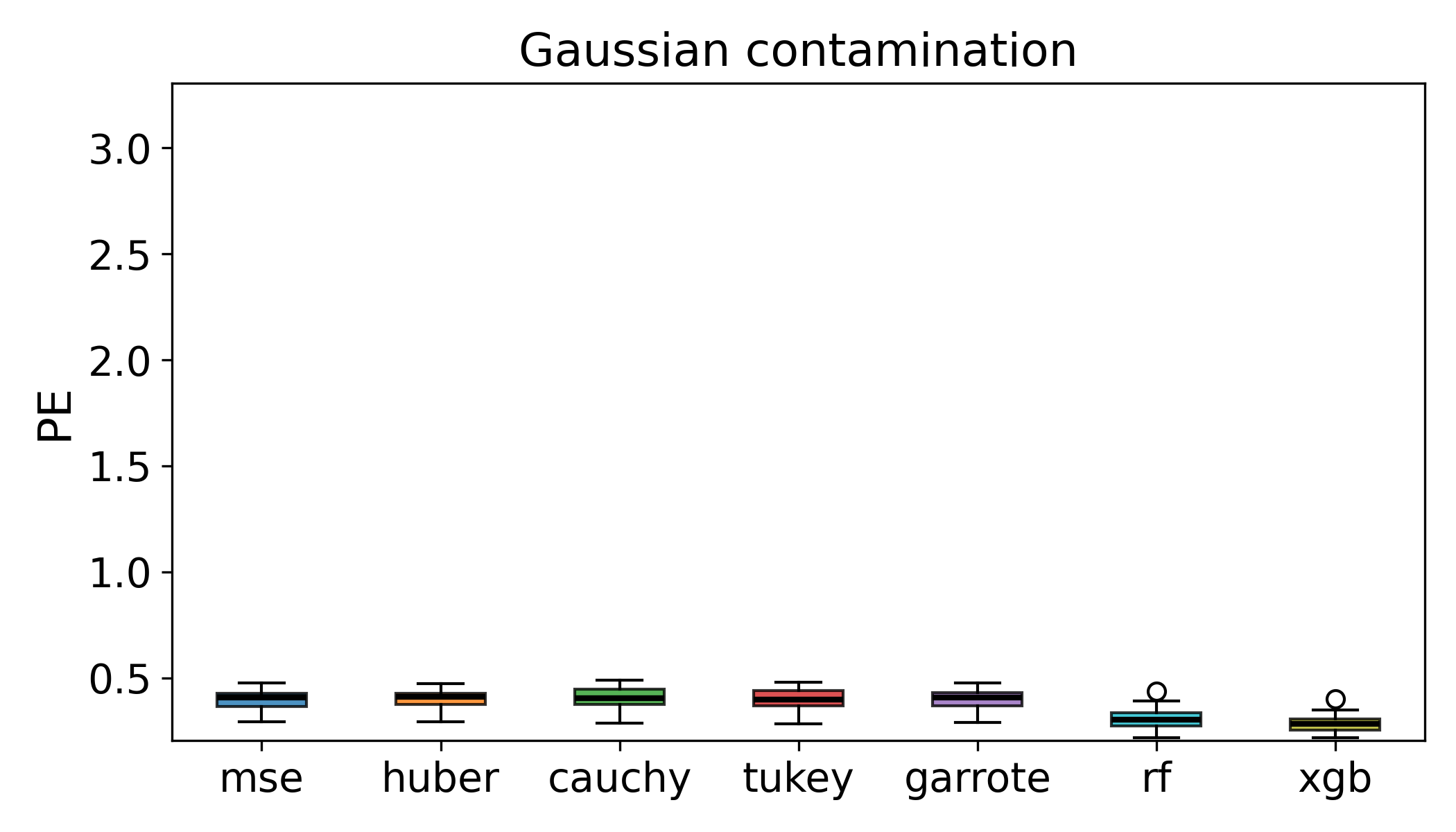}
	\includegraphics[width=0.32\textwidth,height=0.25\textwidth]{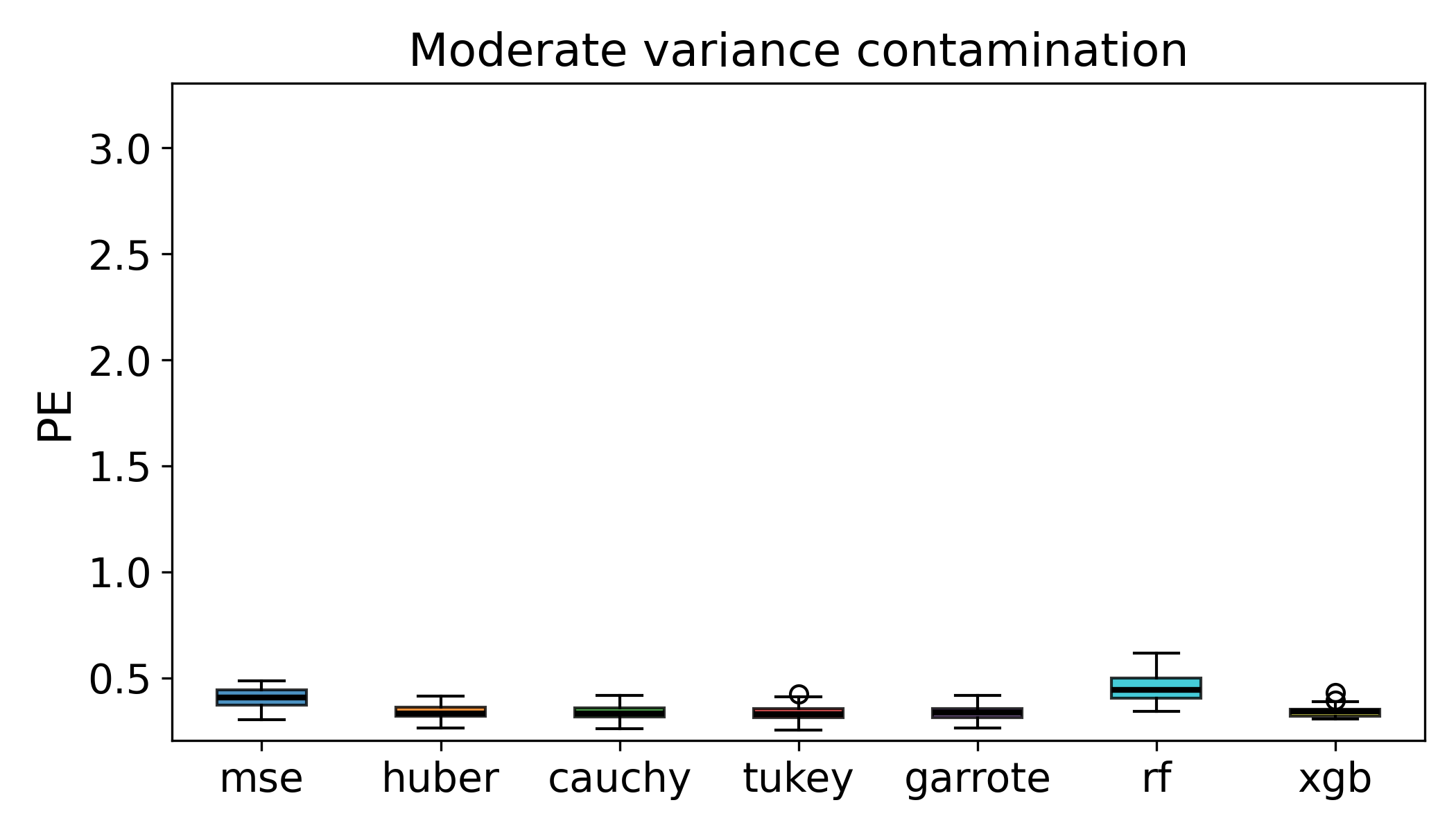}
	\includegraphics[width=0.32\textwidth,height=0.25\textwidth]{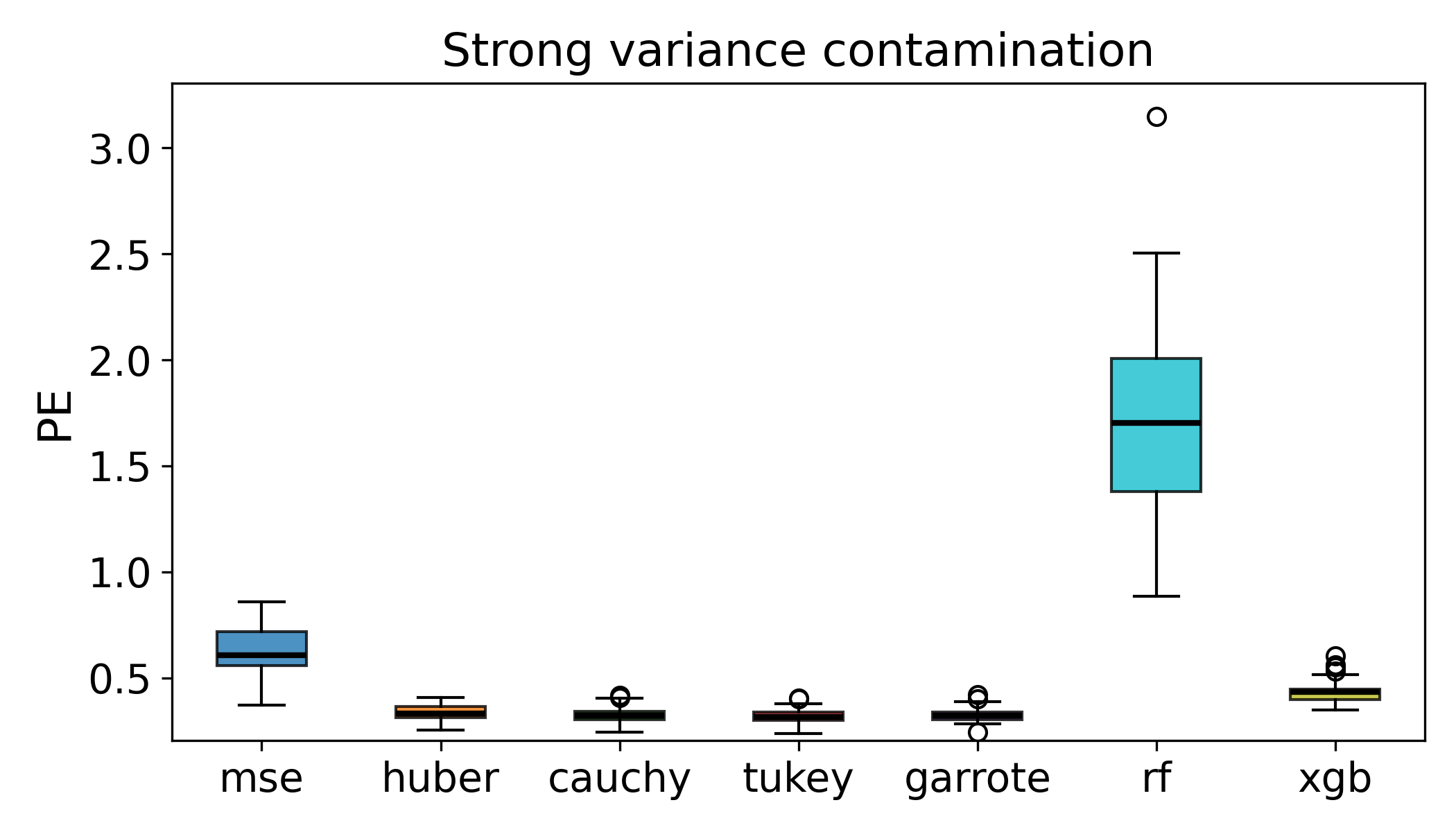}
	
	\vspace{0.2cm}
	
	% --- Row 3 ---
	\includegraphics[width=0.32\textwidth,height=0.25\textwidth]{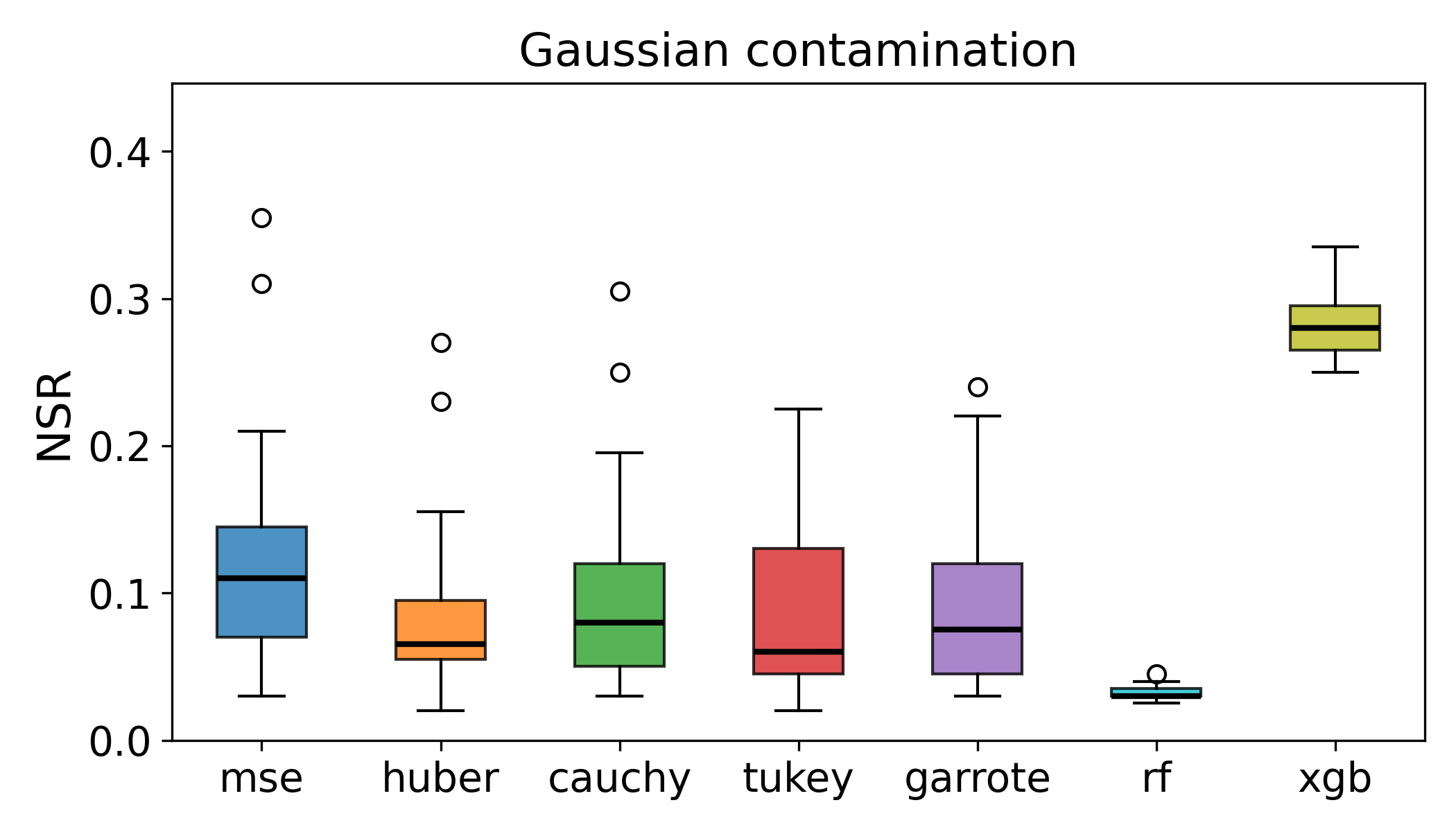}
	\includegraphics[width=0.32\textwidth,height=0.25\textwidth]{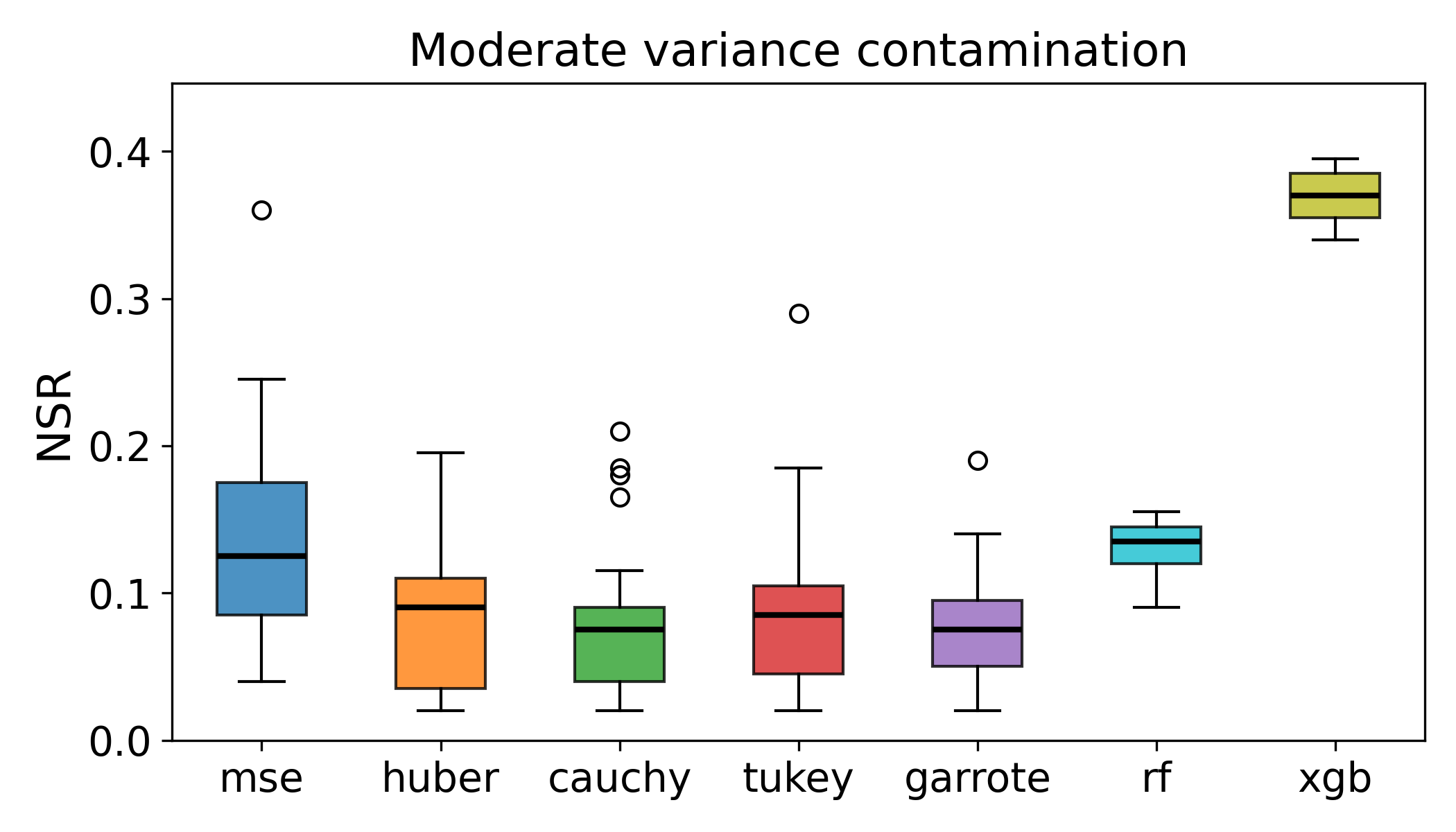}
	\includegraphics[width=0.32\textwidth,height=0.25\textwidth]{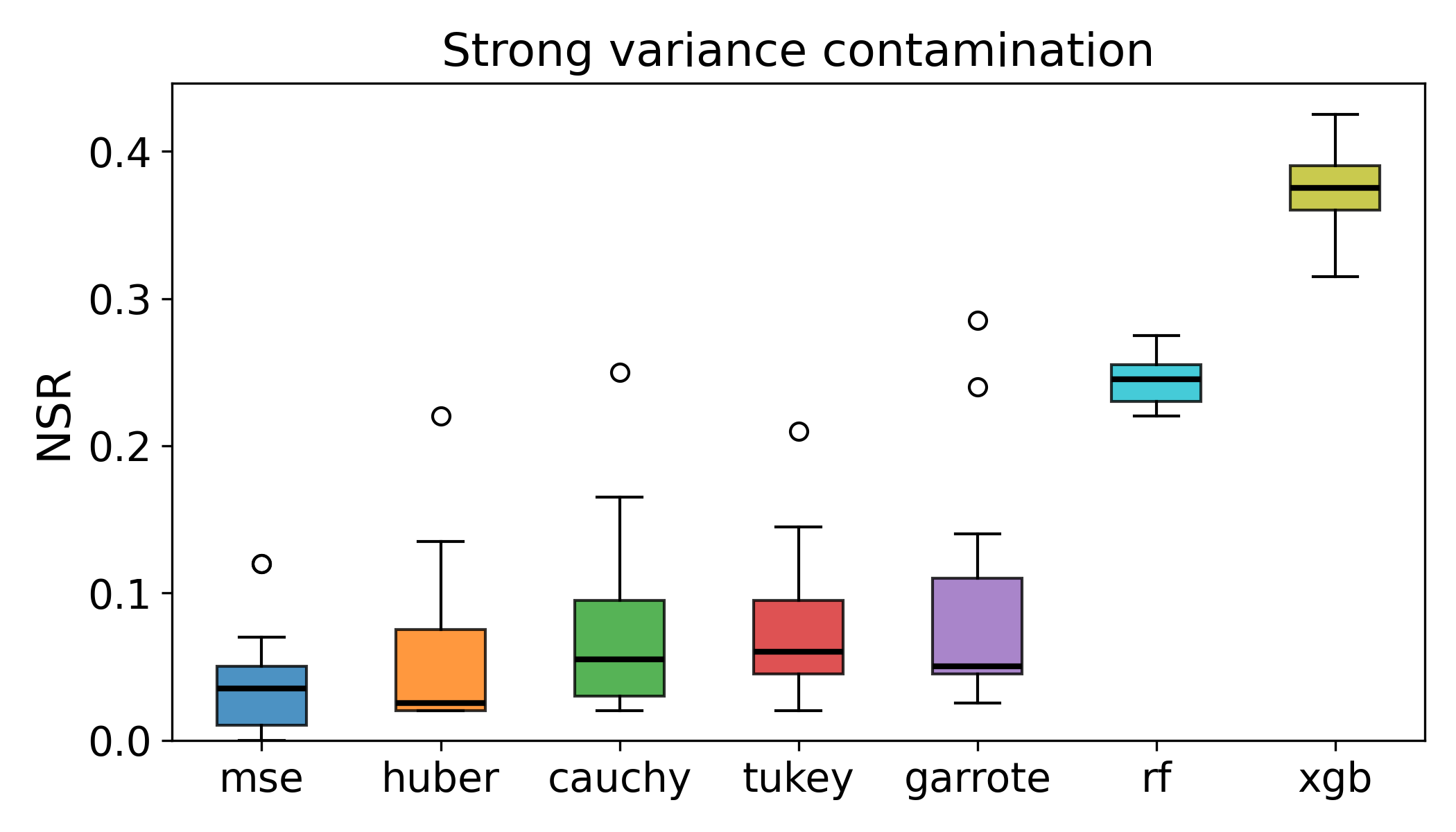}
	
	\vspace{0.2cm}
	
	% --- Row 4 ---
	\includegraphics[width=0.32\textwidth,height=0.25\textwidth]{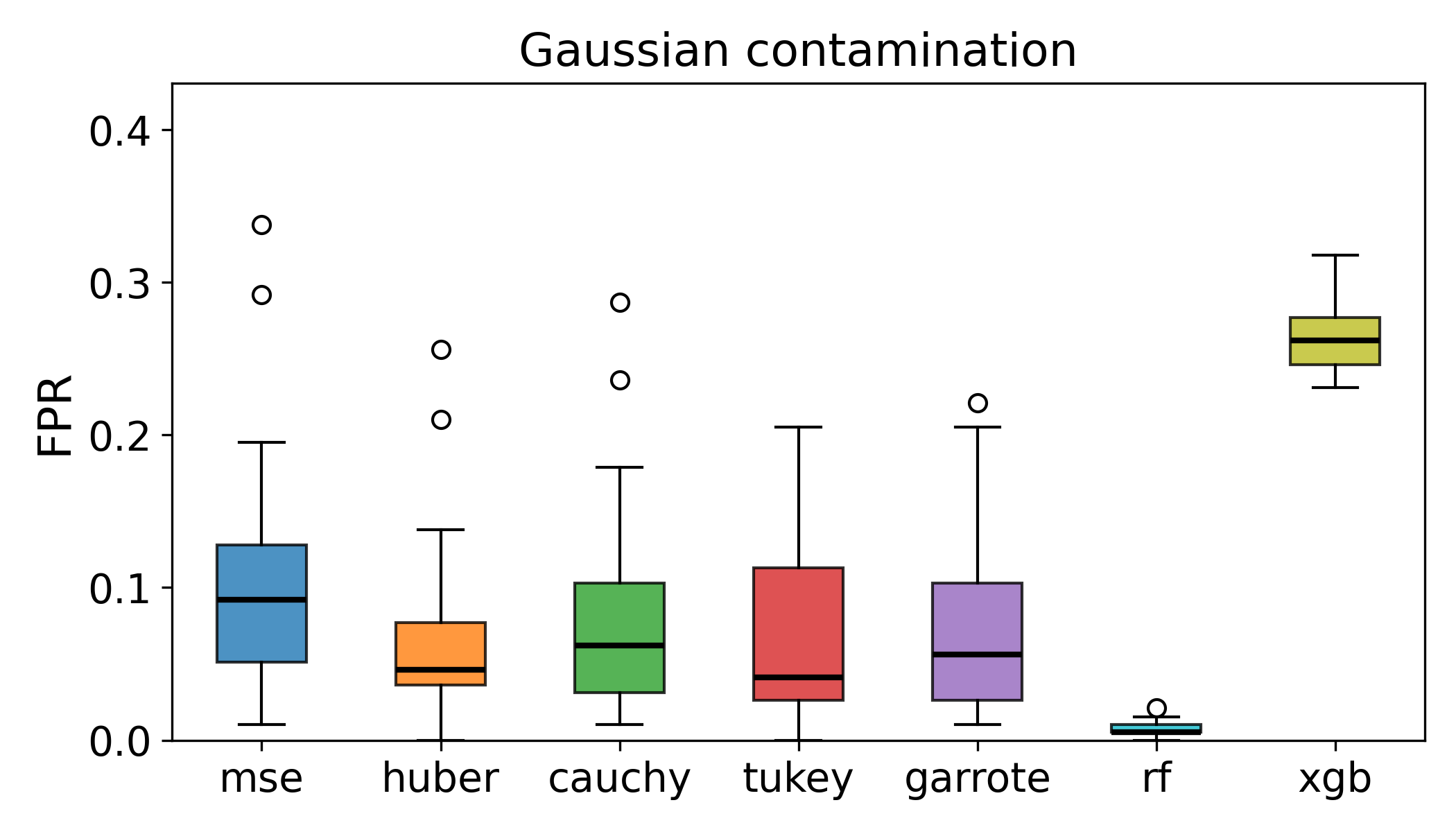}
	\includegraphics[width=0.32\textwidth,height=0.25\textwidth]{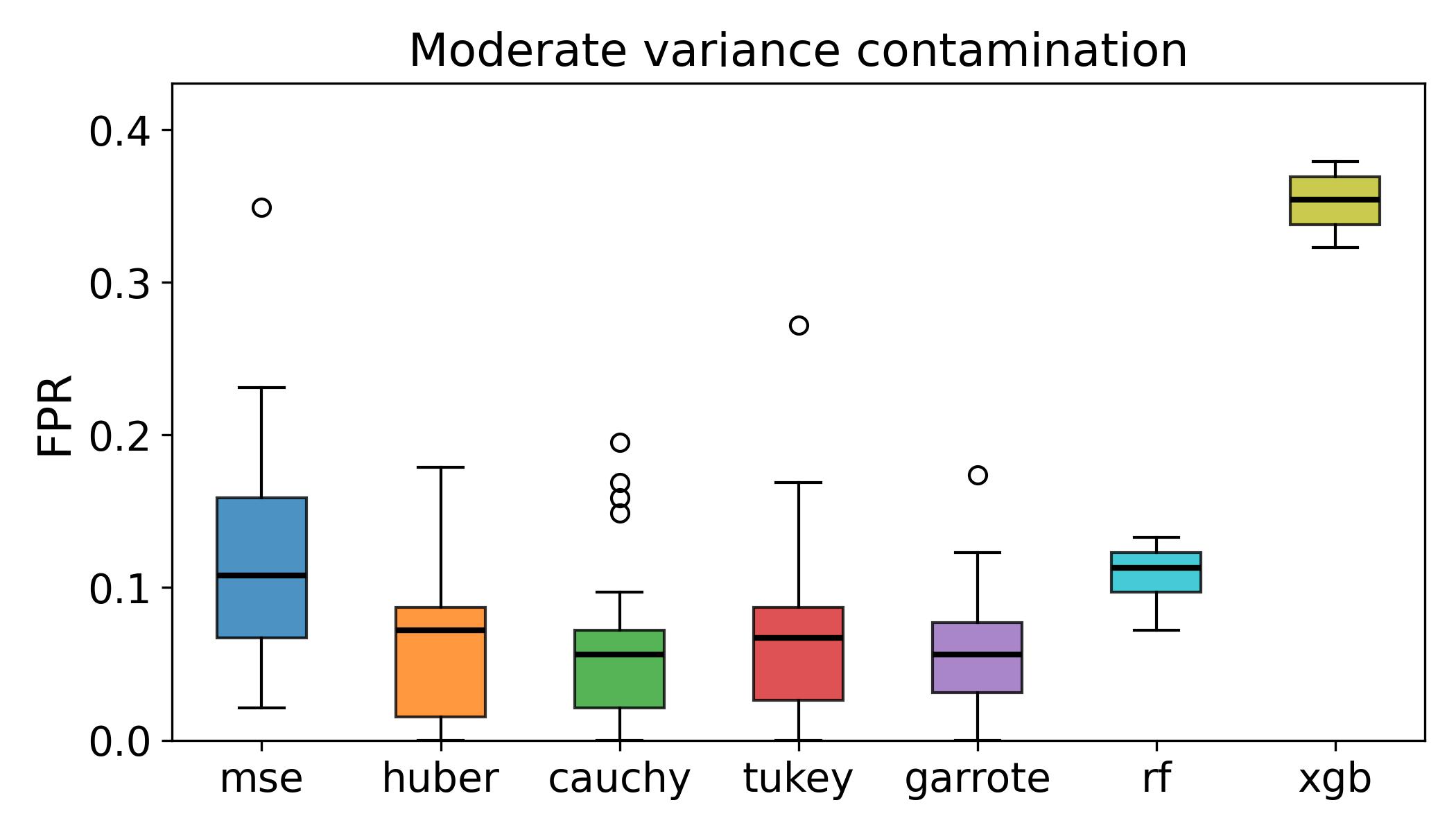}
	\includegraphics[width=0.32\textwidth,height=0.25\textwidth]{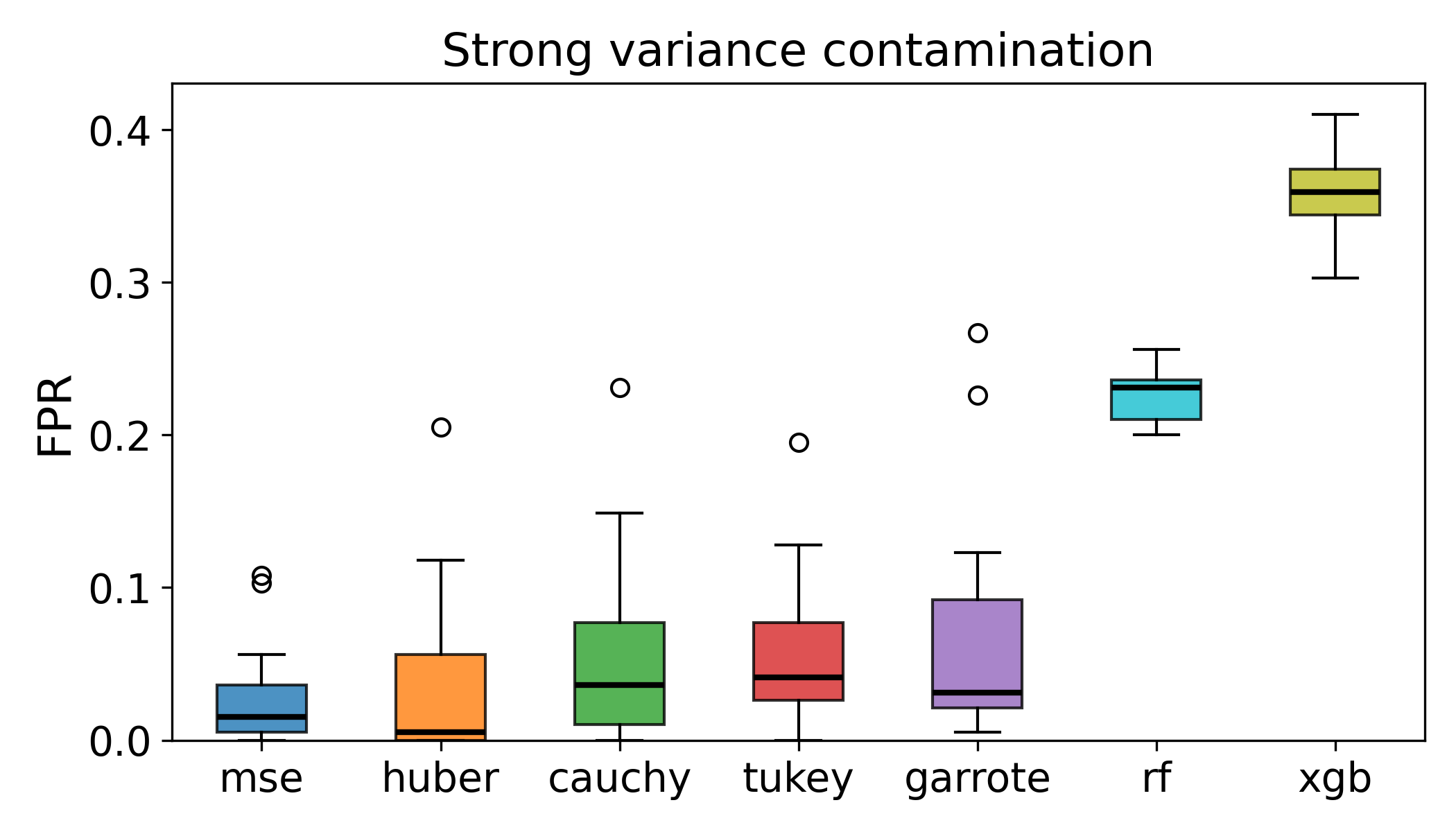}
	
	\vspace{0.2cm}
	
	% --- Row 5 ---
	\includegraphics[width=0.32\textwidth,height=0.25\textwidth]{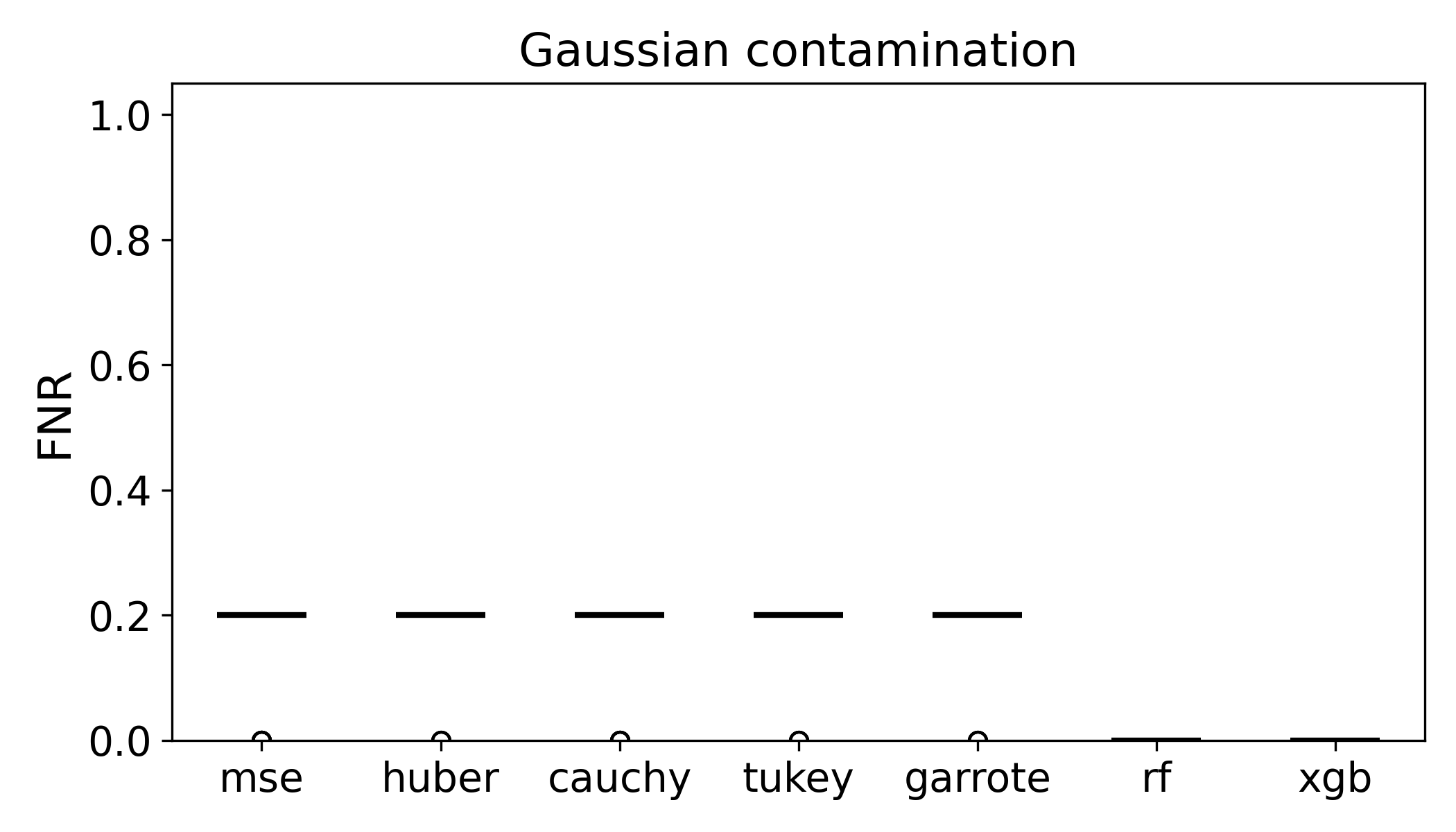}
	\includegraphics[width=0.32\textwidth,height=0.25\textwidth]{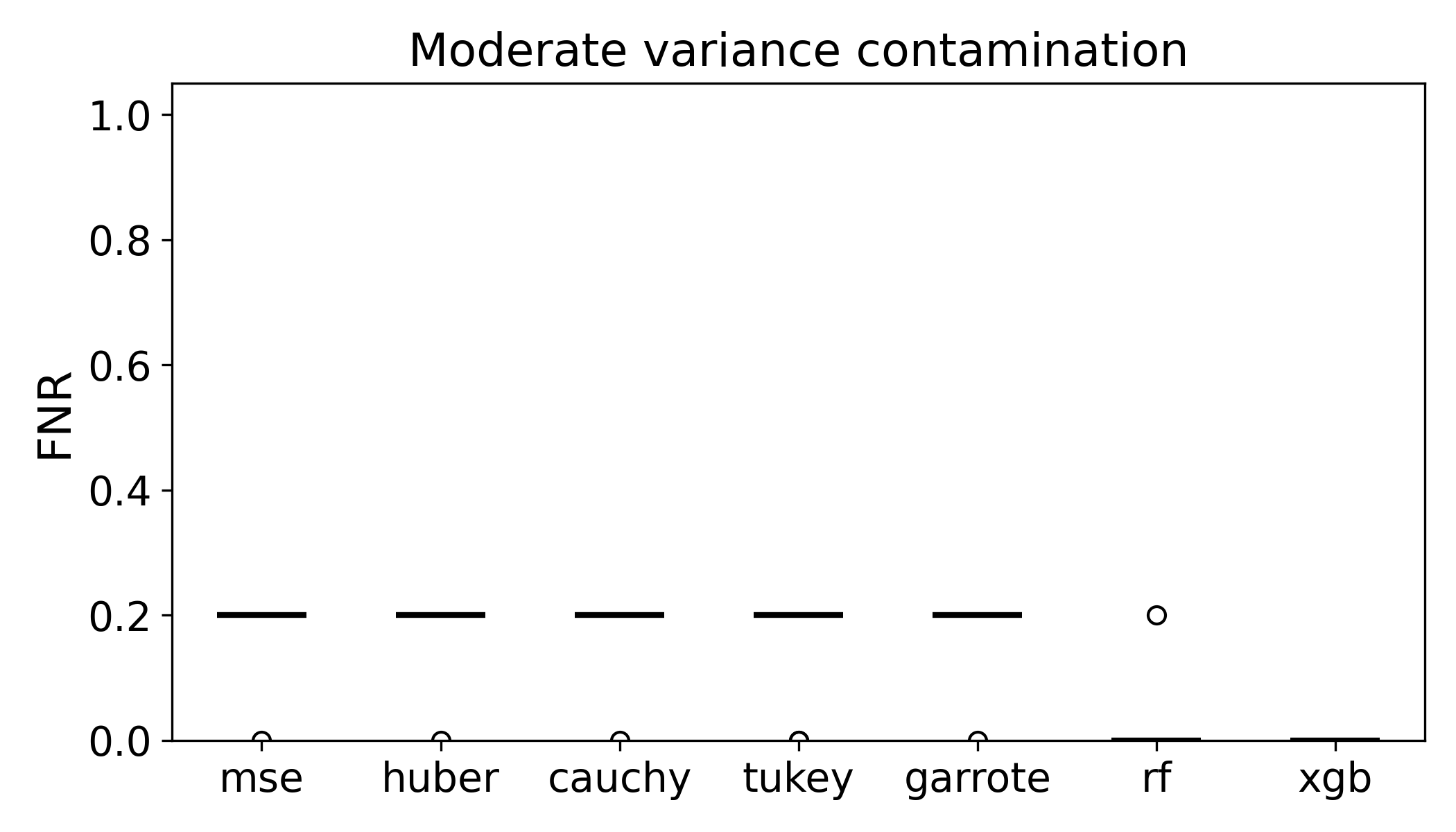}
	\includegraphics[width=0.32\textwidth,height=0.25\textwidth]{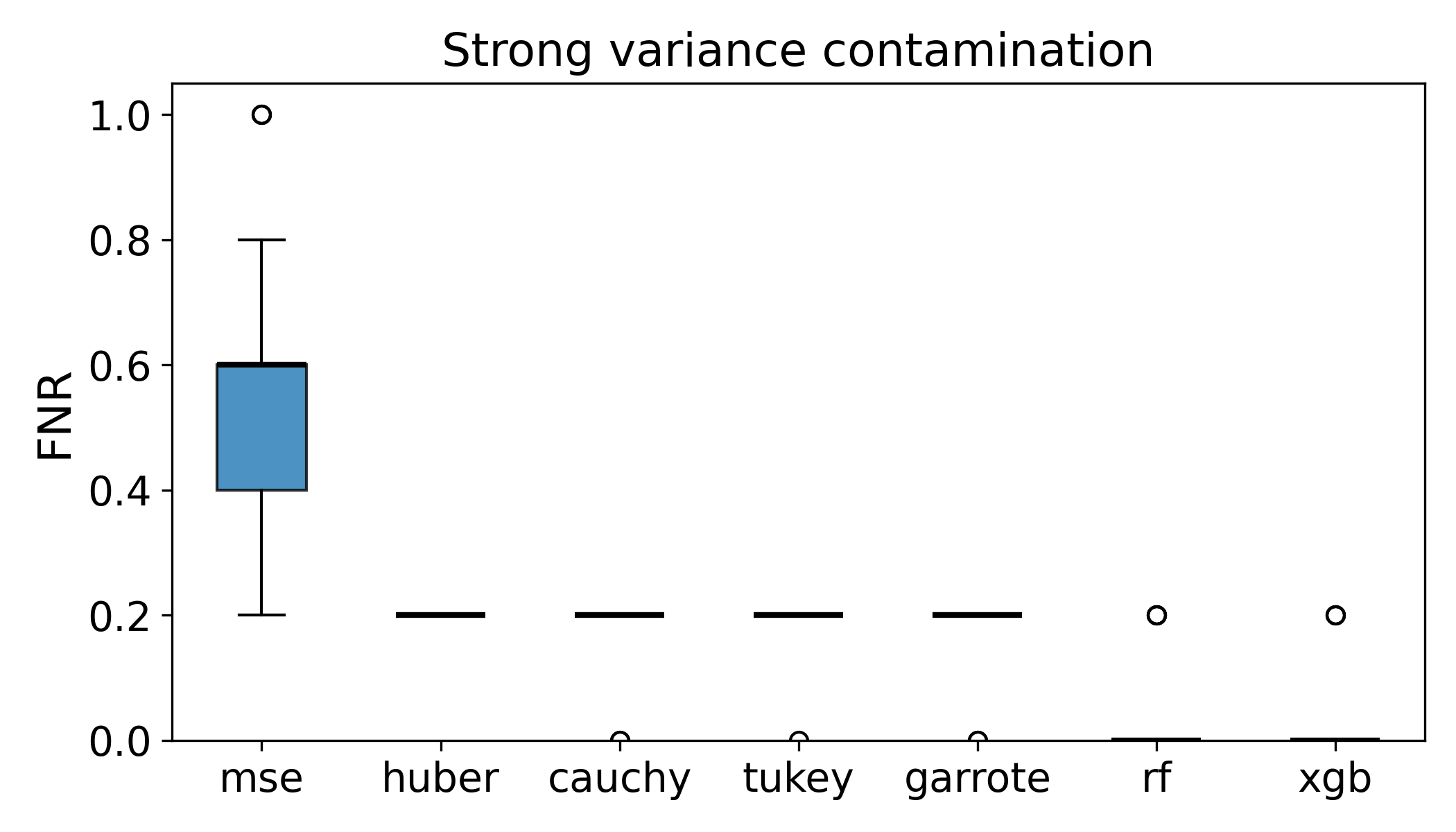}
	
	\caption{Boxplots of the performance metrics results for {\tt model 1} with $10\%$ contamination. First column refers to {\tt gaussian} scenario; second column to {\tt moderate variance} scenario; third column to {\tt strong variance} scenario. The first row shows a realization of the true and contaminated training response $\mathbf{y}$ for all the scenarios.}
	\label{fig_boxplot_model1_variance}
	
\end{figure*}

% boxplots model 2 mixture variance
\begin{figure*}[t]
	\centering
	
	% --- Row 1 ---
	\includegraphics[width=0.32\textwidth,height=0.25\textwidth]{model2/response_n1000_p200_N_RUNS25_M10_gaussian10.png}
	\includegraphics[width=0.32\textwidth,height=0.25\textwidth]{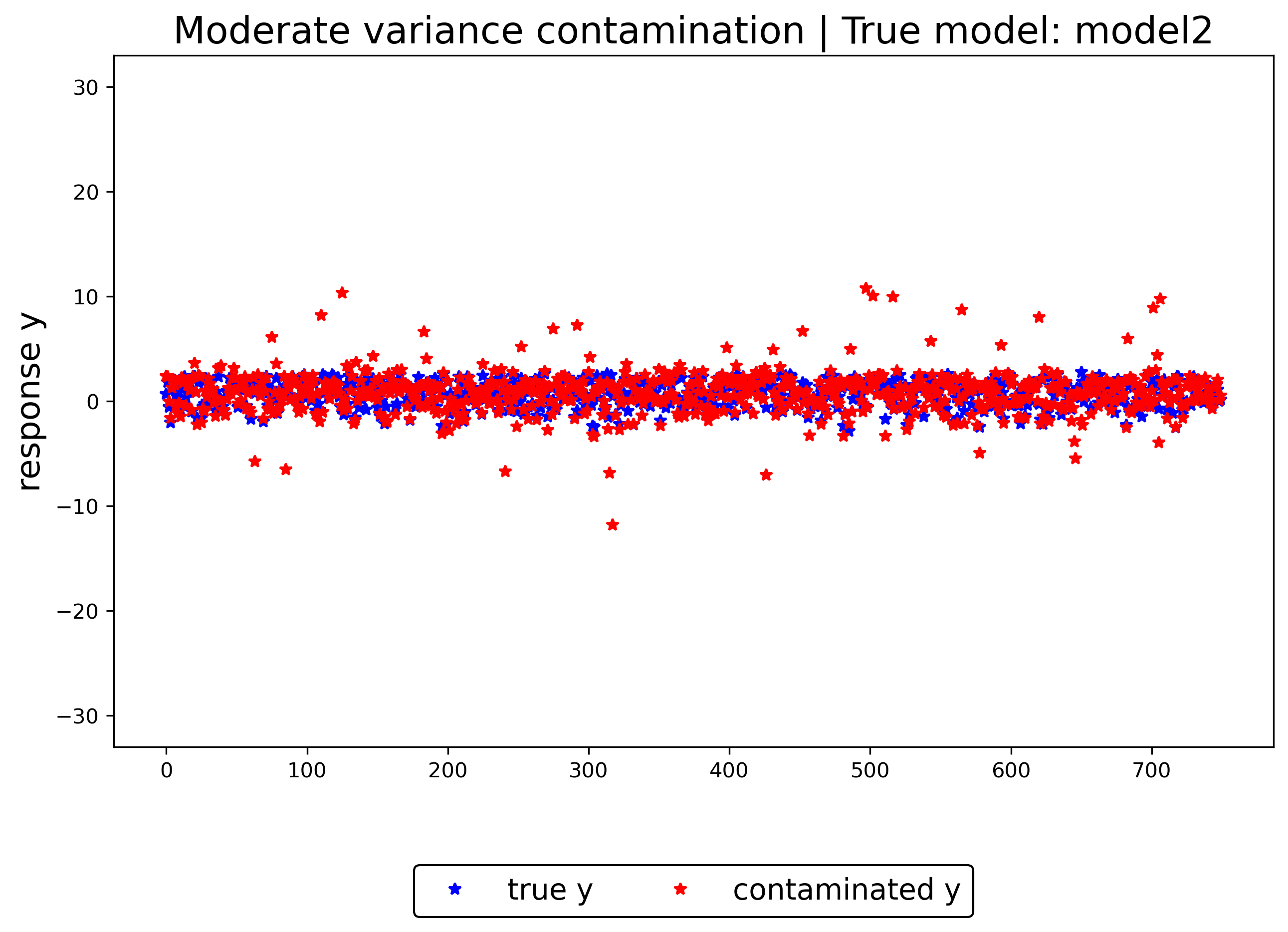}
	\includegraphics[width=0.32\textwidth,height=0.25\textwidth]{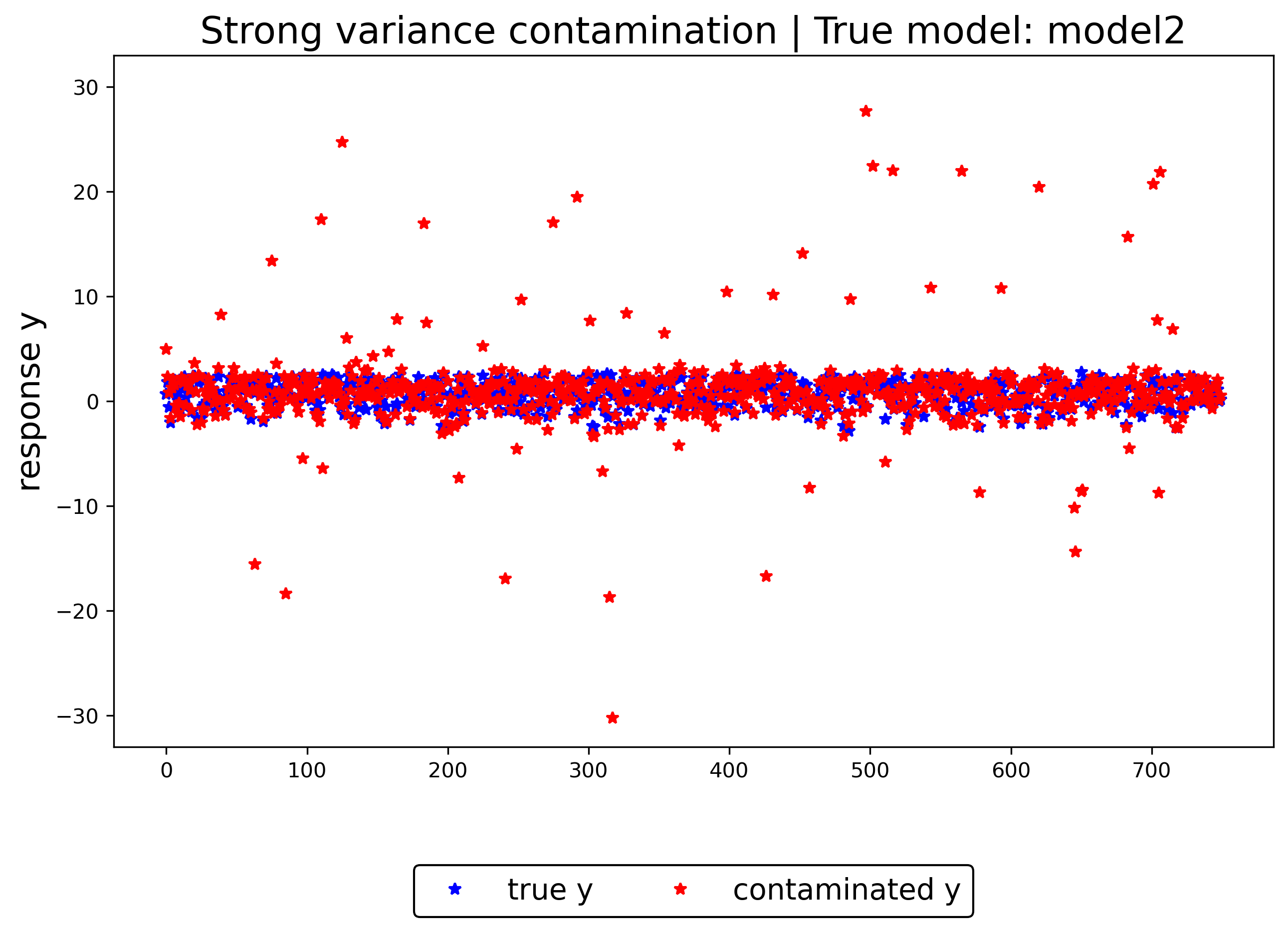}

	% --- Row 2 ---
	\includegraphics[width=0.32\textwidth,height=0.25\textwidth]{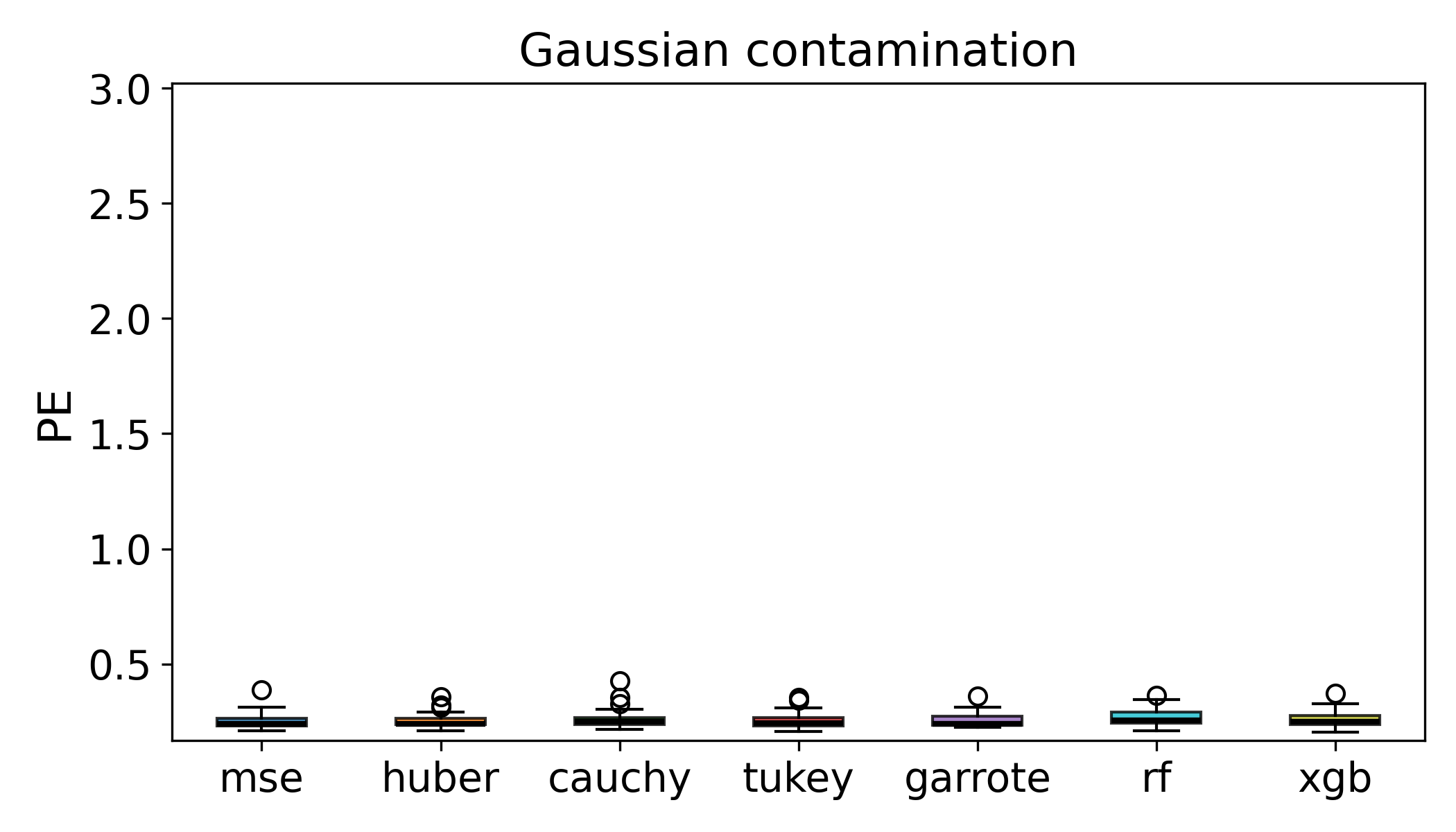}
	\includegraphics[width=0.32\textwidth,height=0.25\textwidth]{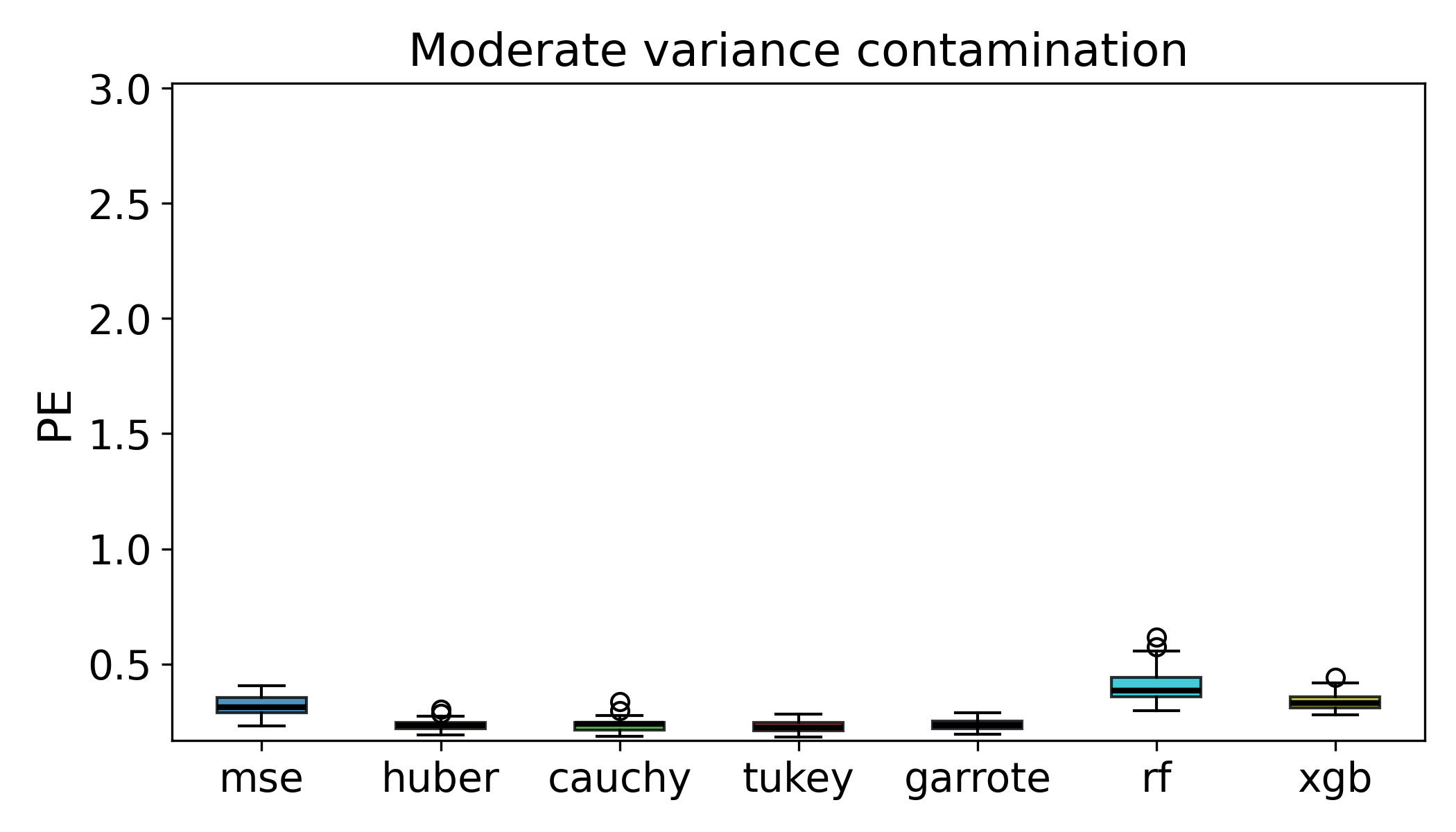}
	\includegraphics[width=0.32\textwidth,height=0.25\textwidth]{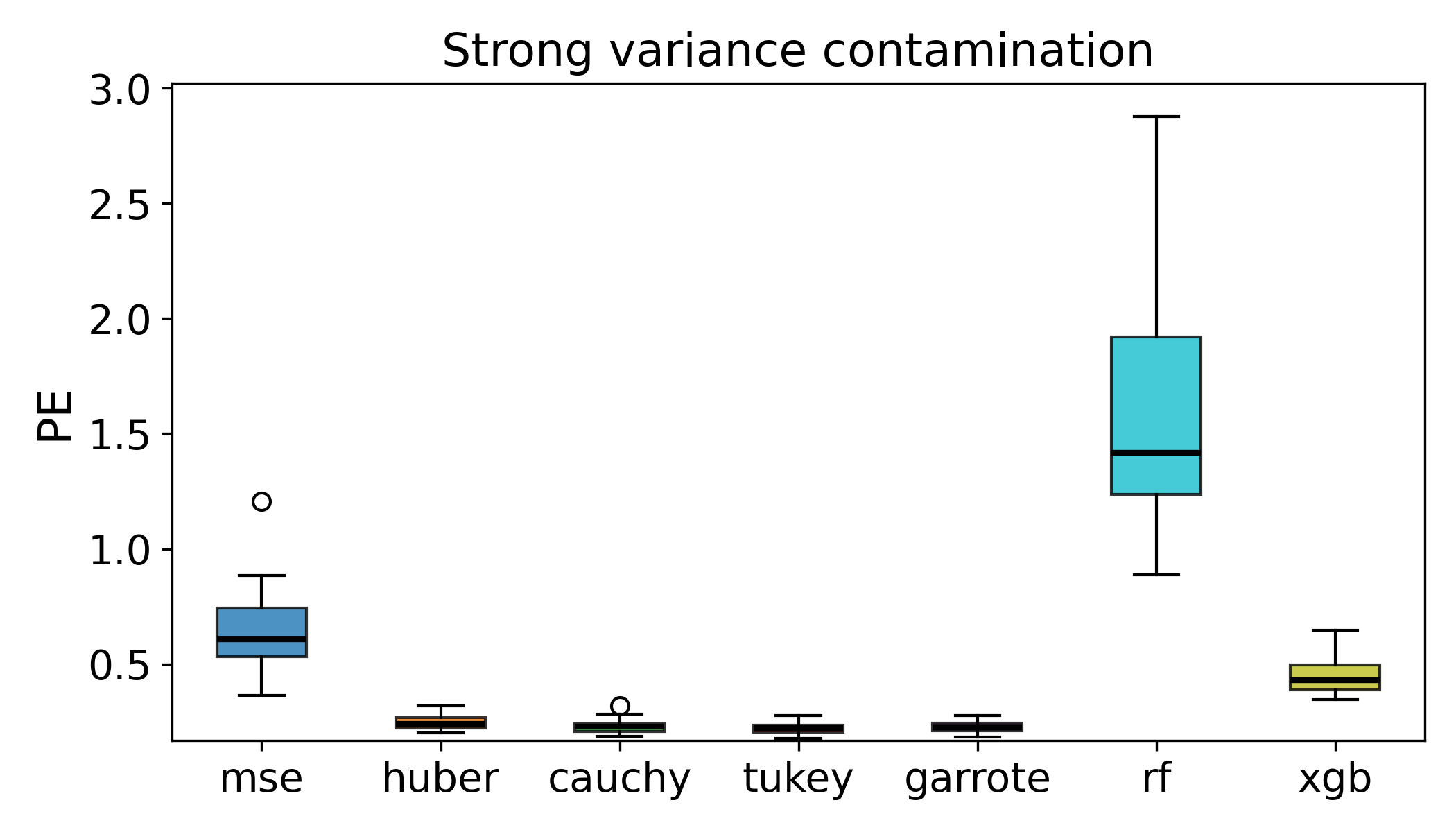}
	
	\vspace{0.2cm}
	
	% --- Row 3 ---
	\includegraphics[width=0.32\textwidth,height=0.25\textwidth]{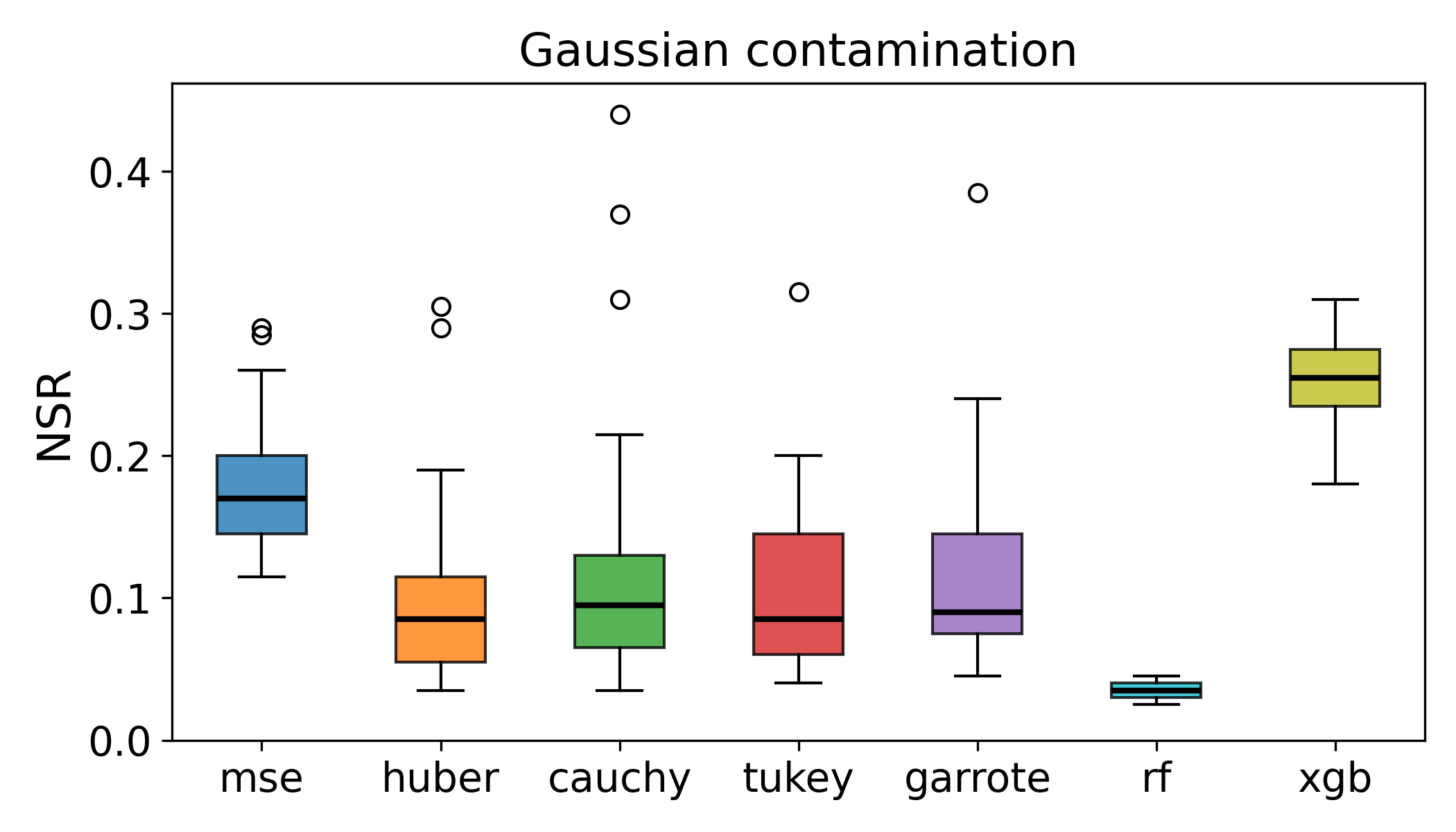}
	\includegraphics[width=0.32\textwidth,height=0.25\textwidth]{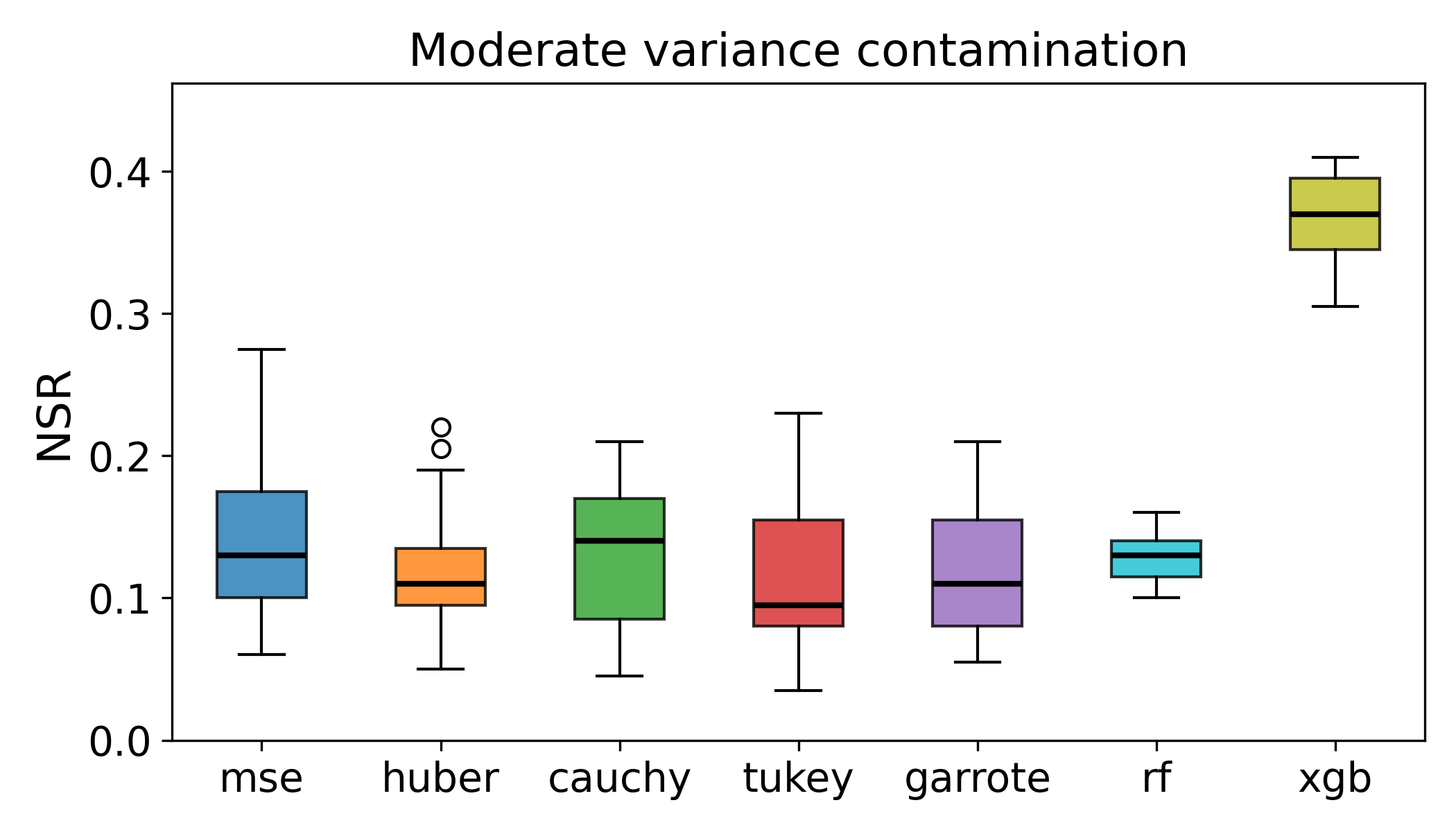}
	\includegraphics[width=0.32\textwidth,height=0.25\textwidth]{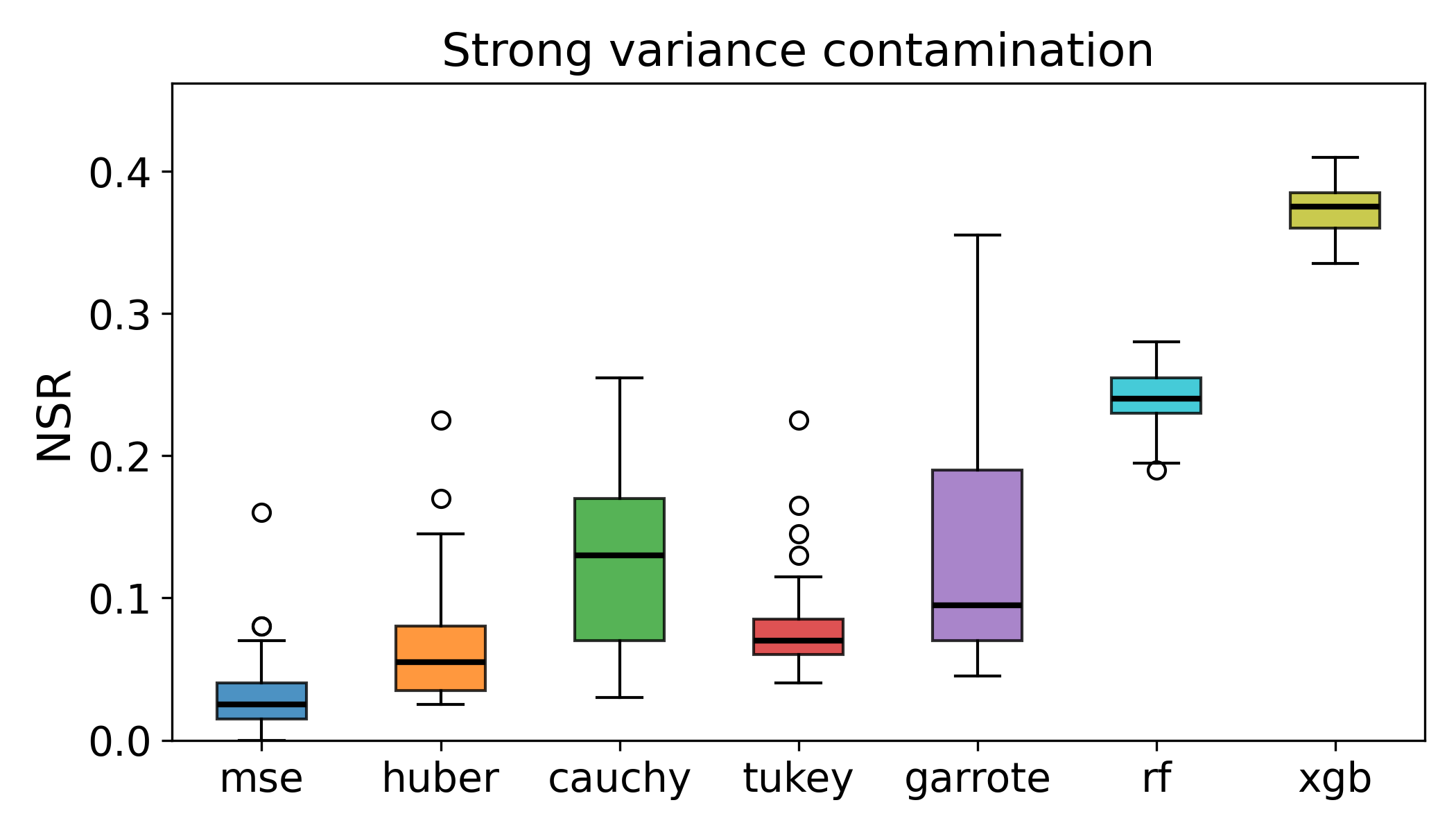}
	
	\vspace{0.2cm}
	
	% --- Row 4 ---
	\includegraphics[width=0.32\textwidth,height=0.25\textwidth]{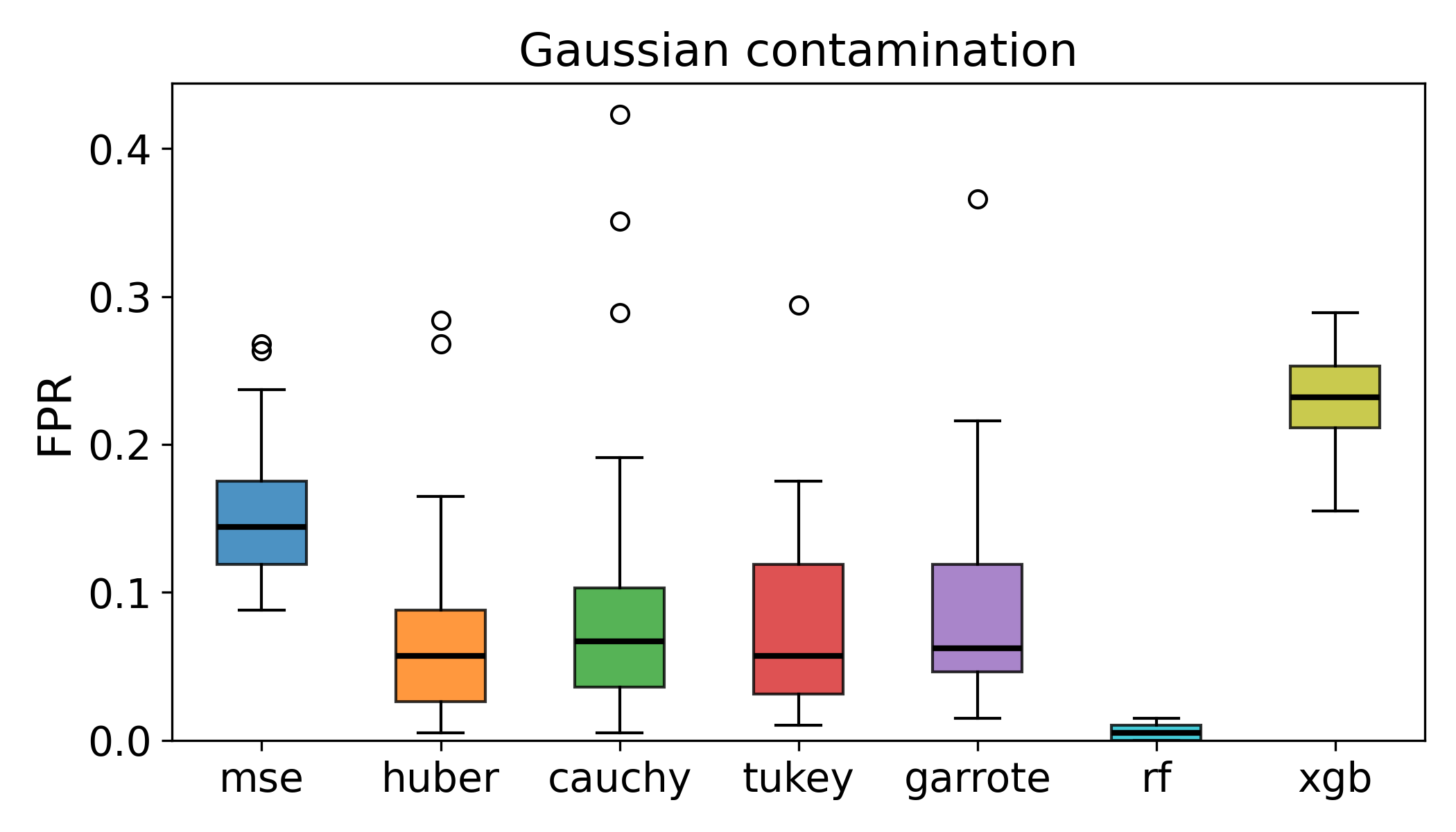}
	\includegraphics[width=0.32\textwidth,height=0.25\textwidth]{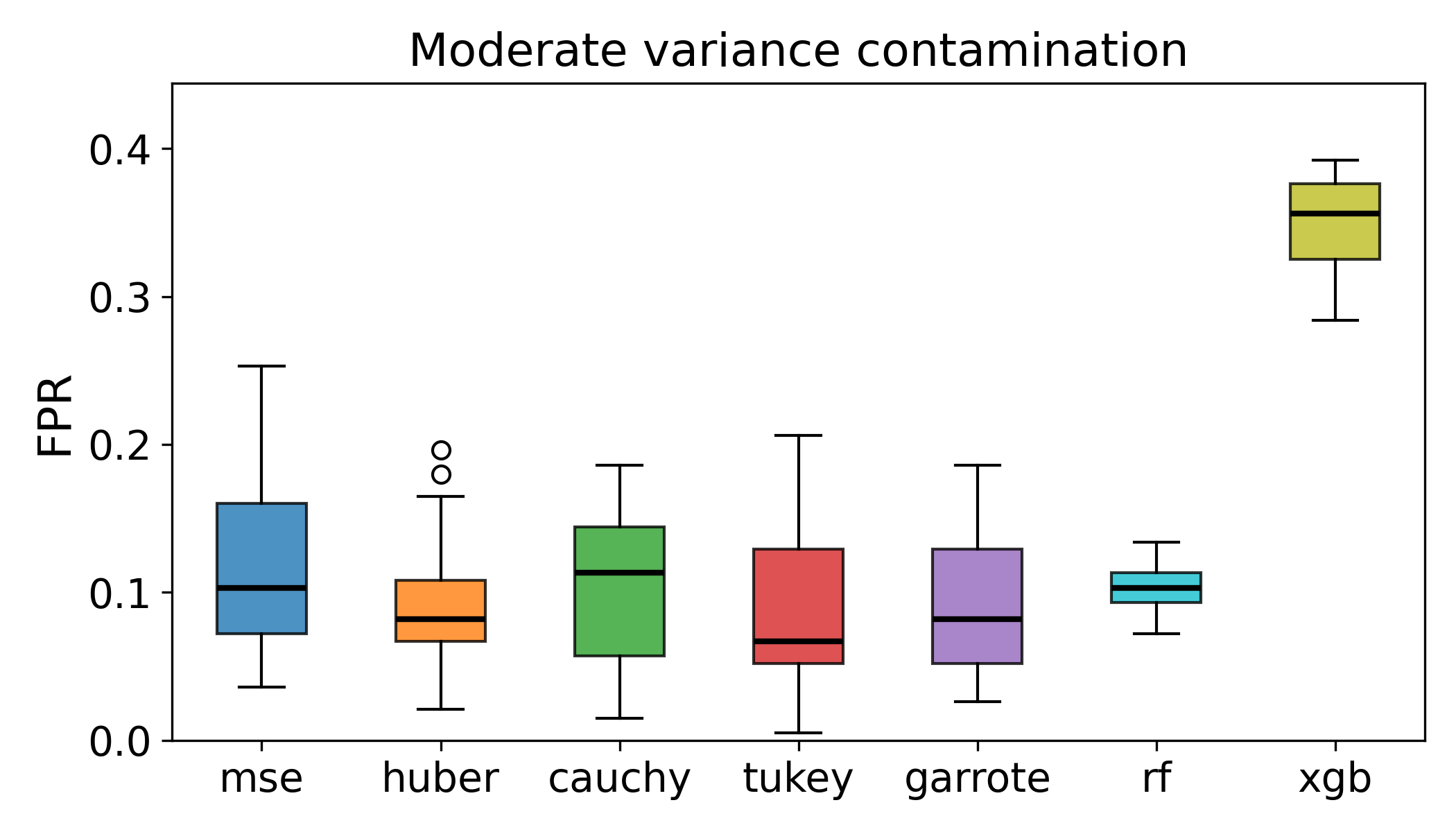}
	\includegraphics[width=0.32\textwidth,height=0.25\textwidth]{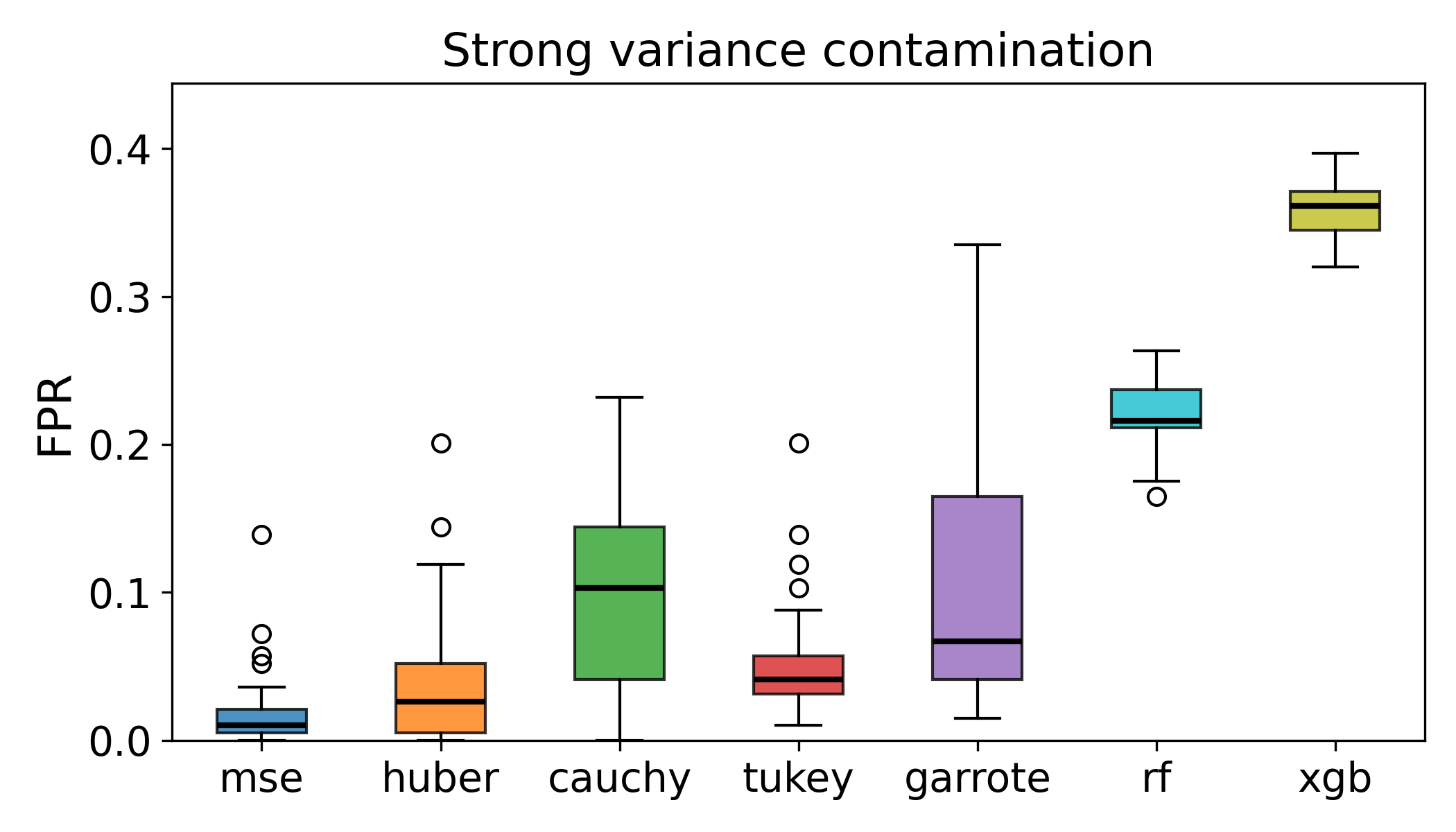}
	
	\vspace{0.2cm}
	
	% --- Row 5 ---
	\includegraphics[width=0.32\textwidth,height=0.25\textwidth]{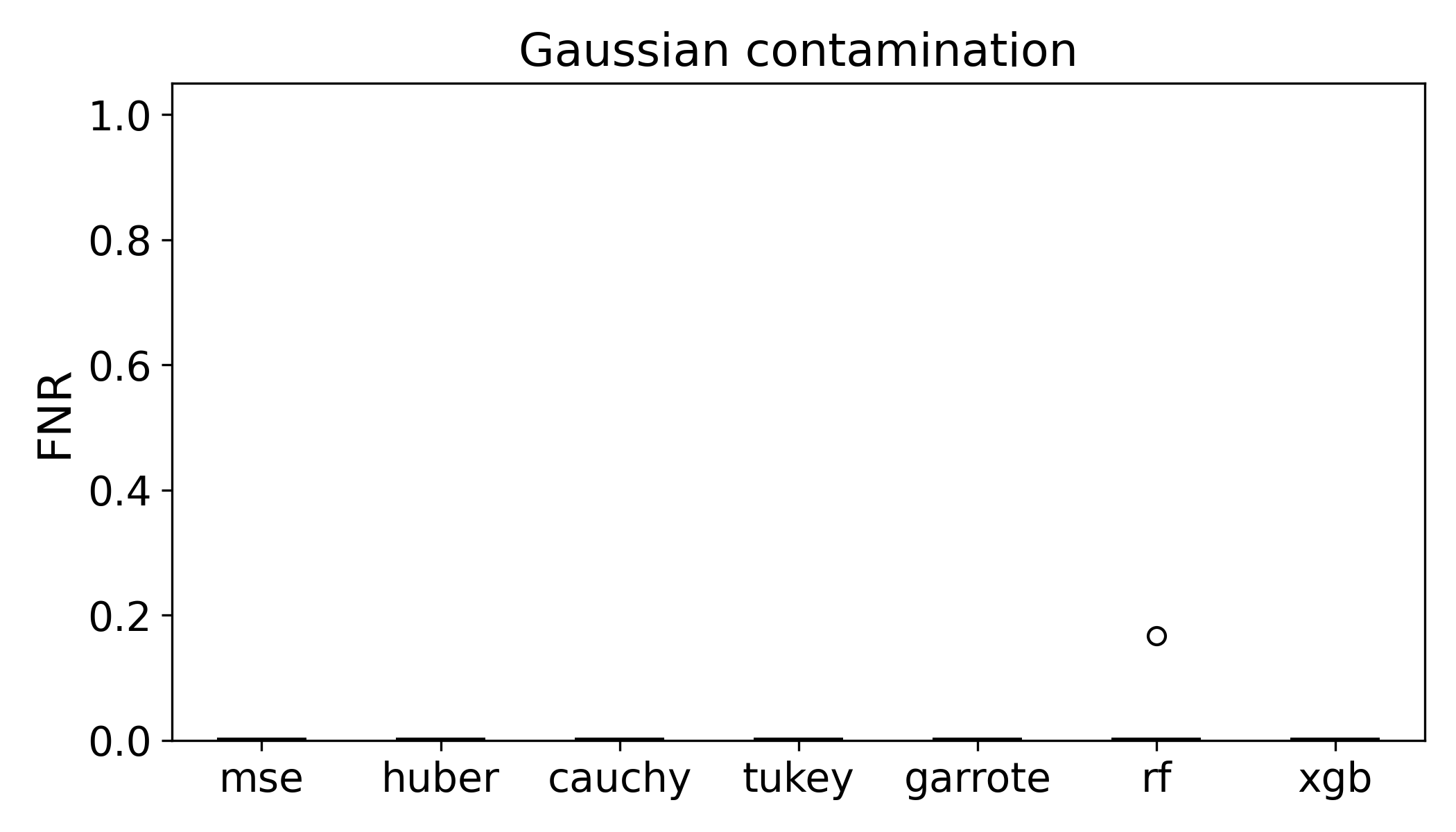}
	\includegraphics[width=0.32\textwidth,height=0.25\textwidth]{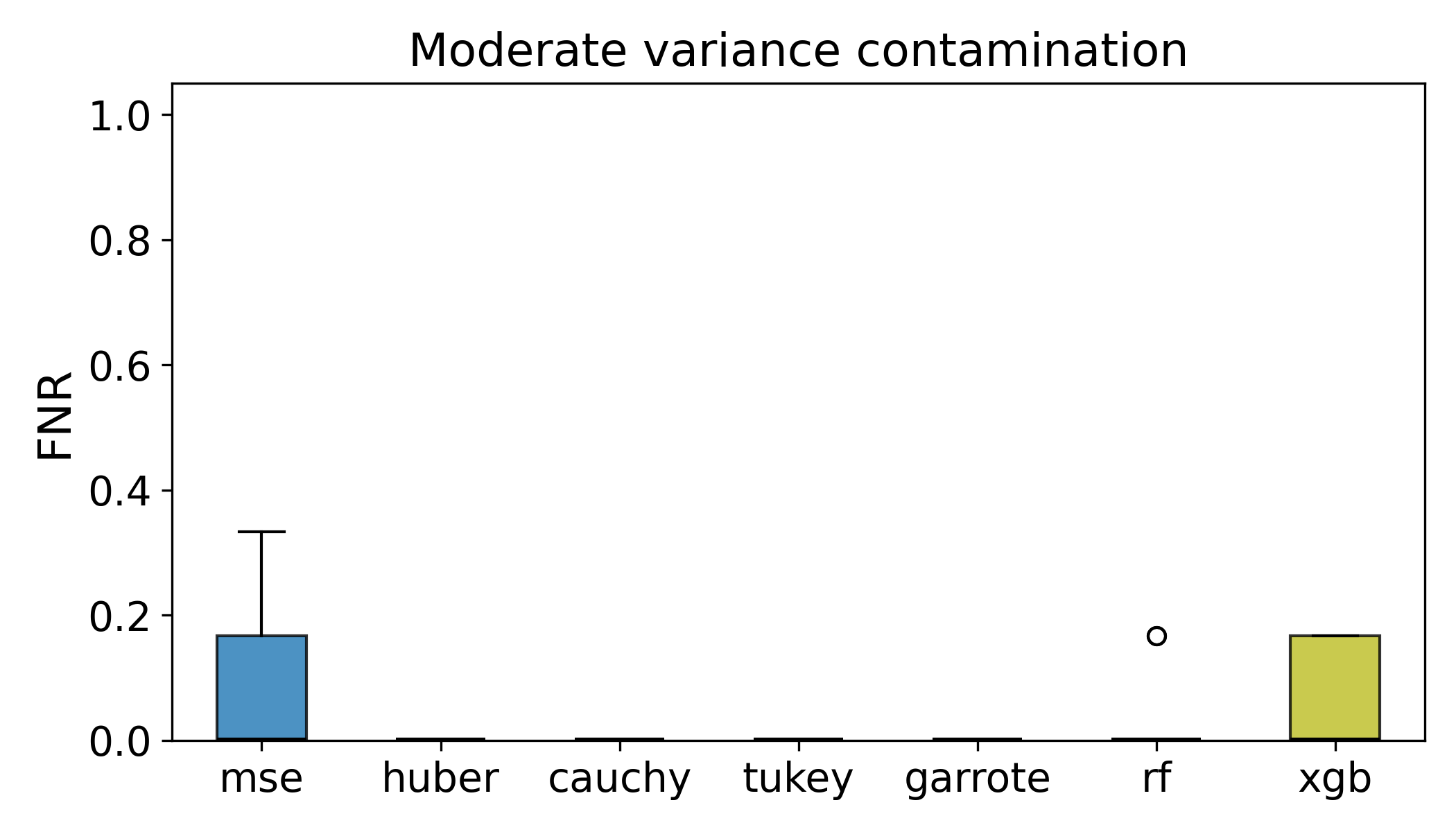}
	\includegraphics[width=0.32\textwidth,height=0.25\textwidth]{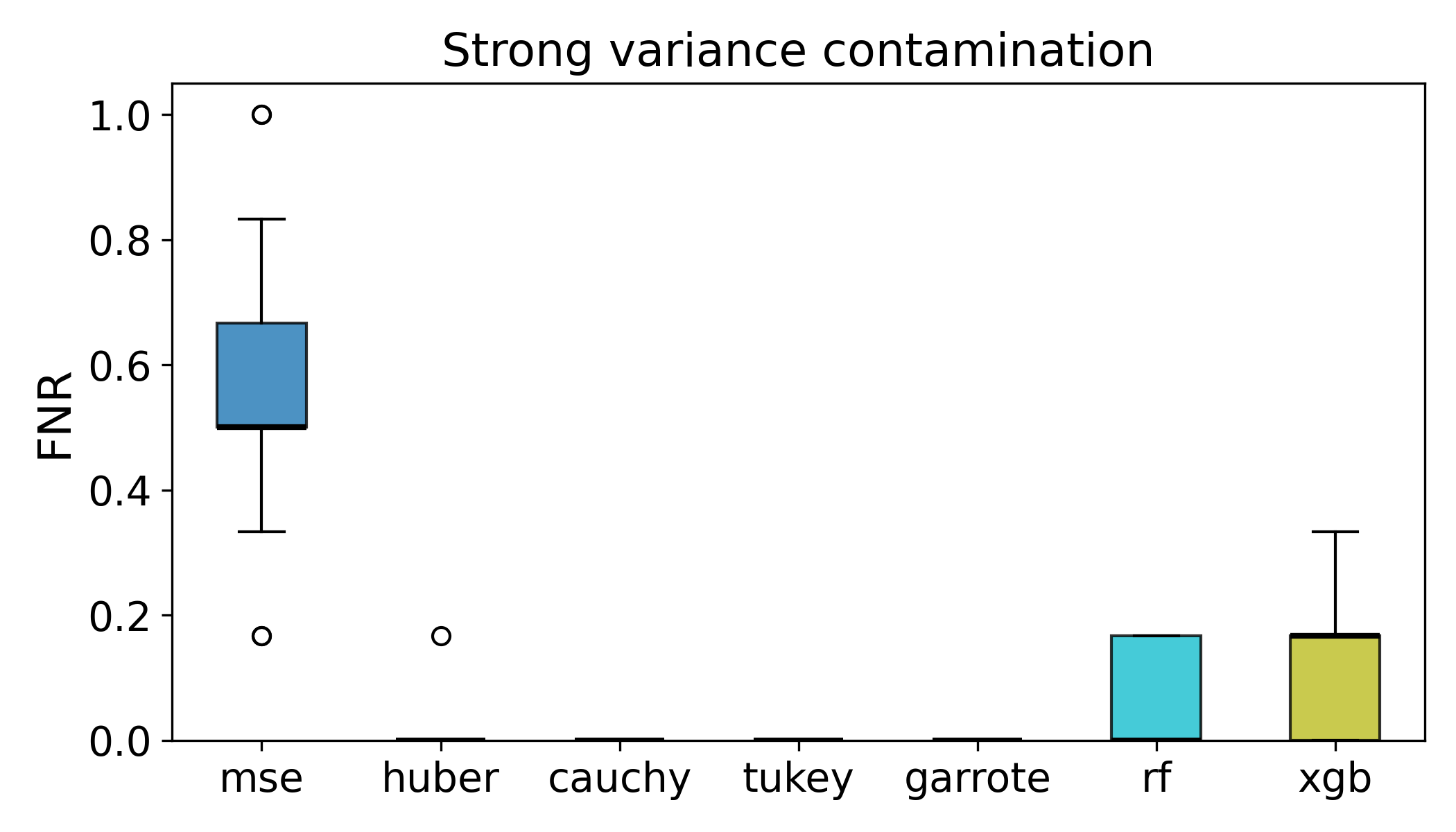}
	
	\caption{Boxplots of the performance metrics results for {\tt model 2} with $10\%$ contamination. First column refers to {\tt gaussian} scenario; second column to {\tt moderate variance} scenario; third column to {\tt strong variance} scenario. The first row shows a realization of the true and contaminated training response $\mathbf{y}$ for all the scenarios.}
	\label{fig_boxplot_model2_variance}
	
\end{figure*}

% boxplots model 3 mixture variance
\begin{figure*}[t]
	\centering
	
	% --- Row 1 ---
	\includegraphics[width=0.32\textwidth,height=0.25\textwidth]{model3/response_n1000_p200_N_RUNS25_M10_gaussian10.png}
	\includegraphics[width=0.32\textwidth,height=0.25\textwidth]{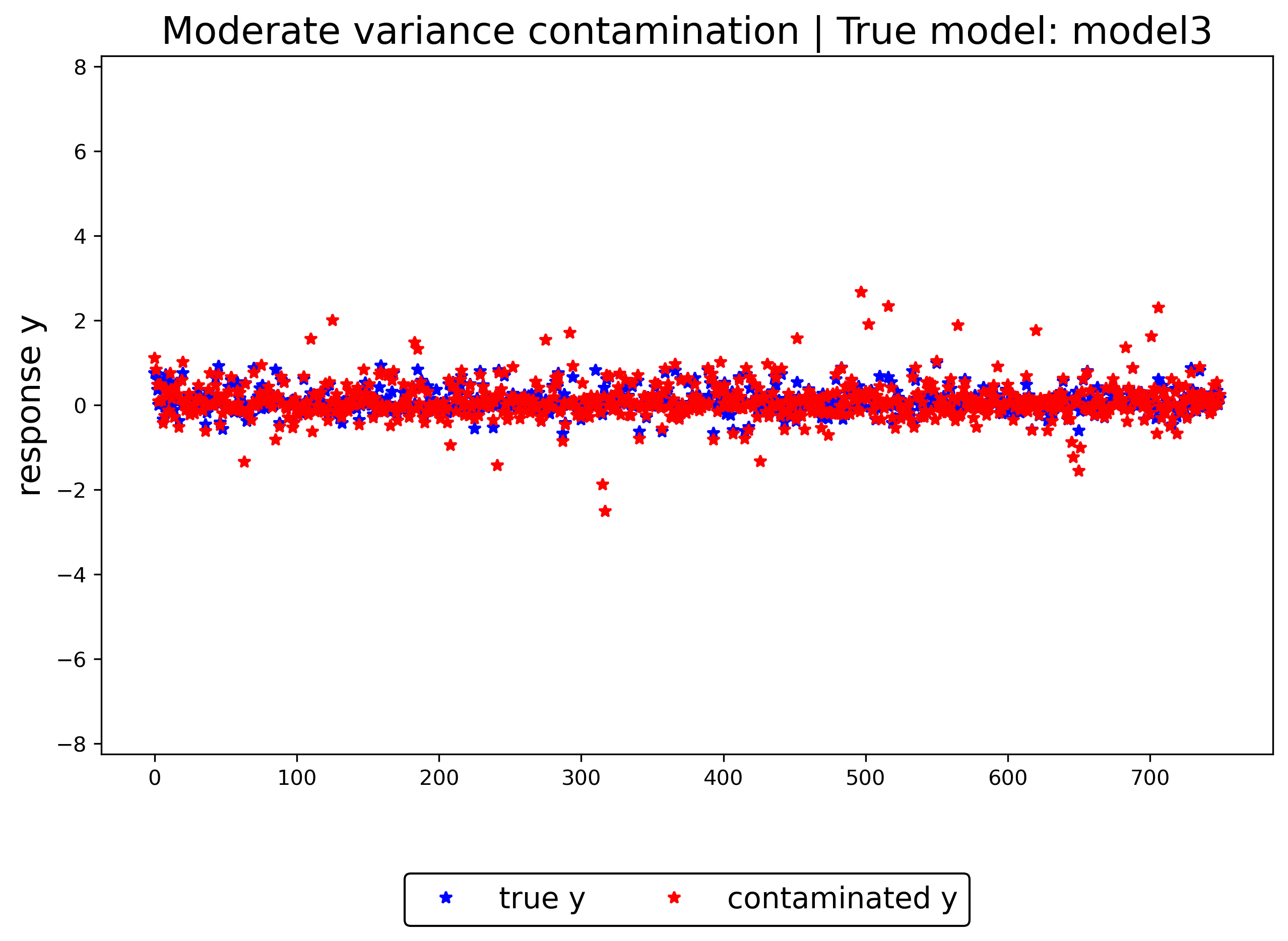}
	\includegraphics[width=0.32\textwidth,height=0.25\textwidth]{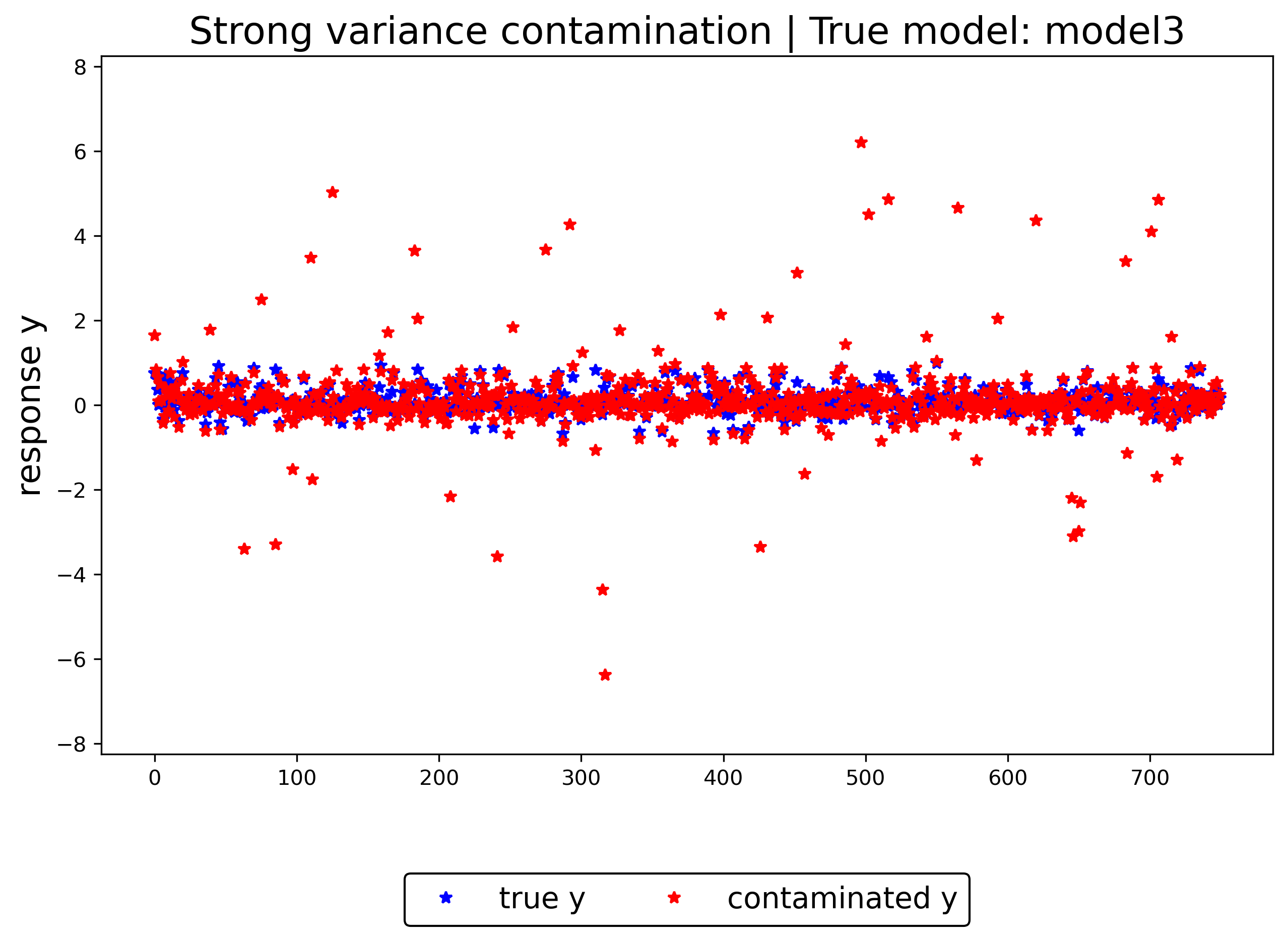}

	% --- Row 2 ---
	\includegraphics[width=0.32\textwidth,height=0.25\textwidth]{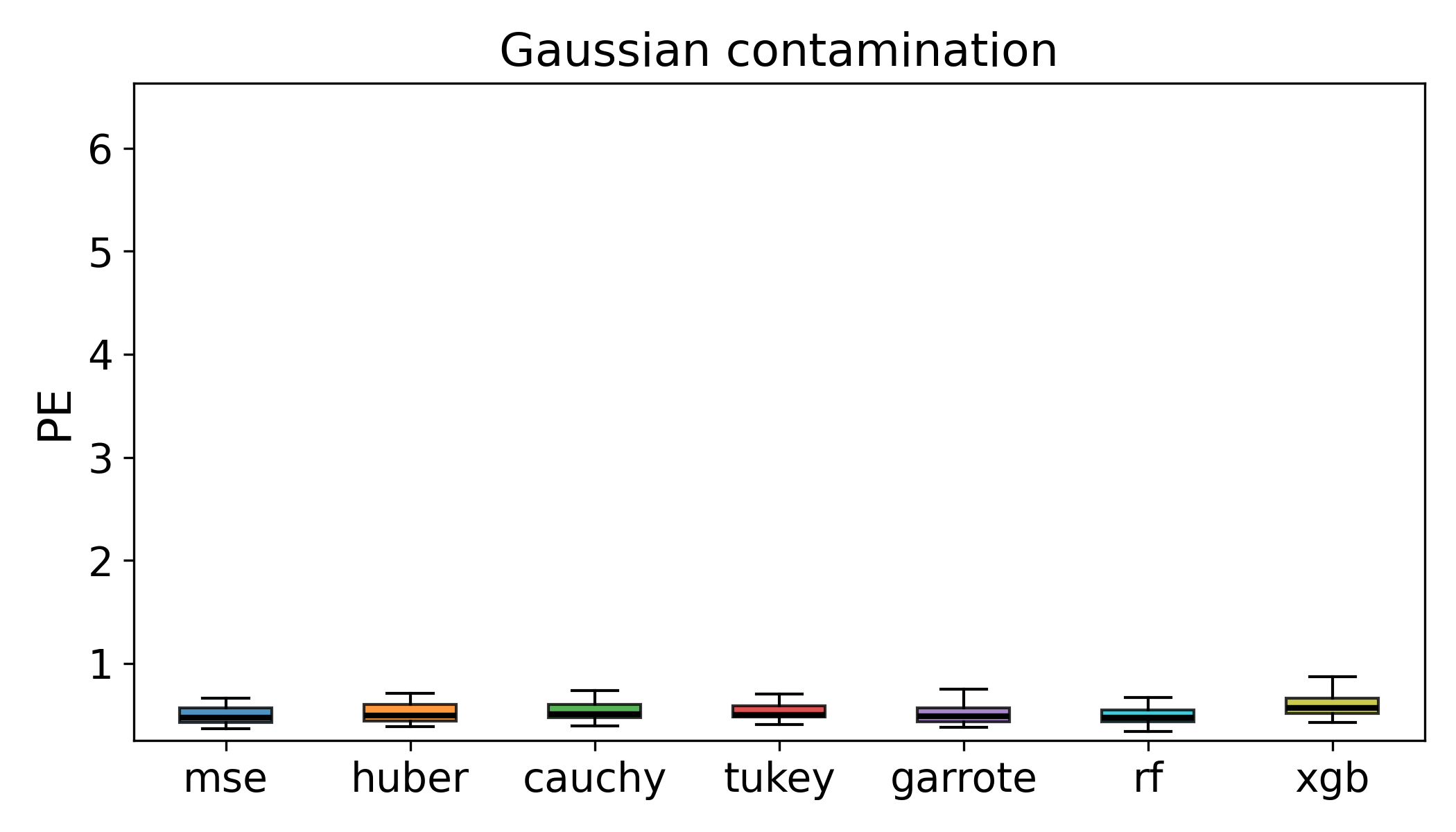}
	\includegraphics[width=0.32\textwidth,height=0.25\textwidth]{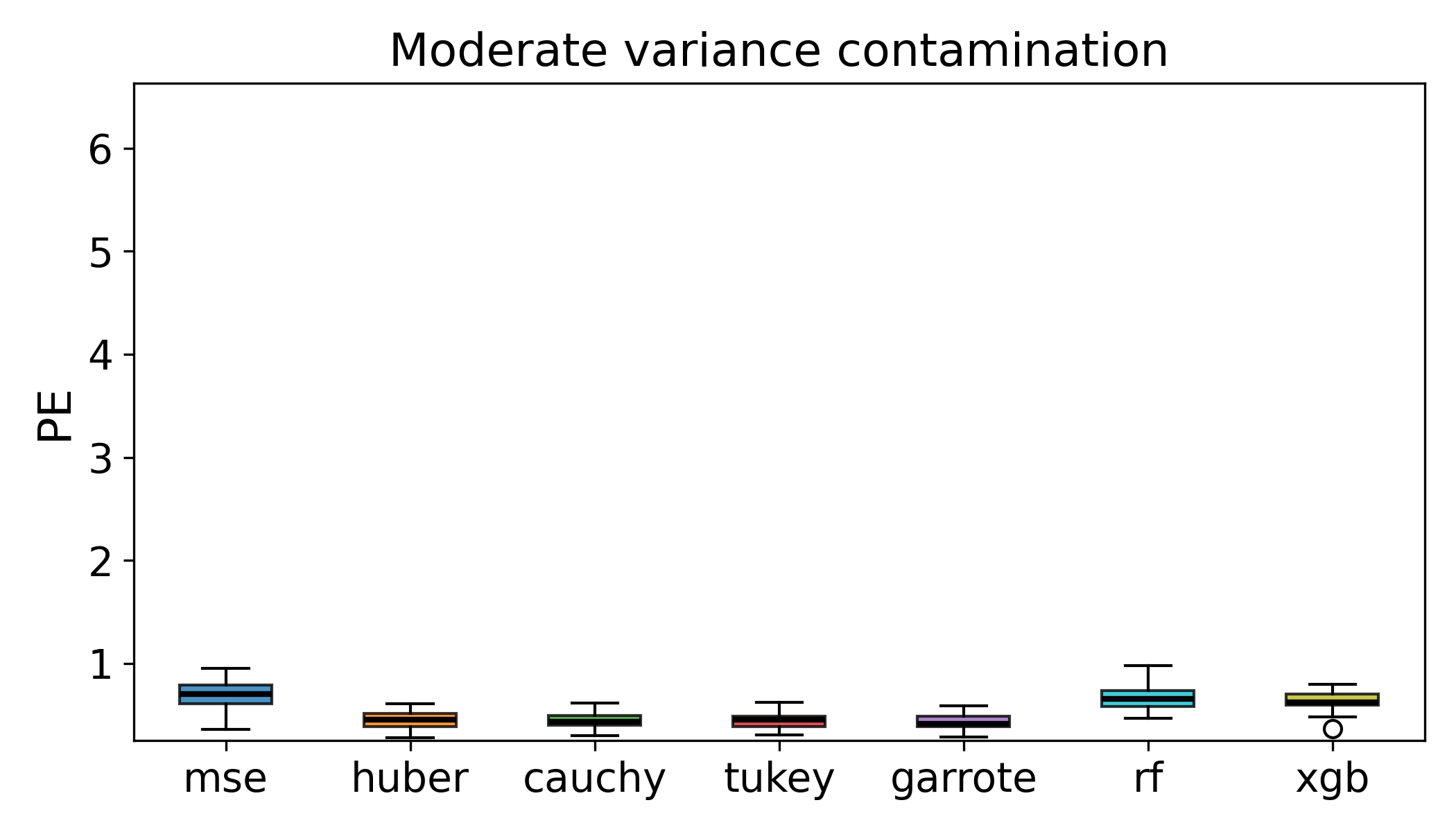}
	\includegraphics[width=0.32\textwidth,height=0.25\textwidth]{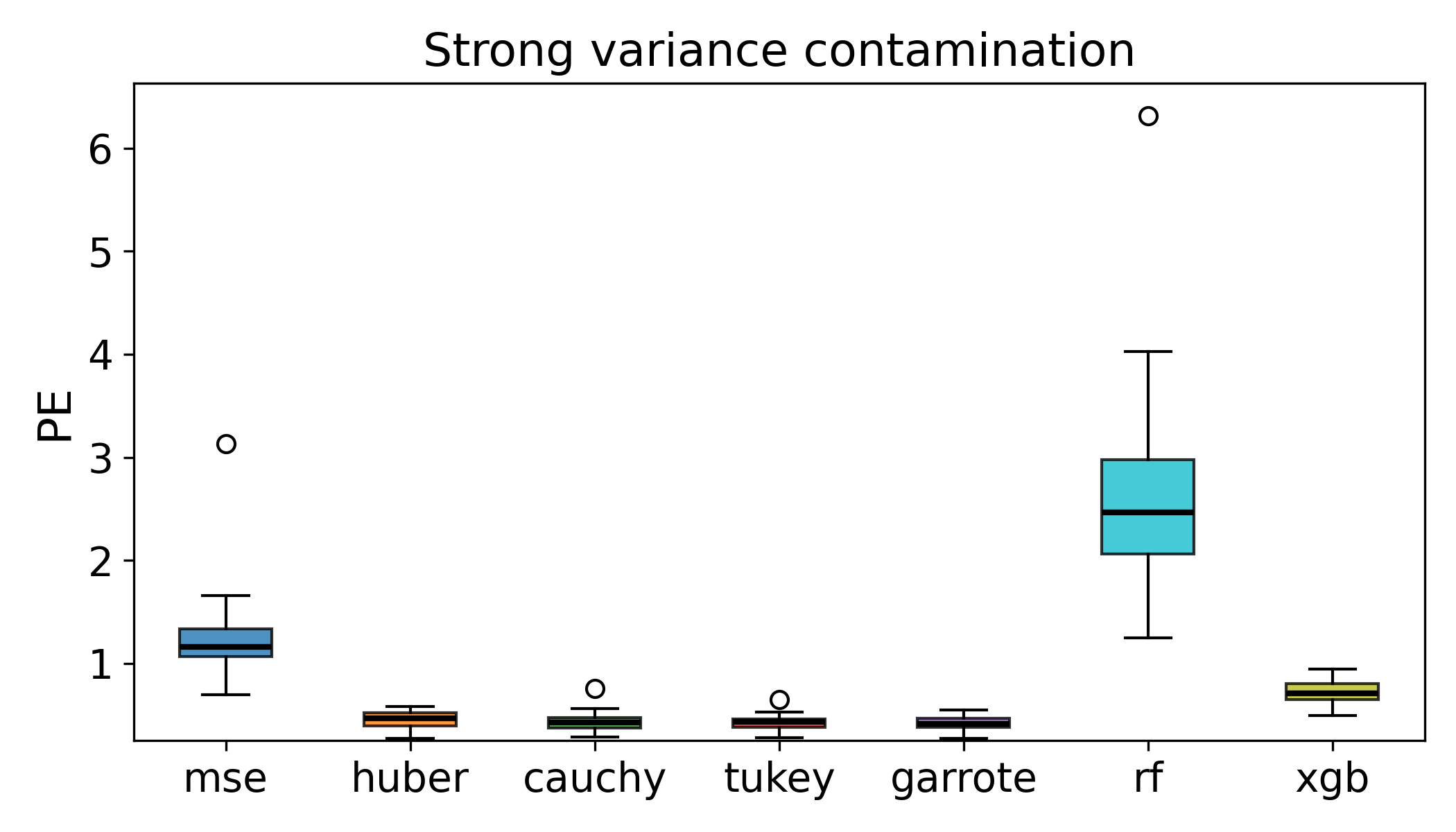}
	
	\vspace{0.2cm}
	
	% --- Row 3 ---
	\includegraphics[width=0.32\textwidth,height=0.25\textwidth]{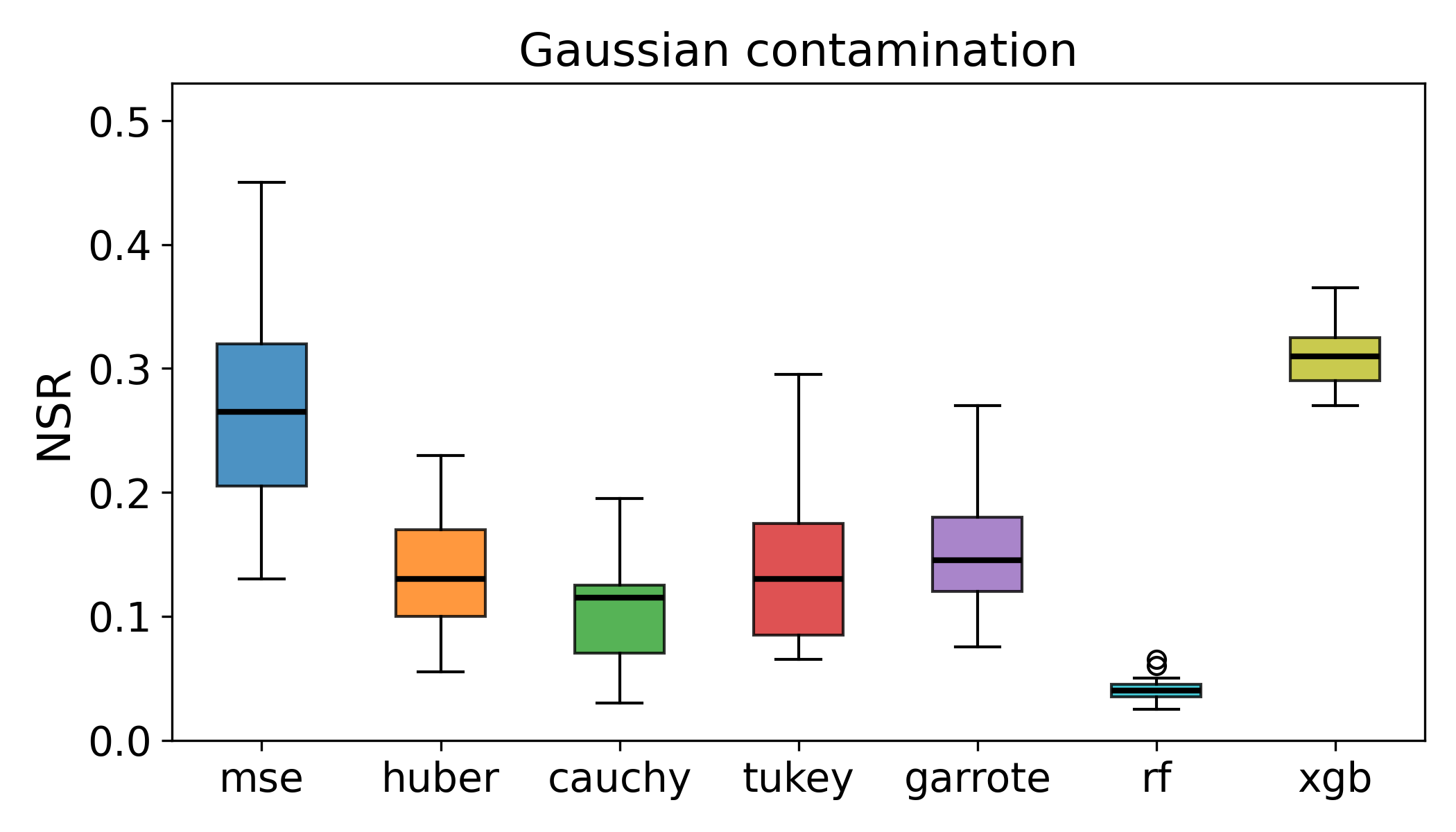}
	\includegraphics[width=0.32\textwidth,height=0.25\textwidth]{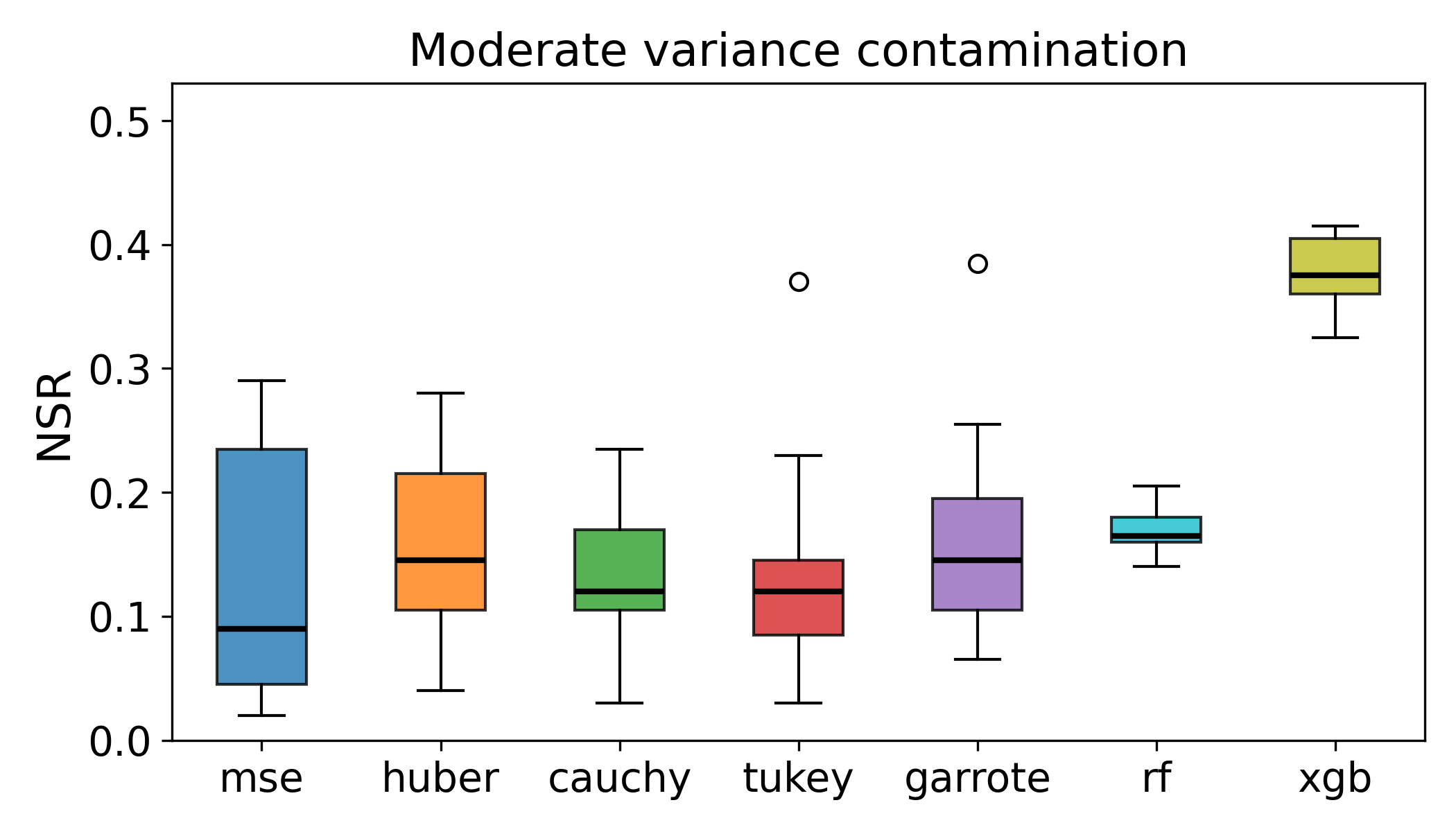}
	\includegraphics[width=0.32\textwidth,height=0.25\textwidth]{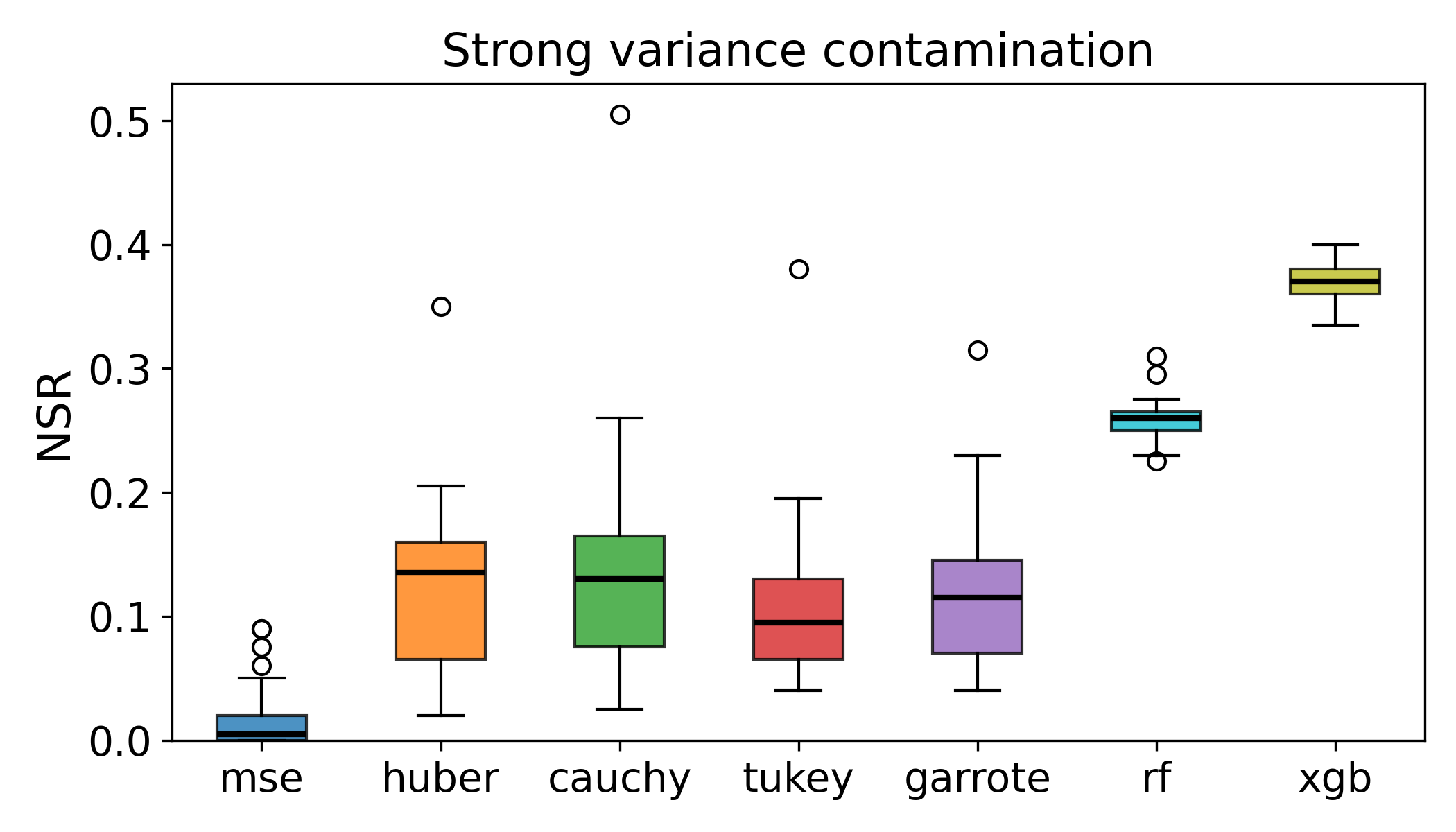}
	
	\vspace{0.2cm}
	
	% --- Row 4 ---
	\includegraphics[width=0.32\textwidth,height=0.25\textwidth]{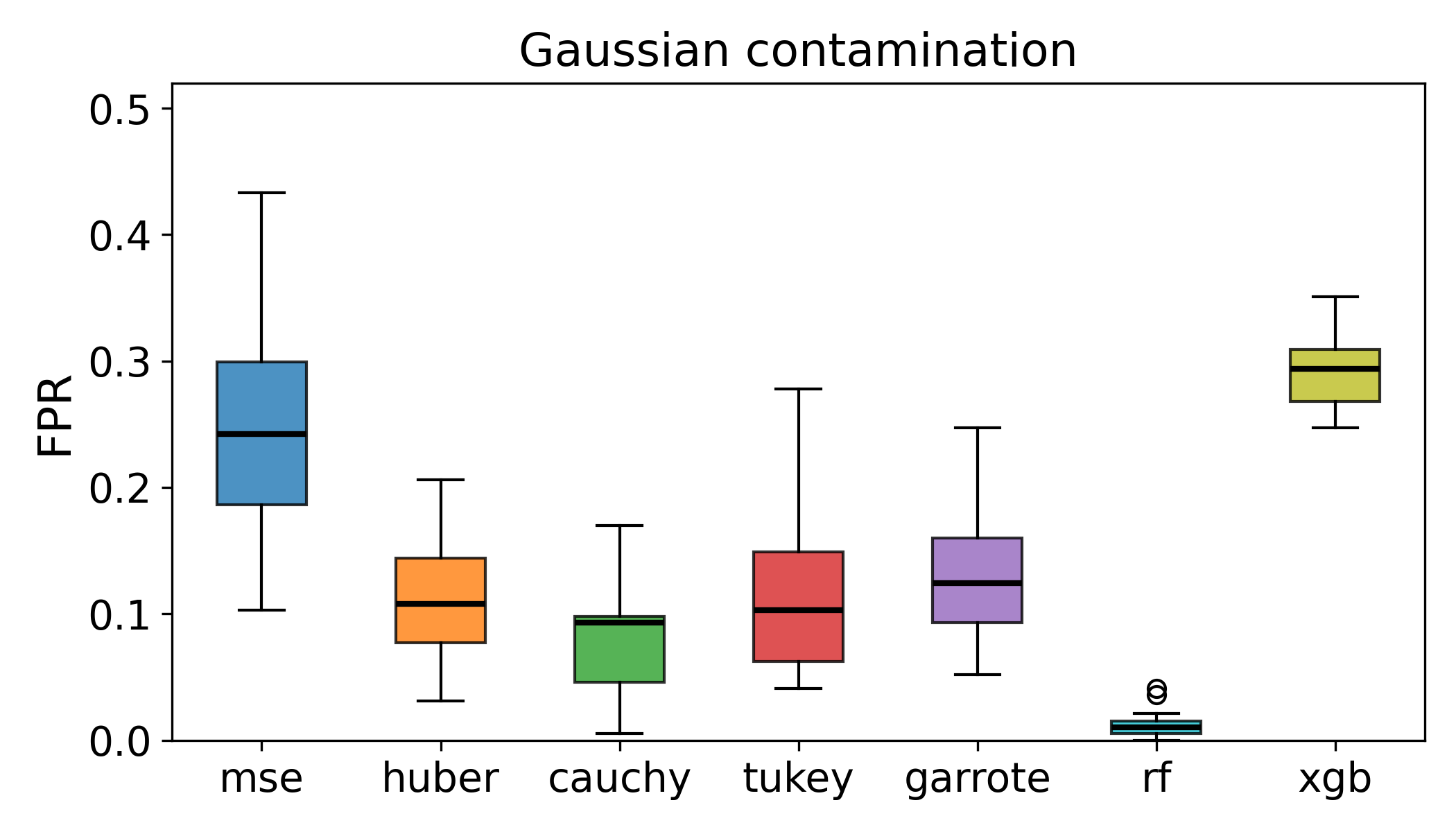}
	\includegraphics[width=0.32\textwidth,height=0.25\textwidth]{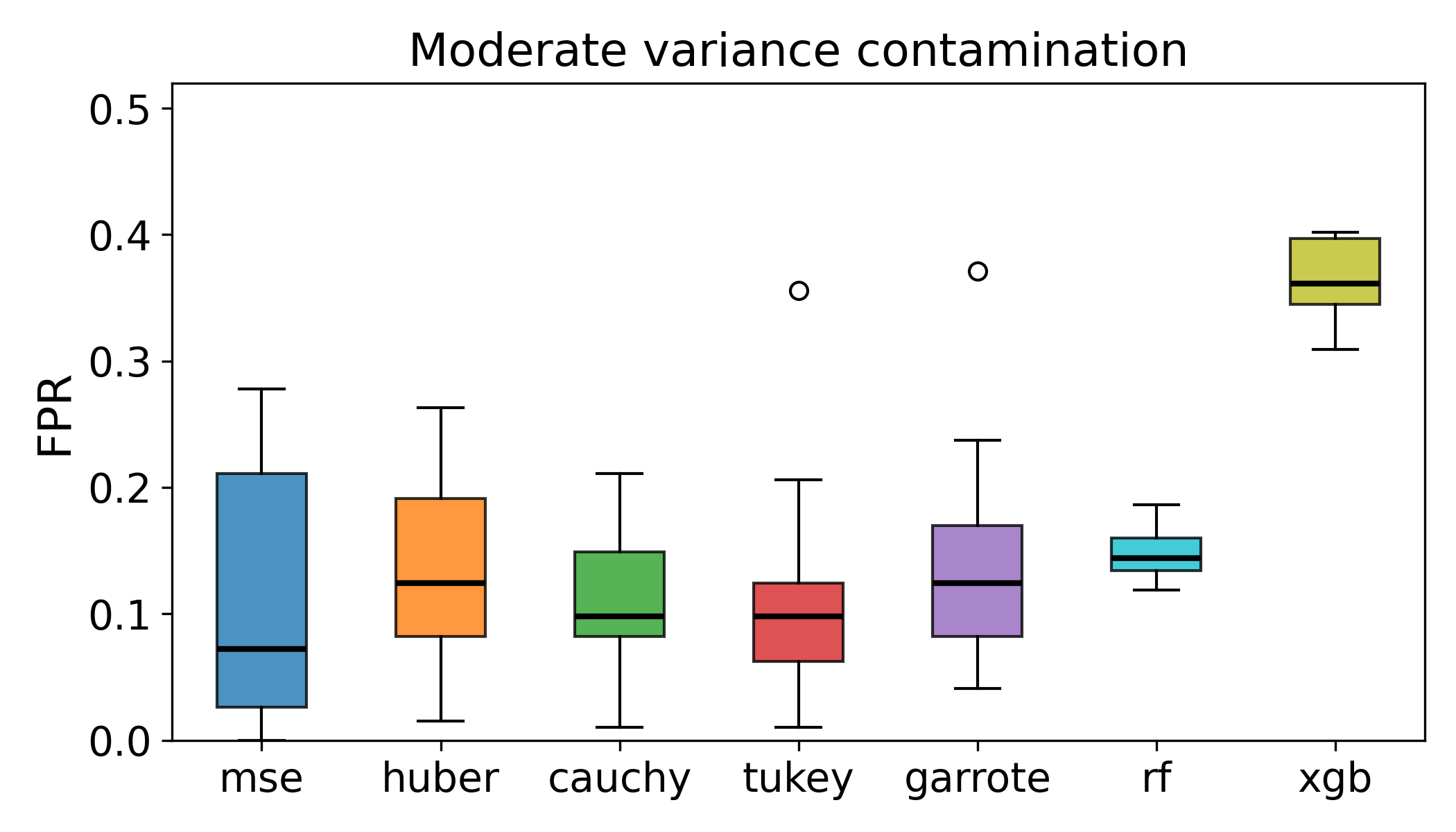}
	\includegraphics[width=0.32\textwidth,height=0.25\textwidth]{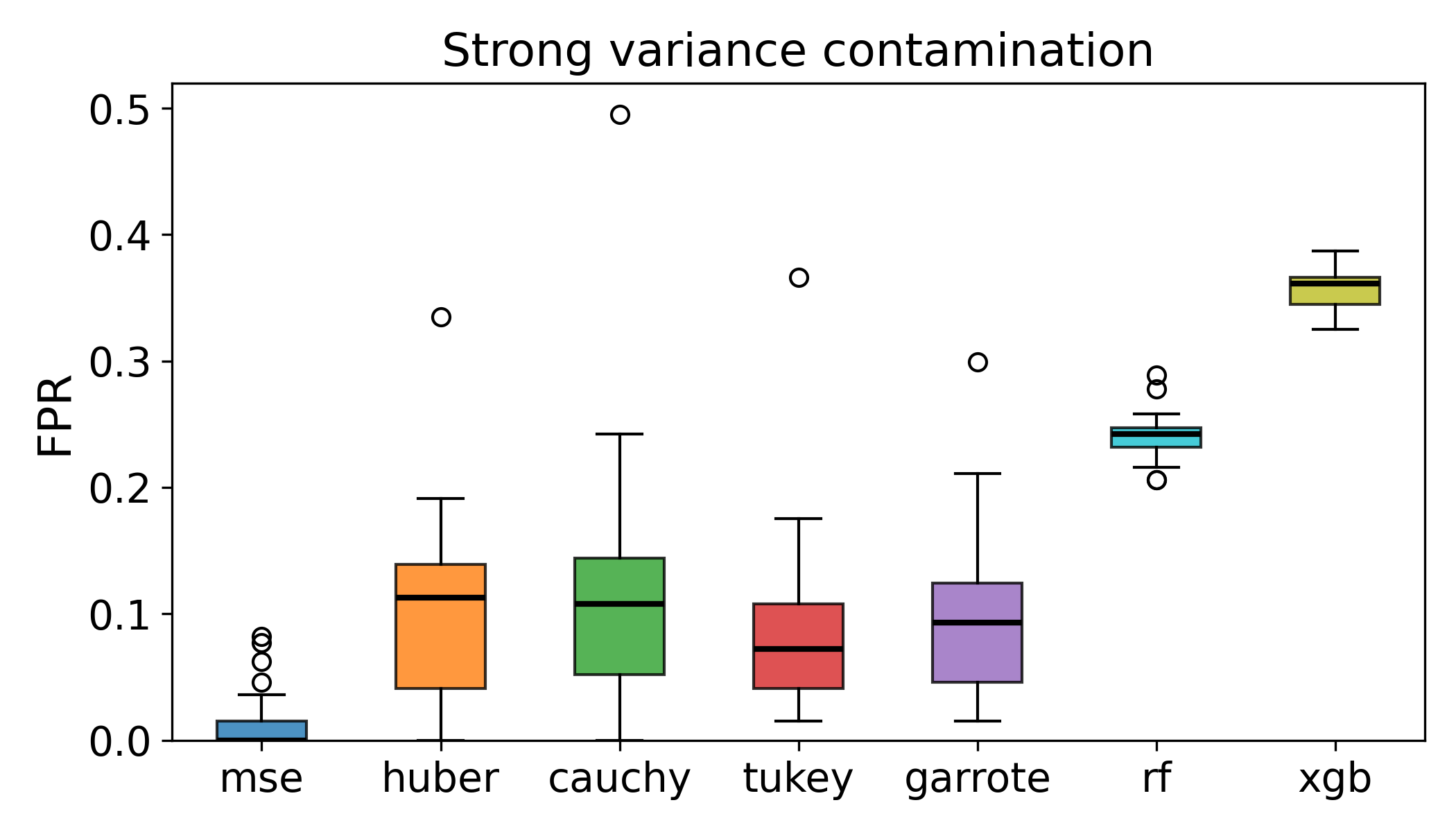}
	
	\vspace{0.2cm}
	
	% --- Row 5 ---
	\includegraphics[width=0.32\textwidth,height=0.25\textwidth]{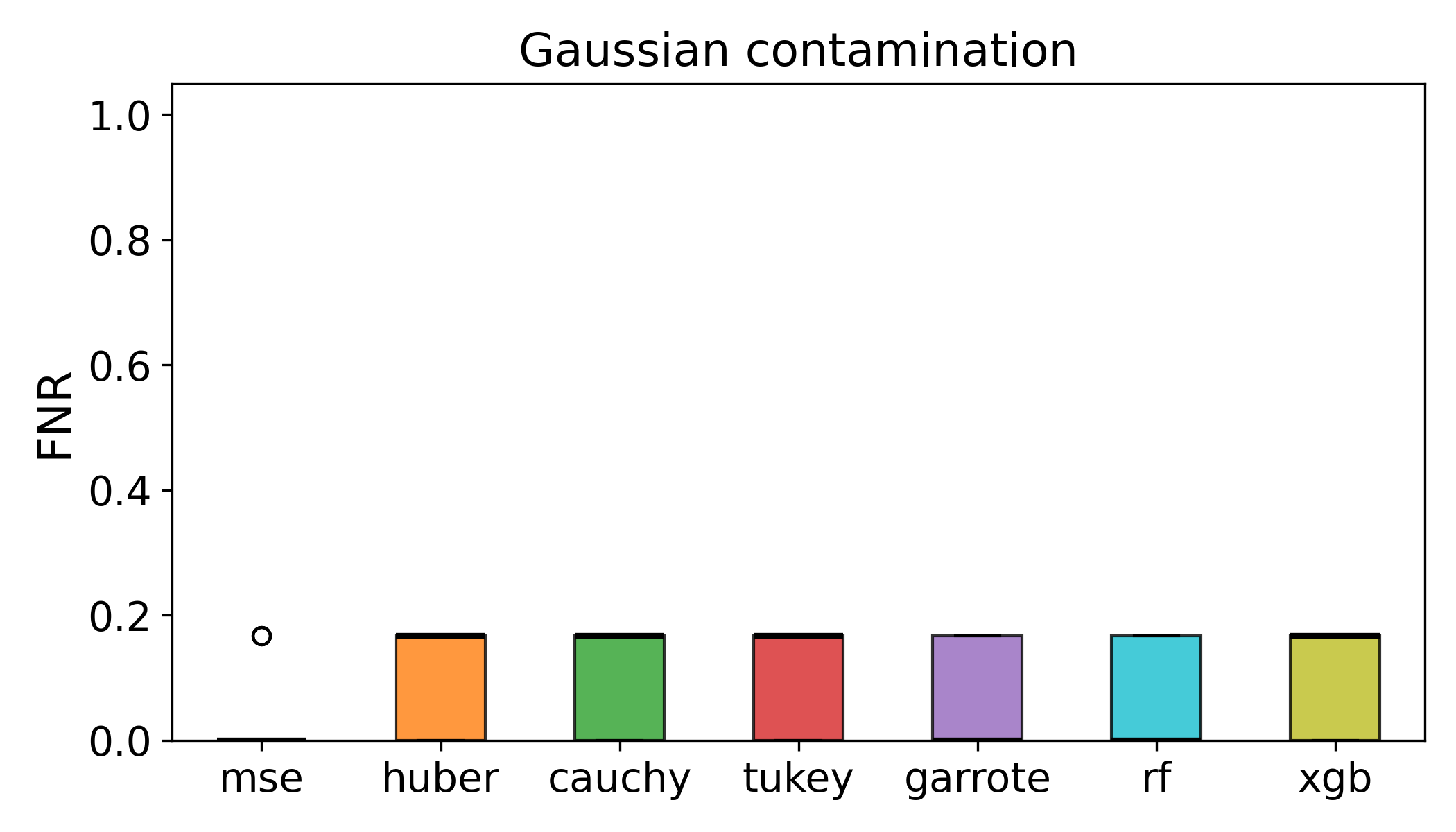}
	\includegraphics[width=0.32\textwidth,height=0.25\textwidth]{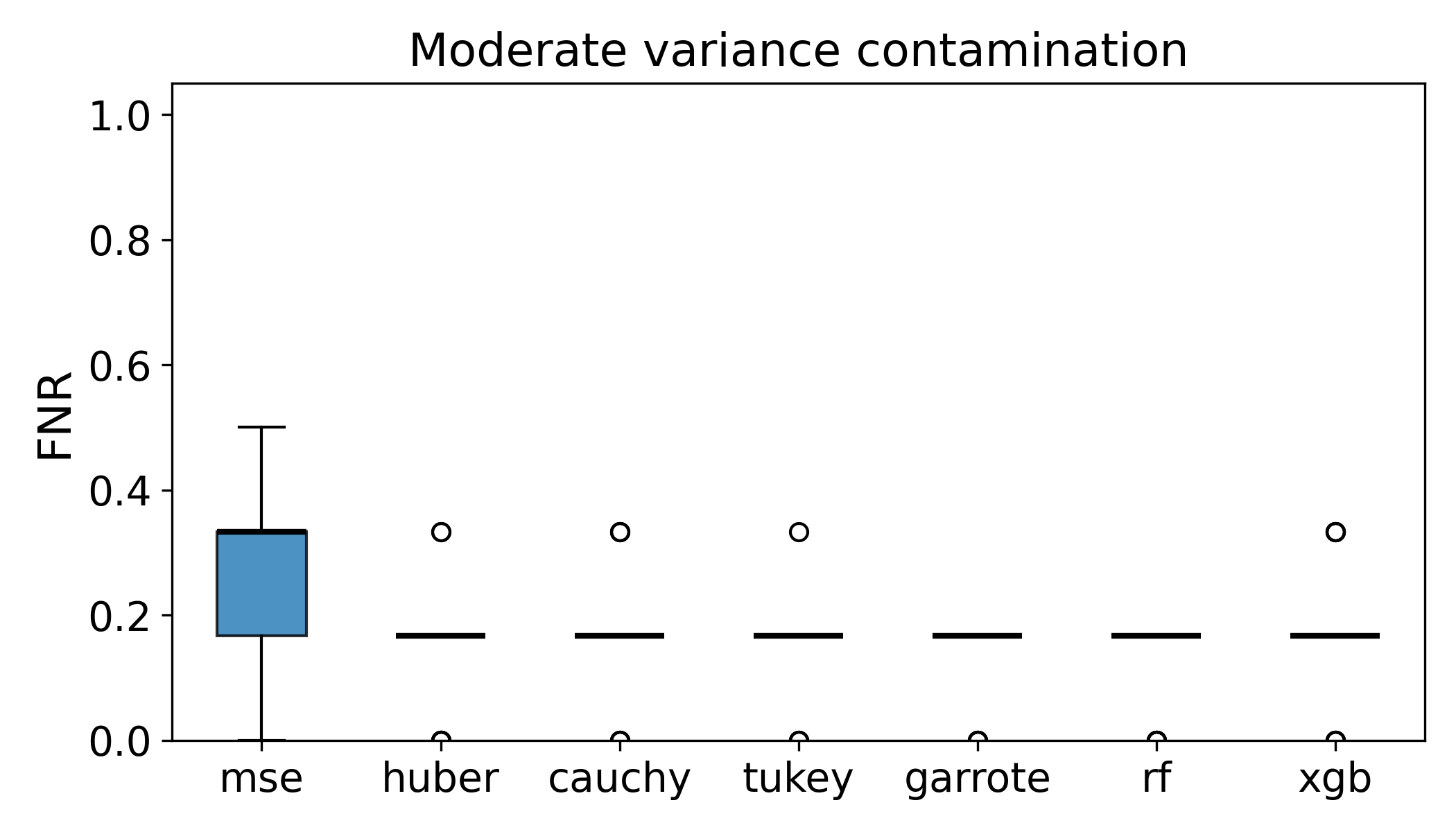}
	\includegraphics[width=0.32\textwidth,height=0.25\textwidth]{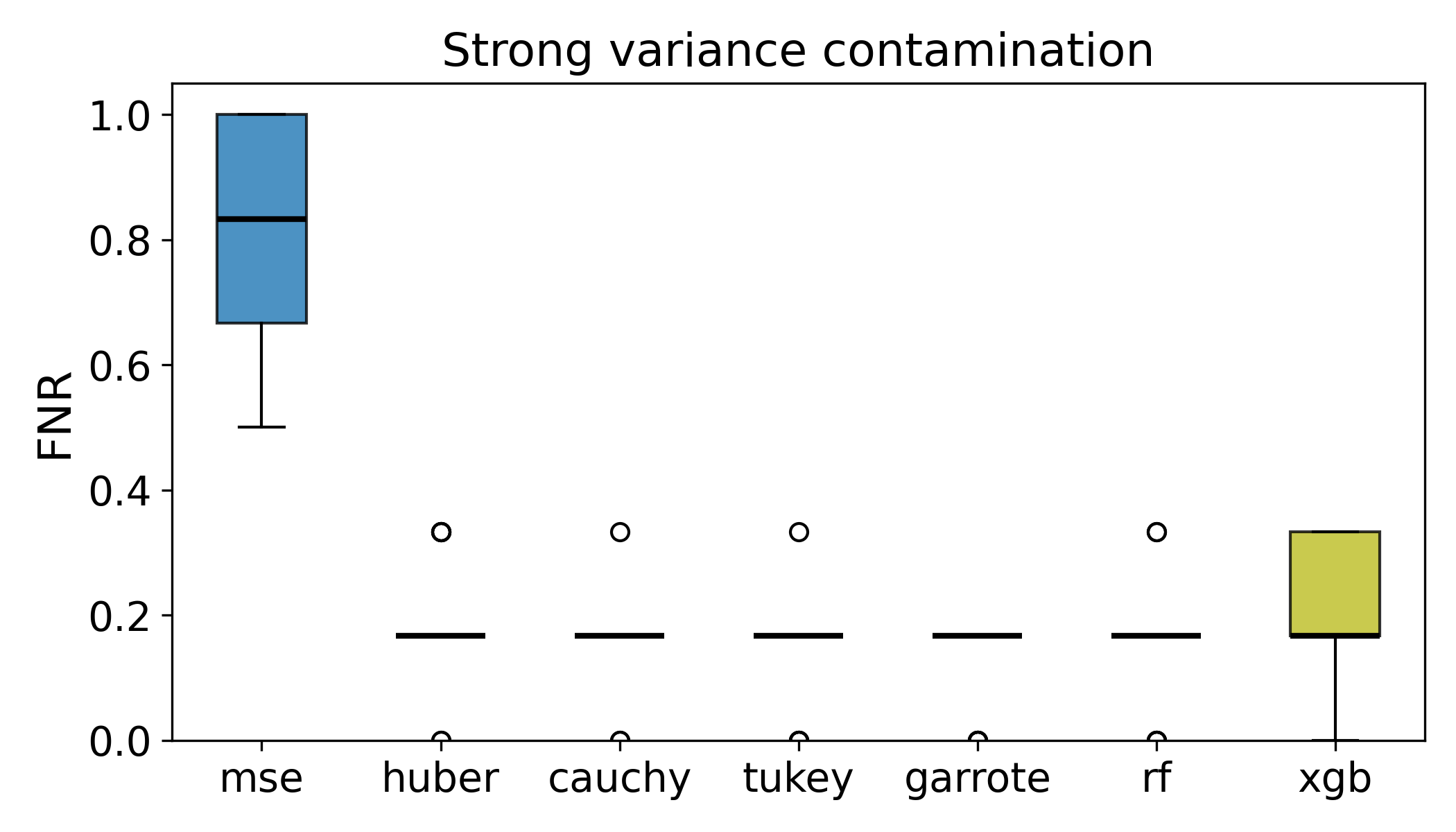}
	
	\caption{Boxplots of the performance metrics results for {\tt model 3} with $10\%$ contamination. First column refers to {\tt gaussian} scenario; second column to {\tt moderate variance} scenario; third column to {\tt strong variance} scenario. The first row shows a realization of the true and contaminated training response $\mathbf{y}$ for all the scenarios.}
	\label{fig_boxplot_model3_variance}
	
\end{figure*}

% boxplots model 4 mixture variance
\begin{figure*}[t]
	\centering
	
	% --- Row 1 ---
	\includegraphics[width=0.32\textwidth,height=0.25\textwidth]{model4/response_n1000_p200_N_RUNS25_M10_gaussian10.png}
	\includegraphics[width=0.32\textwidth,height=0.25\textwidth]{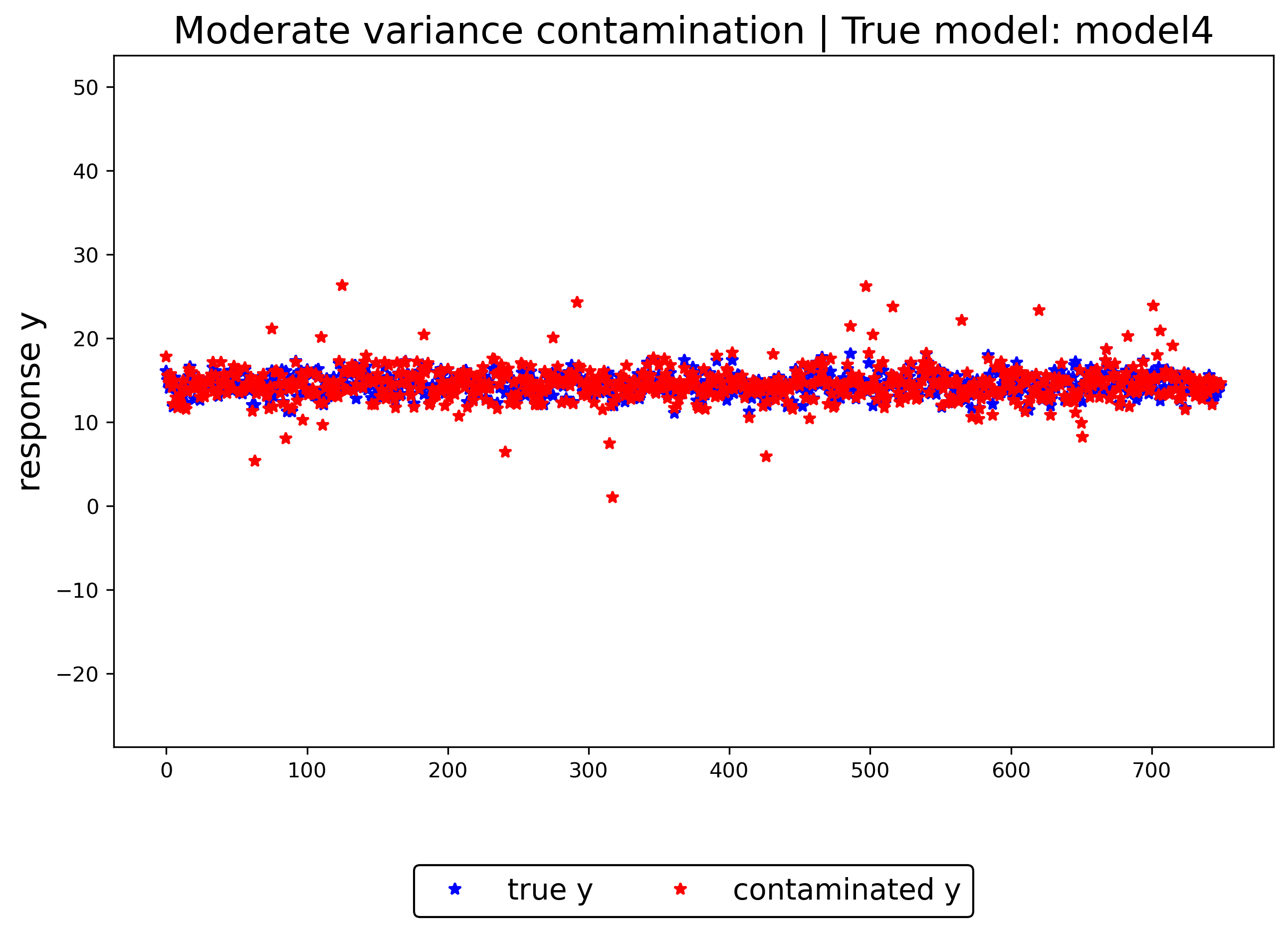}
	\includegraphics[width=0.32\textwidth,height=0.25\textwidth]{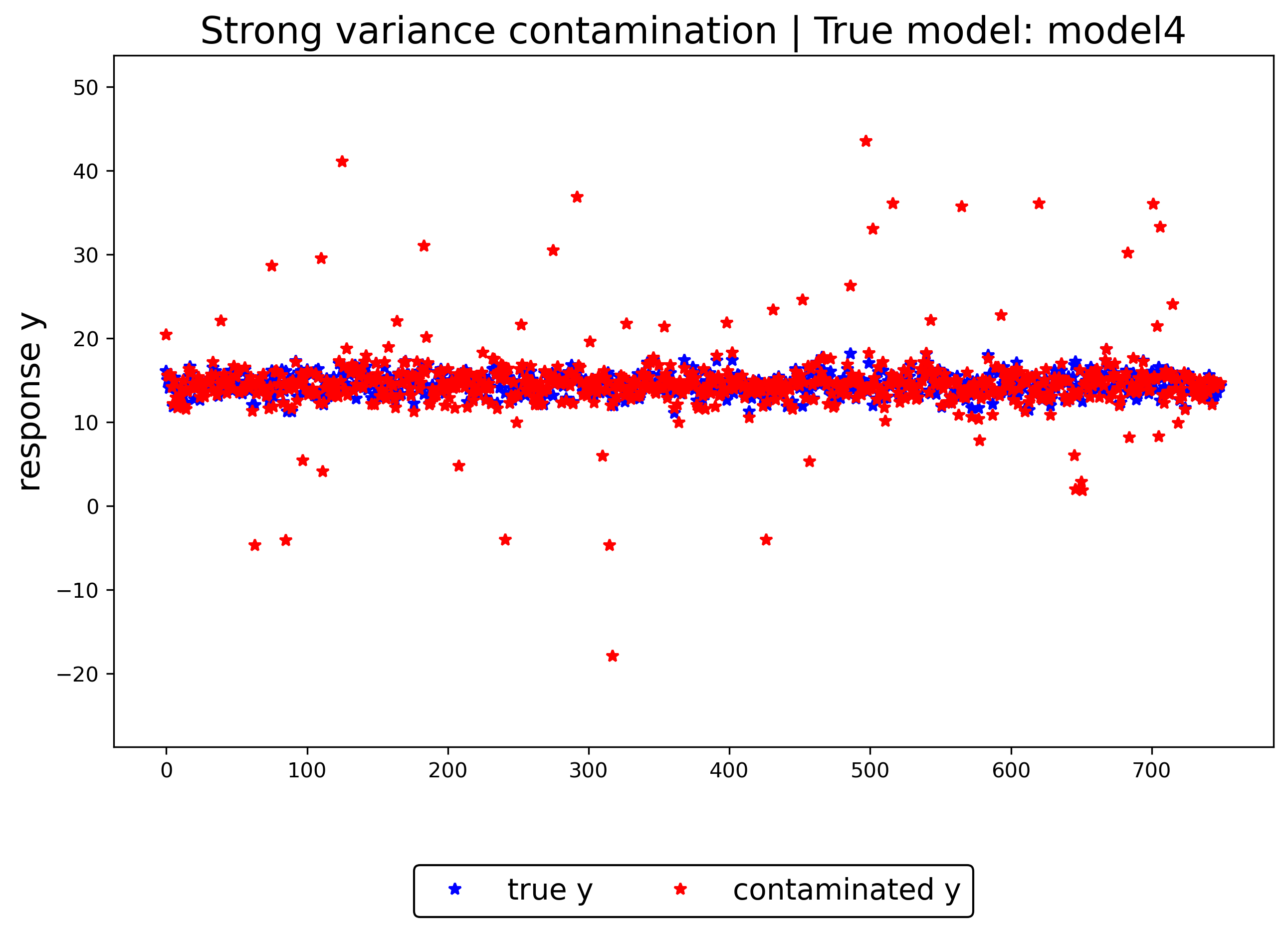}

	% --- Row 2 ---
	\includegraphics[width=0.32\textwidth,height=0.25\textwidth]{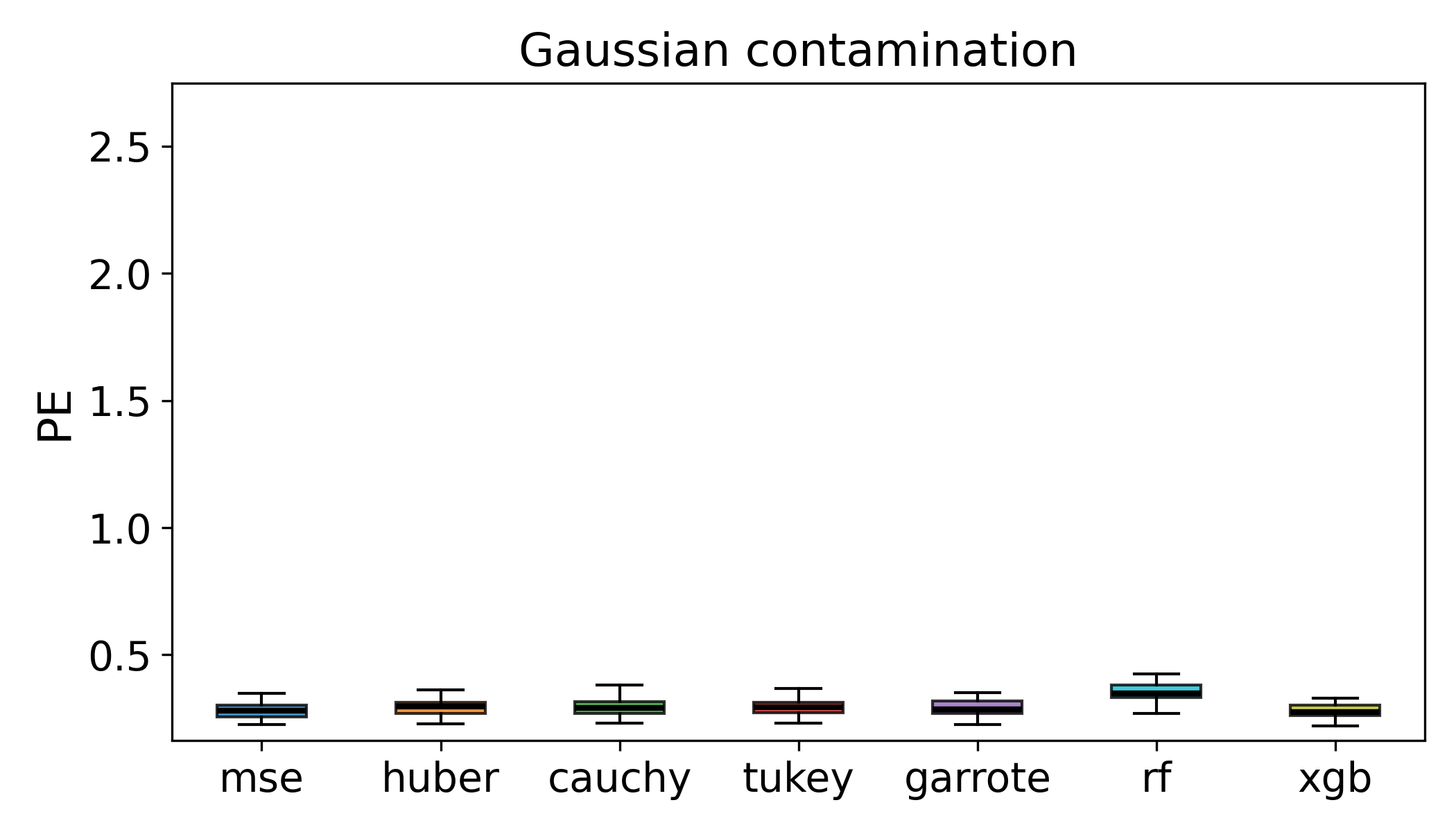}
	\includegraphics[width=0.32\textwidth,height=0.25\textwidth]{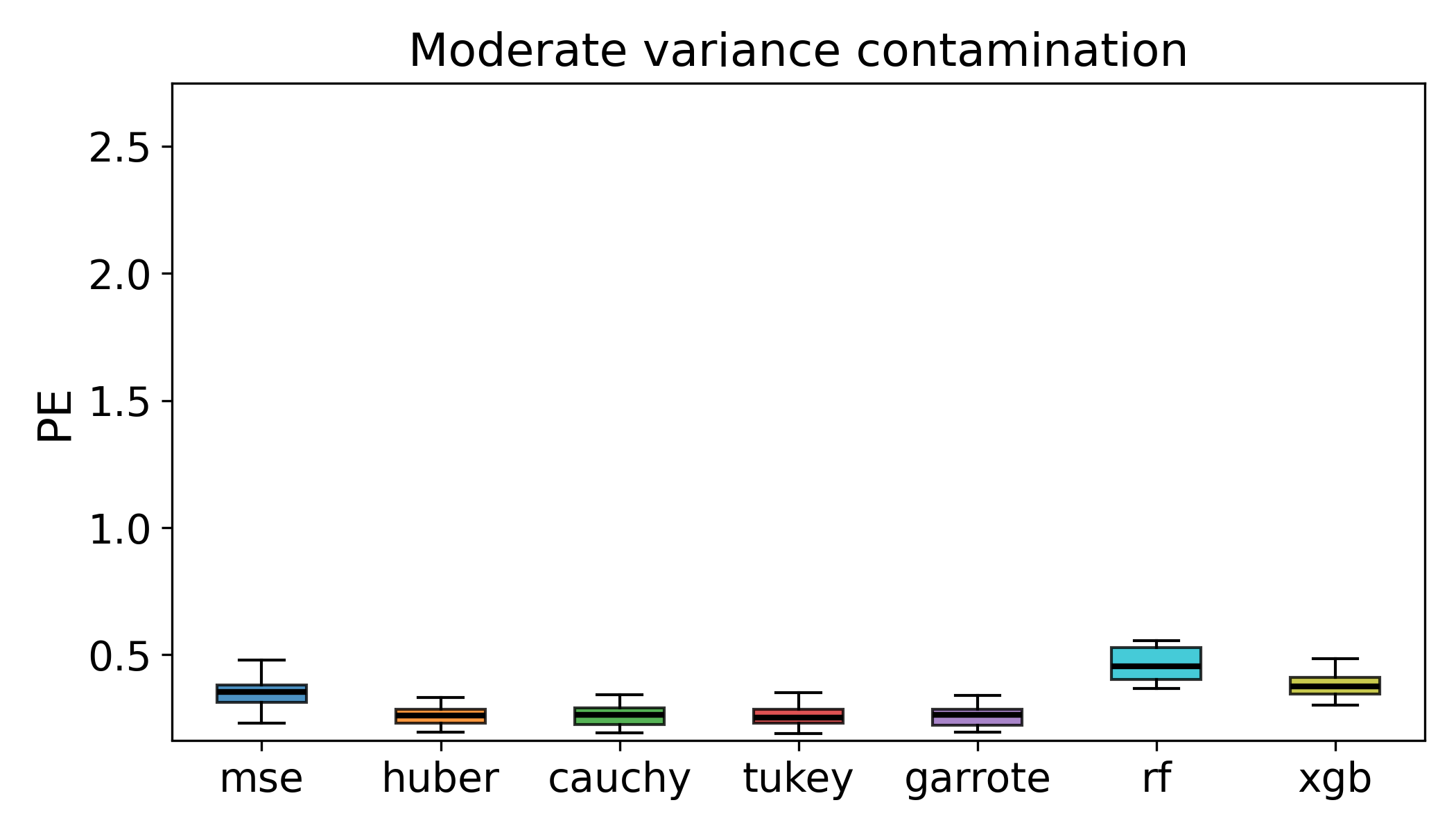}
	\includegraphics[width=0.32\textwidth,height=0.25\textwidth]{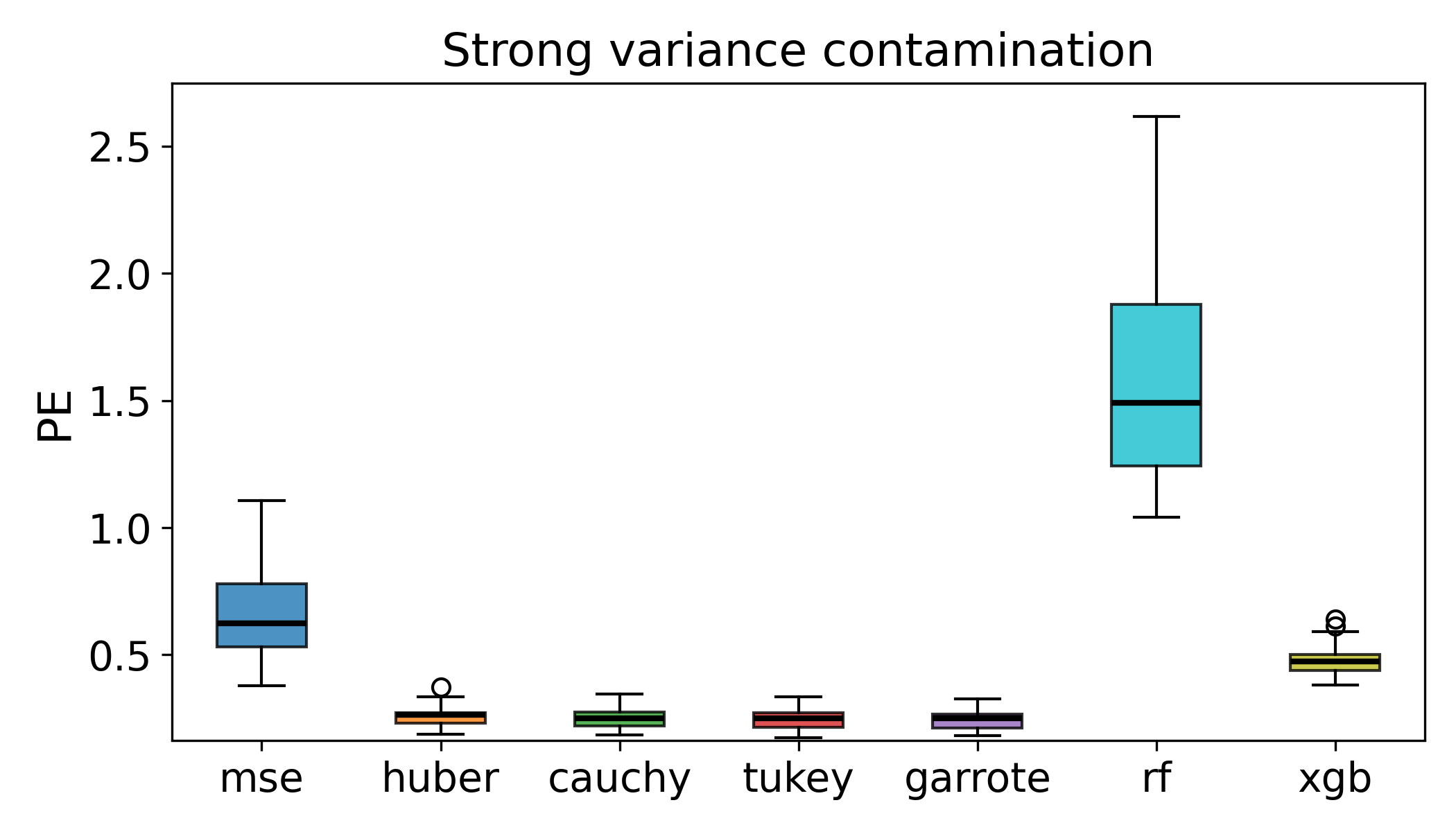}
	
	\vspace{0.2cm}
	
	% --- Row 3 ---
	\includegraphics[width=0.32\textwidth,height=0.25\textwidth]{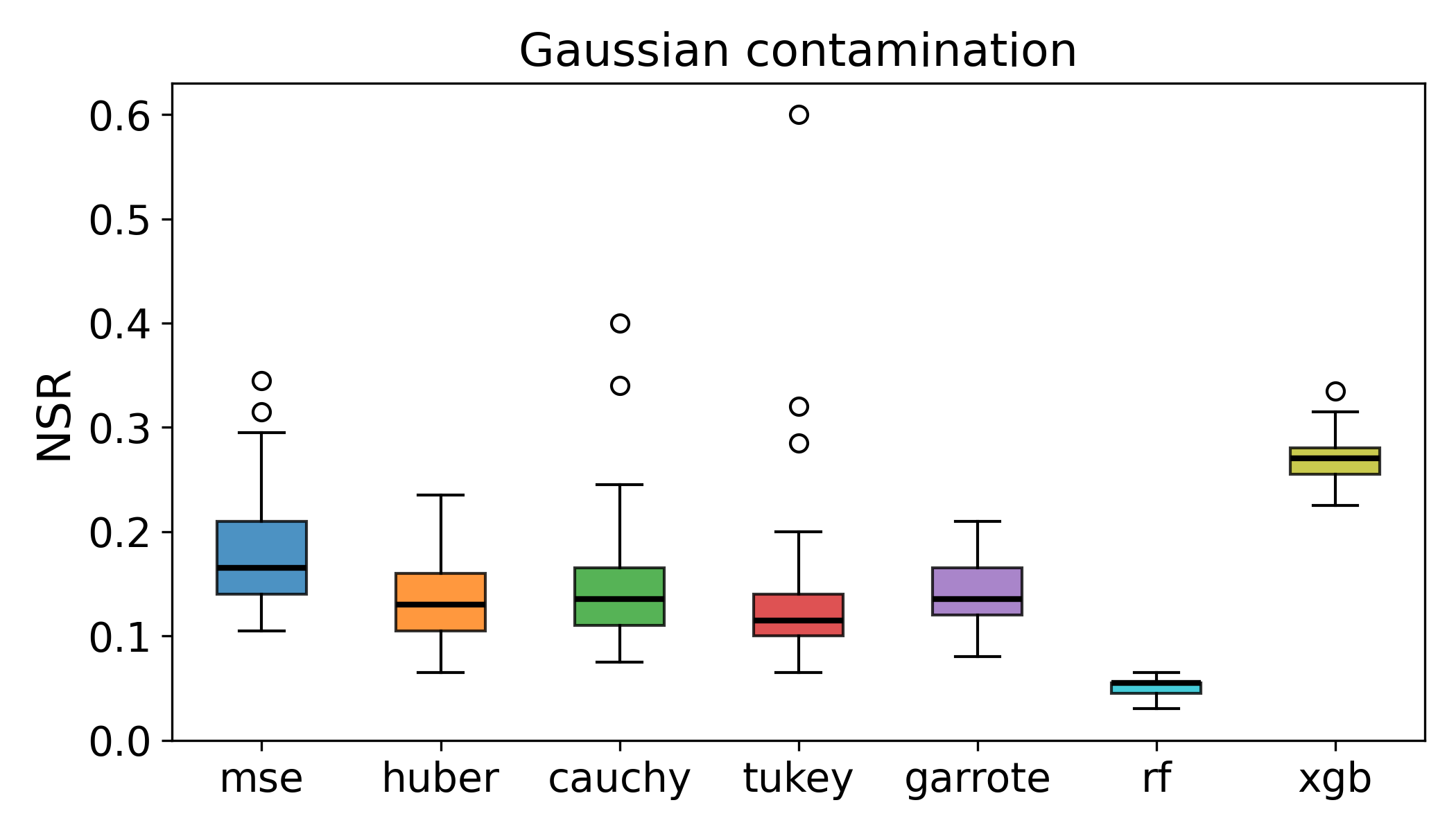}
	\includegraphics[width=0.32\textwidth,height=0.25\textwidth]{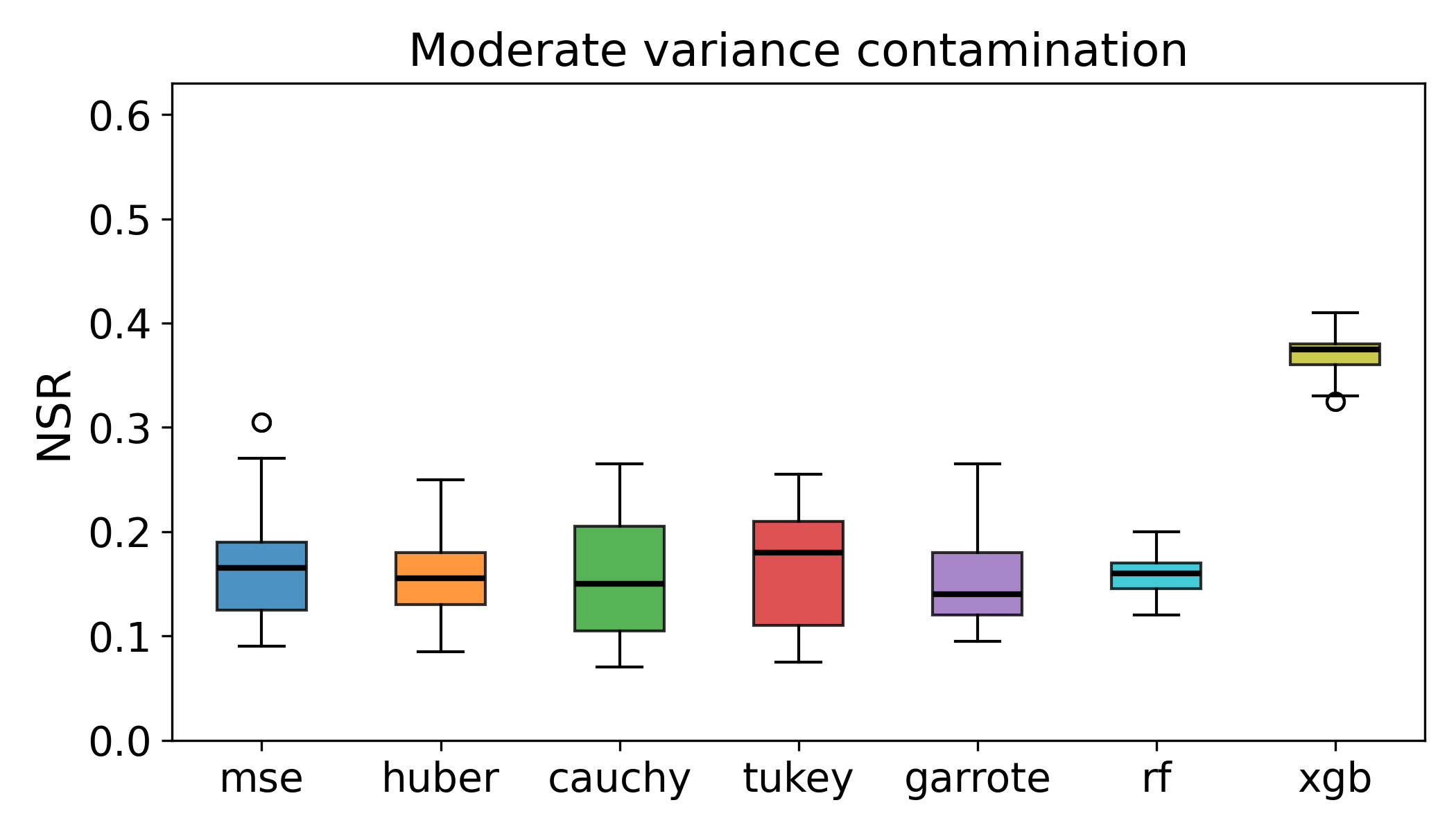}
	\includegraphics[width=0.32\textwidth,height=0.25\textwidth]{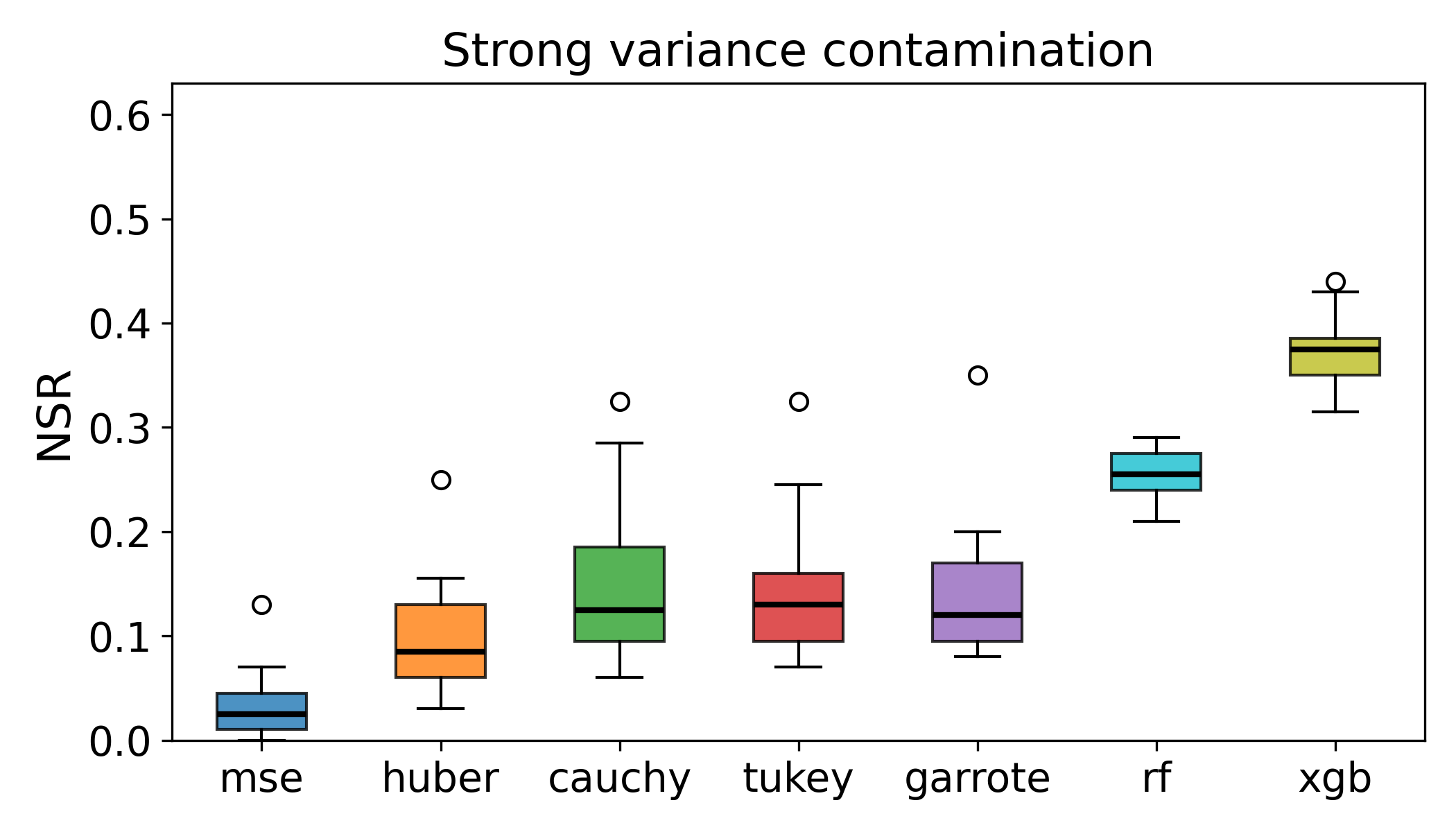}
	
	\vspace{0.2cm}
	
	% --- Row 4 ---
	\includegraphics[width=0.32\textwidth,height=0.25\textwidth]{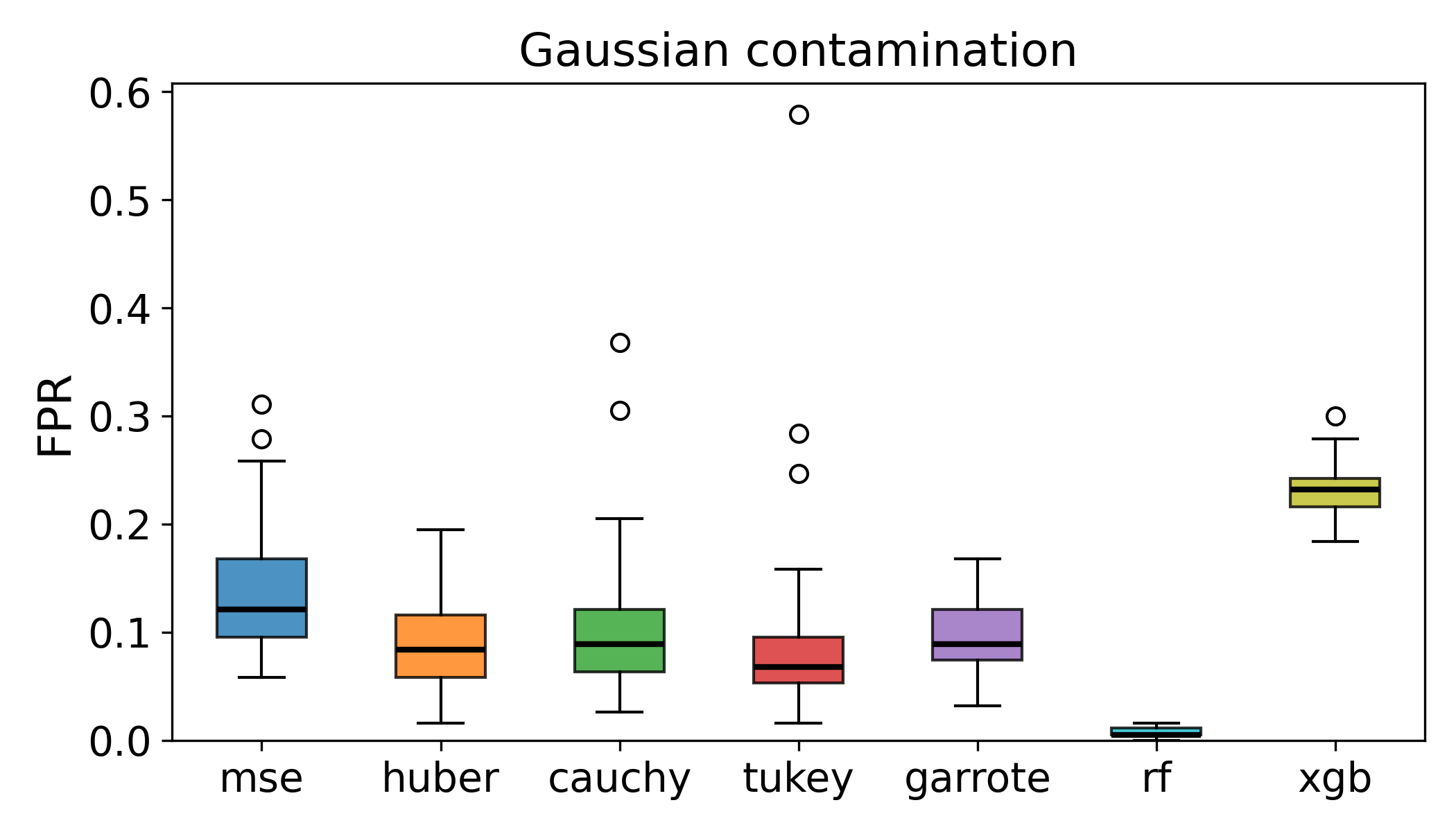}
	\includegraphics[width=0.32\textwidth,height=0.25\textwidth]{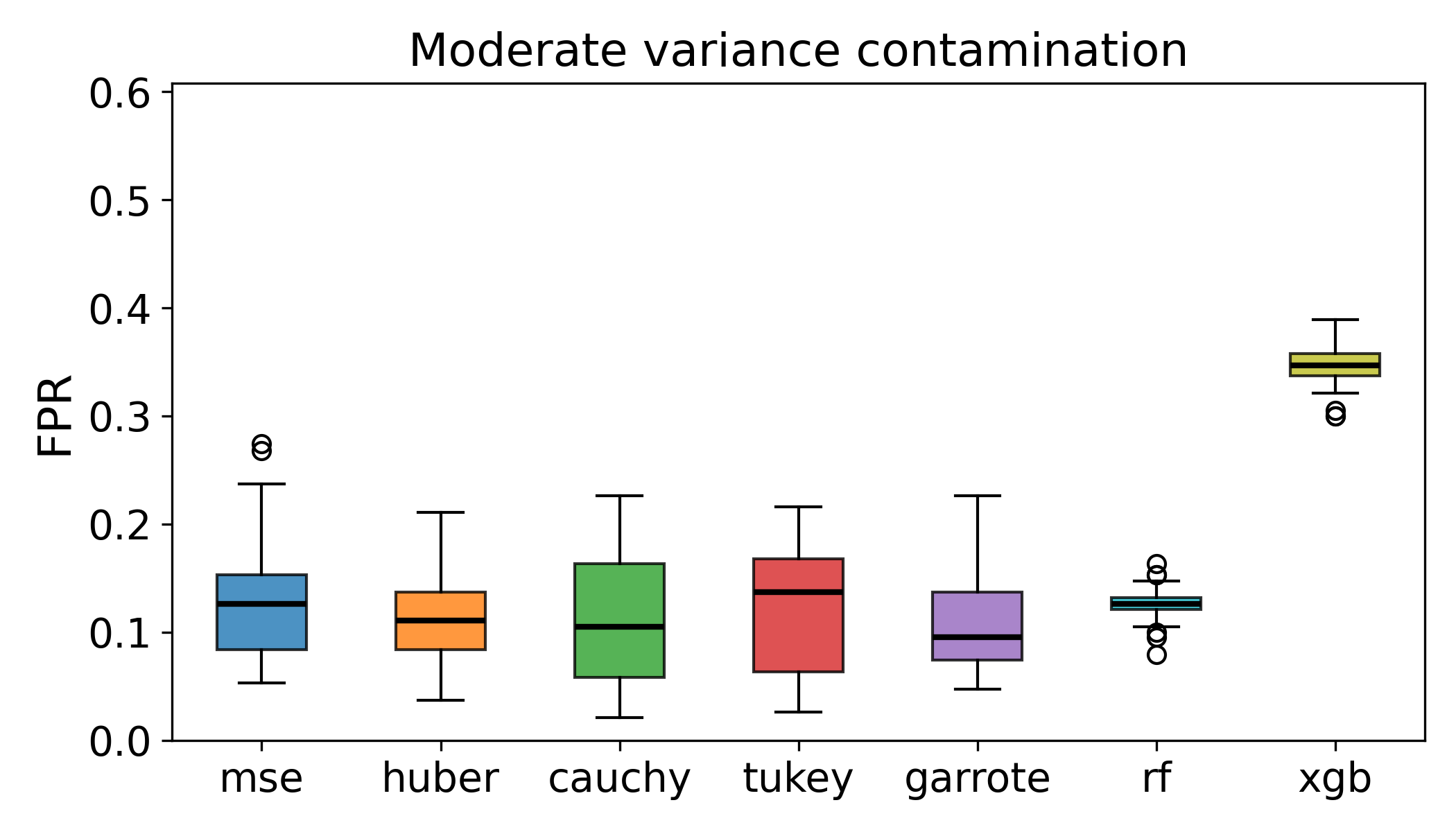}
	\includegraphics[width=0.32\textwidth,height=0.25\textwidth]{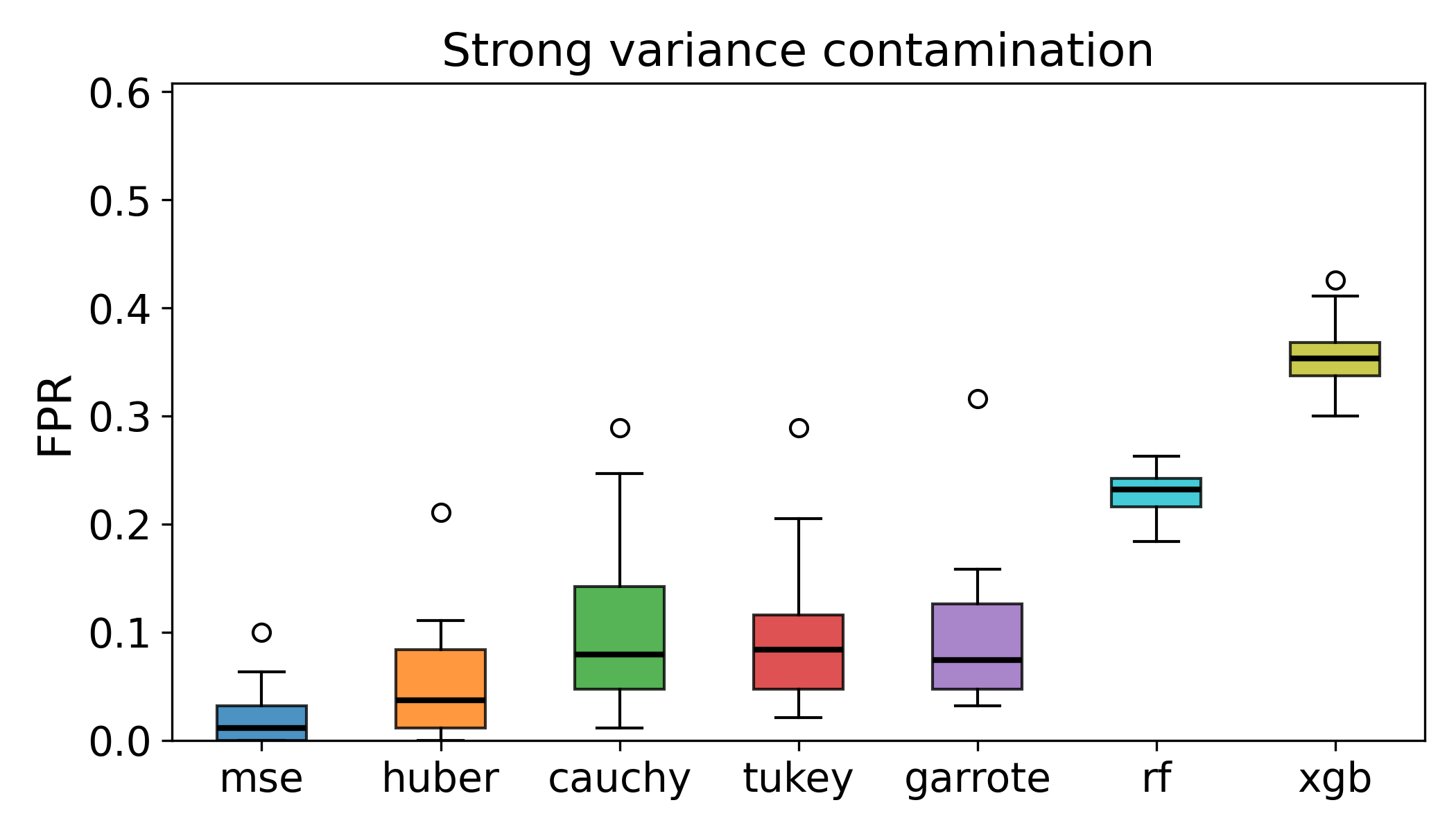}
	
	\vspace{0.2cm}
	
	% --- Row 5 ---
	\includegraphics[width=0.32\textwidth,height=0.25\textwidth]{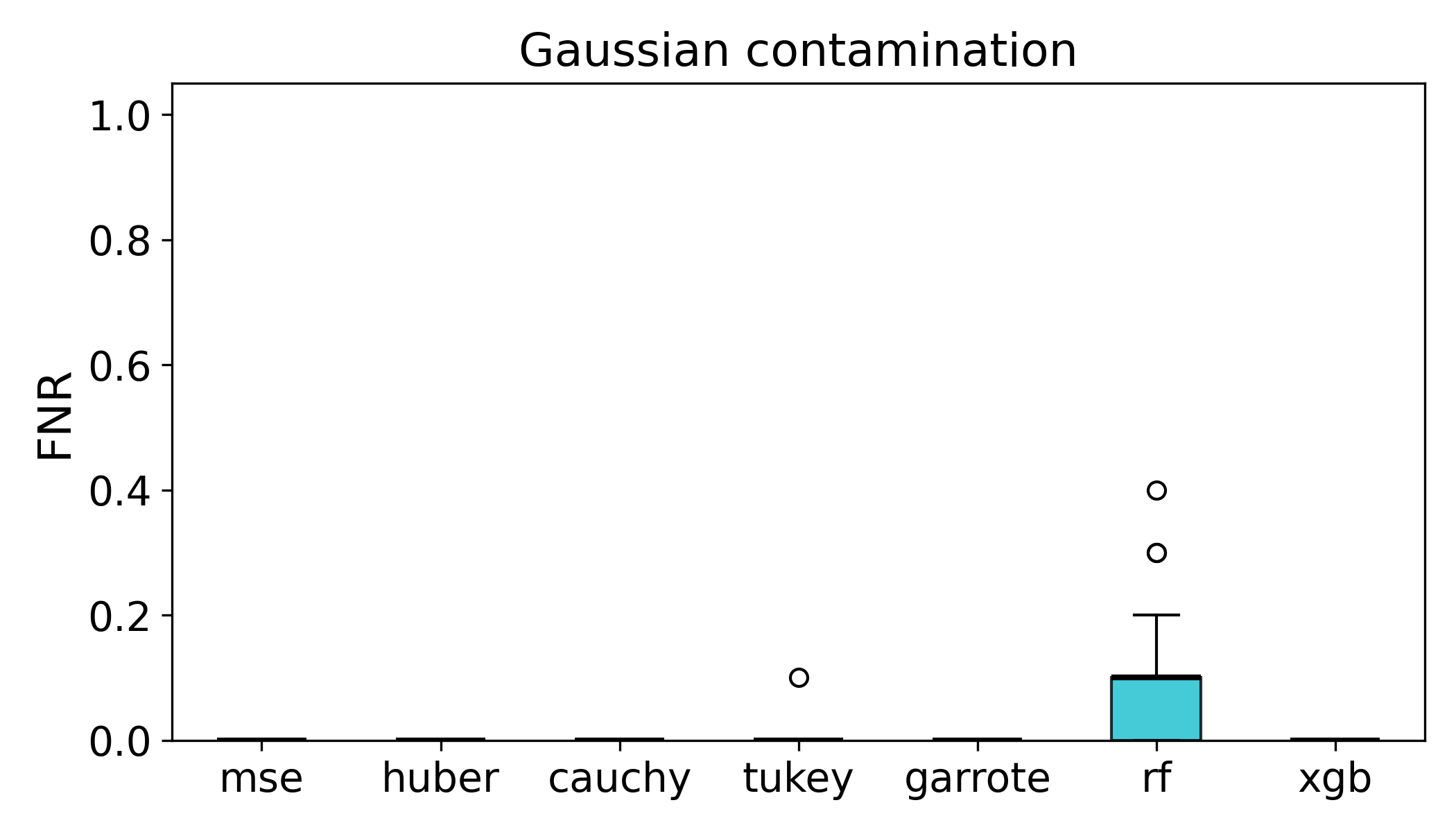}
	\includegraphics[width=0.32\textwidth,height=0.25\textwidth]{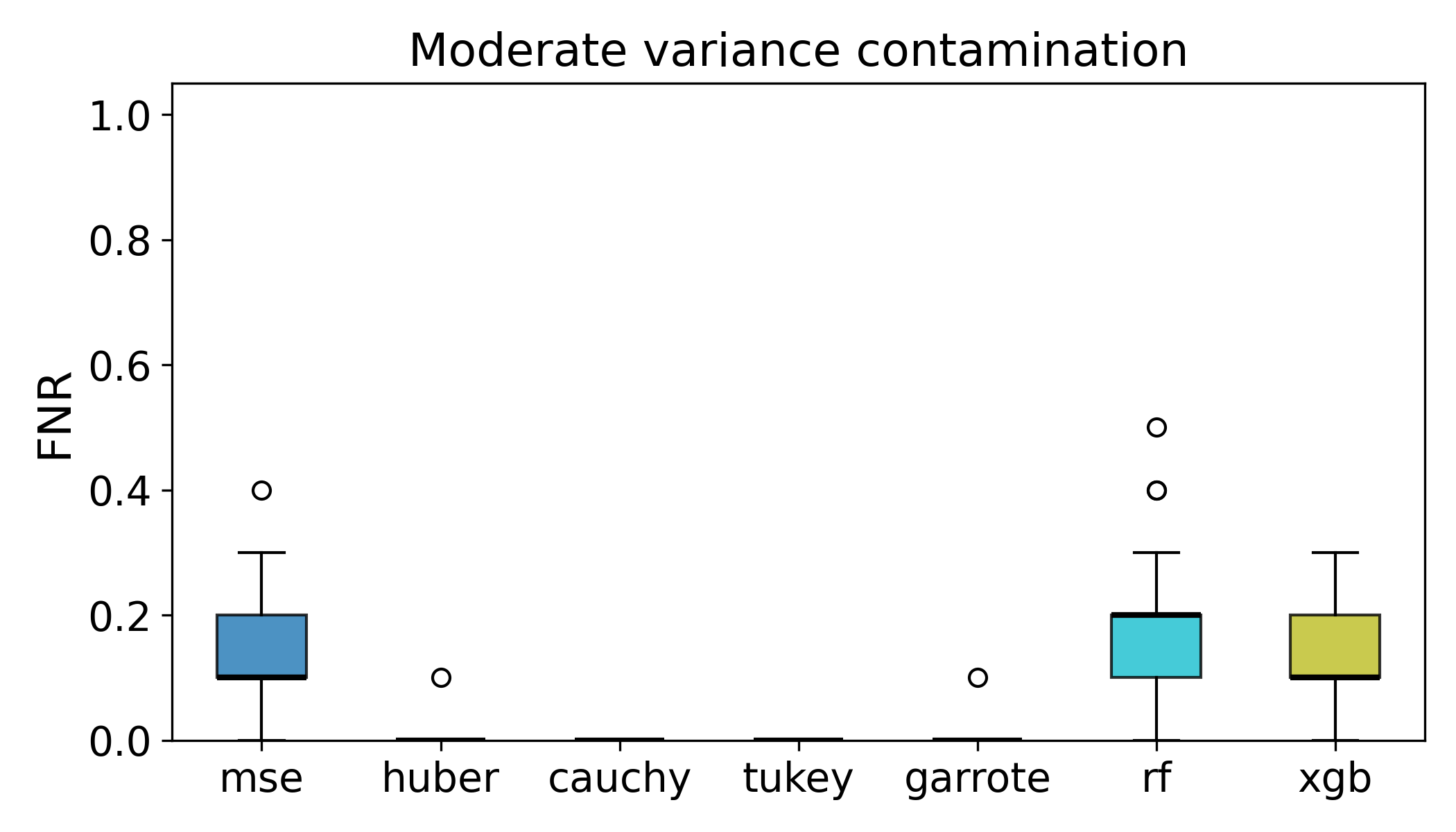}
	\includegraphics[width=0.32\textwidth,height=0.25\textwidth]{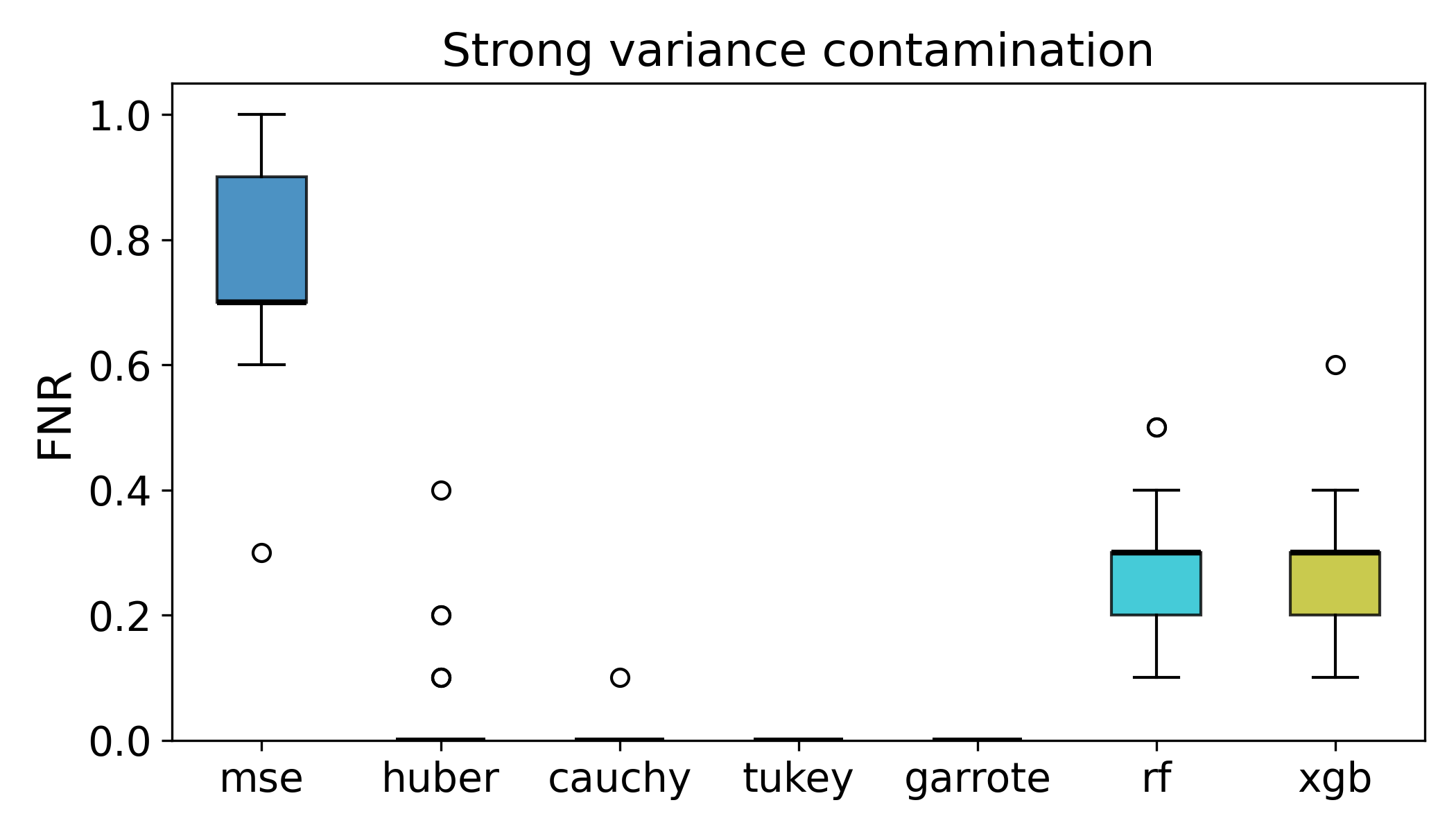}
	
	\caption{Boxplots of the performance metrics results for {\tt model 4} with $10\%$ contamination. First column refers to {\tt gaussian} scenario; second column to {\tt moderate variance} scenario; third column to {\tt strong variance} scenario. The first row shows a realization of the true and contaminated training response $\mathbf{y}$ for all the scenarios.}
	\label{fig_boxplot_model4_variance}	
\end{figure*}

\clearpage

\section{Hyperparameter choice}

In this section, we report Table 13, which summarizes the choices made for the implementation of the Random Forest and XGBoost methods. All other parameters are left at their default value.
%By default the baselines use a fixed, lightly-regularised configuration
%(Table~\ref{tab:hparams}). These are deliberately \emph{not} the library
%defaults: XGBoost in particular uses shallower trees, a lower learning rate and
%row/column subsampling, because the out-of-the-box configuration
%($\code{max\_depth}=6$, $\code{learning\_rate}=0.3$, no subsampling) tends to
%overfit on the small sample sizes of these datasets ($n=242$ and $n=119$). The method-defining parameters (the RF criterion; the XGB objective and, for Top Gear, the native-categorical settings) are always applied.

\begin{table}[h!]
\centering
\caption{Fixed hyperparameters used by the tree baselines.}
\label{tab:hparams}
\small
\begin{tabular}{lll}
\toprule
Learner & Parameter & Value \\
\midrule
RandomForest & \code{criterion}        & \code{absolute\_error} \\
             & \code{n\_estimators}    & 300 \\
             & (all others)            & scikit-learn defaults \\
\midrule
XGBoost      & \code{objective}        & \code{reg:pseudohubererror} \\
             & \code{huber\_slope}     & 1.345 \\
             & \code{n\_estimators}    & 300 \\
             & \code{max\_depth}       & 3 \\
             & \code{learning\_rate}   & 0.05 \\
             & \code{subsample}        & 0.8 \\
             & \code{colsample\_bytree}& 0.8 \\
             & \code{reg\_lambda}      & 1.0 \\
             & \code{importance\_type} & \code{gain} (used for selection) \\
             & \code{enable\_categorical}, \code{tree\_method} & \code{True}, \code{hist} (Top Gear only) \\
\bottomrule
\end{tabular}
\end{table}

\clearpage

%\section{Author Contributions} All authors contributed equally to this work.


%\bibliographystyle{cas-model2-names}

\bibliography{biblio_robustLassoNet}



\end{document}